%% file: main.tex
\documentclass{article}

\usepackage{arxiv}
\usepackage[dvipsnames]{xcolor}
\usepackage[utf8]{inputenc} 
\usepackage[T1]{fontenc}    
\usepackage[colorlinks=true]{hyperref}
\hypersetup{
colorlinks=true,
linkcolor={red},
urlcolor={blue},
filecolor={black},
citecolor={Plum},
}
\usepackage{float}
\usepackage{url}            
\usepackage{booktabs}       
\usepackage{amsfonts}       
\usepackage{nicefrac}       
\usepackage{microtype}      
\usepackage{graphicx}
\usepackage[square, colon, numbers]{natbib}
\usepackage{doi}
\usepackage[labelfont=bf]{caption}
\usepackage{siunitx}
\usepackage{amsmath, amsfonts, amssymb}
\usepackage{bbm}
\usepackage{bbold}
\usepackage{tikz}
\usetikzlibrary{calc, shapes.geometric, shapes.arrows, arrows.meta, shapes.symbols, positioning, fit, backgrounds}
\usepackage{pgfplots}
\usepgfplotslibrary{fillbetween}
\usepgfplotslibrary{groupplots}
\usepackage{pgfplotstable}
\pgfplotsset{compat=1.18}
\usepackage[table]{xcolor}
\usepackage[acronym, nonumberlist]{glossaries}

\usepackage{fontawesome5}
\usepackage{adjustbox}
\usepackage{longtable}
\usepackage{pdflscape}
\usepackage{booktabs}
\usepackage{marvosym}

\definecolor{eggshell}{HTML}{f4f1de}
\definecolor{peach_fuzz}{HTML}{eab69f}
\definecolor{burnt_peach}{HTML}{e07a5f}
\definecolor{twilight_indigo}{HTML}{3d405b}
\definecolor{muted_teal}{HTML}{81b29a}
\definecolor{apricot_cream}{HTML}{f2cc8f}
\definecolor{light_indigo_purple}{HTML}{5F4B8B}
\definecolor{light_light_apricot}{HTML}{E69A8D}
\definecolor{spicy_paprika}{HTML}{E4572E}
\definecolor{twilight_indigo}{HTML}{29335C}
\definecolor{orange}{HTML}{F3A712}
\definecolor{teal}{HTML}{177e89}
\definecolor{dark_teal}{HTML}{084c61}
\definecolor{scarlet_rush}{HTML}{db3a34}
\definecolor{sunflower_gold}{HTML}{ffc857}
\definecolor{dusk_blue}{HTML}{355070}
\definecolor{dusty_lavender}{HTML}{6d597a}
\definecolor{rosewood}{HTML}{b56576}
\definecolor{light_coral}{HTML}{e56b6f}
\definecolor{light_bronze}{HTML}{eaac8b}

\newcommand{\getErrorDataFilePath}[5]{Data/#1/#2/#3/#2_rounds_#4_standard_#5_#1.txt}
\newcommand{\getFullErrorDataFilePath}[5]{Data/#1/#2/#3/#2_full_#4_standard_#5_#1.txt}
\newcommand{\getCrisisErrorDataFilePath}[6]{Data/#1/#2/#3/#2_rounds_#4_standard_#5_#1_#6.txt}
\newcommand{\getSelectedFeaturesDataFilePath}[5]{Data/#1/#2/#3/#2_rounds_#4_standard_#1_#5.txt}
\newcommand{\getFeaturesImportanceDataFilePath}[4]{Data/#1/#2/#3/#2_rounds_#4_standard_feature_importance_#1.txt}
\newcommand{\getImportanceDataFilePath}[4]{Data/#1/#2/#3/#2_rounds_#4_standard_context_importance_#1.txt}
\newcommand{\getContextAblationDataFilePath}[5]{Data/#1/#2/#3/#2_rounds_#4_standard_context_ablation_#5_#1.txt}
\newcommand{\resolutionMonth}{month}
\newcommand{\resolutionDay}{day}
\newcommand{\linear}{linear}
\newcommand{\xgboost}{xgboost}
\newcommand{\tabicl}{tabICL}
\newcommand{\errorFolder}{error}
\newcommand{\ablationFolder}{ablation_study}

\newcommand{\contextimpFolder}{context_importance}

\newcommand{\Econ}{econ}
\newcommand{\noEcon}{no_econ}
\newcommand{\EconNoVehicles}{econ_no_vehicles}
\newcommand{\EconNoHeatPumps}{econ_no_heatpumps}
\newcommand{\EconNoVehiclesNoHeatPumps}{econ_no_vehicles_no_heatpumps}
\newcommand{\Covid}{covid}
\newcommand{\Sobriety}{sobriety}
\newcommand{\rmse}{rmseh}
\newcommand{\rmsess}{rmsessh}
\newcommand{\mape}{mapeh}
\newcommand{\mapess}{mapessh}
\newcommand{\bias}{biash}
\newcommand{\variance}{varianceh}

\def\ridgewidth{8cm}
\def\ridgeheight{3cm}
\def\vsep{3.5cm}      
\def\hsep{2cm}        
\def\xMin{0}
\def\xMax{15}          

\pgfplotsset{
  ridge base/.style={
    width=\ridgewidth,
    height=\ridgeheight,
    scale only axis,
    ybar,
    bar width=10pt,
    xmin=\xMin, xmax=\xMax,
    ymin=-1,
    ymax=1,
    clip=false,
    axis background/.style={fill=white},   
    axis line style={draw=none},
    ytick=\empty,
    axis y line=none,
    every axis plot/.append style={fill=orange, draw=none},
  }
}

\makenoidxglossaries
\newacronym{stlf}{STLF}{Short-Term Load Forecasting}
\newacronym{mtlf}{MTLF}{Medium-Term Load Forecasting}
\newacronym{ltlf}{LTLF}{Long-Term Load Forecasting}
\newacronym{tso}{TSO}{Transmission System Operator}
\newacronym{dso}{DSO}{Distribution System Operator}
\newacronym{sarimax}{SARIMAX}{Seasonal Auto-Regressive Integrated Moving Average with eXogenous}
\newacronym{gam}{GAM}{Generalized Additive Model}
\newacronym{ann}{ANN}{Artificial Neural Network}
\newacronym{fm}{FM}{Foundation Model}
\newacronym{fms}{FMs}{Foundation Models}
\newacronym{tsfm}{TSFM}{Time-Series Foundation Model}
\newacronym{tsfms}{TSFMs}{Time-Series Foundation Models}
\newacronym{tfm}{TFM}{Tabular Foundation Model}
\newacronym{tfms}{TFMs}{Tabular Foundation Models}
\newacronym{rf}{RF}{Random Forest}
\newacronym{gbt}{GBT}{Gradient Boosted Trees}
\newacronym{ml}{ML}{Machine Learning}
\newacronym{lstm}{LSTM}{Long-Short-Term Memory}
\newacronym{llm}{LLM}{Large Language Model}
\newacronym{llms}{LLMs}{Large Language Models}
\newacronym{insee}{INSEE}{Institut National de la Statistique et des Etudes Economiques}
\newacronym{sdes}{SDES}{Service de la Donnée et des Etudes Statistiques}
\newacronym{cpi}{CPI}{Consumer Price Index}
\newacronym{ipi}{IPI}{Industrial Production Index}
\newacronym{spi}{SPI}{Service Production Index}
\newacronym{gdp}{GDP}{Gross Domestic Product}
\newacronym{ev}{EV}{Electric Vehicles}
\newacronym{era5}{ERA5}{European Re-Analysis v5}
\newacronym{nuts}{NUTS}{Nomenclature of Territorial Units for Statistics}
\newacronym{mse}{MSE}{Mean Squared Error}
\newacronym{mae}{MAE}{Mean Absolute Error}
\newacronym{rmse}{RMSE}{Root Mean Squared Error}
\newacronym{mape}{MAPE}{Mean Absolute Percentage Error}
\newacronym{inria}{INRIA}{Institut National de la Recherche en Informatique et en Automatique}
\newacronym{hdd}{HDD}{Heating Degree-Days}
\newacronym{cdd}{CDD}{Cooling Degree-Days}

\tikzstyle{signal_node} = [shape=signal, rounded corners, text centered, text=black,minimum width = 2cm, minimum height = 1cm, signal to = right]
\tikzstyle{round_rectangle} = [rectangle, rounded corners, minimum width = 2cm, minimum height = 1cm, text centered]
\tikzstyle{trapezium_node} = [trapezium, trapezium left angle=70, trapezium right angle=110, minimum width = 2cm,  minimum height = 1cm, text centered]
\tikzstyle{diamond_node} = [diamond, minimum width = 2cm,  minimum height = 1cm, text centered]
\tikzstyle{arrow}= [thick, ->, >=stealth]
\tikzstyle{arrow_dashed}= [-,dashed, >=stealth]

\title{Medium-Term Multi-Resolution Electric Load Forecasting using Economic Data and Foundation Model}

\date{} 					

\author{\href{https://orcid.org/0009-0003-3947-2101}{\includegraphics[scale=0.06]{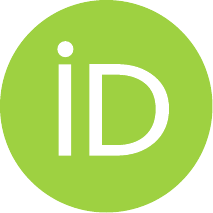}\hspace{1mm}Eloi LINDAS}\textsuperscript{1,3}\thanks{Corresponding author : \href{mailto:eloi.lindas@lsce.ipsl.fr}{eloi.lindas@lsce.ipsl.fr}}\hspace{3cm} Yannig GOUDE\textsuperscript{2,4} \hspace{3cm} Philippe CIAIS\textsuperscript{3}\\
\\
\textsuperscript{1}Atos Inno'Lab TS Bezons, Bezons, France\\
\textsuperscript{2}EDF R\&D Lab, EDF, Palaiseau, France\\
\textsuperscript{3}Laboratoire des Sciences du Climat \& de l'Environnement, CEA, CNRS, UVSQ, \\
Université Paris-Saclay, Gif-sur-Yvette, France\\
\textsuperscript{4}Laboratoire de Mathématiques d'Orsay, CNRS, Université Paris-Saclay, Orsay, France}

\renewcommand{\headeright}{Preprint}
\renewcommand{\undertitle}{Preprint}
\renewcommand{\shorttitle}{Mid-Term Multi-Resolution Load Forecasts using Economic Data and Foundation Model}

\hypersetup{
pdftitle={Paper on Demand Model},
pdfsubject={cs.LG, stat.ML},
pdfauthor={Eloi LINDAS, Yannig GOUDE, Philippe CIAIS},
pdfkeywords={forecasting, electricity load, medium-term, machine learning, economic},
}

\begin{document}

\maketitle
\setcounter{footnote}{0}
\begin{abstract}
Accurate medium-term, from a few months to a few years, electricity load forecasts are crucial for informed decision-making in power plant maintenance scheduling, load dispatch and price settlement. Being comprised between \acrfull{ltlf} which uses mostly economic projections and appliances development scenarios, and \acrfull{stlf} driven by weather, calendar and autoregressive patterns, \acrfull{mtlf} requires both extrapolation capabilities and variability modeling. Yet, it remains unclear if  \acrshort{mtlf} can benefit from economic indicators, and especially at which forecast horizon and resolution. To address these challenges we investigated the impact of socioeconomic data on predictions issued $1$ month and up to $48$ months in advance for France at monthly and daily resolution using a tabular \acrfull{fm}. A dataset covering 20 years of observations of electricity load, weather variables and economic features such as consumer price and production indices, electric vehicle counts or employment is created for the study. To avoid noisy data, we used a new feature selection pipeline, creating ensemble of expert models with diverse feature subsets, to demonstrate that selected economic covariates improve forecast skill by $\SI{20}{\percent}$ over 2015-2025. This enhancement is steady across lead times and resolutions limiting the \acrlong{mape} to $\SI{4}{\percent}$ for monthly granularity and $\SI{5}{\percent}$ for daily granularity. Explainability of the models is investigated through feature and context importance. Results showed that the \acrshort{fm} is limited in the context it leverages pointing towards potential computational savings with a reduced context, while feature importance of economic predictors grows with the forecast horizon. This suggests that including economic data in \acrshort{mtlf} could bridge the gap with \acrshort{ltlf} leading to seamless forecasts while also being suitable for demand projections using climate data and socioeconomic narratives.

\end{abstract}

\keywords{forecasting \and electricity load \and medium-term \and economic \and machine learning}
\paragraph{Data Availability} The datasets curated and developed in this study can be accessed : \href{https://doi.org/10.5281/zenodo.22232294}{here}

\section{Introduction}\label{sec:introduction}
Electric load forecasting is a key task to ensure the efficiency and reliability of modern energy systems. Ensuring that energy is available whenever a switch is activated is taken for granted nowadays despite involving complex mechanisms \cite{hong_2014}. As part of this ecosystem load forecasting is critical to allow smooth operations and management of the energy grid by the \acrfull{tso} and \acrfull{dso}. Not only are load forecasts required for balancing the network at any time and avoid energy demand shortage, but they are also useful for maintenance scheduling, infrastructure planning or price settlement \cite{khuntia_forecasting_2016, nti_electricity_2020}. The diversity of use-cases requiring accurate load forecasts makes it essential and highly valuable for informed decision-making.

Depending on the forecast horizon and thus the scenario, load forecast can be broadly classified into \textbf{three main categories} \cite{khuntia_forecasting_2016, nti_electricity_2020,mamun_comprehensive_2020}. \acrfull{stlf} issues predictions from a few hours to a few days and up to a week in advance. They are mainly used for generator unit commitment, short-term balancing and trading. \acrfull{mtlf} where horizon ranges from a few weeks to a few years is useful for maintenance planning, load dispatch and price settlement as well as risk management. Finally, \acrfull{ltlf} targets predictions from a year to decades in advance to plan for new infrastructure for the power grid, \textit{i.e.}, generation, transmission, and distribution. These different load forecasting classes lead to different forecasting paradigm as the task of predicting load for tomorrow is not the same as predicting the load for that same day next year. Thus, differences in the data, models, techniques as well as temporal granularities being used arise.

As \acrshort{mtlf} is between \acrshort{stlf} and \acrshort{ltlf} it has been understudied compared to the other two forecasting regimes \cite{son_short-term_2017, solyali_comparative_2020,amara_ouali_daily_2021,abumohsen_electrical_2023, bahman_long-term_2025}. It can be attributed to the better knowledge of the factors influencing load at short-term and long-term. Indeed, many studies showed the impact of calendar patterns and weather on load \cite{hong_2014, mamun_comprehensive_2020,  apadula_relationships_2012, staffell_global_2023}. Temperature, which is the main driver for heating and cooling in residential demand, with calendar patterns such as seasonalities, holidays, specific events and load lags already explains most of the short-term variance \cite{son_2020, goude_local_2014}. However, when going for long-term predictions, accurate weather predictions are not available and the short-term patterns of the load, \textit{i.e.}, the autoregressive effect, are least important than the global trend and seasonalities. That's why socioeconomic factors such as economic growth indicators, housing characteristics, policies, and scenario of appliances deployment are leveraged to obtain predictions \cite{elkamel_long-term_2020, grigoryan_electricity_2021, tete_determinants_2024,han_impact_2014, gunay_forecasting_2016}. 

Being comprised in between \acrshort{stlf} and \acrshort{ltlf}, the task of predicting load at medium-term involves learning to extrapolate the global characteristics of the time series while also retaining most of its variability. Therefore, developping a \acrshort{mtlf} model benefits from calendar, weather and socioeconomic data despite significant discrepancies in temporal frequency and coverage between the mutiple sources \cite{zimmermann_efficient_2024,mirasgedis_models_2006,grigoryan_electricity_2021}. Still, it remains unclear which factors are the most helpful for predicting load and if including them would actually improve the accuracy of the model \cite{mir_review_2020,zhao_feature_2012, bouktif_optimal_2018}. Most of the studies use a fixed set of covariates to predict load at medium-term based on literature and expert knowledge. Forecast with horizon ranging in the \acrshort{mtlf} and \acrshort{ltlf} regimes is much more based on power system knowledge than \acrshort{stlf} where purely data-driven methods prove to reach great levels of performance.

Since \acrshort{mtlf} bridges short-term and long-term predictions, one can also ask if the factors that need to be considered in the predictive model change with the forecast lead time and/or resolution. Depending on the horizon considered as well as the temporal granularity which can be fine or coarse, it might prove more efficient to use only certain covariates. Weather and calendar variables might prove to be more useful for smaller lead times than socioeconomic predictors which would only contribute for longer horizons.

Having the right data can prove to be useless without the right model. Since the dependance of load on external variables is highly non-linear, having the right level of model complexity is mandatory in order to take advantage of the predictive information contained in the data. Methods for \acrlong{mtlf} ranges from classical linear and time-series models such as \acrfull{sarimax} and its variant \cite{sen_forecasting_2019, vu_variance_2015, shibayama_design_2023} to \acrfull{ann} such as \acrfull{lstm} \cite{abumohsen_electrical_2023, islam_forecasting_nodate, bashiri_behmiri_incorporating_2023}, encompassing parametric models like \acrfull{gam} \cite{zimmermann_efficient_2024,goude_local_2014} and traditional machine learning techniques like \acrfull{rf}, \acrfull{gbt} or support vectors \cite{sen_forecasting_2021,cordeiro-costas_load_2023, song_2025, dudek_2022}. Of course combining different approaches through hybrid modeling to take the best of different paradigm has also been thoroughly studied \cite{grigoryan_electricity_2021,gonzalez_grandon_electricity_2024}.

When looking into \acrshort{mtlf}, it might seem reasonable to model the different components of the load time-serie independantly. It would facilitate grasping the  short and long-term varying patterns. Decomposing the data into trend and seasonality or multiple frequencies that are modeled independantly can help to grasp the full spectrum of drivers \cite{shibayama_design_2023, gonzalez_grandon_electricity_2024, mo_mrgnn_2025}. Signal processing decomposition methods such as wavelet or empirical mode decomposition have been sucessfully applied to load prediction \cite{liu_monthly_2018, safari_analysis_2020, gupta_2020, bunnoon_2012}. Yet, they lack interpretability as the obtained modes do not equate to explicable frequencies. Conversely, constraining the decomposition into specific frequencies is purely based on expert knowledge and might mix the true underlying processes driving load.

The growth in computing capabilities and data availability in recent years led to a shift of paradigm from machine learning to deep learning and now towards \acrfull{fms}. A \acrlong{fm} is a specific type of deep neural network with billions of parameters which have been pre-trained on millions of datasets both real and synthetic so that they are ready for inference right out of the box \cite{schneider_foundation_2024, awais_2025, yang2023foundation}. The goal of the pre-training step is to avoid the traditional training step on specific data. Once the \acrshort{fm} is pre-trained it can directly infer on new data without the need to tune its parameters. It is called zero-shot prediction. Since the model already learned how to behave on a vast number of datasets during pre-training, it can adapt to a new task which is close to the ones it learned to accomplish during its pre-training. To achieve good results in a zero-shot manner, most \acrshort{fms} leverage in-context learning \cite{dong2024survey, xie2021explanation}. The context being a sample of new data which was not seen during pre-training and is only used at inference time for predicting new data points. It is named after the concept that \acrfull{llms} provide better answers when given context inside the prompt \cite{mei2025survey, zhu_2024}. Even, if \acrshort{fms} offer suprisingly good prediction in zero-shot settings, they can still be fine-tuned to a specific task by updating the model parameters on data.

A diversity of \acrlong{fms} exists, each suited for different predictive endeavors such as language or computer vision specific tasks. Forecasting the electric demand is a time-series predictive problem. Hence, \acrfull{tsfm} built to handle time-series data such as Chronos, TiRex or Moirai among others are suitable for this application \cite{ansari2024chronos,ansari2025chronos2,auer2025tirex, woo2024moirai, liu2026moirai2}. Some of these models are purely autogressive and do not include external covariates in their inputs (Chronos or TiRex\footnote{The newer versions Chronos-2 and TiRex-2 added the possibility to use past and future knwon covariates.}). Additionally, practitioners do not control how much past context is utilized to make predictions making their explainability difficult. That said, load modeling can also be treated within a tabular regression framework meaning that \acrfull{tfms} designed for tabular data can be applied, given that suitable time dependant features are incorporated. Among tabular \acrlong{fm} one can turn towards prior-fitted networks such as TabPFN or TabICL which have a bayesian interpretation \cite{hollmann2023tabpfn, grinsztajn2026tabpfn2, qu2025tabicl, qu2026tabiclv2}. The pre-training phase can be understood as learning a prior of distributions over the predictive task. Then at inference the model uses the context data to adjust the posterior predictions accordingly. Moreover, such tabular models provide better interpretability capabilities through feature importances and can easily be adapted to time-series prediction making them highly suitable for our research \cite{hoo2026, fonseca2026explainerpfn}. 

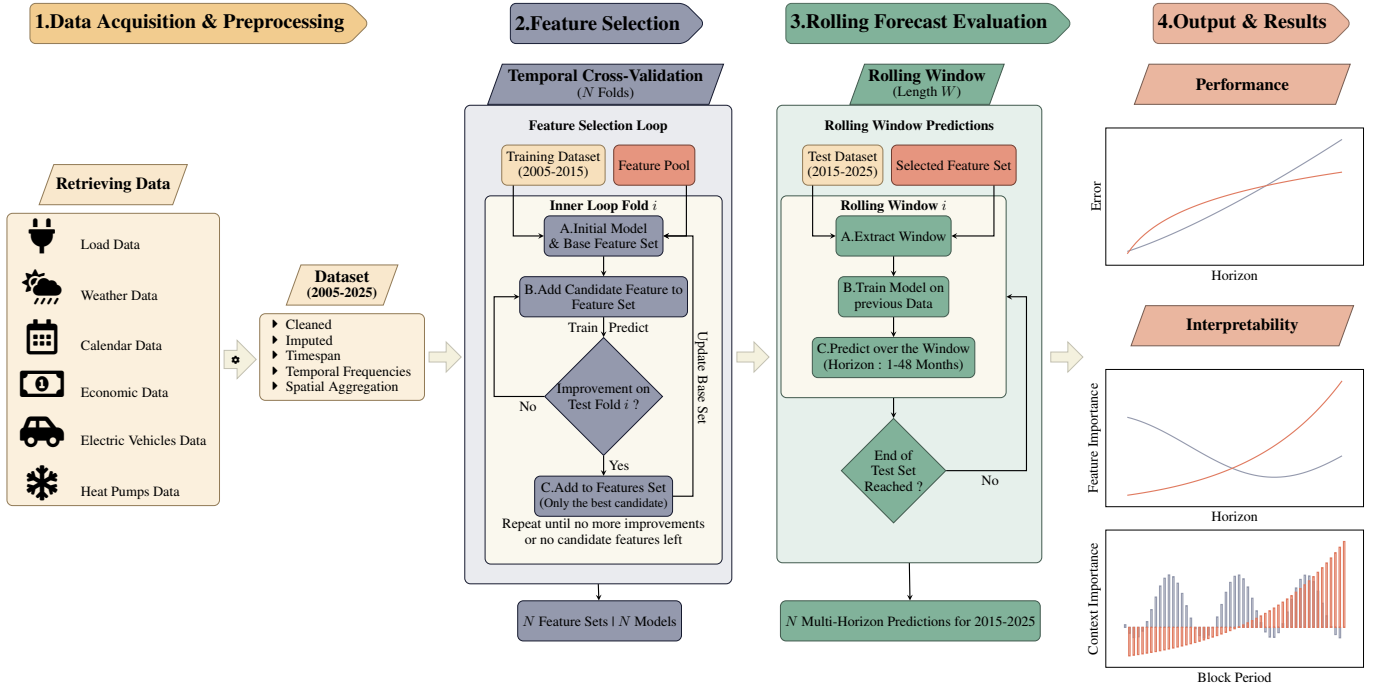
\begin{figure}[h!]
  \centering
  \begin{adjustbox}{max width=\textwidth}
	\input{Scheme/Tex/workflow.tex}
  \end{adjustbox}
  \caption{Modelling framework of this study presented schematically. The raw data are preprocessed prior to being included into a feature selection procedure with the model. Once the best feature sets are obtained the model is evaluated using a rolling forecast methodology to retrieve forecasts for different horizons. This allows to compute the performance as well as the interpretabiity metrics of the model with regard to the forecast horizon.}
  \label{fig:workflow}
\end{figure}

Therefore, the main objective of the paper is to study the impact of socioeconomic covariates on medium-term load predictions using a tabular \acrlong{fm}. We will assess whether such data bears predictive power and if yes, at which horizon and resolution by issuing forecasts at multiple lead times ranging from $1$ months to $48$ months for two resolutions : daily and monthly. Tabular \acrlong{fm} suitability for \acrshort{mtlf} will also be investigated through performance evaluation and comparison with a linear model and a tree-based gradient boosting scheme, which was among the best methods available for tabular data before the arrival of \acrshort{tfms}. Attention will also be paid to the interpretability of the models in order to provide recommandations to practitioners on which factors should or should not be included in the modeling framework. Section \ref{sec:data} describes the data collection and transformation process while the methods designed for the study are presented in section \ref{sec:methods}. Model performance and interpretability through  feature and context importance will be discussed in section \ref{sec:results} before summarizing the findings of this work in section \ref{sec:conclusion}. Figure \ref{fig:workflow} schematically represents the framework of the study.

\section{Data}\label{sec:data}
The following sections present the data collection and preprocessing steps. Load, calendar, weather and socioeconomic data are described in sections \ref{sec:demand_data} to \ref{sec:economic_data} while the pretreatment of the data leading to a suitable dataset for our modelisation pipeline is explained in section \ref{sec:preprocessing}.

\subsection{Electricity Demand Data}\label{sec:demand_data}
RTE, the French \acrshort{tso} provides real time data at 15 minutes resolution openly on their eCO2mix\footnote{The data can be accessed here : \url{https://www.rte-france.com/en/data-publications/eco2mix}} website for power generation, transmission, and demand. A history starting in 2012 can be accessed, with different quality guarantees depending on the recency of the data, offering $12$ years of data. However, as it is ideal for short-term context it proves to be too short for a medium-term prediction framework. Further past data can be collected, but without the quality guarantees from the \acrshort{tso}. Both monthly\footnote{The monthly data can be accessed here : \url{https://analysesetdonnees.rte-france.com/en/consumption/consumption-global} by clicking on "Graph Download Data" of the first graph. Accessed in May 2026.} and hourly\footnote{The hourly data can be downloaded here : \url{https://www.services-rte.com/en/download-data-published-by-rte.html?category=consumption&type=short_term} with one file per year. Accessed in May 2026.} data can be obtained since 1995. Doing so allowed us to double the history length by creating a 25 years dataset of electricity load for France at monthly and daily resolution. Daily granularity comes from the resampling of hourly data which helped to smooth the errors that could be found in the fine resolution data. 

\subsection{Socio-Economic Data}\label{sec:economic_data}
To assess the utility of socioeconomic covariates for \acrshort{mtlf} we gathered several indicators from \acrshort{insee} (the French national Census Bureau), Eurostat (the European Statistics Office) and \acrshort{sdes} (a French data service for energy, transports, and environment). We chose to retrieve indicators that sensibly gave information about the economic situation of the country such as \acrfull{gdp}, \acrfull{cpi} for inflation measure and production indices including \acrfull{ipi} and \acrfull{spi}. Even if industries are still heavily relying on fossil fuels, some like train transportation or information and broadcasting use electricity as their main energy source. Hence, these indicators act as a proxy of the industrial consumption. We also included population and employment data as they can be drivers of the load for longer timescales. France is one of the most visited country in the world, and tourist travel can occur an increase in electricity demand especially for summer and winter months due to an amplification of heating and cooling needs. So to account for the tourism effect we collected the number of nights spent in tourism accomodations such as hotels, campsites, vacation rentals \dots

Since the  energy consumption of a household is directly linked to its energy expense and purchasing power we added the household disposable income. Finally, to try to take into account the rapid growth of electric and hybrid (pluggable) vehicles and heatpumps during the last decade we also included indicators of the number of \acrfull{ev} registered in France and the number of installed heatpumps. In summary, we gathered around $40$ socioeconomic variables with $13$ being linked to the production in diverse industrial and service fields, $8$ main indicators of inflation, growth and population, $1$ for tourism, $4$ for heatpumps and $8$ on transportations. More detailed information are given in appendix \ref{appendixB} and table \ref{tab:features}.\\
Due to the various sources of economic data and the diversity of the variables retrieved, we ended up with various temporal granularities, mostly monthly, quarterly, and yearly. To match the history length of the electricity demand data we had to limit ourselves to features that are measured consistently for decades. Despite being available, housing characteristics such as dwellings surface, construction years, heating technology or other census derived indicators did not meet this criterion. In fact due to modifications of the variables or of their measurement methodology, we could not retrieve a coherent history for more than ten years. 

\subsection{Weather Data}\label{sec:weather_data}
Our weather data comes from \acrshort{era5} dataset on single levels. We downloaded hourly variables over the $\left[-4.55\si{\degree};42.5\si{\degree}; 7.95\si{\degree};51\si{\degree}\right]$ domain at $0.25\si{\degree} \times 0.25\si{\degree}$ spatial resolution. In addition to temperature, which is known to be the main driver of heating and cooling, we retrieved additional weather parameters than can be used to define an apparent temperature. It is the temperature that someone would feel, and is a better proxy for identifying residential heating and cooling regimes \cite{staffell_global_2023,yang_energy_2008}. For instance, solar radiation and cloud cover as well as wind speeds and relative humidity can have an impact on the apparent temperature. Relative humidity in percent was computed from the $2\si{\meter}$ temperature $T$ and $2\si{\meter}$ dewpoint temperature $T_{d}$ available in \acrshort{era5} using the following equation :
\begin{equation}
	RH = 100\exp\left(\dfrac{A_{1}B_{1}\left(T_{d}-T\right)}{\left(B_{1}+T\right)\left(B_{1} + T_{d}\right)}\right)
\end{equation}
with $A_{1} = 17.625$ and $B_{1} = \SI{243.04}{\degreeCelsius}$  coefficients from the Magnus formula \cite{lawrence_relationship_2005}.

Moreover, practitioners often define degree days to quantify the need for heating and cooling based on temperature \cite{staffell_global_2023, mirasgedis_models_2006,gelegenis_simplified_2009, al-hadhrami_comprehensive_2013}. It stems from the U-shaped curve that one can visualize when plotting the electricity demand against the temperature \cite{hong_2014,mirasgedis_models_2006}. The concept defines two threshold temperatures, one for heating $T_{h}$ and one for cooling $T_{c}$ with $T_{h}<T_{c}$ to represent the thresholds below/above which heating/cooling is needed. In between those two temperatures, it is considered that the temperature is comfortable enough not to require any modifications. Usually, heating and cooling degree days are defined at the daily scale using the following equations:
\begin{align}
	HDD &= \max\left(T_{h} - T\right) \\
	CDD &= \max\left(T - T_{c}\right)
\end{align}
with $T$ the daily average temperature. These variables are frequently used to quantify building heating and cooling needs and thus several definitions of $T_{h}$ and $T_{c}$ can be found in the literature, trying to account for the apparent temperature with wind chills or the building thermal inertia \cite{day_new_1999,kheiri_split-degree_2023}. Eurostat uses the same thresholds for all european countries with $T_{h} = \SI{15}{\degreeCelsius}$ and $T_{c} = \SI{24}{\degreeCelsius}$ with a correction on the temperature deltas\footnote{The explanation can be found in the metadata of the European degree-days dataset: \url{https://ec.europa.eu/eurostat/cache/metadata/en/nrg_chdd_esms.htm}}. 

France experiences different climates depending on the region not to mention that the air-conditioning penetration rate is highly variable through areas. For example south-east of France faces meditteranean climate with a good deployment of air-conditioning while east and north-east has a semi-continental climate with close to no cooling installations. We felt that using the same threshold temperatures would be too restrictive. That's why, we defined $T_{h}$ and $T_{c}$ for each of the 12 regions of continental France. To do so we computed a population averaged temperature over the corresponding \acrshort{nuts} boundaries using \acrshort{era5} and population data. Despite experiencing a huge range of temperature, some areas are not as densely populated as others meaning a lower impact on electricity consumption. We then modeled the relationship between daily regional load and daily temperature using a piecewise linear model with 2 nodes and 3 pieces. One for heating, one for cooling and one to meet both ends. We used an exponentially smoothed daily temperature to account for the weather inertia, \textit{i.e.}, that heating do not occur on a single cold day but rather during winter. We also distinguished working days and holidays. Holidays includes both weekends and public holidays. The final model specification is:
\begin{equation}
	P = \alpha_{1} \max\left(T_{h} - T_{pop}^{smooth}\right) + \alpha_{2}\max\left(T_{pop}^{smooth} - T_{c}\right) + \alpha_{3}\mathbb{1}_{holidays} + \alpha_{0}
\end{equation} 
with $P$ the load in \SI{}{\mega\watt}, $T_{pop}^{smooth}$ the population weighted and exponentially smoothed temperature, $\mathbb{1}_{holidays}$ an indicator function being $1$ if the day is a holiday and $0$ if not and $T_{h}$, $T_{c}$ and $\alpha_{i}$ the parameters of the model to estimate. We fitted the model using regional data available since 2012 as the regional granularity is only available in RTE eCO\textsubscript{2}mix. The estimated parameters for each region are given in appendix \ref{appendixC}. Once we had the threshold temperatures for each of the $12$ regions of continental France, we computed the degree days accordingly using the average daily temperature from \acrshort{era5} for each grid cell assuming that the cooling and heating behaviors are stable over time.
Ultimately, we have $9$ weather predictors among which $3$ were computed using the temperature, and thus are correlated with it. Yet, it can help the model to learn the non-linear transfer function between load and temperature by already including non-linear transformation of the temperature.

\subsection{Calendar Data}\label{sec:calendar_data}
In the load forecasting literature it is common to use calendar covariates to assist the model in grasping seasonalities and temporal patterns along with some specific events. We added the day of the week, day of the year, month of the year, quarter of the year and season covariates as integer representing in which day/month/quarter/season the current day or month being predicted belongs. For instance, day of the week is a categorical variable with $1$ for Mondays and $7$ for Sundays. Month of the year encodes January as ones and December as twelves. Yet, to ensure that January and December or Monday and Sunday are correctly represented as neighboring months/days we encoded them into two distinct features with a sine/cosine encoding. Each periodic calendar feature $V$ with period $\mathcal{T}$ is then transformed into:
\begin{align}
	\begin{split}
		 V_{cos} = \cos(\dfrac{2\pi V}{\mathcal{T}})
	\end{split}
	\begin{split}
		V_{sin} = \sin(\dfrac{2\pi V}{\mathcal{T}}).
	\end{split}
\end{align}

Substantial drop in load occurs on weekends compared to weekdays. We created a boolean feature which is $1$ if the day is a weekend and $0$ otherwise. At monthly resolution this feature counts the number of weekend days inside the month. These drops also occurs during public holidays and the summer break of August and July specific to summer school vacations in France. Thus, we obtained the list of public holidays\footnote{There is a python module for public holidays \url{https://github.com/etalab/jours-feries-france-data}} dates as well as school vacations\footnote{For school vacations we referred to the following dataset \url{https://www.data.gouv.fr/datasets/le-calendrier-scolaire}. Accessed in May 2026.} dates since 1990. In France, besides the summer and Christmas break, school vacations depends on the location of the school. The country is split into three zones each with a different schedule of school vacations. We encoded the school holidays into maps with the same spatial resolution as \acrshort{era5}. At monthly resolution, those two variables counts the number of public or school holidays a month has.

\subsection{Preprocessing}\label{sec:preprocessing}
The diversity of data types and sources inevitably leads to mismatch in temporal and/or spatial frequencies and temporal coverage that we need to adress. Firstly, load data is available from 1995 onwards but not all data features have the same history length. Weather and calendar data are also available with a long history, but it is not always the case for  socioeconomic data. Despite having limited ourselves to economic indicators that had a long history, there is a data issue preventing us from retrieving \acrlong{spi} before March 2005. The unit of the variable and the methodology changed leading to missing data prior to this date. That's why our dataset covers the period from March 2005 to December 2025, roughly a twenty years history.

Registered electric and hybrid vehicles are yearly predictors available at city scale but only since 2011. However, prior to 2016 and the exponential boom of \acrshort{ev} sales the number of registered vehicles was really low and can be approximated with a linear function. We did this for each city before clipping the extrapolation to $0$ to prevent negative number of vehicles. We assumed that filling the number of registered electric vehicles was possible as they represent a significant share of vehicles only in the last five years of the data. Before 2011, the share is too small to impact electricity consumption at national scale. Though, our idea was to include this covariate along with the number of installed heatpumps to see if they were becoming useful in the recent part of the modeling period. For long-term forecasts deployment scenarios of \acrshort{ev} and heatpumps are now always considered.

After matching the timespan we matched the temporal frequencies. We will developp models for mid-term forecasts at daily and monthly resolution independantly. Depending on the covariate the granularity differs. \acrshort{gdp} is available every quarter, \acrshort{cpi} every month while the number of installed heatpumps is only updated every year. To keep things simple we chose to keep the raw frequency of each feature and duplicate data points to match the desired daily or monthly resolution.

Some data bear more information through spatial patterns. Obviously, weather is different from an area to another, but it is also the case for population and household income or school holidays. For instance, tourism is much more important in coastal areas in summer and in the Alps and Pyrenees in winter. Even if spatial patterns could carry substantial information for prediction we decided to focus on a pure regression framework with tabular models and leave the spatial aspect for further studies. Hence, we aggregated all spatially resolved data to the country scale. All the detailed information about covariates and their spatial aggregation rules are given in appendix \ref{appendixB}.

Our final dataset is made out of 20 years of daily or monthly samples covering the March 2005 to December 2025 period. We chose to divide the dataset in two equal parts. One for selecting the most useful covariates (see section \ref{sec:feature_selection}) for each model where we use the oldest part of the data from March 2005 to December 2015. The other part, from January 2016 to December 2025, will be used for model evaluation through a rolling forecast procedure described in section \ref{sec:evaluation}. Even though the first part is not properly used only for training and the second one for inference we will still refer to them as training and test in the remainder of the paper.

\section{Methods}\label{sec:methods}
This section describes the modelling framework that we used including the feature selection procedure and the rolling forecast evaluation. The models that we used are also presented. A recap of the pipeline is given in figure \ref{fig:workflow}.

\subsection{Horizon \& Resolution}\label{sec:horizon_resolution}
Models will be developped for both daily and monthly resolution. Attention will be paid so that all training, test, evaluation, and cross-validation blocks are the same in between the resolution to ensure temporal consistency and coherence in the analysis of the results. The idea of having two temporal resolution is to verify if socioeconomic covariates bear predictive information for multiple resolution. Indeed, at monthly resolution, most of the economic covariates will share the same granularity hence the transfer function being easier to learn. At daily resolution it can be more difficult as weather and calendar feature will drive the predictor variability. Economic covariates with their coarser granularity will appear more static. There is also a data size effect with more samples available at daily resolution which could help the learning process when monthly resolution will be limited in terms of data and thus model complexity.

When dealing with medium-term forecasts a key question is the horizon. As previously presented the horizon can range from a few weeks to a few years depending on the goal or needs of the task. For our study we will use a broad range of forecast lead times to analyze if economic data utility depends on the horizon. Predictions will be issued with a lead time of $1$ month and up to $4$ years thanks to our data history of 20 years. No autoregressive effect will be used as they become less and less useful as the forecast horizon increase, and they make multi-horizon prediction difficult. We will focus on learning a transfer function converting inputs into load. Since we used tabular regression models, the forecast horizon is defined as the temporal period between the last training timestep and the timestep one wish to predict for. Therefore, a prediction for March 2025 is made one month in advance if the model was trained up until February 2025 but $2$ years in advance if the training stopped in March 2023. 

It is worth mentioning that for meteorologists, our horizon range covers a diverse set of forecasting regime involving subseasonal, seasonal and decadal predictions which lead to different predictability and drivers of said predictability. For example, subseasonal forecasts are more dependent on initial atmospheric conditions coupled to oceanic interactions than seasonal or decadal predictions where climate patterns and teleconnections are key mechanisms. Due to our hindcast setting, we only use observational data for both weather and socioeconomic data, we avoided considerations about forecast availability, accuracy and seamless forecasting for the covariates to focus on the transfer function and see if it can be improved. 

\subsection{Feature Selection}\label{sec:feature_selection}
The final dataset gathers $63$ features that can be leveraged to predict load for mid-term horizons. To assess whether economic data can improve the predictions and especially which economic indicators have significant predictive power, we implemented a feature selection methodology. This selection step is done over the training set prior to evaluating the obtained model with its covariate subset on the test set. This should boost the signal-to-noise ratio concurrently to improving the interpretability and performance of the model by diminishing model complexity.  

There are three categories of feature selection methods \cite{kumar_2014, jovic_2015}. Filtering methods allow to select the features using a criterion prior to any modelisation step. They are model agnostic. Common methods include removing covariate with low variance, selecting features with the highest correlation with the target or use the mutual information or maximum information minimum redundancy criteria \cite{doumeche_human_nodate}. Such methods are not suited in our case because covariates have different temporal granularities and the higher resolved features will dominate the selection process. Embedded methods are a group of techniques that performs the selection directly during model training. The lasso methodology penalize regression coefficients with an $L1$ penalty shrinking useless terms towards $0$. Tree-based models implement a feature importance criterion based on the impurity or prediction error reduction. However, not all models implements an embedded feature selection methods, in particular \acrlong{fms} do not rely on such methods for their interpretability. Last, wrapper method use the model to rank the feature sets. Each feature subset is given to the model for training and inference on a held-out set of data to evaluate which one is the best. It is the most computationally intensive method but usually the one providing the best results.

With $63$ covariates testing every possible subset is intractable computationally as it would imply trying $2^{63} \approx 9.2\times10^{18}$ feature subsets. Thus, we decided to use a forward feature selection procedure. We start with the smallest set of covariates that we know are useful for load forecasting based on the literature. We add each of the remaining features independantly and score them on a held-out set using a metric. Once all the predictors are ranked we add to the set of features the best one, \textit{i.e.}, with the best score. The procedure iterates with the remaining covariates until no more improvement is achieved. Our starting feature subset consisted of temperature, day of the week, day of the year and public holidays for the daily model and temperature, month of the year, number of weekends and number of public holidays for the monthly model. The details can be found in appendix \ref{appendixB}.

As already presented, economic data have a coarser temporal granularity than weather and calendar data. To avoid missing them in the foward feature selection process due to their low frequency, hiding their utility, we divided the procedure in rounds based on the frequencies of the covariates. During the first round, only features with the highest frequency can be selected. This is mostly weather and calendar based predictors. Once we have a subset made from these feature categories we try to add the coarser ones such as economic data. In total, we divided the procedure into three rounds. The advantages are twofolds. In addition to reducing the risk of economic data being obscured by covariates with higher variance it also reduces the computational cost of the feature selection procedure. Three rounds with small search spaces are quicker to complete than one with the entire set of available predictors.

To improve the diversity of the features being selected by the procedure we used the aggregated hold-out scheme \cite{maillard2019aggregatedholdout}. In standard feature selection the feature subsets are scored on a hold-out set of data or multiple sets in case of cross-validation. To rank the subsets, the scores of all sets are averaged. It does encourage specific covariate being selected for specific periods of time. It fosters the selection of a set which is performing well across different situation. In the aggregated hold-out methodology, instead of using the average of the scores to rank the subsets, the feature selection is run on each set or fold. If a single hold-out set is used it does not change anything but paired with cross-validation it leads to an ensemble of selected covariates subsets, one for each fold, that define an ensemble of models. This allows to have $\mathcal{N}$ models, one for each fold of the cross-validation, with the same hyperparameters but different feature sets, increasing the diversity of the prediction compared to a single model with a best in average predictor set.  

We coupled the aggregated hold-out framework with a cross-validation method. Having only one hold-out set of data for ranking the features can lead to overconfidence in the selection of some predictors. Since our load data is temporally ordered we used a time-series specific cross-validation based on a rolling window. Bear in mind that we would like to focus on mid-term predictions. Thus, we split our training data into $10$ folds. Each fold is obtained by shifting the entire set by one month. The test part of the fold has 4 years to cover medium-term forecasting horizons and the training made out of the past data that remains.

To rank the covariates subsets during feature selection we used the standard \acrfull{rmse}. The metrics that are used in this study either for feature selection or model evaluation are defined in appendix \ref{appendixD}.

\subsection{Models}\label{sec:model}
Depending on the task at hand there are several \acrlong{fms} that one can turn to. In our load forecasting setting the two most suitable are \acrlong{tsfms} and \acrlong{tfms}. \acrshort{tsfms} are dedicated for time-series problems however they inevitably include autoregressive effects with no control over the look-back period as it is a meta-parameter of the model. This is not ideal for our framework as we want the model to learn a transfer function from the inputs to the load target so that we can evaluate the impact of including some covariates or not. Therefore, we turned towards \acrshort{tfms} which are firstly designed for tabular prediction tasks but can be adapted for time-series predictive tasks as with any traditional machine learning algorithm.

We chose to use TabICL\footnote{Access the model through its GitHub : \url{https://github.com/soda-inria/tabicl}} \cite{qu2025tabicl,qu2026tabiclv2} as our \acrlong{fm} because it is among the best performer on TabArena \cite{erickson2025tabarena} for both the accuracy and rapidity of its predictions. It is fully open-source and developped by a team of researchers at \acrfull{inria}. This way we would have more control over the model as opposed to commercial solutions. TabICL is a prior-fitted network which have been pre-trained over a large corpus of real and synthetic datasets. The second version TabICLv2 is an upgrade on both preditive performance and inference speed thanks to an improved architecture and pre-training \cite{qu2026tabiclv2}. It also increased the size of the datasets that can be handled with the team claiming that  $50,000$ samples and $100$ features can be operated in under $10$ seconds on a H100 GPU. This is particularly appropriate for our feature selection procedure and medium-term predictive task. While we used TabICLv2 we will refer to it as TabICL in the following paragraphs.

For the sake of comparison we included two other models. They needed to be fast for both training and inference as our workflow, both feature selection and evaluation requires multiple cycles of training/inference either to find the best subset of covariates or make predictions. Neural networks are not appropriate for this study as their training is computationally heavy and requires considerate monitoring. Tree based models on the other hand are quick to both train and infer with while often reaching state-of-the-art performance in tabular regression problems. In fact, XGBoost gradient boosting scheme is among the main traditionnal machine learning algorithm competing with deep neural networks and \acrlong{fms} in TabArena. Furthermore, many load forecasting studies showed great results with XGBoost. That's why we selected it for our study. To be in line with the \acrlong{fm} paradigm we did not do any hyperparameter tuning of the model and used default values.

Finally, as it is usually used for long-term forecasts, we considered a linear model. It is much simpler, and will be more sensible to the noise of the economic data allowing us to verify our hypothesis. We could have used a \acrshort{gam} which outperforms linear models for non-linear tasks, but their development requires careful design of the link equation by the modeler which is out of scope of this study. Indeed, our greedy and automated feature selection process do not leave space to expert knowledge. Still, using our results to know which covariates are important prior to creating specific models with domain expertise is an interesting lead for further research.

In this study, we will focus on point forecast, but the study could be replicated with probabilistic forecasts. TabICL can output a predictive distribution instead of a single value. As a matter of fact the model is intrisically probabilistic, and the point forecast it outputs is in fact the average of the distribution.

\subsection{Evaluation \& Interpretability}\label{sec:evaluation}
Once the models selected the best subsets of covariates on the training part there are $10$ different feature subsets leading to $10$ different models. It must not be seen as a probabilistic ensemble because the members do not represent a distribution but rather an ensemble of expert making different predictions of the same thing. Our final prediction is the average of the $10$ different predictions. Then, standard regression metrics such as \acrshort{mse} and \acrshort{mape} can be computed. We also looked at the variance and bias decomposition of the \acrshort{mse}. The definition of the metrics can be found in appendix \ref{appendixD}.

Over the test part of the data, ranging from 2016 to 2025, we adopted a rolling forecast scheme to be able to retrieve forecasts for a same day or month with multiple lead time. We made predictions for $1$ month up to $4$ years in advance using a shift of one month every time as in the cross-validation procedure used in the feature selection process. Every time, all available past data are used to train the model meaning that it first starts with only the training set accessible and ends with the entire 2005 to 2021 history. Doing so provided us with $85$ forecasts for each lead time. The performance metrics are then computed per forecast horizon to analyse error against prediction horizon patterns.

In addition to performance metrics monitoring, we also dwelved into the interpretability of the models. To begin with we looked at the selected variables during the feature selection procedure to see if some stood out from the crowd. With the aggregated hold-out framework we can compute a selection frequency for each feature based on the number of occurences it has in the $10$ different feature subsets obtained.
Then we computed two types of interpretability metrics. The first one is feature importance based on Shapley values to check if certain covariates are better drivers of load for specific lead time or period of time. We computed them for each expert and each prediction window. Shapley values originate from game theory and enable to distribute the payout, \textit{i.e.}, deviation from the average prediction, among the input features. As they can not always be computed exactly, several approximations methods exist. Depending on the model, their calculation can be expensive even with approximations. Fortunately, TabICL has a specific module to compute Shapley values efficiently.

Since TabICL is a \acrlong{fm} which leverages in-context learning it needs a sample of the new data to be able to infer on it. In our case the sample referred to the training data and future values of the covariates of each forecasting window. The context only serves as adjusting the posterior predictive distribution. Therefore, modifying the context modifies the prediction as with traditional training data. Hence, one can wonder what is important to cover with the context to obtain optimal performance. In fact, when the context length, \textit{i.e.}, number of samples in it, increases the inference time grow rapidly. The complexity of the algorithm being $\mathcal{O}\left(n^{2} + nm^{2}\right)$ with $n$ the number of rows and $m$ the number of columns \cite{qu2026tabiclv2}. Practitioners might want to have the smallest context possible to reduce inference time while retaining most of the accuracy of the predictions. Appendix \ref{appendixL} presents complexity computations on our datasets to illustrate that reducing the context and the number of features can save computation time which can be critical when making operational forecasts. To estimate what is critical in the context and what can be removed we designed a context importance experiment.

The idea is simple, we apply the same rolling window evaluation procedure however we remove a part of the training data from the so-called context. We then have forecasts with the entire context and without. Using these new predictions and the one made with the entire context we can compute the usual metrics. We denote $S$ the score computed when leveraging all the context available and $S_{i}$ when block $i$ has been removed. The context importance of block $i$ is then defined as the difference between the two scores:

\begin{equation}
	CI_{i} = \dfrac{\Delta S_{i}}{n_{i}} = \dfrac{S_{i} - S}{n_{i}}. 
\end{equation}
Here $n_{i}$ is the number of samples remaining in the training set after removing block $i$. In our case $S$ is the \acrshort{rmse}.

Our metric imply that lower is better, meaning that if $CI_{i}>0 \Leftrightarrow S_i>S$ then removing block $i$ from the context actually altered the predictions. The block is important. Consequently, if $CI_{i}<0 \Leftrightarrow S_i<S$ it means that the block $i$ can be removed from the context without any regrets because it alters the model. The different blocks of context are indiced by $i$ because with the rolling window, blocks change with each prediction window. Yet, the $i$ part of the data always refers to the same position. Of course the size of the block $l$ must be chosen wisely as a block too wide will always be significant while a block too small will not have an impact on the predictions of the model. We went with blocks of length $l=1$ year such that there are $16$ blocks for each forecast horizon. 

Bear in mind that our final model is an average of $10$ models with different covariate sets. So the Shapley values were computed for each model before averaging them to get the average contribution of each covariate to the final prediction. For context importance the score is calculated on the average of the different models output directly.

\section{Results \& Discussions}\label{sec:results}
\begin{figure}[h!]
	\centering
	\begin{adjustbox}{max width=\textwidth}
		\input{Plots/RMSE_horizon}	
	\end{adjustbox}
	\caption{\acrshort{rmse} evolution with forecast horizon. Each column presents results for a different model. Top row shows performance for the monthly resolution while bottom row depicts daily resolution. For both resolution the forecast horizon is given in months from $1$ to $48$, \textit{i.e.}, $4$ years. The shaded area represents the $\SI{95}{\percent}$ bootstrapped confidence interval. The $y$ axis has been clipped for the linear models for better visibility.}
	\label{fig:rmse_horizon}
\end{figure}
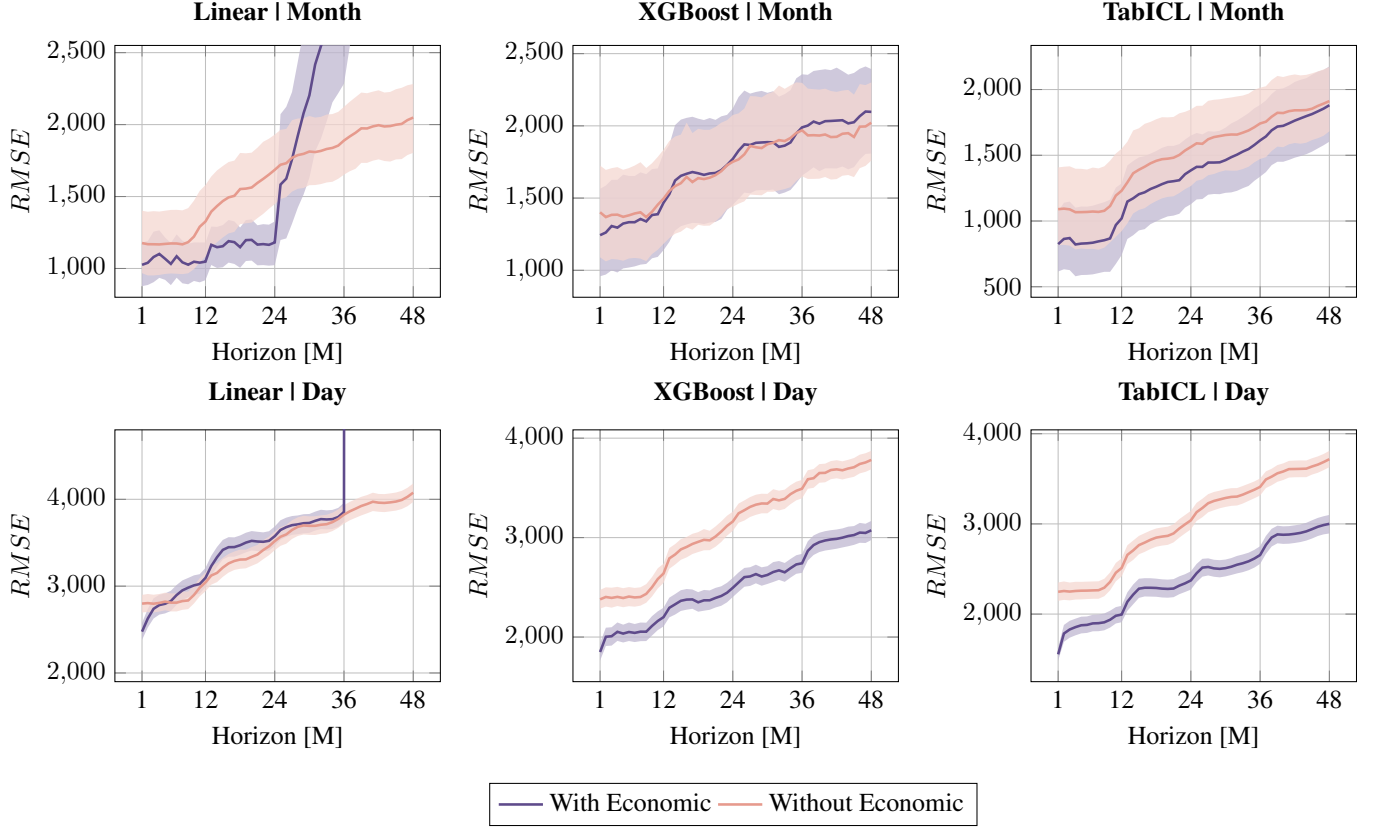

This section presents the performance of our models for both daily and monthly resolution  as well as the interpretability measures such as feature and context importance. The impact of \acrshort{ev} and heatpumps predictors and the selected features are also discussed. Performance is translated to prediction error evolution with forecast horizon graphics. See figure \ref{fig:rmse_horizon} for \acrshort{rmse} and appendix \ref{appendixE} for other metrics. A discussion on the advantage of the feature selection procedure is given in appendix \ref{appendixJ}.

To assess wether socioeconomic variables should be considered for \acrshort{mtlf}, we run the feature selection procedure with and without socioeconomic predictors in the feature search space. The graphics in figure \ref{fig:rmse_horizon} show that including socioeconomic predictors helps to improve the accuracy of the models, but only under certain conditions.\\
TabICL shows better performance with socioeconomic data at both monthly and daily resolution while XGBoost and linear model are more contrasted. Economic features allow to steadily improves the performance of TabICL at daily resolution by $\SI{20}{\percent}$ over the entire range of forecast horizon studied here. On coarser resolution, the improvement decreases as the horizon increases, reaching close to $\SI{0}{\percent}$ for lead time $48$ months as shown in figures \ref{fig:rmse_skill_score_horizon} and \ref{fig:mape_skill_score_horizon}. It means that socioeconomic data helps more for shorter than longer horizons which is counterintuitive. We associate this behavior to the \acrshort{fm} itself more than to the use of socioeconomic data. Especially since it is the only model showing enhancements at such resolution when including economic predictors.
Indeed, at monthly resolution there is close to no differences when including economic covariates for XGBoost. However, at daily resolution it is clear that they bring additionnal information translating to a gain in performance. As for TabICL the gain is steady over the horizon range at around $\SI{20}{\percent}$. For The linear model, there is no difference between the two setup at daily resolution and including economic data is only better for forecasts issued less than $2$ years in advance at monthly resolution. 

The different models enable to vizualize that model architecture should be complex enough, otherwise it can not grasp the relationship between load and economic data. Moreover, it appears that the gap in performance between the two configurations is wider for daily models than monthly models. It suggests that resolution of the predictive target must not match resolution of the economic inputs. It could be induced by the law of large numbers, \textit{i.e.}, having more data helps in unraveling the complex patterns between load and economic predictors. In that case having a long data history is mandatory in order to leverage exotic covariates contributions to the transfer function.

When comparing models, see figures \ref{fig:rmse_model_comparison} and \ref{fig:mape_model_comparison} for easier vizualization, we can see that TabICL proves its superiority for every resolution with and without the economic covariates. To be precise, only the linear model beats it, by a small margin, when economic data are included at monthly resolution for horizon between $1$ and $2$ years. Suprisingly, the performance of the linear model comes close to more complex models at monthly resolution for both experiments except for its overshoot for horizon above $3$ years. It is even keeping up with XGBoost when only calendar and weather predictors are considered. \\
At daily resolution, when data samples are abundant, it can not compete. Having more data also improves the results of XGBoost as the gap between TabICL and XGBoost shrinks. Of course, this is observed with default XGBoost hyperparameters, if tuning was performed we could expect the performance to be even closer. The small gap in prediction error between the two models stresses out the capability of TabICL in zero-shot prediction as well as why XGBoost is among the only traditional \acrshort{ml} models in the TabArena leaderboad. With a smaller dataset, \textit{i.e.}, monthly frequency, we can see the performance gap widening between the two indicating that TabICL is particularly suited for low data regimes, where \acrshort{ml} models are challenged. 

The monthly \acrlong{fm} with economic data limits the \acrshort{rmse} to less than $\SI{1,000}{\mega\watt}$ or $\SI{2}{\percent}$ of \acrshort{mape} (see figure \ref{fig:mape_horizon}) for horizon below $1$ year, while keeping the error below $\SI{2,000}{\mega\watt}-\SI{4}{\percent}$ at longer lead time. Forecasts with horizon comprised between $1$ and $3$ years have an error ranging from $\SI{1,200}{\mega\watt}-\SI{2}{\percent}$ to $\SI{1,800}{\mega\watt}-\SI{3.5}{\percent}$. In comparison, monthly economic XGBoost makes an error of around $\SI{1400}{\mega\watt}-\SI{2.5}{\percent}$ for horizons below a year, between $\SI{1,500}{\mega\watt}-\SI{3}{\percent}$ and $\SI{2,000}{\mega\watt}-\SI{4}{\percent}$ for horizons comprised between a year and $3$ years and around $\SI{2,200}{\mega\watt}-\SI{4.3}{\percent}$ for longer-terms. The linear model can not compete for lead times bigger than $2$ years as its error explodes. \\
For daily predictions, the error is bigger but stays below $\SI{3,000}{\mega\watt}-\SI{5}{\percent}$ at horizon $48$ months for both TabICL and XGBoost. Forecasts with lead time comprised between $1$ and $3$ years have an error of $\SI{2,000}{\mega\watt}-\SI{2.5}{\percent}$ to $\SI{2,800}{\mega\watt}-\SI{4}{\percent}$. Those numbers emphasize the prediction accuracy of the models with economic data for medium-term horizons. A grain of salt needs to be taken though, as those metrics are obtained with observed inputs. In a true operational setup, inputs would be provided through their own forecasting pipeline leading to an increased error margin. 

The superiority of models including economic covariates for TabICL and XGBoost can be further explained through bias and variance (see metrics definition in appendix \ref{appendixC}). In figure \ref{fig:bias_horizon}, we can notice than the monthly models have the same bias with and without the economic data while the daily models exhibits a much lower bias with economic data. This partly explain why they outperform their counterpart without. The other part of the explanation is given by the variance plots in figure \ref{fig:variance_horizon}. Models with economic predictors have much higher variance than models without for both temporal granularities. Having a higher variance is usually associated with overfitting and poor generalisation error, however it is not what's observed here. In our case, a higher variance is actually benefitting the results as it helps to fit complex pattern of load using non-linear relationships with economic indicators.

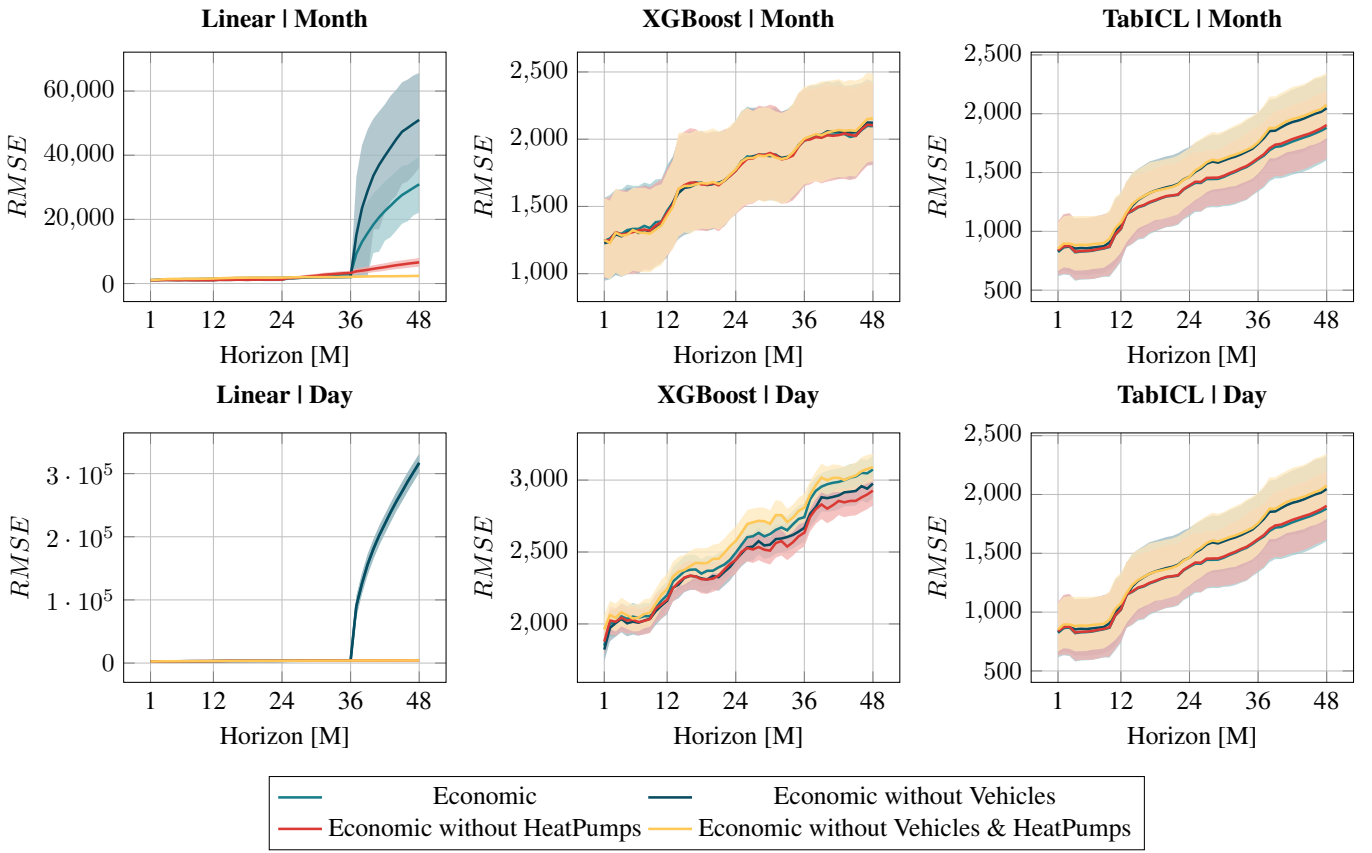
\begin{figure}[h!]
	\centering
	\begin{adjustbox}{max width=\textwidth}
		\input{Plots/rmse_ablation.tex}
	\end{adjustbox}
	\caption{Ablation study of vehicles and heatpumps predictors with evolution of the RMSE with forecast horizon. Each column presents results for a different model. Top row shows performance for the monthly resolution while bottom row depicts daily resolution. For both resolution the forecast horizon is given in months from 1 to 48, \textit{i.e.}, 4 years. The shaded area represents the $\SI{95}{\percent}$ bootstrapped confidence interval. The y axis has been clipped for the linear models for better visibility.}
	\label{fig:rmse_ablation}
\end{figure}

Including economic data is actually improving the performance of the predictive models for mid-term horizons. But, where it shines is during recession periods as they carry information about the contraction of the economy. To evaluate the gain during an economic crisis we looked at the same metrics but computed during COVID crisis and the sobriety period following the start of the Ukraine war in winter 2021-2022. The COVID crisis period was determined using the Oxford Stringency index tracker with a value above $60$ which encompassed the main lockdowns period \cite{hale_global_2021}. The results are presented in appendix \ref{appendixH} in figures \ref{fig:rmse_horizon_covid} and \ref{fig:rmse_horizon_sobriety} for the \acrshort{rmse}. We observe that for both crisis models with only calendar and weather covariates fail to predict the change in load induced by the recession leading to a plateau in performance for all horizons. Models considering socioeconomic indicators in their inputs avoid this plateau and predict better those periods. Again, this is more visible on the daily resolution than the monthly where the gap is small. Even the linear models are better with economic data when including economic predictors except for the daily resolution during the pandemic. These results show that in addition to improving mid-term forecasts, economic data is also enabling better prediction during specific events like economic crisis. Even if our analysis is based on realized data, we still think that the results would stay the same with economic forecasts and expert advice. In case forecasts are not achievable we could turn towards contrasted economic scenarios to propose a diverse set of load projections.

In this work we focused on point forecast but it is worth mentioning that the study could be replicated in a probabilistic setting using probabilistic predictive predictions or specific quantiles through quantile loss.

On every graphic of metric evolution with regard to the forecast horizon, an overshoot of error arises for the linear model with economic predictors at lead time $36$ for the monthly and the daily resolution. As it is not observable when only calendar and weather inputs are used it must stem from some socioeconomic predictors. In fact, this phenomenom is attributed to the electric vehicles and heatpumps covariates that were selected because there is a boom in the number of registered electric vehicles as well as installed heatpumps in recent years, especially after 2020. It leads to a sharp rise of the predictors values, breaking the linearity of the relationship learned on past data. This analysis was made thanks to an ablation experiment whose results are presented in figure \ref{fig:rmse_ablation}. We run the feature selection and evaluation procedures while removing heatpumps and/or electric vehicles covariates from the feature search space. Only the linear model suffers from this problem. This is another example that linear extrapolation hypothesis are not suited for data regimes change. 

Moreover, this ablation study shows that adding the electric vehicle and the heatpump covariates does not help in getting better performance for XGBoost and TabICL. The difference in terms of metric is small at monthly resolution but grows with the finer daily granularity. Our idea was to introduce these covariates as they are markers of the energy transition, representing the recent changes induced by climate change onto societal behaviors. But, it seems to still be too recent to be useful to the models for medium-term forecasting. For \acrlong{ltlf} though, it can not be ignored as the vehicle fleet and heating and cooling means are going to continue to evolve in the next decade.

\begin{figure}[h!]
	\centering
	\begin{adjustbox}{max width=\textwidth}
		\input{Plots/selected_features_month.tex}
	\end{adjustbox}
	\caption{Selection ratio for each feature appearing in one the feature subsets obtained after the feature selection process for the monthly models. Covariates are grouped by their category for easier visualisation. For the monthly resolution, the procedure started from a reduced subset made from : t2m, moy\_cos, moy\_sin, is\_weekend and is\_public\_holiday.}
	\label{fig:selected_features_month}
\end{figure}
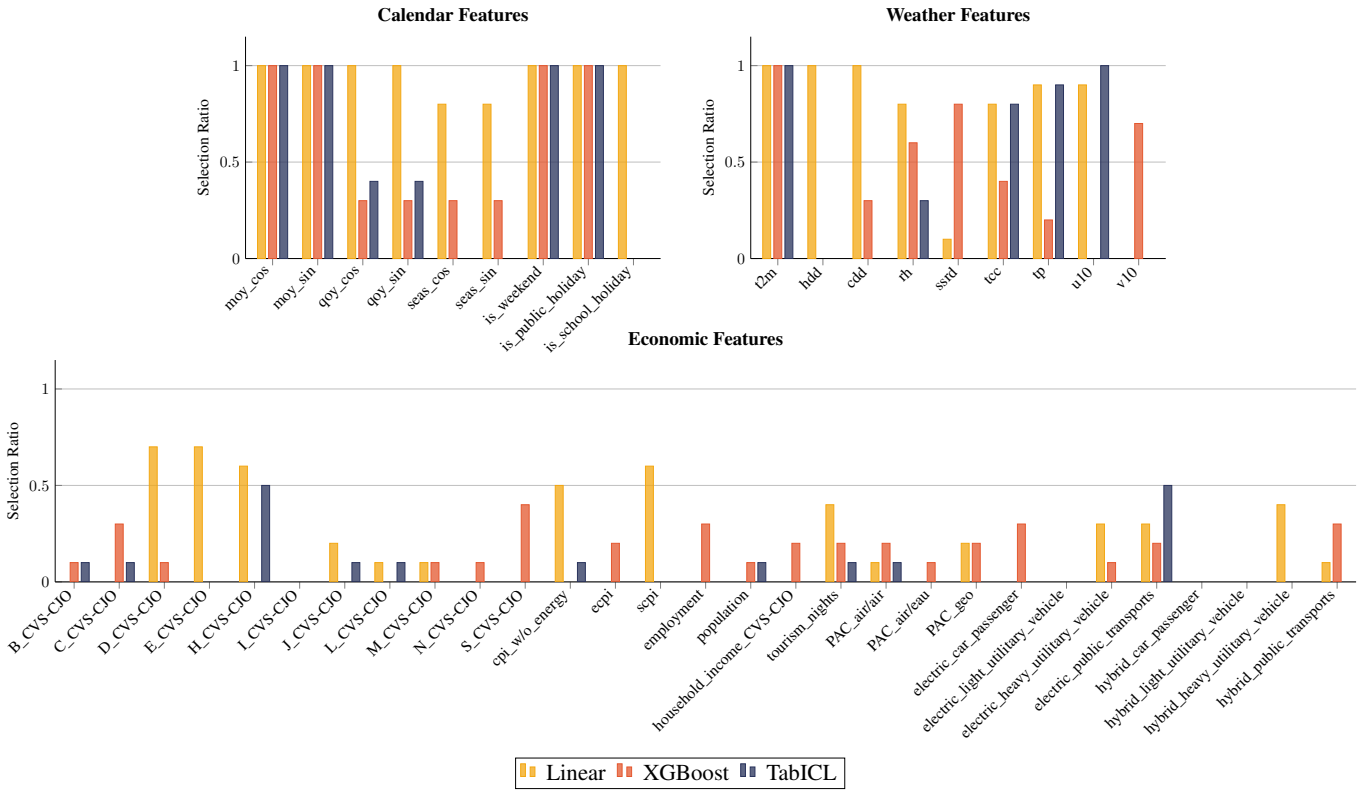

Once we gauged the impact of socioeconomic data on model prediction performance we can dive into the interpretability. First, we can look at the feature subsets that we obtain from the feature selection procedure. There are $10$ different subsets, one for each cross-validation fold, allowing us to compute a selection ratio for each of the predictor involved. It is calculated as the number of occurences of the predictors over the number of folds. The results for the monthly resolution are displayed in figure \ref{fig:selected_features_month} while daily results can be found in figure \ref{fig:selected_features_day}.

The selection ratios show that there are no real consensus among the models and resolutions on which economic predictors are worth including into the model inputs. This is quite surprising considering that they did indeed improve the model predictions. Only a few covariates are selected more than $\SI{50}{\percent}$ of the time and most of them are chosen by the linear model. TabICL and XGBoost are more conservative and select fewer economic covariates. At monthly resolution only the \acrlong{ipi} for transportation and storage (H\_CVS-CJO) and the number of electric public transports are included by TabICL in $\SI{50}{\percent}$ of the subsets. It switches towards the electricity, gas and steam index (D\_CVS-CJO) and the hybrid public transports at daily granularity. Otherwise, the predictors are only selected by one expert in the ensemble. In fact, both XGBoost and TabICL selects most of their economic covariates less than $\SI{30}{\percent}$ of the time. The absence of concordance between models on some features makes it difficult to give recommandation on which data should or should not be included. Still, we can conclude that creating an ensemble of experts from the feature selection increased the diversity in the selected inputs with some covariates performing well on specific periods of time and timescales which led to global enhancements of the performance.

For other covariate categories, \textit{i.e.}, calendar and weather, it is more clear which feature are actually useful. For example the linear model needed more calendar features such as month, quarter, season or weekend indicators, which are selected in more than $\SI{80}{\percent}$ of subsets, for extrapolating at mid-term horizons. More complex models were less reliant on such predictors with XGBoost including month of the year $\SI{60}{\percent}$ of the time at daily resolution and quarter and season indicators in $\SI{30}{\percent}$ of experts for monthly predictions. In comparison, TabICL only used quarter of the year in $\SI{40}{\percent}$ of the subsets at monthly resolution. But, it is aggreeing with the linear model on school holidays for the daily resolution with a selection frequency of $\SI{100}{\percent}$. It could be expected as school holidays are linked with changes in the economic activity in France, especially for the summer break and helps modelize the winter/summer cycle.

Weather predictors exhibit different conclusions. Unsurprisingly, the non-linear transformation of temperature into degree-days is mandatory for linear model with a $\SI{100}{\percent}$ selection rate. It is the only way to model the non-linearity in the load-temperature relationship. \acrshort{hdd} and \acrshort{cdd} are also useful for more complicated models. For instance TabICL selects them in $\SI{80}{\percent}$ and $\SI{90}{\percent}$ of the experts at daily resolution. It is not the case for the monthly resolution though meaning that at coarser resolution temperature itself is sufficient. For other weather variables, wind, precipitation, and nebulosity of which solar radiation and cloud cover are a proxy should be included into the model inputs as they are selected in more than $\SI{50}{\percent}$ of the subsets depending on the algorithm. It is most likely that the models build their own apparent temperature from such covariates. But, again it is model sensitive and thus difficult to provide recommandations.

Despite not having a consensus in covariates between models we were able to look into the feature importance by computing Shapley values. As it is computationally expensive for \acrlong{fm} like TabICL we were able to retrieve them only for the monthly resolution. Figure \ref{fig:feature_importance_month} shows the evolution of the feature importance with the forecast horizon. We can see that some feature contributes uniformly across the range of horizons. It is visible for weather and calendar features even if some cycles appear due to the seasonality. For example, heating degree days are useful in winter but not in summer when their value is zero by design. The results confirm our intuition that socioeconomic variables are more helpful to predict longer horizons than short-horizons. For instance, population, some industrial production indices, consumer price index without energy and electric vehicles covariates have an increasing trend with the lead time for all the models. As expected, this increasing trend is much more accentuated for the linear model. Yet, we can observe the same behavior for TabICL and XGBoost. It is worth mentioning that despite being important, economic predictors have a much lower feature importance than weather or calendar predictors. B\_CVS-CJO, C\_CVS-CJO, L\_CVS-CJO, N\_CVS-CJO, employment or population  have a feature importance range in the hundreds while the same models value the temperature in the thousands. Still, they can be more influential than precipitation (tp), wind speeds (u10) or calendar indicators such as season or number of weekends. H\_CVS-CJO, J\_CVS-CJO and tourism\_nights even range in the thousands. It shows that socioeconomic predictors contribute significantly to the model output when compared to other non-standard calendar or weather predictors. In another experiment we explored second order shapley values, which represents interaction terms between variables, to check if some cross phenomena arise between economic and calendar or weather feature. It did not provide any valuable insights as the second order feature importances are marginals.

\begin{figure}[H]
	\centering
	\begin{adjustbox}{max width=\textwidth}
		\input{Plots/feature_importance_part1.tex}
	\end{adjustbox}
	\caption*{}
\end{figure}

\begin{figure}[H]
	\centering
	\begin{adjustbox}{max width=\textwidth}
		\input{Plots/feature_importance_part2.tex}
	\end{adjustbox}
	\caption{Feature Importance evolution with the forecast horizon. Each subplot presents a covariate with the different model that have selected it in their ensemble. If a model does not appear onto a graphic it means that it did not select the covariate. The feature importance is computed through the Shapley values of each ensemble expert before averaging them to get the contribution. The shaded are represents the $\SI{95}{\percent}$ bootstrapped confidence interval.}
	\label{fig:feature_importance_month}
\end{figure}
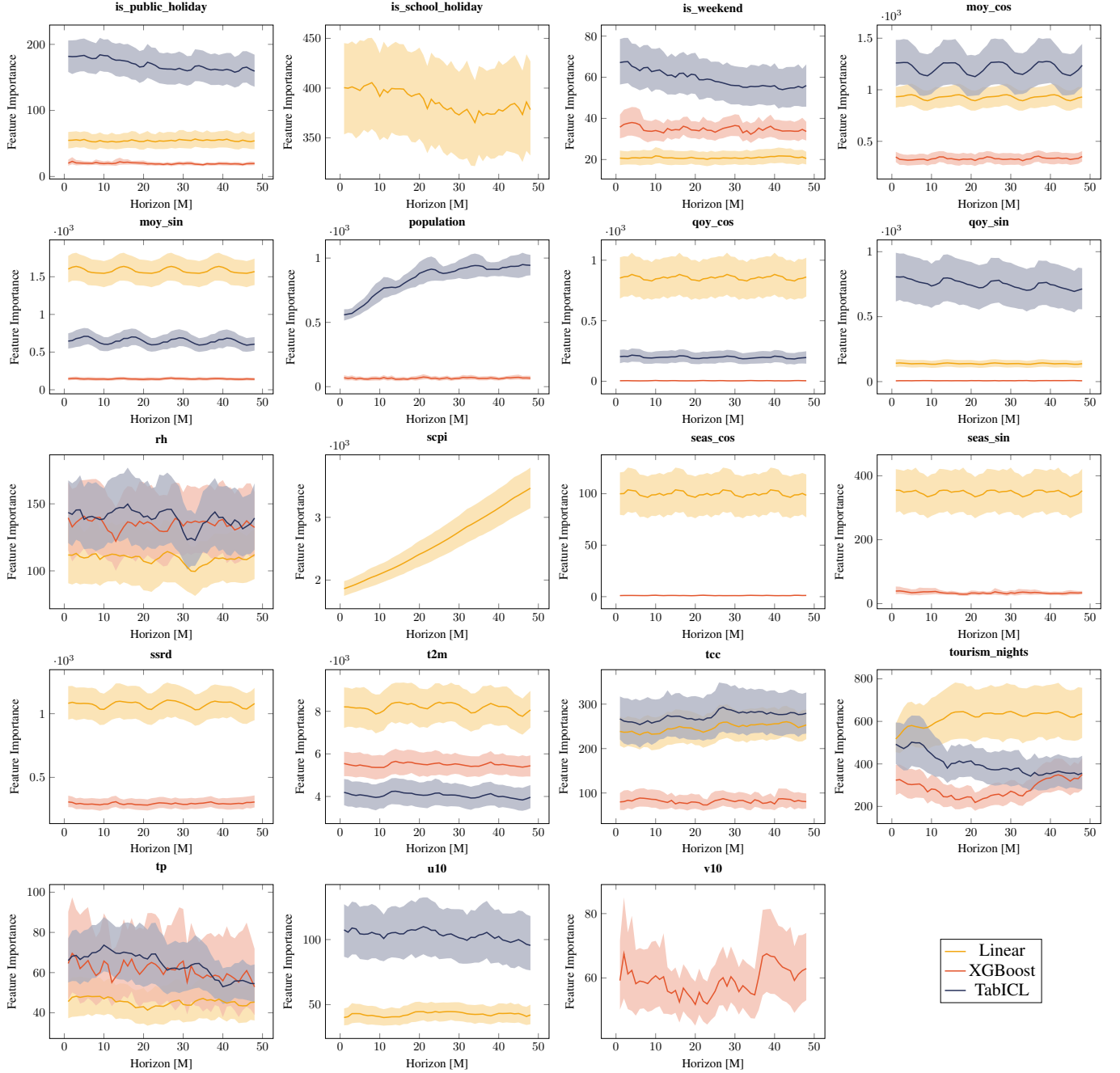

\begin{figure}[h!]
	\centering
	\begin{adjustbox}{max width=\textwidth}
		\input{Plots/context_importance.tex}
	\end{adjustbox}
	\caption{Context Importance for several forecast horizons. Each column shows the results for a different model while each row is a different lead time. Forecast horizons are spaced by $6$ months and increase when we go down a column. The bars are centered around zero. A postive context importance for a block means that this block is useful for the model predictive performance while a negative bar means that it is not. Blocks of context are numbered. The higher the number the more recent is the data.}
	\label{fig:context_importance_month}
\end{figure}
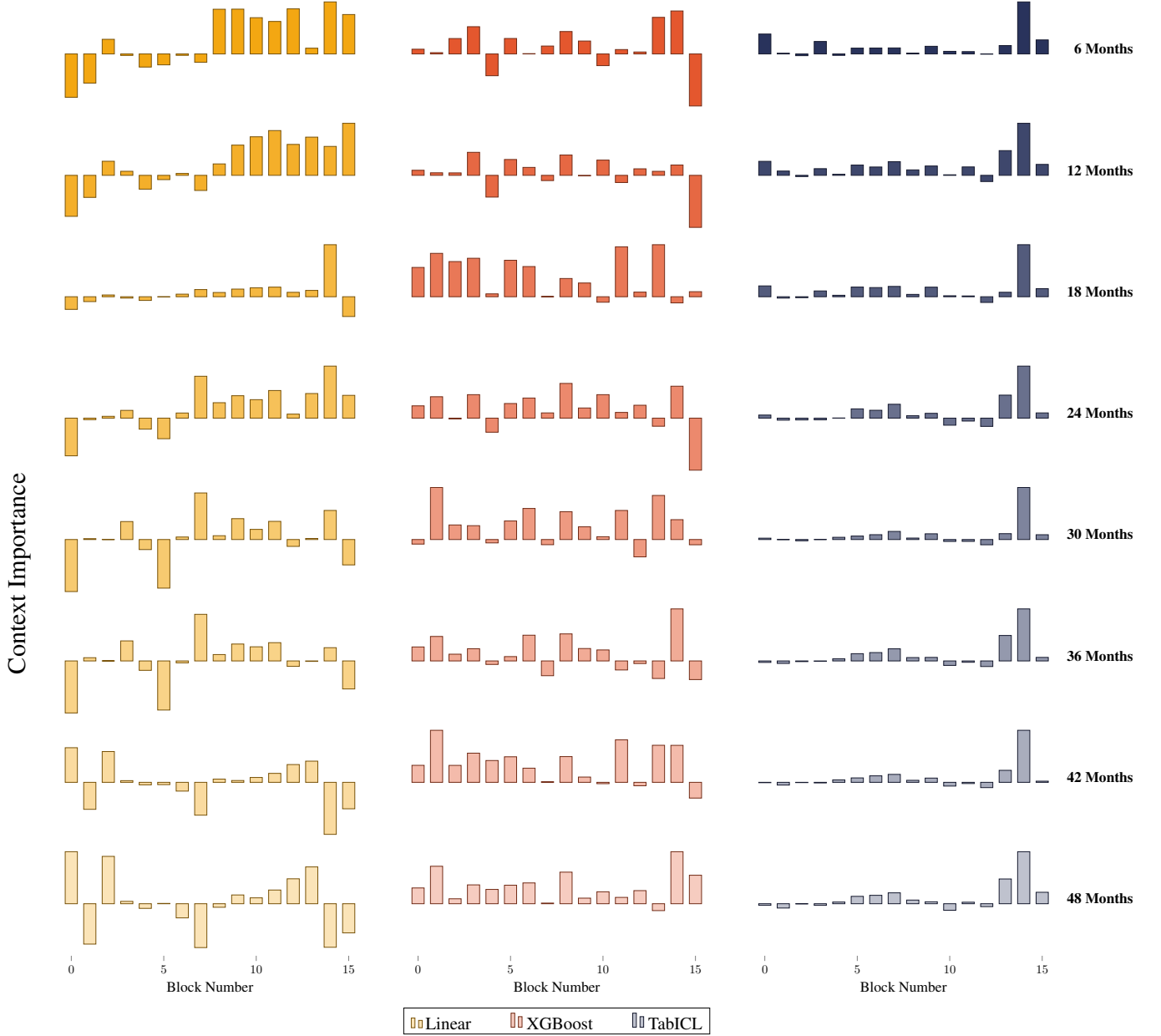

Finally, we investigated the context importance to quantify what was important in the context or training data during the rolling forecast evaluation. We chose to look at different horizon in figure \ref{fig:context_importance_month} spaced by $6$ months every time. We can see that the most recent blocks dominate the context importance for TabICL even when increasing the prediction lead time. It is due to the attention mechanism that \acrshort{fms} relies on. Most of the modern architecture are transformer-based neural networks involving a mechanism called attention to attend to other columns (features) and rows (samples) in the data. It is well known in the literature that attention has a range over which it cannot grasp any effects. It seems that even for medium-term load predictions, this behavior restricts the information that TabICL can extract from the long history given as context. It also means that, we could have used a shorter history as context and still get around the same level of prediction accuracy. To verify this hypothesis we repeated the rolling window evaluation while limiting the available context. The results are given in figure \ref{fig:context_ablation} in appendix \ref{appendixK}.

Other models do not exhibit the same kind of behavior. XGBoost as a much more uniform context importance across the blocks. The long history provided in the training data is entirely leveraged and not only the most recent part of the data. Only a few blocks have a negative context importance troughout the different lead times, showing that the algorithm is able to draw information from the entire history. The linear model offers an in between. Shorter horizons emphasize the recent data blocks though less heavily than TabICL, but the effect smoothes out when increasing the forecasting horizon. Above $24$ months in advance the context importance is distributed more evenly across data blocks. It confirms that forecasting with long horizons requires a long training history.

It is worth mentioning that some data blocks have a small context importance across every model. It can be attributed to the fact that they provide few useful information for prediction or that the information is redundant with other blocks. In both cases, we can conclude that we need less training or context data for \acrshort{mtlf} than what we intially thought. Indeed, if negative context importance means that including the block of data actually hurted the prediction of the model then it might as well be removed. Consequently, the training, or inference time in case of \acrshort{fms}, would be reduced.

\section{Conclusion}\label{sec:conclusion}
In this work, we demonstrate that it is possible to leverage socioeconomic data for \acrlong{mtlf} at both daily and monthly resolution. We curated a dataset covering around $20$ years of load data combined with calendar, weather and economic predictors suitable for electricity demand modeling. To assess the relevance of economic data for \acrshort{mtlf} we designed a feature selection pipeline using an aggregated hold-out framework to create ensemble of experts models with different covariate subsets. The obtained models were then averaged before evaluation with a rolling forecast approach to quantify the prediction accuracy across lead times. Both steps used $10$ years of data with forecasts horizon ranging from $1$ month and up to $48$ months and proved that carefully selected economic covariates improve the forecasting accuracy across all lead times and resolution.

Our selection of three model architecture enabled us to show that models must be complex enough to take advantage of the exotic predictors. In addition, we confirmed the superiority of \acrlong{fm} for load prediction in both settings: with and without the economic data. TabICL with economic indicators did not exceed $\SI{2,000}{\mega\watt}$ and $\SI{3,000}{\mega\watt}$ in \acrshort{rmse} or $\SI{4}{\percent}$ and $\SI{5}{\percent}$ in \acrshort{mape} for monthly and daily resolution respectively. Monthly forecasts issued with horizon comprised between $1$ and $3$ years had a prediction error limited between $\SI{1,200}{\mega\watt}$ or $\SI{2}{\percent}$ and $\SI{1,800}{\mega\watt}$ or $\SI{3,5}{\percent}$ while daily was confined between $\SI{2,000}{\mega\watt}$-$\SI{2,5}{\percent}$ and $\SI{2,800}{\mega\watt}$-$\SI{4}{\percent}$. This corresponds to an improvement of around $\SI{20}{\percent}$ over the model without economic predictors across all lead times, demonstrating the gains that could be obtained. Moreover, the models are also better suited for forecasting during recession periods as we analyzed for two crisis : COVID and the sobriety period induced by the war in Ukraine.

Thorough attention was paid to the explainability of the models with a look on the selected predictors by the feature selection procedure, feature importance and context importance. Altough, no consensus was found on which economic predictors needed to be included, we attributed the better performance not to a few covariates but to the diversity of the feature subsets used by the ensemble of experts. Feature importance confirmed that economic predictors have a significant contribution to the model which increases with the forecast horizon, validating their utility for longer-term predictions. Last, we showed through context importance that we can leverage a long training history as long as it covers diverse events representative of the forecasting period, except for \acrshort{fms} which suffers from an attention bottleneck limiting the value of older data.

These results imply that including socioeconomic data is useful for improving \acrshort{mtlf} while also bridging the gap with \acrshort{ltlf} were economic indicators and projection scenarios are commonly used for accurate predictions. It also suggests that such models could be used with climate projections data to obtain load projections up to 2100. Using the Shared Socioeconomic Pathways narratives to have socioeconomic scenarios in addition to climate projections data could lead to finer and unified projections of electricity demand.

\noindent 

\paragraph{Competing Interests} The authors declare that they have no competing interests.

\paragraph{Funding} This research was supported by a grant from the Association Nationale de la Recherche et de la Technologie (ANRT) No. 2024/0010.

\clearpage

\begin{appendix}
\section{Appendix A: List of Abbreviations}\label{appendixA}
\printnoidxglossary[type=\acronymtype]

\clearpage
\section{Appendix B : Description of covariates considered}\label{appendixB}
The table below presents all the covarietes considered in this study with their description, source, unit, granularities, and the aggregation rules used.
The starting covariate sets used for each model are : 

\vspace{0.5cm}
\hspace{2cm}
\begin{minipage}[t]{0.5\textwidth}
\textbf{Daily Model}
\begin{itemize}
	\item t2m
	\item doy\_cos
	\item doy\_sin
	\item dow\_cos
	\item dow\_sin
	\item is\_public\_holidays
\end{itemize}
\end{minipage}%
\begin{minipage}[t]{0.5\textwidth}
\textbf{Monthly Model}
\begin{itemize}
	\item t2m
	\item moy\_cos
	\item moy\_sin
	\item is\_weekend
	\item is\_public\_holidays
\end{itemize}
\end{minipage}

\newgeometry{left=0.5cm,right = 0.5cm, top = 0.5cm, bottom = 0.5cm}
\input{Tables/table_features.tex}
\newgeometry{
    textheight=9in,
    textwidth=18cm,
    top=1in,
    headheight=24pt,
    headsep=25pt,
    footskip=30pt,
    bottom = 1in,
  }

\clearpage

\section{Appendix C : Degree Days Regional Model Parameters}\label{appendixC}
\begin{figure}[h!]
	\centering
	\includegraphics[width=0.7\textwidth]{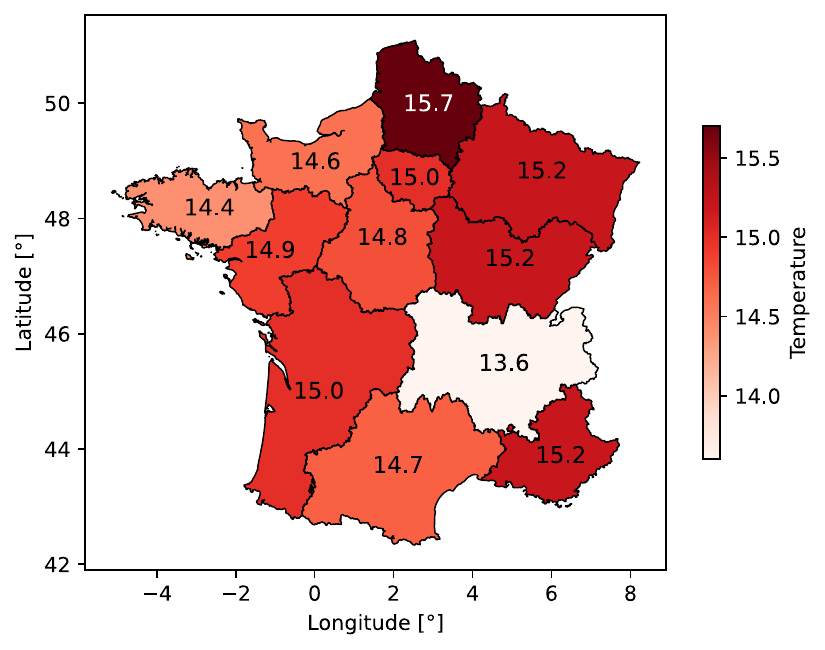}
	\caption{Heating threshold temperature $T_{h}$ estimated for each of the $12$ region of continental France.}
	\label{fig:heating_temp_map}
\end{figure}
\vspace{-0.4cm}
\begin{figure}[h!]
	\centering
	\includegraphics[width=0.7\textwidth]{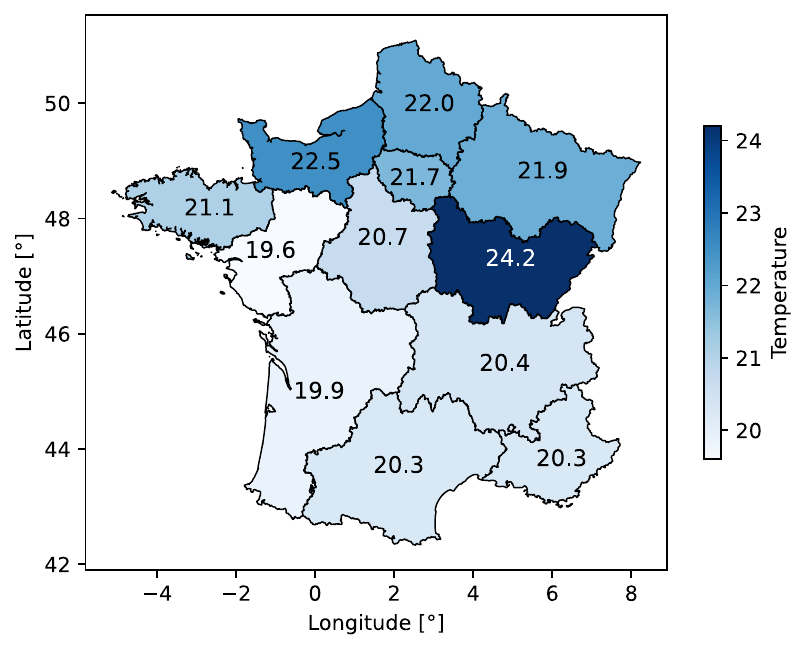}
	\caption{Cooling threshold temperature $T_{c}$ estimated for each of the $12$ region of continental France.}
	\label{fig:cooling_temp_map}
\end{figure}
\clearpage

\section{Appendix D : Metric Definition}\label{appendixD}
In this work, the following metrics were used for feature selection and/or evaluation of the models. Regarding the notations,
$N$ refers to the number of samples, $y$ the true value and $\hat{y}$ the predicted value. 
\begin{align}
		MSE &= \dfrac{1}{N}\sum_{i=1}^{N}\left(y_{i} - \hat{y}_{i}\right)^{2} \\
		RMSE &= \sqrt{MSE} =\sqrt{\dfrac{1}{N}\sum_{i=1}^{N}\left(y_{i} - \hat{y}_{i}\right)^{2}}\\
		MAE  &= \sum_{i=1}^{N}|y_{i} - \hat{y}_{i}| \\
		MAPE &= \sum_{i=1}^{N}|\dfrac{y_{i} - \hat{y}_{i}}{y_{i}}|
\end{align}
We also defined the bias and variance as the \acrshort{mse} decomposition :
\begin{align}
	MSE &= Bias^2 + Variance + \sigma^2 \\
	Bias &= \dfrac{1}{N}\sum_{i=1}^{N} \left(y_{i} - \hat{y}_{i}\right)\\
	Variance &= \dfrac{1}{N}\sum_{i=1}^{N}\left(y_{i} - \bar{y}\right)^{2}
\end{align}
with $\bar{y}$ the average of the true value over the $N$ samples and $\sigma$ the irreducible modelisation error of the residuals $\varepsilon$. Indeed, the regression problem is defined as $y = f(x) + \varepsilon$ with $x$ the inputs, $f$ the learned transfer function and $\varepsilon \sim \mathcal{N}(0, \sigma)$ the modelisation error.
\clearpage

\section{Appendix E: Performance evolution with forecast horizon for other metrics}\label{appendixE}
\begin{figure}[h!]
	\centering
	\begin{adjustbox}{max width=\textwidth}
		\input{Plots/MAPE_horizon.tex}
	\end{adjustbox}
	\caption{\acrshort{mape} evolution with forecast horizon. Each column presents results for a different model. Top row shows performance for the monthly resolution while bottom row depicts daily resolution. For both resolution the forecast horizon is given in months from $1$ to $48$, \textit{i.e.}, $4$ years. The shaded area represents the $\SI{95}{\percent}$ bootstrapped confidence interval. The $y$ axis has been clipped for the linear models for better visibility.}
	\label{fig:mape_horizon}
\end{figure}

\begin{figure}[h!]
	\centering
	\begin{adjustbox}{max width=\textwidth}
		\input{Plots/BIAS_horizon.tex}
	\end{adjustbox}
	\caption{Bias evolution with forecast horizon. Each column presents results for a different model. Top row shows performance for the monthly resolution while bottom row depicts daily resolution. For both resolution the forecast horizon is given in months from $1$ to $48$, \textit{i.e.}, $4$ years. The shaded area represents the $\SI{95}{\percent}$ bootstrapped confidence interval. The $y$ axis has been clipped for the linear models for better visibility.}
	\label{fig:bias_horizon}
\end{figure}

\begin{figure}[h!]
	\centering
	\begin{adjustbox}{max width=\textwidth}
		\input{Plots/VARIANCE_horizon.tex}
	\end{adjustbox}
	\caption{Variance evolution with forecast horizon. Each column presents results for a different model. Top row shows performance for the monthly resolution while bottom row depicts daily resolution. For both resolution the forecast horizon is given in months from $1$ to $48$, \textit{i.e.}, $4$ years. The shaded area represents the $\SI{95}{\percent}$ bootstrapped confidence interval. The $y$ axis has been clipped for the linear models for better visibility.}
	\label{fig:variance_horizon}
\end{figure}
\clearpage

\section{Appendix F: Skill Scores evolution with forecast horizon}\label{appendixF}
For easier comparison we computed the skill score $SS$ which is defined as :
\begin{equation}
	SS = \dfrac{S - S_{reference}}{S_{perfect} - S_{reference}}
\end{equation}
with $S$ a metric or score like \acrshort{rmse} or \acrshort{mape} computed for the given predictions, $S_{reference}$ the same metric computed for a reference forecast and $S_{perfect}$ the metric in case of a perfect forecast. For \acrshort{rmse} and \acrshort{mape} the perfect forecast implies zero error leading skill score to be written :
\begin{equation}
	SS = 1 - \dfrac{S}{S_{reference}}.
\end{equation}
We also introduce the following notations:
\begin{align}
	\begin{split}
		RMSESS = 1 - \dfrac{RMSE}{RMSE_{reference}}
	\end{split}
	\begin{split}
		MAPESS = 1 - \dfrac{MAPE}{MAPE_{reference}}.
	\end{split}
\end{align}
It is a way to quantify enhancement or degradation of the forecast when comparing to a known reference. If it is positive then the forecast improves the reference but if negative it is worse. In the following plots (figure \ref{fig:rmse_skill_score_horizon} and \ref{fig:mape_skill_score_horizon}) the reference is taken to be the models without socioeconomic predictors.
\begin{figure}[h!]
	\centering
	\begin{adjustbox}{max width=\textwidth}
		\input{Plots/RMSE_skill_score.tex}
	\end{adjustbox}
	\caption{$RMSESS$ Skill Score evolution with forecast horizon. Each column presents results for a different resolution. For both resolution the forecast horizon is given in months from 1 to 48, \textit{i.e.}, 4 years. The shaded area represents the $\SI{95}{\percent}$ bootstrapped confidence interval. The $y$ axis has been clipped for better visibility due to linear model overshoot.}
	\label{fig:rmse_skill_score_horizon}
\end{figure}

\begin{figure}[h!]
	\centering
	\begin{adjustbox}{max width=\textwidth}
		\input{Plots/MAPE_skill_score.tex}
	\end{adjustbox}
	\caption{$MAPESS$ Skill Score evolution with forecast horizon. Each column presents results for a different resolution. For both resolution the forecast horizon is given in months from 1 to 48, \textit{i.e.}, 4 years. The shaded area represents the $\SI{95}{\percent}$ bootstrapped confidence interval. The $y$ axis has been clipped for better visibility due to linear model overshoot.}
	\label{fig:mape_skill_score_horizon}
\end{figure}
\clearpage

\section{Appendix G: Inter Model Performance Comparison}
\begin{figure}[h!]
	\centering
	\begin{adjustbox}{max width=\textwidth}
		\input{Plots/RMSE_horizon_comparison.tex}
	\end{adjustbox}
	\caption{$RMSE$ evolution with forecast horizon. Left column presents results when only calendar and weather covariates are included and right column the results when socioeconomic data are added. The rows displays the results for both monthly and daily resolution. Both resolution show forecast horizon in months from 1 to 48, \textit{i.e.}, 4 years. The shaded area represents the $\SI{95}{\percent}$ bootstrapped confidence interval. The $y$ axis has been clipped for better visibility due to linear model overshoot.}
	\label{fig:rmse_model_comparison}
\end{figure}

\begin{figure}[h!]
	\centering
	\begin{adjustbox}{max width=\textwidth}
		\input{Plots/MAPE_horizon_comparison.tex}
	\end{adjustbox}
	\caption{$MAPE$ evolution with forecast horizon. Left column presents results when only calendar and weather covariates are included and right column the results when socioeconomic data are added. The rows displays the results for both monthly and daily resolution. Both resolution show forecast horizon in months from 1 to 48, \textit{i.e.}, 4 years. The shaded area represents the $\SI{95}{\percent}$ bootstrapped confidence interval. The $y$ axis has been clipped for better visibility due to linear model overshoot.}
	\label{fig:mape_model_comparison}
\end{figure}
\clearpage

\section{Appendix H: Performance evolution with forecast horizon during economic crisis}\label{appendixH}

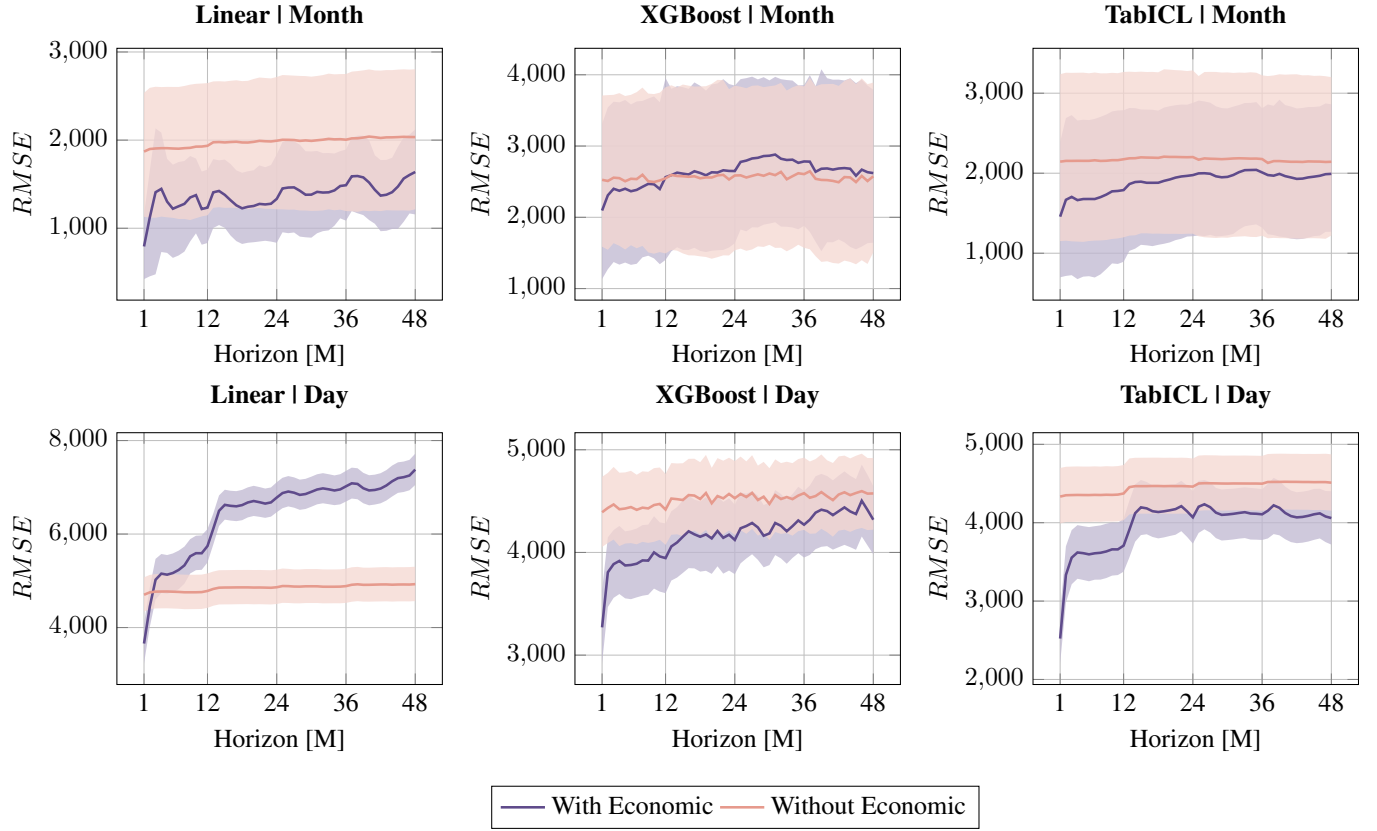
\begin{figure}[h!]
	\centering
	\begin{adjustbox}{max width=\textwidth}
		\input{Plots/RMSE_horizon_covid}
	\end{adjustbox}
	\caption{$RMSE$ evolution with forecast horizon during COVID lockdowns. Each column presents results for a different model. Top row shows performance for the monthly resolution while bottom row depicts daily resolution. For both resolution the forecast horizon is given in months from 1 to 48, \textit{i.e.}, 4 years. The shaded area represents the $\SI{95}{\percent}$ bootstrapped confidence interval.}
	\label{fig:rmse_horizon_covid}
\end{figure}

\begin{figure}[h!]
	\centering
	\begin{adjustbox}{max width=\textwidth}
		\input{Plots/RMSE_horizon_sobriety}
	\end{adjustbox}
	\caption{$RMSE$ evolution with forecast horizon during the sobriety period induced by the Ukraine war. Each column presents results for a different model. Top row shows performance for the monthly resolution while bottom row depicts daily resolution. For both resolution the forecast horizon is given in months from 1 to 48, \textit{i.e.}, 4 years. The shaded area represents the $\SI{95}{\percent}$ bootstrapped confidence interval.}
	\label{fig:rmse_horizon_sobriety}
\end{figure}
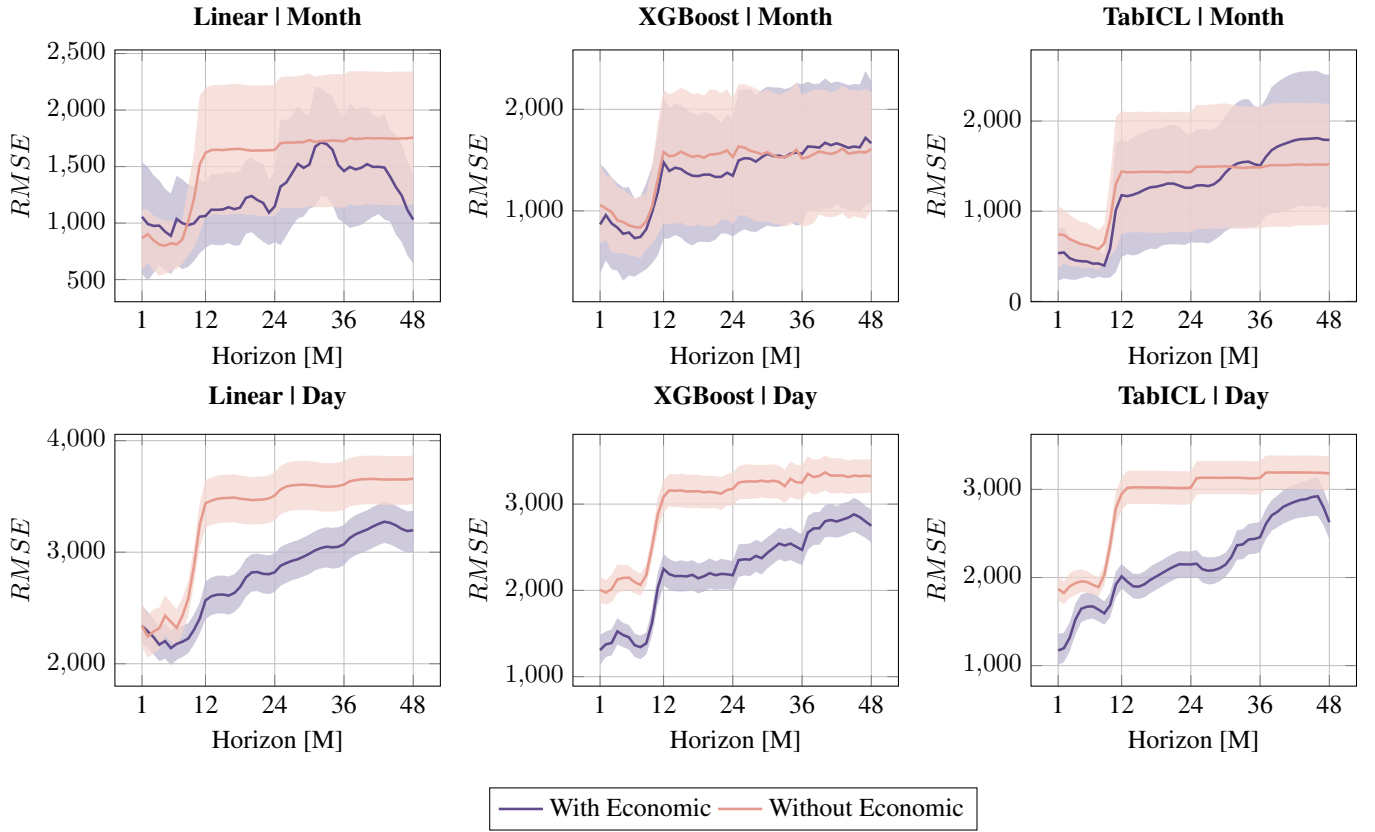

\clearpage
\section{Appendix I: Selected Features for the Daily Model}\label{appendixI}
\begin{figure}[h!]
	\centering
	\begin{adjustbox}{max width=\textwidth}
		\input{Plots/selected_features_day.tex}
	\end{adjustbox}
	\caption{Selection ratio for each feature appearing in one the feature subsets obtained after the feature selection process for the daily models. Covariates are grouped by their category for easier visualisation. For the monthly resolution, the procedure started a reduced subset made from : t2m, doy\_cos, doy\_sin, dow\_cos, dow\_sin, and is\_public\_holiday.}
	\label{fig:selected_features_day}
\end{figure}
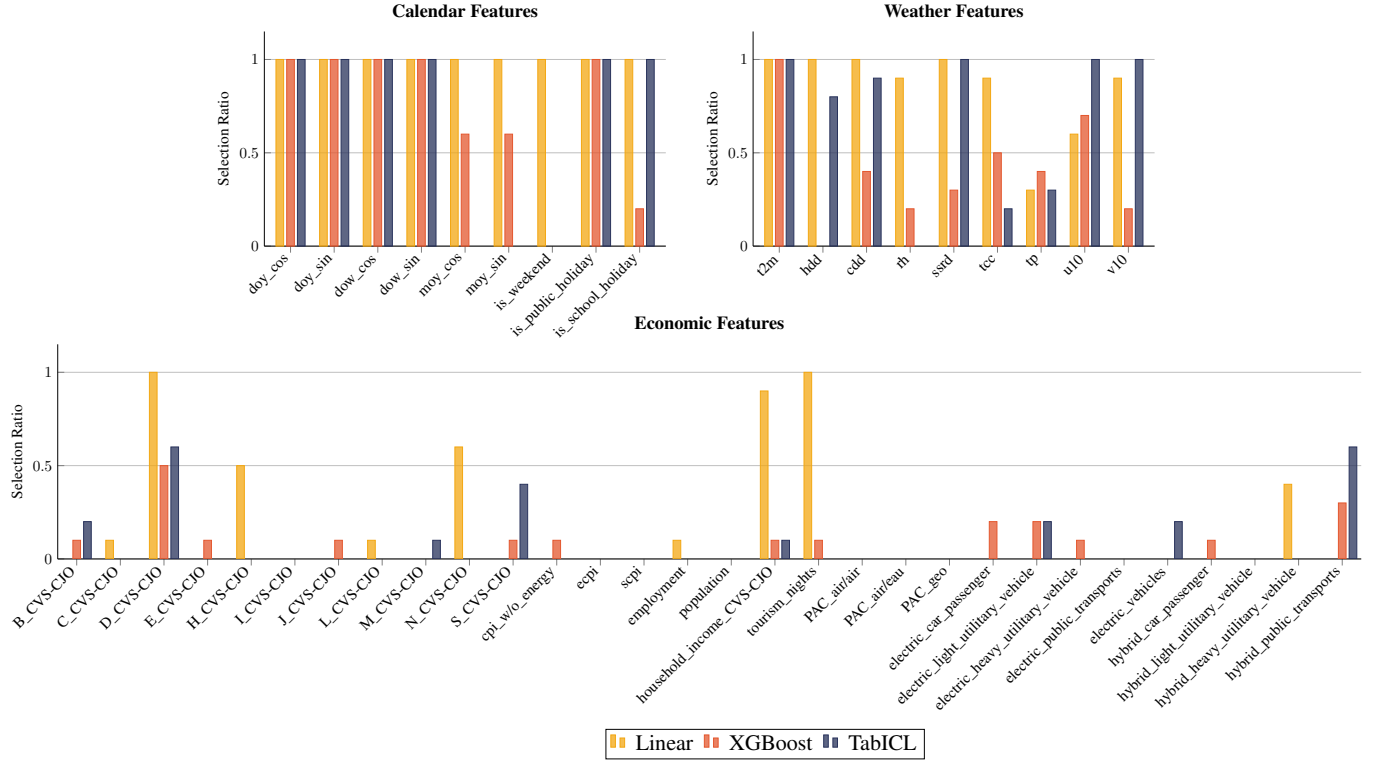
\clearpage

\section{Appendix J: Feature Selection Utility}\label{appendixJ}
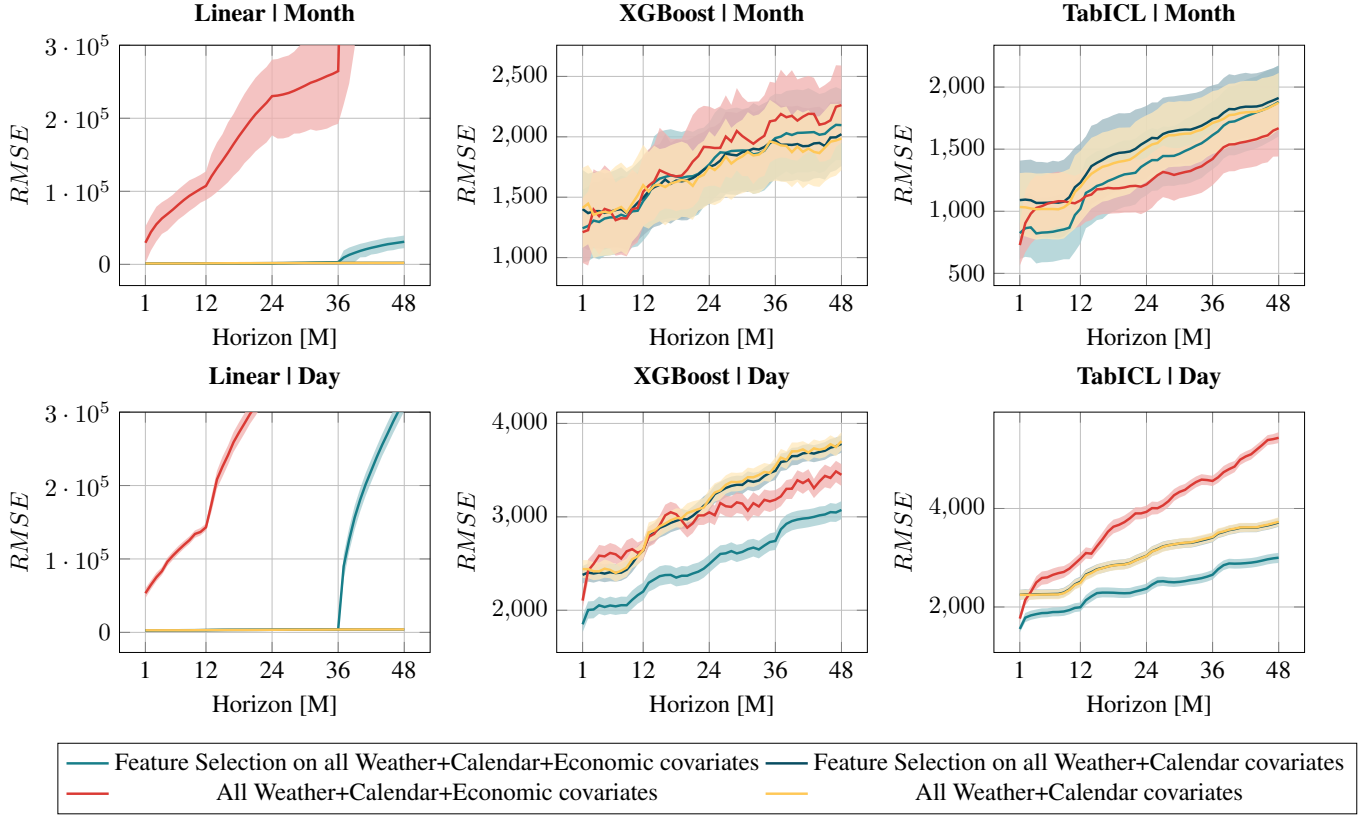
\begin{figure}[h!]
	\centering
	\begin{adjustbox}{max width=\textwidth}
		\input{Plots/RMSE_feature_selection.tex}
	\end{adjustbox}
	\caption{$RMSE$ evolution with forecast horizon with or without the feature selection procedure. We consider different set of features on which we applied or not the feature selection method.
	Each column presents results for a different model. Top row shows performance for the monthly resolution while bottom row depicts daily resolution.
	For both resolution the forecast horizon is given in months from 1 to 48, i.e., 4 years. The shaded area represents the $\SI{95}{\percent}$	bootstrapped confidence interval. The $y$ axis has been clipped for the linear models for better visibility.}
	\label{fig:feature_selection_comparison}
\end{figure}

The feature selection procedure that we implemented can be debated, especially for \acrshort{fm} which are claimed to handle multiple covariates easily. Hence, we compared the models for which we applied the feature selection with the same models using all available covariates. We repeated the experiments when including and not including economic predictors. Performance results are illustrated by the figure \ref{fig:feature_selection_comparison}.

At first glance we can notice that feature selection is improving the performance of the model at all horizons. On second thoughts we can see that it is true for the daily resolution when considering economic data, where both TabICL and XGBoost benefits highly from this step. Indeed, daily TabICL with feature selection predictions with economic covariates have a \acrshort{rmse} of $\SI{1,800}{\mega\watt}$ to $\SI{3,000}{\mega\watt}$ troughout the horizon range. Its counterpart without any feature selection ranges between $\SI{1,900}{\mega\watt}$ and $\SI{5,000}{\mega\watt}$, nearly twice as much error. XGBoost follows the same pattern with an error increase of $\SI{800}{\mega\watt}$ when discarding the feature selection step.
At monthly resolution the gain is less clear for XGBoost with both curves being close on the entire covariate set. TabICL without feature selection is also better by a few hundred $\SI{}{\mega\watt}$. This again show that enough data samples are required to make the feature selection step worthwhile. When focusing only on weather and calendar covariates there is no advantage in applying the feature selection step as for both TabICL and XGBoost at daily resolution the curves superpose. At monthly resolution though it seems that there is little to no differences, making the higher computational burden of the feature selection not worth it.
\clearpage

\section{Appendix K: Context Ablation}\label{appendixK}
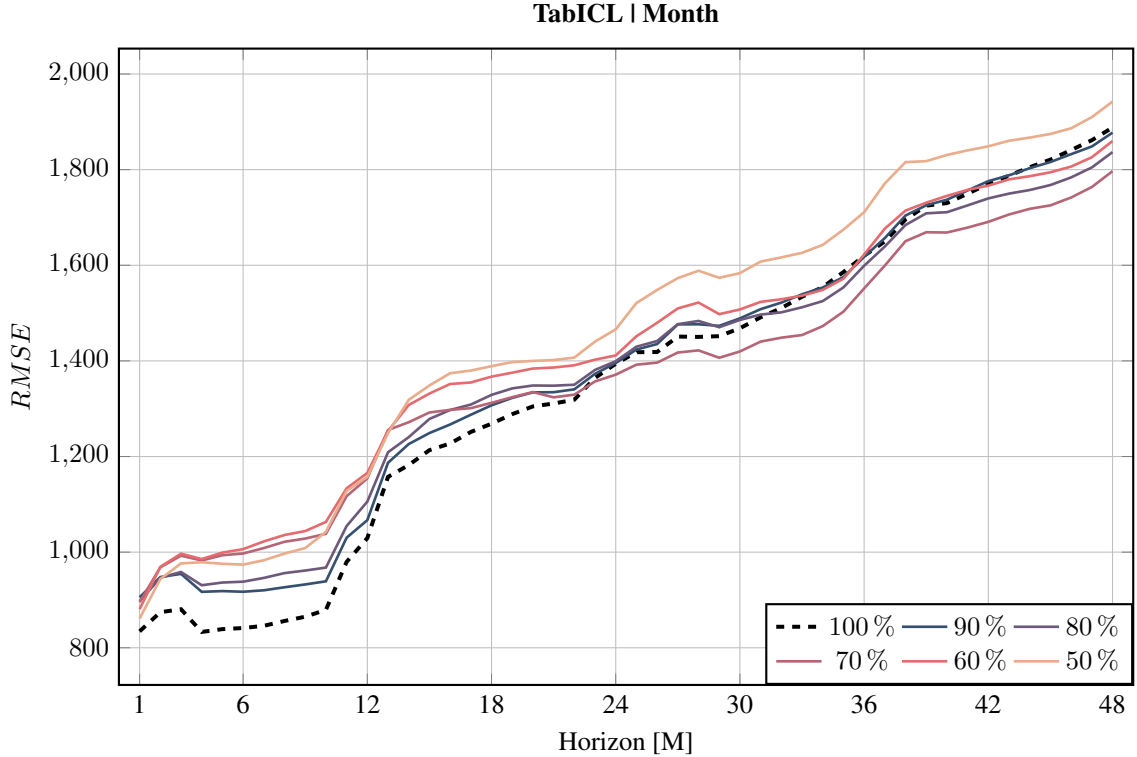
\begin{figure}[h!]
	\centering
	\begin{adjustbox}{max width=\textwidth}
		\input{Plots/RMSE_context_ablation.tex}
	\end{adjustbox}
	\caption{$RMSE$ evolution with forecast horizon for different context size for our monthly TabICL model with economic covariates. Context size is defined as a percentage of the initial context where only the most recent data is kept. The performance of the model with the entire context are represented by the black dashed line.}
	\label{fig:context_ablation}
\end{figure}

Our context importance experiment results of figure \ref{fig:context_importance_month} suggested that TabICL does not need the entire context as it can only attend to the most recent part of the context due to the attention range bottleneck. To verify this idea we repeated the rolling forecast evaluation while limiting the context available to the model. We removed the oldest part of the data and kept only a certain percentage of the original context. The results are given in figure \ref{fig:context_ablation}.

First, we can notice that reducing the context is not harming the performance of the model for every horizon. Shorter lead times are more affected than longer horizons. Indeed, model with only $\SI{70}{\percent}$ of the context has a lower \acrshort{rmse} than the original model starting at horizon $22$ months. TabICL with $\SI{80}{\percent}$ is also better from horizon $32$ months. Second, we can see that the degradation of performance is tolerable with around $\SI{10}{\percent}$ increase in error  which stays consistent across forecast horizon for model with $\SI{70}{\percent}$ to $\SI{90}{\percent}$ of the initial context. Reducing even more the context to only $\SI{50}{\percent}$ of its original size leads to a decrease of around $\SI{15}{\percent}$ which depending on the application can or can not be acceptable considering the potential computation savings. 

\clearpage
\section{Appendix L: TabICL Complexity}\label{appendixL}
The complexity of TabICL algorithm is given to be $\mathcal{O}\left(n^{2} + nm^{2}\right)$ where $n$ refers to the number of rows in the context and inference data and $m$ the number of columns. As computation time increase drastically for big datasets we illustrated how reducing the number of features and the context can help save computation time which can be crucial in an operational setup. Of course, this is done as a theoretical illustration as it is well known that compute time is equally dependent on the hardware and software.

Our initial datasets are made with $m=55$ features which we reduced trough feature selection to between $9$ and $16$ features depending on the expert. Indeed, with the aggregated hold-out procedure that we used in the feature selection process, we created an ensemble of expert with different feature subsets. Thus, to illustrate the savings a reduction of $m$ can occur we plotted the complexity for different number of covariates.

The other main driver of complexity is the number of rows $n$. To evaluate our models, we carried a rolling forecast evaluation window meaning that the number of samples grows during evaluation. Hence, we chose to illustrate the complexity on the range of values $n$ took during this step for both monthly and daily resolution. Our experiments on context showed that we can reduce the context without losing too much predictive performance while allowing to accelerate the inference. Thus, a context reduction case was added to the complexity plots.

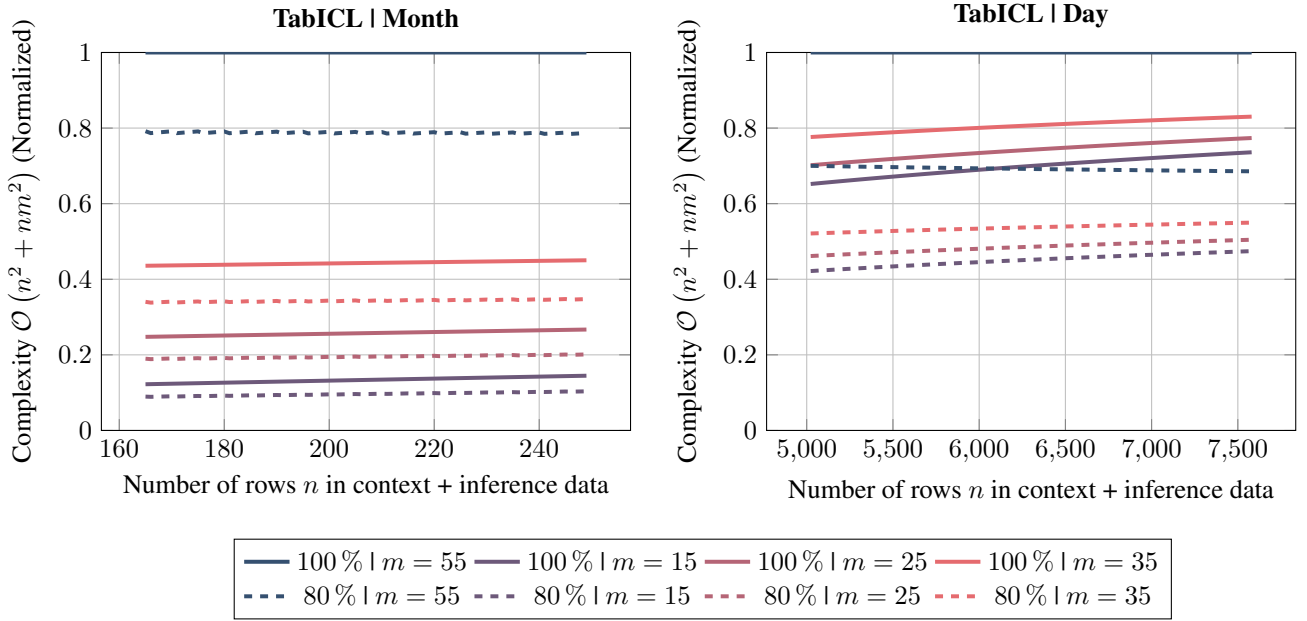
\begin{figure}[h!]
	\centering
	\begin{adjustbox}{max width=\textwidth}
		\input{Plots/complexity.tex}
	\end{adjustbox}
	\caption{Complexity of TabICL evolution with the number of samples $n$ for different context size and number of features $m$ for the monthly resolution on the left panel and daily resolution on the right panel. The curves are normalized by the complexity of the model without any features nor rows reduction.}
	\label{fig:complexity}
\end{figure}

Figure \ref{fig:complexity} presents the complexity curves over the range of $n$ encountered during rolling window evaluation for a sensible set of features $m$. We normalized the curves by the complexity of the model without any feature nor context reduction, \textit{i.e}, a model with $m=55$ features and full context. The idea is to give order of magnitudes on the savings easily achievable even on a small dataset like us.

We can notice that reducing the number of features allow to save between $\SI{55}{\percent}$ and $\SI{85}{\percent}$ of complexity for the monthly resolution which is a small dataset with atmost $250$ rows. The gain is lowered when the dataset size increases, with a saving of $\SI{20}{\percent}$ to $\SI{35}{\percent}$ at the daily resolution. In this case, it is the number of rows which drives the complexity. Limiting the context to only $\SI{80}{\percent}$ of its original length offers huge savings with already a $\SI{30}{\percent}$ improvement. Combined with a column reduction it leads to a total of $\SI{50}{\percent}$ to $\SI{60}{\percent}$ of complexity saved over the original model. At monthly resolution reducing the number of rows improves only for $\SI{20}{\percent}$ and add $\SI{5}{\percent}$ to $\SI{10}{\percent}$ in addition to the feature reduction leading to a total saving of $\SI{65}{\percent}$ to $\SI{90}{\percent}$ depending on the number of columns kept. 

This illustration highlights two points. First the order of magnitudes of computation savings that can be achieved in our case with small datasets and many covariates when combining feature and context reduction. The enhancements range between $\SI{50}{\percent}$ to $\SI{90}{\percent}$ when compared to the raw dataset. Second, it shows that depending on the situation it might be more cost-effective to spend time reducing the number of columns and keep the entire context as in our monthly resolution or to diminish the size of the context and leave all features available. Of course, these conclusions do not take into account the predictive performance and focus only on the complexity and the computation cost. 
\end{appendix}
\clearpage

\newpage

\end{document}

%% file: Scheme/Tex/workflow.tex



\begin{tikzpicture}[node distance = 2cm]

\node[signal_node, fill = apricot_cream, draw = apricot_cream!60!black] (signal1) {\Large \textbf{1.Data Acquisition \& Preprocessing}};  
\node[signal_node, fill = twilight_indigo!50, draw = twilight_indigo!60!black, right = 3.5cm of signal1] (signal2) {\Large \textbf{2.Feature Selection}};  
\node[signal_node, fill = muted_teal, draw = muted_teal!60!black, right = 2cm of signal2] (signal3) {\Large \textbf{3.Rolling Forecast Evaluation}};  
\node[signal_node, fill = peach_fuzz, draw = peach_fuzz!60!black, right = 2cm of signal3] (signal4) {\Large \textbf{4.Output \& Results}};  

\node[trapezium_node, fill = apricot_cream!45, draw = apricot_cream!60!black, below=3cm of signal1, inner xsep=-15pt, xshift=-1.9cm] (trapezium1) {\large \textbf{Retrieving Data}};
\node[round_rectangle, fill = apricot_cream!30, draw = apricot_cream!60!black, below=0.2cm of trapezium1] (round_rectangle2) 
    {\begin{tabular}{c l}
      {\Huge \faPlug}    & Load Data \\
      \\
      {\Huge \Huge \faCloudSunRain}       & Weather Data \\
      \\
      {\Huge \faCalendar*[regular]} & Calendar Data \\
      \\
      {\Huge \faMoneyBill*[regular]} & Economic Data \\
      \\
      {\Huge \faCarSide} & Electric Vehicles Data \\
      \\
      {\Huge \faSnowflake} & Heat Pumps Data \\

    \end{tabular}};
\node[trapezium_node, fill = apricot_cream!45, draw = apricot_cream!60!black, below=5.5cm of signal1, align=center, xshift = 3.8cm] (trapezium2) {\large \textbf{Dataset}\\\textbf{(2005-2025)}};
\node[round_rectangle, fill = apricot_cream!30, draw = apricot_cream!60!black, below=0.2cm of trapezium2] (round_rectangle3)
{\begin{tabular}{l}
      \faCaretRight \hspace{0.1cm} Cleaned\\
      \faCaretRight \hspace{0.1cm} Imputed\\
      \faCaretRight \hspace{0.1cm} Timespan\\
      \faCaretRight \hspace{0.1cm} Temporal Frequencies\\
      \faCaretRight \hspace{0.1cm} Spatial Aggregation \\
\end{tabular}};

\node[
    single arrow, 
    draw=eggshell!80!black, 
    fill=eggshell, 
    minimum height=0.8cm, 
    minimum width=0.1cm,
    align=center,
    right=0.1cm of round_rectangle2,
  ] (single_arrow1){\tiny\faCog};
\node[
    single arrow, 
    draw=eggshell!80!black, 
    fill=eggshell, 
    minimum height=0.8cm, 
    minimum width=0.1cm,
    align=center,
    right=0.1cm of round_rectangle3,
  ] (single_arrow2) {};

\node[trapezium_node, fill=twilight_indigo!50, draw = twilight_indigo!50!black, below=0.5cm of signal2, inner xsep=2pt, xshift=0.2cm, align = center] (trapezium3) {\large \textbf{Temporal Cross-Validation}\\ ($N$ Folds)};
\node[round_rectangle, fill=apricot_cream!60, draw=apricot_cream!60!black, below=1cm of trapezium3, align=center, xshift=-1.3cm] (round_rectangle4) {Training Dataset\\(2005-2015)};
\node[round_rectangle, fill=burnt_peach!80, draw=burnt_peach!50!black, right=0.2cm of round_rectangle4, align=center] (round_rectangle5) {Feature Pool};
\node[round_rectangle, fill=twilight_indigo!50, draw=twilight_indigo!50!black, below=0.75cm of round_rectangle4, xshift = 1.25cm, align=center] (round_rectangle6) {A.Initial Model \\ \& Base Feature Set};
\node[round_rectangle, fill=twilight_indigo!50, draw=twilight_indigo!50!black, below=0.5cm of round_rectangle6, align=center] (round_rectangle7) {B.Add Candidate Feature to\\Feature Set};
\node[diamond_node, fill=twilight_indigo!50, draw=twilight_indigo!50!black, below=0.5cm of round_rectangle7, align=center, inner xsep=0pt, inner ysep=0pt] (diamond1) {Improvement on\\Test Fold $i$ ?};
\node[round_rectangle, fill=twilight_indigo!50, draw=twilight_indigo!50!black, below=0.5cm of diamond1, align=center] (round_rectangle8) {C.Add to Features Set \\(\small Only the best candidate)};
\node[align=center, below=0.05cm of round_rectangle8, inner xsep=-10pt, inner ysep=0pt] (floating_text1) {Repeat until no more improvements\\or no candidate features left};
\node[round_rectangle, fill=twilight_indigo!50, draw=twilight_indigo!50!black, below=1.4cm of floating_text1, xshift = -0.1cm] (round_rectangle9) {$N$ Feature Sets | $N$ Models };
\begin{scope}[on background layer]
        \node (feature_selection_bg) [
            round_rectangle, fill = twilight_indigo!10, draw = twilight_indigo!50!black,
            fit={(round_rectangle4) (round_rectangle5) (diamond1) (round_rectangle6) (round_rectangle7) (round_rectangle8) (floating_text1)},
            inner xsep=26pt,
            inner ysep=28pt
        ] {};
        \node[anchor=north, yshift=-0.25cm, xshift =0cm] at (feature_selection_bg.north) {\textbf{Feature Selection Loop}};
\end{scope}
\begin{scope}[on background layer, yshift=-10cm]
        \node (inner_loop_bg) [
            round_rectangle, fill = eggshell!50, draw = twilight_indigo!50!black,
            fit={(round_rectangle6) (round_rectangle7) (diamond1) (round_rectangle8) (floating_text1)},
            inner xsep=22pt,
            inner ysep=14pt
        ] {};
    \node[anchor=north, ] at (inner_loop_bg.north) {\textbf{Inner Loop Fold $i$}};
\end{scope}
\draw[arrow] (round_rectangle6) -- (round_rectangle7);
\draw[arrow] (round_rectangle7) -- node[left] {Train} node[right] {Predict}(diamond1);
\draw[arrow] (diamond1) -- node[right] {Yes} (round_rectangle8);
\draw[arrow] (feature_selection_bg.south) -- (round_rectangle9);

\draw[arrow] (round_rectangle4.south)  ++(-1,0) |- (round_rectangle6.west);
\draw[arrow] (round_rectangle5.south)  ++(0.8,0) |- (round_rectangle6.east);
\draw[arrow] (diamond1.west) -- ++(-1.25,0) node[below right, xshift=0.5cm] {No} |- (round_rectangle7.west);
\draw[arrow] (round_rectangle8.east) -- ++(0.5,0) node[below right, xshift=0.5cm, yshift=4.25cm, rotate=-90] {Update Base Set} |- (round_rectangle6.east);

\node[
    single arrow, 
    draw=eggshell!80!black, 
    fill=eggshell, 
    minimum height=0.8cm, 
    minimum width=0.1cm,
    align=center,
    right=6.85cm of single_arrow2,
  ] (single_arrow3) {};

\node[trapezium_node, fill=muted_teal, draw = muted_teal!50!black, below=0.5cm of signal3, inner xsep=2pt, xshift=0.2cm, align = center] (trapezium4) {\large \textbf{Rolling Window} \\ (Length $W$)};
\node[round_rectangle, fill=apricot_cream!60, draw=apricot_cream!60!black, below=1cm of trapezium4, align=center, xshift=-2.1cm] (round_rectangle10) {Test Dataset\\(2015-2025)};
\node[round_rectangle, fill=burnt_peach!80, draw=burnt_peach!50!black, right=0.2cm of round_rectangle10, align=center] (round_rectangle11) {Selected Feature Set};
\node[round_rectangle, fill=muted_teal, draw=muted_teal!50!black, below=0.75cm of round_rectangle11, xshift = -1.5cm, align=center] (round_rectangle12) {A.Extract Window};
\node[round_rectangle, fill=muted_teal, draw=muted_teal!50!black, below=0.5cm of round_rectangle12, align=center] (round_rectangle13) {B.Train Model on\\previous Data};
\node[round_rectangle, fill=muted_teal, draw=muted_teal!50!black, below=0.5cm of round_rectangle13, align=center] (round_rectangle14) {C.Predict over the Window\\(Horizon : 1-48 Months)};
\node[diamond_node, fill=muted_teal, draw=muted_teal!50!black, below=1cm of round_rectangle14, align=center, inner xsep=0pt, inner ysep=2pt] (diamond2) {End of\\Test Set\\Reached ?};
\node[round_rectangle, fill=muted_teal, draw=muted_teal!50!black, right=2.45cm of round_rectangle9, align=center] (round_rectangle15) {$N$ Multi-Horizon Predictions for 2015-2025};
\begin{scope}[on background layer]
        \node (rolling_window_bg) [
            round_rectangle, fill = muted_teal!20, draw = muted_teal!50!black,
            fit={(round_rectangle10) (round_rectangle11) (round_rectangle12) (round_rectangle13) (round_rectangle14) (diamond2)},
            inner xsep=18pt,
            inner ysep=28pt
        ] {};
        \node[anchor=north, yshift=-0.25cm, xshift=0cm] at (rolling_window_bg.north) {\textbf{Rolling Window Predictions}};
\end{scope}
\begin{scope}[on background layer, yshift=-10cm]
        \node (inner_window_bg) [
            round_rectangle, fill = eggshell!50, draw = muted_teal!50!black,
            fit={(round_rectangle12) (round_rectangle13) (round_rectangle14)},
            inner xsep=22pt,
            inner ysep=14pt
        ] {};
    \node[anchor=north, ] at (inner_window_bg.north) {\textbf{Rolling Window $i$}};
\end{scope}

\draw[arrow] (round_rectangle12) -- (round_rectangle13);
\draw[arrow] (round_rectangle13) -- (round_rectangle14);
\draw[arrow] (inner_window_bg.south) -- (diamond2);
\draw[arrow] (rolling_window_bg.south) -- (round_rectangle15);

\draw[arrow] (round_rectangle10.south)  ++(-0.75,0) |- (round_rectangle12.west);
\draw[arrow] (round_rectangle11.south)  ++(1,0) |- (round_rectangle12.east);
\draw[arrow] (diamond2.east) -- ++(2,0) node[below right, xshift=-1.25cm] {No} |- (inner_window_bg.east);

\node[
    single arrow, 
    draw=eggshell!80!black, 
    fill=eggshell, 
    minimum height=0.8cm, 
    minimum width=0.1cm,
    align=center,
    right=7cm of single_arrow3,
  ] (single_arrow4) {};

\node[trapezium_node, fill=peach_fuzz, draw = peach_fuzz!50!black, below=0.5cm of signal4, inner xsep=2pt, xshift=0cm, align = center] (trapezium5) {\large \textbf{Performance}};
\begin{axis}[
    name=fakeErrorPlot,
    xlabel=Horizon,
    ylabel=Error,
    width=8cm,
    height=5cm,
    at={(trapezium5.south)},
    xshift=-3.4cm,
    yshift=-4cm,
    ticks=none,
    xticklabels={},
    yticklabels={},
]
    \addplot[domain=1:10, samples=50, color=twilight_indigo!50, mark=none, thick] {5 * x^1.3};
    \addplot[domain=1:10, samples=50, color=burnt_peach, mark=none, thick] {30 * ln(x) + 3};
\end{axis}

\node[trapezium_node, fill=peach_fuzz, draw = peach_fuzz!50!black, below=5cm of trapezium5, inner xsep=2pt, xshift=0cm, align = center] (trapezium6) {\large \textbf{Interpretability}};
\begin{axis}[
    name=fakeFeatureImpPlot,
    xlabel=Horizon,
    ylabel=Feature Importance,
    width=8cm,
    height=5cm,
    at={(trapezium6.south)},
    xshift=-3.4cm,
    yshift=-4cm,
    ticks=none,
    xticklabels={},
    yticklabels={},
]
    \addplot[domain=1:10, samples=50, color=twilight_indigo!50, mark=none, thick] {3*cos(x*2*pi/0.25)+2};
    \addplot[domain=1:10, samples=50, color=burnt_peach, mark=none, thick] {exp(x/4)-4};
\end{axis}

\begin{axis}[
    ybar,
    bar width=1pt,
    name=fakeContextImpPlot,
    xlabel=Block Period,
    ylabel=Context Importance,
    width=8cm,
    height=5cm,
    at={(trapezium6.south)},
    xshift=-3.4cm,
    yshift=-8cm,
    ticks=none,
    xticklabels={},
    yticklabels={},
]
    \addplot[domain=1:10, samples=50, color=twilight_indigo!50, mark=none, thick] {3*cos(x*2*pi/0.05)+2};
    \addplot[domain=1:10, samples=50, color=burnt_peach, mark=none, thick] {exp(x/4)-4};
\end{axis}
\end{tikzpicture}

%% file: Plots/RMSE_horizon.tex




\pgfplotstableread[col sep=comma]{\getErrorDataFilePath{\resolutionMonth}{\linear}{\errorFolder}{\Econ}{\rmse}}\dataRMSELinearMonthEcon
\pgfplotstableread[col sep=comma]{\getErrorDataFilePath{\resolutionMonth}{\linear}{\errorFolder}{\noEcon}{\rmse}}\dataRMSELinearMonthNoEcon

\pgfplotstableread[col sep=comma]{\getErrorDataFilePath{\resolutionDay}{\linear}{\errorFolder}{\Econ}{\rmse}}\dataRMSELinearDayEcon
\pgfplotstableread[col sep=comma]{\getErrorDataFilePath{\resolutionDay}{\linear}{\errorFolder}{\noEcon}{\rmse}}\dataRMSELinearDayNoEcon

\pgfplotstableread[col sep=comma]{\getErrorDataFilePath{\resolutionMonth}{\xgboost}{\errorFolder}{\Econ}{\rmse}}\dataRMSEXGBoostMonthEcon
\pgfplotstableread[col sep=comma]{\getErrorDataFilePath{\resolutionMonth}{\xgboost}{\errorFolder}{\noEcon}{\rmse}}\dataRMSEXGBoostMonthNoEcon

\pgfplotstableread[col sep=comma]{\getErrorDataFilePath{\resolutionDay}{\xgboost}{\errorFolder}{\Econ}{\rmse}}\dataRMSEXGBoostDayEcon
\pgfplotstableread[col sep=comma]{\getErrorDataFilePath{\resolutionDay}{\xgboost}{\errorFolder}{\noEcon}{\rmse}}\dataRMSEXGBoostDayNoEcon

\pgfplotstableread[col sep=comma]{\getErrorDataFilePath{\resolutionMonth}{\tabicl}{\errorFolder}{\Econ}{\rmse}}\dataRMSETabICLMonthEcon
\pgfplotstableread[col sep=comma]{\getErrorDataFilePath{\resolutionMonth}{\tabicl}{\errorFolder}{\noEcon}{\rmse}}\dataRMSETabICLMonthNoEcon

\pgfplotstableread[col sep=comma]{\getErrorDataFilePath{\resolutionDay}{\tabicl}{\errorFolder}{\Econ}{\rmse}}\dataRMSETabICLDayEcon
\pgfplotstableread[col sep=comma]{\getErrorDataFilePath{\resolutionDay}{\tabicl}{\errorFolder}{\noEcon}{\rmse}}\dataRMSETabICLDayNoEcon

\begin{tikzpicture}
    \begin{groupplot}[
    group style={
        group size=3 by 2,        
        horizontal sep=1.8cm,
        vertical sep=1.8cm,
    },
    width=6cm,
    height=5cm,
    grid=major,
    scaled ticks=false,
    xtick={1, 12, 24, 36, 48},
    xticklabels={1, 12, 24, 36, 48},
    legend columns = 2,
    legend style={at={(1.15,-2.1)}, anchor=south west},
]
 
\nextgroupplot[ymin=800, ymax=2550, title={\textbf{Linear | Month}}, xlabel={Horizon [M]}, ylabel={$RMSE$}]
\addplot[mark=none, line width=1pt, color=light_indigo_purple]
        table [x=horizon, y=mean, col sep=comma] {\dataRMSELinearMonthEcon};
\addlegendentry{With Economic};
\addplot[mark=none, line width=1pt, color=light_light_apricot]
        table [x=horizon, y=mean, col sep=comma] {\dataRMSELinearMonthNoEcon};
\addlegendentry{Without Economic};
\addplot[name path=lower_econ_linear_month, fill=none, draw=none, mark=none, forget plot]
    table[x=horizon, y=low_0.95, col sep=comma, header=true]{\dataRMSELinearMonthEcon};
\addplot[name path=upper_econ_linear_month, fill=none, draw=none, mark=none, forget plot]
    table[x=horizon, y=high_0.95, col sep=comma, header=true]{\dataRMSELinearMonthEcon};
\addplot[light_indigo_purple!40, fill opacity=0.75, forget plot] fill between[of=lower_econ_linear_month and upper_econ_linear_month];

\addplot[name path=lower_no_econ_linear_month, fill=none, draw=none, mark=none, forget plot]
    table[x=horizon, y=low_0.95, col sep=comma, header=true]{\dataRMSELinearMonthNoEcon};
\addplot[name path=upper_no_econ_linear_month, fill=none, draw=none, mark=none, forget plot]
    table[x=horizon, y=high_0.95, col sep=comma, header=true]{\dataRMSELinearMonthNoEcon};
\addplot[light_light_apricot!40, fill opacity=0.75, forget plot] fill between[of=lower_no_econ_linear_month and upper_no_econ_linear_month];

\nextgroupplot[title={\textbf{XGBoost | Month}}, xlabel={Horizon [M]}, ylabel={$RMSE$}]
\addplot[mark=none, line width=1pt, color=light_indigo_purple]
        table [x=horizon, y=mean, col sep=comma] {\dataRMSEXGBoostMonthEcon};
\addplot[mark=none, line width=1pt, color=light_light_apricot]
        table [x=horizon, y=mean, col sep=comma] {\dataRMSEXGBoostMonthNoEcon};
\addplot[name path=lower_econ_xgboost_month, fill=none, draw=none, mark=none, forget plot]
    table[x=horizon, y=low_0.95, col sep=comma, header=true]{\dataRMSEXGBoostMonthEcon};
\addplot[name path=upper_econ_xgboost_month, fill=none, draw=none, mark=none, forget plot]
    table[x=horizon, y=high_0.95, col sep=comma, header=true]{\dataRMSEXGBoostMonthEcon};
\addplot[light_indigo_purple!40, fill opacity=0.75, forget plot] fill between[of=lower_econ_xgboost_month and upper_econ_xgboost_month];

\addplot[name path=lower_no_econ_xgboost_month, fill=none, draw=none, mark=none, forget plot]
    table[x=horizon, y=low_0.95, col sep=comma, header=true]{\dataRMSEXGBoostMonthNoEcon};
\addplot[name path=upper_no_econ_xgboost_month, fill=none, draw=none, mark=none, forget plot]
    table[x=horizon, y=high_0.95, col sep=comma, header=true]{\dataRMSEXGBoostMonthNoEcon};
\addplot[light_light_apricot!40, fill opacity=0.75, forget plot] fill between[of=lower_no_econ_xgboost_month and upper_no_econ_xgboost_month];

\nextgroupplot[title={\textbf{TabICL | Month}}, xlabel={Horizon [M]}, ylabel={$RMSE$}]
\addplot[mark=none, line width=1pt, color=light_indigo_purple]
        table [x=horizon, y=mean, col sep=comma] {\dataRMSETabICLMonthEcon};
\addplot[mark=none, line width=1pt, color=light_light_apricot]
        table [x=horizon, y=mean, col sep=comma] {\dataRMSETabICLMonthNoEcon};
\addplot[name path=lower_econ_tabICL_month, fill=none, draw=none, mark=none, forget plot]
    table[x=horizon, y=low_0.95, col sep=comma, header=true]{\dataRMSETabICLMonthEcon};
\addplot[name path=upper_econ_tabICL_month, fill=none, draw=none, mark=none, forget plot]
    table[x=horizon, y=high_0.95, col sep=comma, header=true]{\dataRMSETabICLMonthEcon};
\addplot[light_indigo_purple!40, fill opacity=0.75, forget plot] fill between[of=lower_econ_tabICL_month and upper_econ_tabICL_month];

\addplot[name path=lower_no_econ_tabICL_month, fill=none, draw=none, mark=none, forget plot]
    table[x=horizon, y=low_0.95, col sep=comma, header=true]{\dataRMSETabICLMonthNoEcon};
\addplot[name path=upper_no_econ_tabICL_month, fill=none, draw=none, mark=none, forget plot]
    table[x=horizon, y=high_0.95, col sep=comma, header=true]{\dataRMSETabICLMonthNoEcon};
\addplot[light_light_apricot!40, fill opacity=0.75, forget plot] fill between[of=lower_no_econ_tabICL_month and upper_no_econ_tabICL_month];

\nextgroupplot[ymin = 1900, ymax=4800, title={\textbf{Linear | Day}}, xlabel={Horizon [M]}, ylabel={$RMSE$}]
\addplot[mark=none, line width=1pt, color=light_indigo_purple]
        table [x=horizon, y=mean, col sep=comma] {\dataRMSELinearDayEcon};
\addplot[mark=none, line width=1pt, color=light_light_apricot]
        table [x=horizon, y=mean, col sep=comma] {\dataRMSELinearDayNoEcon};
\addplot[name path=lower_econ_linear_day, fill=none, draw=none, mark=none, forget plot]
    table[x=horizon, y=low_0.95, col sep=comma, header=true]{\dataRMSELinearDayEcon};
\addplot[name path=upper_econ_linear_day, fill=none, draw=none, mark=none, forget plot]
    table[x=horizon, y=high_0.95, col sep=comma, header=true]{\dataRMSELinearDayEcon};
\addplot[light_indigo_purple!40, fill opacity=0.75, forget plot] fill between[of=lower_econ_linear_day and upper_econ_linear_day];

\addplot[name path=lower_no_econ_linear_day, fill=none, draw=none, mark=none, forget plot]
    table[x=horizon, y=low_0.95, col sep=comma, header=true]{\dataRMSELinearDayNoEcon};
\addplot[name path=upper_no_econ_linear_day, fill=none, draw=none, mark=none, forget plot]
    table[x=horizon, y=high_0.95, col sep=comma, header=true]{\dataRMSELinearDayNoEcon};
\addplot[light_light_apricot!40, fill opacity=0.75, forget plot] fill between[of=lower_no_econ_linear_day and upper_no_econ_linear_day];

\nextgroupplot[title={\textbf{XGBoost | Day}}, xlabel={Horizon [M]}, ylabel={$RMSE$}]
\addplot[mark=none, line width=1pt, color=light_indigo_purple]
        table [x=horizon, y=mean, col sep=comma] {\dataRMSEXGBoostDayEcon};
\addplot[mark=none, line width=1pt, color=light_light_apricot]
        table [x=horizon, y=mean, col sep=comma] {\dataRMSEXGBoostDayNoEcon};
\addplot[name path=lower_econ_xgboost_day, fill=none, draw=none, mark=none, forget plot]
    table[x=horizon, y=low_0.95, col sep=comma, header=true]{\dataRMSEXGBoostDayEcon};
\addplot[name path=upper_econ_xgboost_day, fill=none, draw=none, mark=none, forget plot]
    table[x=horizon, y=high_0.95, col sep=comma, header=true]{\dataRMSEXGBoostDayEcon};
\addplot[light_indigo_purple!40, fill opacity=0.75, forget plot] fill between[of=lower_econ_xgboost_day and upper_econ_xgboost_day];

\addplot[name path=lower_no_econ_xgboost_day, fill=none, draw=none, mark=none, forget plot]
    table[x=horizon, y=low_0.95, col sep=comma, header=true]{\dataRMSEXGBoostDayNoEcon};
\addplot[name path=upper_no_econ_xgboost_day, fill=none, draw=none, mark=none, forget plot]
    table[x=horizon, y=high_0.95, col sep=comma, header=true]{\dataRMSEXGBoostDayNoEcon};
\addplot[light_light_apricot!40, fill opacity=0.75, forget plot] fill between[of=lower_no_econ_xgboost_day and upper_no_econ_xgboost_day];

\nextgroupplot[title={\textbf{TabICL | Day}}, xlabel={Horizon [M]}, ylabel={$RMSE$}]
\addplot[mark=none, line width=1pt, color=light_indigo_purple]
        table [x=horizon, y=mean, col sep=comma] {\dataRMSETabICLDayEcon};
\addplot[mark=none, line width=1pt, color=light_light_apricot]
        table [x=horizon, y=mean, col sep=comma] {\dataRMSETabICLDayNoEcon};
\addplot[name path=lower_econ_tabICL_day, fill=none, draw=none, mark=none, forget plot]
    table[x=horizon, y=low_0.95, col sep=comma, header=true]{\dataRMSETabICLDayEcon};
\addplot[name path=upper_econ_tabICL_day, fill=none, draw=none, mark=none, forget plot]
    table[x=horizon, y=high_0.95, col sep=comma, header=true]{\dataRMSETabICLDayEcon};
\addplot[light_indigo_purple!40, fill opacity=0.75, forget plot] fill between[of=lower_econ_tabICL_day and upper_econ_tabICL_day];

\addplot[name path=lower_no_econ_tabICL_day, fill=none, draw=none, mark=none, forget plot]
    table[x=horizon, y=low_0.95, col sep=comma, header=true]{\dataRMSETabICLDayNoEcon};
\addplot[name path=upper_no_econ_tabICL_day, fill=none, draw=none, mark=none, forget plot]
    table[x=horizon, y=high_0.95, col sep=comma, header=true]{\dataRMSETabICLDayNoEcon};
\addplot[light_light_apricot!40, fill opacity=0.75, forget plot] fill between[of=lower_no_econ_tabICL_day and upper_no_econ_tabICL_day];
 
\end{groupplot}

\end{tikzpicture}

%% file: Plots/RMSE_ablation.tex




\pgfplotstableread[col sep=comma]{\getErrorDataFilePath{\resolutionMonth}{\linear}{\ablationFolder}{\Econ}{\rmse}}\dataRMSELinearMonthEcon
\pgfplotstableread[col sep=comma]{\getErrorDataFilePath{\resolutionMonth}{\linear}{\ablationFolder}{\EconNoVehicles}{\rmse}}\dataRMSELinearMonthEconNoVehicles
\pgfplotstableread[col sep=comma]{\getErrorDataFilePath{\resolutionMonth}{\linear}{\ablationFolder}{\EconNoHeatPumps}{\rmse}}\dataRMSELinearMonthEconNoHeatPumps
\pgfplotstableread[col sep=comma]{\getErrorDataFilePath{\resolutionMonth}{\linear}{\ablationFolder}{\EconNoVehiclesNoHeatPumps}{\rmse}}\dataRMSELinearMonthEconNoVehiclesNoHeatPumps

\pgfplotstableread[col sep=comma]{\getErrorDataFilePath{\resolutionMonth}{\xgboost}{\ablationFolder}{\Econ}{\rmse}}\dataRMSEXGBoostMonthEcon
\pgfplotstableread[col sep=comma]{\getErrorDataFilePath{\resolutionMonth}{\xgboost}{\ablationFolder}{\EconNoVehicles}{\rmse}}\dataRMSEXGBoostMonthEconNoVehicles
\pgfplotstableread[col sep=comma]{\getErrorDataFilePath{\resolutionMonth}{\xgboost}{\ablationFolder}{\EconNoHeatPumps}{\rmse}}\dataRMSEXGBoostMonthEconNoHeatPumps
\pgfplotstableread[col sep=comma]{\getErrorDataFilePath{\resolutionMonth}{\xgboost}{\ablationFolder}{\EconNoVehiclesNoHeatPumps}{\rmse}}\dataRMSEXGBoostMonthEconNoVehiclesNoHeatPumps

\pgfplotstableread[col sep=comma]{\getErrorDataFilePath{\resolutionMonth}{\tabicl}{\ablationFolder}{\Econ}{\rmse}}\dataRMSETabICLMonthEcon
\pgfplotstableread[col sep=comma]{\getErrorDataFilePath{\resolutionMonth}{\tabicl}{\ablationFolder}{\EconNoVehicles}{\rmse}}\dataRMSETabICLMonthEconNoVehicles
\pgfplotstableread[col sep=comma]{\getErrorDataFilePath{\resolutionMonth}{\tabicl}{\ablationFolder}{\EconNoHeatPumps}{\rmse}}\dataRMSETabICLMonthEconNoHeatPumps
\pgfplotstableread[col sep=comma]{\getErrorDataFilePath{\resolutionMonth}{\tabicl}{\ablationFolder}{\EconNoVehiclesNoHeatPumps}{\rmse}}\dataRMSETabICLMonthEconNoVehiclesNoHeatPumps

\pgfplotstableread[col sep=comma]{\getErrorDataFilePath{\resolutionDay}{\linear}{\ablationFolder}{\Econ}{\rmse}}\dataRMSELinearDayEcon
\pgfplotstableread[col sep=comma]{\getErrorDataFilePath{\resolutionDay}{\linear}{\ablationFolder}{\EconNoVehicles}{\rmse}}\dataRMSELinearDayEconNoVehicles
\pgfplotstableread[col sep=comma]{\getErrorDataFilePath{\resolutionDay}{\linear}{\ablationFolder}{\EconNoHeatPumps}{\rmse}}\dataRMSELinearDayEconNoHeatPumps
\pgfplotstableread[col sep=comma]{\getErrorDataFilePath{\resolutionDay}{\linear}{\ablationFolder}{\EconNoVehiclesNoHeatPumps}{\rmse}}\dataRMSELinearDayEconNoVehiclesNoHeatPumps

\pgfplotstableread[col sep=comma]{\getErrorDataFilePath{\resolutionDay}{\xgboost}{\ablationFolder}{\Econ}{\rmse}}\dataRMSEXGBoostDayEcon
\pgfplotstableread[col sep=comma]{\getErrorDataFilePath{\resolutionDay}{\xgboost}{\ablationFolder}{\EconNoVehicles}{\rmse}}\dataRMSEXGBoostDayEconNoVehicles
\pgfplotstableread[col sep=comma]{\getErrorDataFilePath{\resolutionDay}{\xgboost}{\ablationFolder}{\EconNoHeatPumps}{\rmse}}\dataRMSEXGBoostDayEconNoHeatPumps
\pgfplotstableread[col sep=comma]{\getErrorDataFilePath{\resolutionDay}{\xgboost}{\ablationFolder}{\EconNoVehiclesNoHeatPumps}{\rmse}}\dataRMSEXGBoostDayEconNoVehiclesNoHeatPumps

\pgfplotstableread[col sep=comma]{\getErrorDataFilePath{\resolutionMonth}{\tabicl}{\ablationFolder}{\Econ}{\rmse}}\dataRMSETabICLDayEcon
\pgfplotstableread[col sep=comma]{\getErrorDataFilePath{\resolutionMonth}{\tabicl}{\ablationFolder}{\EconNoVehicles}{\rmse}}\dataRMSETabICLDayEconNoVehicles
\pgfplotstableread[col sep=comma]{\getErrorDataFilePath{\resolutionMonth}{\tabicl}{\ablationFolder}{\EconNoHeatPumps}{\rmse}}\dataRMSETabICLDayEconNoHeatPumps
\pgfplotstableread[col sep=comma]{\getErrorDataFilePath{\resolutionMonth}{\tabicl}{\ablationFolder}{\EconNoVehiclesNoHeatPumps}{\rmse}}\dataRMSETabICLDayEconNoVehiclesNoHeatPumps

\begin{tikzpicture}
    \begin{groupplot}[
    group style={
        group size=3 by 2,        
        horizontal sep=1.8cm,
        vertical sep=1.8cm,
    },
    width=6cm,
    height=5cm,
    grid=major,
    scaled ticks=false,
    xtick={1, 12, 24, 36, 48},
    xticklabels={1, 12, 24, 36, 48},
    legend columns = 2,
    legend style={at={(0.45,-2.2)}, anchor=south west},
]
 
\nextgroupplot[title={\textbf{Linear | Month}}, xlabel={Horizon [M]}, ylabel={$RMSE$}]

\addplot[mark=none, line width=1pt, color=teal]
        table [x=horizon, y=mean, col sep=comma] {\dataRMSELinearMonthEcon};
\addlegendentry{Economic};
\addplot[mark=none, line width=1pt, color=dark_teal]
        table [x=horizon, y=mean, col sep=comma] {\dataRMSELinearMonthEconNoVehicles};
\addlegendentry{Economic without Vehicles};
\addplot[mark=none, line width=1pt, color=scarlet_rush]
        table [x=horizon, y=mean, col sep=comma] {\dataRMSELinearMonthEconNoHeatPumps};
\addlegendentry{Economic without HeatPumps};
\addplot[mark=none, line width=1pt, color=sunflower_gold]
        table [x=horizon, y=mean, col sep=comma] {\dataRMSELinearMonthEconNoVehiclesNoHeatPumps};
\addlegendentry{Economic without Vehicles \& HeatPumps};

\addplot[name path=lower_econ_linear_month, fill=none, draw=none, mark=none, forget plot]
    table[x=horizon, y=low_0.95, col sep=comma, header=true]{\dataRMSELinearMonthEcon};
\addplot[name path=upper_econ_linear_month, fill=none, draw=none, mark=none, forget plot]
    table[x=horizon, y=high_0.95, col sep=comma, header=true]{\dataRMSELinearMonthEcon};
\addplot[teal!40, fill opacity=0.75, forget plot] fill between[of=lower_econ_linear_month and upper_econ_linear_month];

\addplot[name path=lower_econ_no_vehicles_linear_month, fill=none, draw=none, mark=none, forget plot]
    table[x=horizon, y=low_0.95, col sep=comma, header=true]{\dataRMSELinearMonthEconNoVehicles};
\addplot[name path=upper_econ_no_vehicles_linear_month, fill=none, draw=none, mark=none, forget plot]
    table[x=horizon, y=high_0.95, col sep=comma, header=true]{\dataRMSELinearMonthEconNoVehicles};
\addplot[dark_teal!40, fill opacity=0.75, forget plot] fill between[of=lower_econ_no_vehicles_linear_month and upper_econ_no_vehicles_linear_month];

\addplot[name path=lower_econ_no_heatpumps_linear_month, fill=none, draw=none, mark=none, forget plot]
    table[x=horizon, y=low_0.95, col sep=comma, header=true]{\dataRMSELinearMonthEconNoHeatPumps};
\addplot[name path=upper_econ_no_heatpumps_linear_month, fill=none, draw=none, mark=none, forget plot]
    table[x=horizon, y=high_0.95, col sep=comma, header=true]{\dataRMSELinearMonthEconNoHeatPumps};
\addplot[scarlet_rush!40, fill opacity=0.75, forget plot] fill between[of=lower_econ_no_heatpumps_linear_month and upper_econ_no_heatpumps_linear_month];

\addplot[name path=lower_econ_no_vehicles_no_heatpumps_linear_month, fill=none, draw=none, mark=none, forget plot]
    table[x=horizon, y=low_0.95, col sep=comma, header=true]{\dataRMSELinearMonthEconNoVehiclesNoHeatPumps};
\addplot[name path=upper_econ_no_vehicles_no_heatpumps_linear_month, fill=none, draw=none, mark=none, forget plot]
    table[x=horizon, y=high_0.95, col sep=comma, header=true]{\dataRMSELinearMonthEconNoVehiclesNoHeatPumps};
\addplot[sunflower_gold!40, fill opacity=0.75, forget plot] fill between[of=lower_econ_no_vehicles_no_heatpumps_linear_month and upper_econ_no_vehicles_no_heatpumps_linear_month];

\nextgroupplot[title={\textbf{XGBoost | Month}}, xlabel={Horizon [M]}, ylabel={$RMSE$}]
\addplot[mark=none, line width=1pt, color=teal]
        table [x=horizon, y=mean, col sep=comma] {\dataRMSEXGBoostMonthEcon};
\addplot[mark=none, line width=1pt, color=dark_teal]
        table [x=horizon, y=mean, col sep=comma] {\dataRMSEXGBoostMonthEconNoVehicles};
\addplot[mark=none, line width=1pt, color=scarlet_rush]
        table [x=horizon, y=mean, col sep=comma] {\dataRMSEXGBoostMonthEconNoHeatPumps};
\addplot[mark=none, line width=1pt, color=sunflower_gold]
        table [x=horizon, y=mean, col sep=comma] {\dataRMSEXGBoostMonthEconNoVehiclesNoHeatPumps};

\addplot[name path=lower_econ_xgboost_month, fill=none, draw=none, mark=none, forget plot]
    table[x=horizon, y=low_0.95, col sep=comma, header=true]{\dataRMSEXGBoostMonthEcon};
\addplot[name path=upper_econ_xgboost_month, fill=none, draw=none, mark=none, forget plot]
    table[x=horizon, y=high_0.95, col sep=comma, header=true]{\dataRMSEXGBoostMonthEcon};
\addplot[teal!40, fill opacity=0.75, forget plot] fill between[of=lower_econ_xgboost_month and upper_econ_xgboost_month];

\addplot[name path=lower_econ_no_vehicles_xgboost_month, fill=none, draw=none, mark=none, forget plot]
    table[x=horizon, y=low_0.95, col sep=comma, header=true]{\dataRMSEXGBoostMonthEconNoVehicles};
\addplot[name path=upper_econ_no_vehicles_xgboost_month, fill=none, draw=none, mark=none, forget plot]
    table[x=horizon, y=high_0.95, col sep=comma, header=true]{\dataRMSEXGBoostMonthEconNoVehicles};
\addplot[dark_teal!40, fill opacity=0.75, forget plot] fill between[of=lower_econ_no_vehicles_xgboost_month and upper_econ_no_vehicles_xgboost_month];

\addplot[name path=lower_econ_no_heatpumps_xgboost_month, fill=none, draw=none, mark=none, forget plot]
    table[x=horizon, y=low_0.95, col sep=comma, header=true]{\dataRMSEXGBoostMonthEconNoHeatPumps};
\addplot[name path=upper_econ_no_heatpumps_xgboost_month, fill=none, draw=none, mark=none, forget plot]
    table[x=horizon, y=high_0.95, col sep=comma, header=true]{\dataRMSEXGBoostMonthEconNoHeatPumps};
\addplot[scarlet_rush!40, fill opacity=0.75, forget plot] fill between[of=lower_econ_no_heatpumps_xgboost_month and upper_econ_no_heatpumps_xgboost_month];

\addplot[name path=lower_econ_no_vehicles_no_heatpumps_xgboost_month, fill=none, draw=none, mark=none, forget plot]
    table[x=horizon, y=low_0.95, col sep=comma, header=true]{\dataRMSEXGBoostMonthEconNoVehiclesNoHeatPumps};
\addplot[name path=upper_econ_no_vehicles_no_heatpumps_xgboost_month, fill=none, draw=none, mark=none, forget plot]
    table[x=horizon, y=high_0.95, col sep=comma, header=true]{\dataRMSEXGBoostMonthEconNoVehiclesNoHeatPumps};
\addplot[sunflower_gold!40, fill opacity=0.75, forget plot] fill between[of=lower_econ_no_vehicles_no_heatpumps_xgboost_month and upper_econ_no_vehicles_no_heatpumps_xgboost_month];

\nextgroupplot[title={\textbf{TabICL | Month}}, xlabel={Horizon [M]}, ylabel={$RMSE$}]
\addplot[mark=none, line width=1pt, color=teal]
        table [x=horizon, y=mean, col sep=comma] {\dataRMSETabICLMonthEcon};
\addplot[mark=none, line width=1pt, color=dark_teal]
        table [x=horizon, y=mean, col sep=comma] {\dataRMSETabICLMonthEconNoVehicles};
\addplot[mark=none, line width=1pt, color=scarlet_rush]
        table [x=horizon, y=mean, col sep=comma] {\dataRMSETabICLMonthEconNoHeatPumps};
\addplot[mark=none, line width=1pt, color=sunflower_gold]
        table [x=horizon, y=mean, col sep=comma] {\dataRMSETabICLMonthEconNoVehiclesNoHeatPumps};

\addplot[name path=lower_econ_tabicl_month, fill=none, draw=none, mark=none, forget plot]
    table[x=horizon, y=low_0.95, col sep=comma, header=true]{\dataRMSETabICLMonthEcon};
\addplot[name path=upper_econ_tabicl_month, fill=none, draw=none, mark=none, forget plot]
    table[x=horizon, y=high_0.95, col sep=comma, header=true]{\dataRMSETabICLMonthEcon};
\addplot[teal!40, fill opacity=0.75, forget plot] fill between[of=lower_econ_tabicl_month and upper_econ_tabicl_month];

\addplot[name path=lower_econ_no_vehicles_tabicl_month, fill=none, draw=none, mark=none, forget plot]
    table[x=horizon, y=low_0.95, col sep=comma, header=true]{\dataRMSETabICLMonthEconNoVehicles};
\addplot[name path=upper_econ_no_vehicles_tabicl_month, fill=none, draw=none, mark=none, forget plot]
    table[x=horizon, y=high_0.95, col sep=comma, header=true]{\dataRMSETabICLMonthEconNoVehicles};
\addplot[dark_teal!40, fill opacity=0.75, forget plot] fill between[of=lower_econ_no_vehicles_tabicl_month and upper_econ_no_vehicles_tabicl_month];

\addplot[name path=lower_econ_no_heatpumps_tabicl_month, fill=none, draw=none, mark=none, forget plot]
    table[x=horizon, y=low_0.95, col sep=comma, header=true]{\dataRMSETabICLMonthEconNoHeatPumps};
\addplot[name path=upper_econ_no_heatpumps_tabicl_month, fill=none, draw=none, mark=none, forget plot]
    table[x=horizon, y=high_0.95, col sep=comma, header=true]{\dataRMSETabICLMonthEconNoHeatPumps};
\addplot[scarlet_rush!40, fill opacity=0.75, forget plot] fill between[of=lower_econ_no_heatpumps_tabicl_month and upper_econ_no_heatpumps_tabicl_month];

\addplot[name path=lower_econ_no_vehicles_no_heatpumps_tabicl_month, fill=none, draw=none, mark=none, forget plot]
    table[x=horizon, y=low_0.95, col sep=comma, header=true]{\dataRMSETabICLMonthEconNoVehiclesNoHeatPumps};
\addplot[name path=upper_econ_no_vehicles_no_heatpumps_tabicl_month, fill=none, draw=none, mark=none, forget plot]
    table[x=horizon, y=high_0.95, col sep=comma, header=true]{\dataRMSETabICLMonthEconNoVehiclesNoHeatPumps};
\addplot[sunflower_gold!40, fill opacity=0.75, forget plot] fill between[of=lower_econ_no_vehicles_no_heatpumps_tabicl_month and upper_econ_no_vehicles_no_heatpumps_tabicl_month];

\nextgroupplot[title={\textbf{Linear | Day}}, xlabel={Horizon [M]}, ylabel={$RMSE$}]

\addplot[mark=none, line width=1pt, color=teal]
        table [x=horizon, y=mean, col sep=comma] {\dataRMSELinearDayEcon};
\addplot[mark=none, line width=1pt, color=dark_teal]
        table [x=horizon, y=mean, col sep=comma] {\dataRMSELinearDayEconNoVehicles};
\addplot[mark=none, line width=1pt, color=scarlet_rush]
        table [x=horizon, y=mean, col sep=comma] {\dataRMSELinearDayEconNoHeatPumps};
\addplot[mark=none, line width=1pt, color=sunflower_gold]
        table [x=horizon, y=mean, col sep=comma] {\dataRMSELinearDayEconNoVehiclesNoHeatPumps};

\addplot[name path=lower_econ_linear_day, fill=none, draw=none, mark=none, forget plot]
    table[x=horizon, y=low_0.95, col sep=comma, header=true]{\dataRMSELinearDayEcon};
\addplot[name path=upper_econ_linear_day, fill=none, draw=none, mark=none, forget plot]
    table[x=horizon, y=high_0.95, col sep=comma, header=true]{\dataRMSELinearDayEcon};
\addplot[teal!40, fill opacity=0.75, forget plot] fill between[of=lower_econ_linear_day and upper_econ_linear_day];

\addplot[name path=lower_econ_no_vehicles_linear_day, fill=none, draw=none, mark=none, forget plot]
    table[x=horizon, y=low_0.95, col sep=comma, header=true]{\dataRMSELinearDayEconNoVehicles};
\addplot[name path=upper_econ_no_vehicles_linear_day, fill=none, draw=none, mark=none, forget plot]
    table[x=horizon, y=high_0.95, col sep=comma, header=true]{\dataRMSELinearDayEconNoVehicles};
\addplot[dark_teal!40, fill opacity=0.75, forget plot] fill between[of=lower_econ_no_vehicles_linear_day and upper_econ_no_vehicles_linear_day];

\addplot[name path=lower_econ_no_heatpumps_linear_day, fill=none, draw=none, mark=none, forget plot]
    table[x=horizon, y=low_0.95, col sep=comma, header=true]{\dataRMSELinearDayEconNoHeatPumps};
\addplot[name path=upper_econ_no_heatpumps_linear_day, fill=none, draw=none, mark=none, forget plot]
    table[x=horizon, y=high_0.95, col sep=comma, header=true]{\dataRMSELinearDayEconNoHeatPumps};
\addplot[scarlet_rush!40, fill opacity=0.75, forget plot] fill between[of=lower_econ_no_heatpumps_linear_day and upper_econ_no_heatpumps_linear_day];

\addplot[name path=lower_econ_no_vehicles_no_heatpumps_linear_day, fill=none, draw=none, mark=none, forget plot]
    table[x=horizon, y=low_0.95, col sep=comma, header=true]{\dataRMSELinearDayEconNoVehiclesNoHeatPumps};
\addplot[name path=upper_econ_no_vehicles_no_heatpumps_linear_day, fill=none, draw=none, mark=none, forget plot]
    table[x=horizon, y=high_0.95, col sep=comma, header=true]{\dataRMSELinearDayEconNoVehiclesNoHeatPumps};
\addplot[sunflower_gold!40, fill opacity=0.75, forget plot] fill between[of=lower_econ_no_vehicles_no_heatpumps_linear_day and upper_econ_no_vehicles_no_heatpumps_linear_day];

\nextgroupplot[title={\textbf{XGBoost | Day}}, xlabel={Horizon [M]}, ylabel={$RMSE$}]
\addplot[mark=none, line width=1pt, color=teal]
        table [x=horizon, y=mean, col sep=comma] {\dataRMSEXGBoostDayEcon};
\addplot[mark=none, line width=1pt, color=dark_teal]
        table [x=horizon, y=mean, col sep=comma] {\dataRMSEXGBoostDayEconNoVehicles};
\addplot[mark=none, line width=1pt, color=scarlet_rush]
        table [x=horizon, y=mean, col sep=comma] {\dataRMSEXGBoostDayEconNoHeatPumps};
\addplot[mark=none, line width=1pt, color=sunflower_gold]
        table [x=horizon, y=mean, col sep=comma] {\dataRMSEXGBoostDayEconNoVehiclesNoHeatPumps};

\addplot[name path=lower_econ_xgboost_day, fill=none, draw=none, mark=none, forget plot]
    table[x=horizon, y=low_0.95, col sep=comma, header=true]{\dataRMSEXGBoostDayEcon};
\addplot[name path=upper_econ_xgboost_day, fill=none, draw=none, mark=none, forget plot]
    table[x=horizon, y=high_0.95, col sep=comma, header=true]{\dataRMSEXGBoostDayEcon};
\addplot[teal!40, fill opacity=0.75, forget plot] fill between[of=lower_econ_xgboost_day and upper_econ_xgboost_day];

\addplot[name path=lower_econ_no_vehicles_xgboost_day, fill=none, draw=none, mark=none, forget plot]
    table[x=horizon, y=low_0.95, col sep=comma, header=true]{\dataRMSEXGBoostDayEconNoVehicles};
\addplot[name path=upper_econ_no_vehicles_xgboost_day, fill=none, draw=none, mark=none, forget plot]
    table[x=horizon, y=high_0.95, col sep=comma, header=true]{\dataRMSEXGBoostDayEconNoVehicles};
\addplot[dark_teal!40, fill opacity=0.75, forget plot] fill between[of=lower_econ_no_vehicles_xgboost_day and upper_econ_no_vehicles_xgboost_day];

\addplot[name path=lower_econ_no_heatpumps_xgboost_day, fill=none, draw=none, mark=none, forget plot]
    table[x=horizon, y=low_0.95, col sep=comma, header=true]{\dataRMSEXGBoostDayEconNoHeatPumps};
\addplot[name path=upper_econ_no_heatpumps_xgboost_day, fill=none, draw=none, mark=none, forget plot]
    table[x=horizon, y=high_0.95, col sep=comma, header=true]{\dataRMSEXGBoostDayEconNoHeatPumps};
\addplot[scarlet_rush!40, fill opacity=0.75, forget plot] fill between[of=lower_econ_no_heatpumps_xgboost_day and upper_econ_no_heatpumps_xgboost_day];

\addplot[name path=lower_econ_no_vehicles_no_heatpumps_xgboost_day, fill=none, draw=none, mark=none, forget plot]
    table[x=horizon, y=low_0.95, col sep=comma, header=true]{\dataRMSEXGBoostDayEconNoVehiclesNoHeatPumps};
\addplot[name path=upper_econ_no_vehicles_no_heatpumps_xgboost_day, fill=none, draw=none, mark=none, forget plot]
    table[x=horizon, y=high_0.95, col sep=comma, header=true]{\dataRMSEXGBoostDayEconNoVehiclesNoHeatPumps};
\addplot[sunflower_gold!40, fill opacity=0.75, forget plot] fill between[of=lower_econ_no_vehicles_no_heatpumps_xgboost_day and upper_econ_no_vehicles_no_heatpumps_xgboost_day];

\nextgroupplot[title={\textbf{TabICL | Day}}, xlabel={Horizon [M]}, ylabel={$RMSE$}]
\addplot[mark=none, line width=1pt, color=teal]
        table [x=horizon, y=mean, col sep=comma] {\dataRMSETabICLDayEcon};
\addplot[mark=none, line width=1pt, color=dark_teal]
        table [x=horizon, y=mean, col sep=comma] {\dataRMSETabICLDayEconNoVehicles};
\addplot[mark=none, line width=1pt, color=scarlet_rush]
        table [x=horizon, y=mean, col sep=comma] {\dataRMSETabICLDayEconNoHeatPumps};
\addplot[mark=none, line width=1pt, color=sunflower_gold]
        table [x=horizon, y=mean, col sep=comma] {\dataRMSETabICLDayEconNoVehiclesNoHeatPumps};

\addplot[name path=lower_econ_tabicl_day, fill=none, draw=none, mark=none, forget plot]
    table[x=horizon, y=low_0.95, col sep=comma, header=true]{\dataRMSETabICLDayEcon};
\addplot[name path=upper_econ_tabicl_day, fill=none, draw=none, mark=none, forget plot]
    table[x=horizon, y=high_0.95, col sep=comma, header=true]{\dataRMSETabICLDayEcon};
\addplot[teal!40, fill opacity=0.75, forget plot] fill between[of=lower_econ_tabicl_day and upper_econ_tabicl_day];

\addplot[name path=lower_econ_no_vehicles_tabicl_day, fill=none, draw=none, mark=none, forget plot]
    table[x=horizon, y=low_0.95, col sep=comma, header=true]{\dataRMSETabICLDayEconNoVehicles};
\addplot[name path=upper_econ_no_vehicles_tabicl_day, fill=none, draw=none, mark=none, forget plot]
    table[x=horizon, y=high_0.95, col sep=comma, header=true]{\dataRMSETabICLDayEconNoVehicles};
\addplot[dark_teal!40, fill opacity=0.75, forget plot] fill between[of=lower_econ_no_vehicles_tabicl_day and upper_econ_no_vehicles_tabicl_day];

\addplot[name path=lower_econ_no_heatpumps_tabicl_day, fill=none, draw=none, mark=none, forget plot]
    table[x=horizon, y=low_0.95, col sep=comma, header=true]{\dataRMSETabICLDayEconNoHeatPumps};
\addplot[name path=upper_econ_no_heatpumps_tabicl_day, fill=none, draw=none, mark=none, forget plot]
    table[x=horizon, y=high_0.95, col sep=comma, header=true]{\dataRMSETabICLDayEconNoHeatPumps};
\addplot[scarlet_rush!40, fill opacity=0.75, forget plot] fill between[of=lower_econ_no_heatpumps_tabicl_day and upper_econ_no_heatpumps_tabicl_day];

\addplot[name path=lower_econ_no_vehicles_no_heatpumps_tabicl_day, fill=none, draw=none, mark=none, forget plot]
    table[x=horizon, y=low_0.95, col sep=comma, header=true]{\dataRMSETabICLDayEconNoVehiclesNoHeatPumps};
\addplot[name path=upper_econ_no_vehicles_no_heatpumps_tabicl_day, fill=none, draw=none, mark=none, forget plot]
    table[x=horizon, y=high_0.95, col sep=comma, header=true]{\dataRMSETabICLDayEconNoVehiclesNoHeatPumps};
\addplot[sunflower_gold!40, fill opacity=0.75, forget plot] fill between[of=lower_econ_no_vehicles_no_heatpumps_tabicl_day and upper_econ_no_vehicles_no_heatpumps_tabicl_day];

\end{groupplot}

\end{tikzpicture}

%% file: Plots/selected_features_month.tex





\pgfplotstableread[col sep=comma]{\getSelectedFeaturesDataFilePath{month}{linear}{selected_features}{econ}{calendar}}\dataCalendarLinear
\pgfplotstableread[col sep=comma]{\getSelectedFeaturesDataFilePath{month}{xgboost}{selected_features}{econ}{calendar}}\dataCalendarXGBoost
\pgfplotstableread[col sep=comma]{\getSelectedFeaturesDataFilePath{month}{tabICL}{selected_features}{econ}{calendar}}\dataCalendarTabICL
\pgfplotstableread[col sep=comma]{\getSelectedFeaturesDataFilePath{month}{linear}{selected_features}{econ}{weather}}\dataWeatherLinear
\pgfplotstableread[col sep=comma]{\getSelectedFeaturesDataFilePath{month}{xgboost}{selected_features}{econ}{weather}}\dataWeatherXGBoost
\pgfplotstableread[col sep=comma]{\getSelectedFeaturesDataFilePath{month}{tabICL}{selected_features}{econ}{weather}}\dataWeatherTabICL
\pgfplotstableread[col sep=comma]{\getSelectedFeaturesDataFilePath{month}{linear}{selected_features}{econ}{economic}}\dataEconomicLinear
\pgfplotstableread[col sep=comma]{\getSelectedFeaturesDataFilePath{month}{xgboost}{selected_features}{econ}{economic}}\dataEconomicXGBoost
\pgfplotstableread[col sep=comma]{\getSelectedFeaturesDataFilePath{month}{tabICL}{selected_features}{econ}{economic}}\dataEconomicTabICL

\begin{tikzpicture}

\begin{axis}[
    ybar,
    bar width=5pt,
    width=8.5cm,
    height=6.5cm,
    x=1cm,
    ymin=0,
    ymax=1.15,
    ymajorgrids,
    axis lines*=left,
    ylabel={Selection Ratio},
    enlarge x limits=0.075,
    xtick=data,
    x tick label style={rotate=45, anchor=east},
    name=plotCalendar,
    title={\large\textbf{Calendar Features}},
    at={(0,0)},
    anchor=north west,
    symbolic x coords={moy\_cos, moy\_sin, qoy\_cos, qoy\_sin, seas\_cos, seas\_sin, is\_weekend, is\_public\_holiday, is\_school\_holiday},
    legend style={at={(0.65,-2.25)}, font=\Large, legend columns=3, /tikz/column sep=5pt, anchor=north west},
  ]
  \addplot[color = orange, fill=orange, fill opacity=0.75] table[x=features, y=occurence] {\dataCalendarLinear};
  \addplot[color = spicy_paprika, fill=spicy_paprika, fill opacity=0.75]   table[x=features, y=occurence] {\dataCalendarXGBoost};
  \addplot[color = twilight_indigo, , fill=twilight_indigo, fill opacity=0.75]      table[x=features, y=occurence] {\dataCalendarTabICL};
  \legend{Linear, XGBoost, TabICL}
\end{axis}

\begin{axis}[
    ybar,
    bar width=5pt,
    width=8.5cm,
    height=6.5cm,
    x=1cm,
    ymin=0,
    ymax=1.15,
    ymajorgrids,
    axis lines*=left,
    ylabel={Selection Ratio},
    enlarge x limits=0.075,
    xtick=data,
    x tick label style={rotate=45, anchor=east},
    name=plotWeather,
    title={\large\textbf{Weather Features}},
    at={(plotCalendar.north east)},
    anchor=north west,
    xshift=2cm,
    xtick={t2m, hdd, cdd, rh,  ssrd, tcc, tp, u10, v10},
    xticklabels={t2m, hdd, cdd, rh,  ssrd, tcc, tp, u10, v10},
    symbolic x coords={t2m, hdd, cdd, rh,  ssrd, tcc, tp, u10, v10},
  ]
  \addplot[color = orange, fill=orange, fill opacity=0.75]table[x=features, y=occurence] {\dataWeatherLinear};
  \addplot[color = spicy_paprika, fill=spicy_paprika, fill opacity=0.75] table[x=features, y=occurence] {\dataWeatherXGBoost};
  \addplot[color = twilight_indigo, , fill=twilight_indigo, fill opacity=0.75] table[x=features, y=occurence] {\dataWeatherTabICL};
\end{axis}

\begin{axis}[
    ybar,
    bar width=5pt,
    width=8.5cm,
    height=6.5cm,
    x=1cm,
    ymin=0,
    ymax=1.15,
    ymajorgrids,
    axis lines*=left,
    ylabel={Selection Ratio},
    enlarge x limits=0.075,
    xtick=data,
    x tick label style={rotate=45, anchor=east},
    name=plotEconomic,
    title={\large\textbf{Economic Features}},
    at={($(plotCalendar.south west)!0.5!(plotWeather.south east)$)},
    anchor=north,
    yshift=-2.25cm,
    enlarge x limits=0.015,
    xtick={B\_CVS-CJO, C\_CVS-CJO, D\_CVS-CJO, E\_CVS-CJO, H\_CVS-CJO, I\_CVS-CJO, J\_CVS-CJO, L\_CVS-CJO, M\_CVS-CJO, N\_CVS-CJO, S\_CVS-CJO, cpi\_w/o\_energy, ecpi, scpi, employment, population, household\_income\_CVS-CJO, tourism\_nights, PAC\_air/air, PAC\_air/eau, PAC\_geo, electric\_car\_passenger, electric\_light\_utilitary\_vehicle, electric\_heavy\_utilitary\_vehicle, electric\_public\_transports, hybrid\_car\_passenger, hybrid\_light\_utilitary\_vehicle, hybrid\_heavy\_utilitary\_vehicle, hybrid\_public\_transports},
    xticklabels={B\_CVS-CJO, C\_CVS-CJO, D\_CVS-CJO, E\_CVS-CJO, H\_CVS-CJO, I\_CVS-CJO, J\_CVS-CJO, L\_CVS-CJO, M\_CVS-CJO, N\_CVS-CJO, S\_CVS-CJO, cpi\_w/o\_energy, ecpi, scpi, employment, population, household\_income\_CVS-CJO, tourism\_nights, PAC\_air/air, PAC\_air/eau, PAC\_geo, electric\_car\_passenger, electric\_light\_utilitary\_vehicle, electric\_heavy\_utilitary\_vehicle, electric\_public\_transports, hybrid\_car\_passenger, hybrid\_light\_utilitary\_vehicle, hybrid\_heavy\_utilitary\_vehicle, hybrid\_public\_transports},
    symbolic x coords={B\_CVS-CJO, C\_CVS-CJO, D\_CVS-CJO, E\_CVS-CJO, H\_CVS-CJO, I\_CVS-CJO, J\_CVS-CJO, L\_CVS-CJO, M\_CVS-CJO, N\_CVS-CJO, S\_CVS-CJO, cpi\_w/o\_energy, ecpi, scpi, employment, population, household\_income\_CVS-CJO, tourism\_nights, PAC\_air/air, PAC\_air/eau, PAC\_geo, electric\_car\_passenger, electric\_light\_utilitary\_vehicle, electric\_heavy\_utilitary\_vehicle, electric\_public\_transports, hybrid\_car\_passenger, hybrid\_light\_utilitary\_vehicle, hybrid\_heavy\_utilitary\_vehicle, hybrid\_public\_transports},
  ]
  \addplot[color = orange, fill=orange, fill opacity=0.75] table[x=features, y=occurence] {\dataEconomicLinear};
  \addplot[color = spicy_paprika, fill=spicy_paprika, fill opacity=0.75] table[x=features, y=occurence] {\dataEconomicXGBoost};
  \addplot[color = twilight_indigo, , fill=twilight_indigo, fill opacity=0.75] table[x=features, y=occurence] {\dataEconomicTabICL};
\end{axis}

\end{tikzpicture}


%% file: Plots/feature_importance_part1.tex




\pgfplotstableread[col sep=comma]{\getFeaturesImportanceDataFilePath{month}{linear}{feature_importance}{econ}}\dataFeatureImportanceLinear
\pgfplotstableread[col sep=comma]{\getFeaturesImportanceDataFilePath{month}{xgboost}{feature_importance}{econ}}\dataFeatureImportanceXGBoost
\pgfplotstableread[col sep=comma]{\getFeaturesImportanceDataFilePath{month}{tabICL}{feature_importance}{econ}}\dataFeatureImportanceTabICL

\begin{tikzpicture}
\begin{groupplot}[group style={group size=4 by 6, horizontal sep=1.5cm, vertical sep=1.75cm}, width=8cm, height=6cm, xlabel={Horizon [M]}, ylabel={Feature Importance}, legend columns=3,
    legend style={at={(2,-0.4)}, anchor=north west},
    /pgfplots/scale ticks below exponent={-3},
/pgfplots/scale ticks above exponent={2}]
\nextgroupplot[title={\textbf{B\_CVS-CJO}}]
\addplot [spicy_paprika, mark=none, line width=1pt] table [x=horizon, y=B_CVS-CJO_mean] {\dataFeatureImportanceXGBoost};
\addplot [name path=lower_xgboost_feature_0, draw=none, forget plot] table [x=horizon, y=B_CVS-CJO_low_0.95] {\dataFeatureImportanceXGBoost};
\addplot [name path=upper_xgboost_feature_0, draw=none, forget plot] table [x=horizon, y=B_CVS-CJO_high_0.95] {\dataFeatureImportanceXGBoost};
\addplot [spicy_paprika!40, fill opacity=0.75, forget plot] fill between [of=upper_xgboost_feature_0 and lower_xgboost_feature_0];
\addplot [twilight_indigo, mark=none, line width=1pt] table [x=horizon, y=B_CVS-CJO_mean] {\dataFeatureImportanceTabICL};
\addplot [name path=lower_tabicl_feature_0, draw=none, forget plot] table [x=horizon, y=B_CVS-CJO_low_0.95] {\dataFeatureImportanceTabICL};
\addplot [name path=upper_tabicl_feature_0, draw=none, forget plot] table [x=horizon, y=B_CVS-CJO_high_0.95] {\dataFeatureImportanceTabICL};
\addplot [twilight_indigo!40, fill opacity=0.75, forget plot] fill between [of=upper_tabicl_feature_0 and lower_tabicl_feature_0];
\nextgroupplot[title={\textbf{C\_CVS-CJO}}]
\addplot [spicy_paprika, mark=none, line width=1pt] table [x=horizon, y=C_CVS-CJO_mean] {\dataFeatureImportanceXGBoost};
\addplot [name path=lower_xgboost_feature_1, draw=none, forget plot] table [x=horizon, y=C_CVS-CJO_low_0.95] {\dataFeatureImportanceXGBoost};
\addplot [name path=upper_xgboost_feature_1, draw=none, forget plot] table [x=horizon, y=C_CVS-CJO_high_0.95] {\dataFeatureImportanceXGBoost};
\addplot [spicy_paprika!40, fill opacity=0.75, forget plot] fill between [of=upper_xgboost_feature_1 and lower_xgboost_feature_1];
\addplot [twilight_indigo, mark=none, line width=1pt] table [x=horizon, y=C_CVS-CJO_mean] {\dataFeatureImportanceTabICL};
\addplot [name path=lower_tabicl_feature_1, draw=none, forget plot] table [x=horizon, y=C_CVS-CJO_low_0.95] {\dataFeatureImportanceTabICL};
\addplot [name path=upper_tabicl_feature_1, draw=none, forget plot] table [x=horizon, y=C_CVS-CJO_high_0.95] {\dataFeatureImportanceTabICL};
\addplot [twilight_indigo!40, fill opacity=0.75, forget plot] fill between [of=upper_tabicl_feature_1 and lower_tabicl_feature_1];
\nextgroupplot[title={\textbf{D\_CVS-CJO}}]
\addplot [orange, mark=none, line width=1pt] table [x=horizon, y=D_CVS-CJO_mean] {\dataFeatureImportanceLinear};
\addplot [name path=lower_linear_feature_2, draw=none, forget plot] table [x=horizon, y=D_CVS-CJO_low_0.95] {\dataFeatureImportanceLinear};
\addplot [name path=upper_linear_feature_2, draw=none, forget plot] table [x=horizon, y=D_CVS-CJO_high_0.95] {\dataFeatureImportanceLinear};
\addplot [orange!40, fill opacity=0.75, forget plot] fill between [of=upper_linear_feature_2 and lower_linear_feature_2];
\addplot [spicy_paprika, mark=none, line width=1pt] table [x=horizon, y=D_CVS-CJO_mean] {\dataFeatureImportanceXGBoost};
\addplot [name path=lower_xgboost_feature_2, draw=none, forget plot] table [x=horizon, y=D_CVS-CJO_low_0.95] {\dataFeatureImportanceXGBoost};
\addplot [name path=upper_xgboost_feature_2, draw=none, forget plot] table [x=horizon, y=D_CVS-CJO_high_0.95] {\dataFeatureImportanceXGBoost};
\addplot [spicy_paprika!40, fill opacity=0.75, forget plot] fill between [of=upper_xgboost_feature_2 and lower_xgboost_feature_2];
\nextgroupplot[title={\textbf{E\_CVS-CJO}}]
\addplot [orange, mark=none, line width=1pt] table [x=horizon, y=E_CVS-CJO_mean] {\dataFeatureImportanceLinear};
\addplot [name path=lower_linear_feature_3, draw=none, forget plot] table [x=horizon, y=E_CVS-CJO_low_0.95] {\dataFeatureImportanceLinear};
\addplot [name path=upper_linear_feature_3, draw=none, forget plot] table [x=horizon, y=E_CVS-CJO_high_0.95] {\dataFeatureImportanceLinear};
\addplot [orange!40, fill opacity=0.75, forget plot] fill between [of=upper_linear_feature_3 and lower_linear_feature_3];
\nextgroupplot[title={\textbf{H\_CVS-CJO}}]
\addplot [orange, mark=none, line width=1pt] table [x=horizon, y=H_CVS-CJO_mean] {\dataFeatureImportanceLinear};
\addplot [name path=lower_linear_feature_4, draw=none, forget plot] table [x=horizon, y=H_CVS-CJO_low_0.95] {\dataFeatureImportanceLinear};
\addplot [name path=upper_linear_feature_4, draw=none, forget plot] table [x=horizon, y=H_CVS-CJO_high_0.95] {\dataFeatureImportanceLinear};
\addplot [orange!40, fill opacity=0.75, forget plot] fill between [of=upper_linear_feature_4 and lower_linear_feature_4];
\addplot [twilight_indigo, mark=none, line width=1pt] table [x=horizon, y=H_CVS-CJO_mean] {\dataFeatureImportanceTabICL};
\addplot [name path=lower_tabicl_feature_4, draw=none, forget plot] table [x=horizon, y=H_CVS-CJO_low_0.95] {\dataFeatureImportanceTabICL};
\addplot [name path=upper_tabicl_feature_4, draw=none, forget plot] table [x=horizon, y=H_CVS-CJO_high_0.95] {\dataFeatureImportanceTabICL};
\addplot [twilight_indigo!40, fill opacity=0.75, forget plot] fill between [of=upper_tabicl_feature_4 and lower_tabicl_feature_4];
\nextgroupplot[title={\textbf{J\_CVS-CJO}}]
\addplot [orange, mark=none, line width=1pt] table [x=horizon, y=J_CVS-CJO_mean] {\dataFeatureImportanceLinear};
\addplot [name path=lower_linear_feature_5, draw=none, forget plot] table [x=horizon, y=J_CVS-CJO_low_0.95] {\dataFeatureImportanceLinear};
\addplot [name path=upper_linear_feature_5, draw=none, forget plot] table [x=horizon, y=J_CVS-CJO_high_0.95] {\dataFeatureImportanceLinear};
\addplot [orange!40, fill opacity=0.75, forget plot] fill between [of=upper_linear_feature_5 and lower_linear_feature_5];
\addplot [twilight_indigo, mark=none, line width=1pt] table [x=horizon, y=J_CVS-CJO_mean] {\dataFeatureImportanceTabICL};
\addplot [name path=lower_tabicl_feature_5, draw=none, forget plot] table [x=horizon, y=J_CVS-CJO_low_0.95] {\dataFeatureImportanceTabICL};
\addplot [name path=upper_tabicl_feature_5, draw=none, forget plot] table [x=horizon, y=J_CVS-CJO_high_0.95] {\dataFeatureImportanceTabICL};
\addplot [twilight_indigo!40, fill opacity=0.75, forget plot] fill between [of=upper_tabicl_feature_5 and lower_tabicl_feature_5];
\nextgroupplot[title={\textbf{L\_CVS-CJO}}]
\addplot [orange, mark=none, line width=1pt] table [x=horizon, y=L_CVS-CJO_mean] {\dataFeatureImportanceLinear};
\addplot [name path=lower_linear_feature_6, draw=none, forget plot] table [x=horizon, y=L_CVS-CJO_low_0.95] {\dataFeatureImportanceLinear};
\addplot [name path=upper_linear_feature_6, draw=none, forget plot] table [x=horizon, y=L_CVS-CJO_high_0.95] {\dataFeatureImportanceLinear};
\addplot [orange!40, fill opacity=0.75, forget plot] fill between [of=upper_linear_feature_6 and lower_linear_feature_6];
\addplot [twilight_indigo, mark=none, line width=1pt] table [x=horizon, y=L_CVS-CJO_mean] {\dataFeatureImportanceTabICL};
\addplot [name path=lower_tabicl_feature_6, draw=none, forget plot] table [x=horizon, y=L_CVS-CJO_low_0.95] {\dataFeatureImportanceTabICL};
\addplot [name path=upper_tabicl_feature_6, draw=none, forget plot] table [x=horizon, y=L_CVS-CJO_high_0.95] {\dataFeatureImportanceTabICL};
\addplot [twilight_indigo!40, fill opacity=0.75, forget plot] fill between [of=upper_tabicl_feature_6 and lower_tabicl_feature_6];
\nextgroupplot[title={\textbf{M\_CVS-CJO}}]
\addplot [orange, mark=none, line width=1pt] table [x=horizon, y=M_CVS-CJO_mean] {\dataFeatureImportanceLinear};
\addplot [name path=lower_linear_feature_7, draw=none, forget plot] table [x=horizon, y=M_CVS-CJO_low_0.95] {\dataFeatureImportanceLinear};
\addplot [name path=upper_linear_feature_7, draw=none, forget plot] table [x=horizon, y=M_CVS-CJO_high_0.95] {\dataFeatureImportanceLinear};
\addplot [orange!40, fill opacity=0.75, forget plot] fill between [of=upper_linear_feature_7 and lower_linear_feature_7];
\addplot [spicy_paprika, mark=none, line width=1pt] table [x=horizon, y=M_CVS-CJO_mean] {\dataFeatureImportanceXGBoost};
\addplot [name path=lower_xgboost_feature_7, draw=none, forget plot] table [x=horizon, y=M_CVS-CJO_low_0.95] {\dataFeatureImportanceXGBoost};
\addplot [name path=upper_xgboost_feature_7, draw=none, forget plot] table [x=horizon, y=M_CVS-CJO_high_0.95] {\dataFeatureImportanceXGBoost};
\addplot [spicy_paprika!40, fill opacity=0.75, forget plot] fill between [of=upper_xgboost_feature_7 and lower_xgboost_feature_7];
\nextgroupplot[title={\textbf{N\_CVS-CJO}}]
\addplot [spicy_paprika, mark=none, line width=1pt] table [x=horizon, y=N_CVS-CJO_mean] {\dataFeatureImportanceXGBoost};
\addplot [name path=lower_xgboost_feature_8, draw=none, forget plot] table [x=horizon, y=N_CVS-CJO_low_0.95] {\dataFeatureImportanceXGBoost};
\addplot [name path=upper_xgboost_feature_8, draw=none, forget plot] table [x=horizon, y=N_CVS-CJO_high_0.95] {\dataFeatureImportanceXGBoost};
\addplot [spicy_paprika!40, fill opacity=0.75, forget plot] fill between [of=upper_xgboost_feature_8 and lower_xgboost_feature_8];
\nextgroupplot[ymax=1200, title={\textbf{PAC\_air/air}}]
\addplot [orange, mark=none, line width=1pt] table [x=horizon, y=PAC_air/air_mean] {\dataFeatureImportanceLinear};
\addplot [name path=lower_linear_feature_9, draw=none, forget plot] table [x=horizon, y=PAC_air/air_low_0.95] {\dataFeatureImportanceLinear};
\addplot [name path=upper_linear_feature_9, draw=none, forget plot] table [x=horizon, y=PAC_air/air_high_0.95] {\dataFeatureImportanceLinear};
\addplot [orange!40, fill opacity=0.75, forget plot] fill between [of=upper_linear_feature_9 and lower_linear_feature_9];
\addplot [spicy_paprika, mark=none, line width=1pt] table [x=horizon, y=PAC_air/air_mean] {\dataFeatureImportanceXGBoost};
\addplot [name path=lower_xgboost_feature_9, draw=none, forget plot] table [x=horizon, y=PAC_air/air_low_0.95] {\dataFeatureImportanceXGBoost};
\addplot [name path=upper_xgboost_feature_9, draw=none, forget plot] table [x=horizon, y=PAC_air/air_high_0.95] {\dataFeatureImportanceXGBoost};
\addplot [spicy_paprika!40, fill opacity=0.75, forget plot] fill between [of=upper_xgboost_feature_9 and lower_xgboost_feature_9];
\addplot [twilight_indigo, mark=none, line width=1pt] table [x=horizon, y=PAC_air/air_mean] {\dataFeatureImportanceTabICL};
\addplot [name path=lower_tabicl_feature_9, draw=none, forget plot] table [x=horizon, y=PAC_air/air_low_0.95] {\dataFeatureImportanceTabICL};
\addplot [name path=upper_tabicl_feature_9, draw=none, forget plot] table [x=horizon, y=PAC_air/air_high_0.95] {\dataFeatureImportanceTabICL};
\addplot [twilight_indigo!40, fill opacity=0.75, forget plot] fill between [of=upper_tabicl_feature_9 and lower_tabicl_feature_9];
\nextgroupplot[title={\textbf{PAC\_air/eau}}]
\addplot [spicy_paprika, mark=none, line width=1pt] table [x=horizon, y=PAC_air/eau_mean] {\dataFeatureImportanceXGBoost};
\addplot [name path=lower_xgboost_feature_10, draw=none, forget plot] table [x=horizon, y=PAC_air/eau_low_0.95] {\dataFeatureImportanceXGBoost};
\addplot [name path=upper_xgboost_feature_10, draw=none, forget plot] table [x=horizon, y=PAC_air/eau_high_0.95] {\dataFeatureImportanceXGBoost};
\addplot [spicy_paprika!40, fill opacity=0.75, forget plot] fill between [of=upper_xgboost_feature_10 and lower_xgboost_feature_10];
\nextgroupplot[title={\textbf{PAC\_geo}}]
\addplot [orange, mark=none, line width=1pt] table [x=horizon, y=PAC_geo_mean] {\dataFeatureImportanceLinear};
\addplot [name path=lower_linear_feature_11, draw=none, forget plot] table [x=horizon, y=PAC_geo_low_0.95] {\dataFeatureImportanceLinear};
\addplot [name path=upper_linear_feature_11, draw=none, forget plot] table [x=horizon, y=PAC_geo_high_0.95] {\dataFeatureImportanceLinear};
\addplot [orange!40, fill opacity=0.75, forget plot] fill between [of=upper_linear_feature_11 and lower_linear_feature_11];
\addplot [spicy_paprika, mark=none, line width=1pt] table [x=horizon, y=PAC_geo_mean] {\dataFeatureImportanceXGBoost};
\addplot [name path=lower_xgboost_feature_11, draw=none, forget plot] table [x=horizon, y=PAC_geo_low_0.95] {\dataFeatureImportanceXGBoost};
\addplot [name path=upper_xgboost_feature_11, draw=none, forget plot] table [x=horizon, y=PAC_geo_high_0.95] {\dataFeatureImportanceXGBoost};
\addplot [spicy_paprika!40, fill opacity=0.75, forget plot] fill between [of=upper_xgboost_feature_11 and lower_xgboost_feature_11];
\nextgroupplot[title={\textbf{S\_CVS-CJO}}]
\addplot [spicy_paprika, mark=none, line width=1pt] table [x=horizon, y=S_CVS-CJO_mean] {\dataFeatureImportanceXGBoost};
\addplot [name path=lower_xgboost_feature_12, draw=none, forget plot] table [x=horizon, y=S_CVS-CJO_low_0.95] {\dataFeatureImportanceXGBoost};
\addplot [name path=upper_xgboost_feature_12, draw=none, forget plot] table [x=horizon, y=S_CVS-CJO_high_0.95] {\dataFeatureImportanceXGBoost};
\addplot [spicy_paprika!40, fill opacity=0.75, forget plot] fill between [of=upper_xgboost_feature_12 and lower_xgboost_feature_12];
\nextgroupplot[title={\textbf{cdd}}]
\addplot [orange, mark=none, line width=1pt] table [x=horizon, y=cdd_mean] {\dataFeatureImportanceLinear};
\addplot [name path=lower_linear_feature_13, draw=none, forget plot] table [x=horizon, y=cdd_low_0.95] {\dataFeatureImportanceLinear};
\addplot [name path=upper_linear_feature_13, draw=none, forget plot] table [x=horizon, y=cdd_high_0.95] {\dataFeatureImportanceLinear};
\addplot [orange!40, fill opacity=0.75, forget plot] fill between [of=upper_linear_feature_13 and lower_linear_feature_13];
\addplot [spicy_paprika, mark=none, line width=1pt] table [x=horizon, y=cdd_mean] {\dataFeatureImportanceXGBoost};
\addplot [name path=lower_xgboost_feature_13, draw=none, forget plot] table [x=horizon, y=cdd_low_0.95] {\dataFeatureImportanceXGBoost};
\addplot [name path=upper_xgboost_feature_13, draw=none, forget plot] table [x=horizon, y=cdd_high_0.95] {\dataFeatureImportanceXGBoost};
\addplot [spicy_paprika!40, fill opacity=0.75, forget plot] fill between [of=upper_xgboost_feature_13 and lower_xgboost_feature_13];
\nextgroupplot[title={\textbf{cpi\_w/o\_energy}}]
\addplot [orange, mark=none, line width=1pt] table [x=horizon, y=cpi_w/o_energy_mean] {\dataFeatureImportanceLinear};
\addplot [name path=lower_linear_feature_14, draw=none, forget plot] table [x=horizon, y=cpi_w/o_energy_low_0.95] {\dataFeatureImportanceLinear};
\addplot [name path=upper_linear_feature_14, draw=none, forget plot] table [x=horizon, y=cpi_w/o_energy_high_0.95] {\dataFeatureImportanceLinear};
\addplot [orange!40, fill opacity=0.75, forget plot] fill between [of=upper_linear_feature_14 and lower_linear_feature_14];
\addplot [twilight_indigo, mark=none, line width=1pt] table [x=horizon, y=cpi_w/o_energy_mean] {\dataFeatureImportanceTabICL};
\addplot [name path=lower_tabicl_feature_14, draw=none, forget plot] table [x=horizon, y=cpi_w/o_energy_low_0.95] {\dataFeatureImportanceTabICL};
\addplot [name path=upper_tabicl_feature_14, draw=none, forget plot] table [x=horizon, y=cpi_w/o_energy_high_0.95] {\dataFeatureImportanceTabICL};
\addplot [twilight_indigo!40, fill opacity=0.75, forget plot] fill between [of=upper_tabicl_feature_14 and lower_tabicl_feature_14];
\nextgroupplot[title={\textbf{ecpi}}]
\addplot [spicy_paprika, mark=none, line width=1pt] table [x=horizon, y=ecpi_mean] {\dataFeatureImportanceXGBoost};
\addplot [name path=lower_xgboost_feature_15, draw=none, forget plot] table [x=horizon, y=ecpi_low_0.95] {\dataFeatureImportanceXGBoost};
\addplot [name path=upper_xgboost_feature_15, draw=none, forget plot] table [x=horizon, y=ecpi_high_0.95] {\dataFeatureImportanceXGBoost};
\addplot [spicy_paprika!40, fill opacity=0.75, forget plot] fill between [of=upper_xgboost_feature_15 and lower_xgboost_feature_15];
\nextgroupplot[title={\textbf{electric\_car\_passenger}}]
\addplot [spicy_paprika, mark=none, line width=1pt] table [x=horizon, y=electric_car_passenger_mean] {\dataFeatureImportanceXGBoost};
\addplot [name path=lower_xgboost_feature_16, draw=none, forget plot] table [x=horizon, y=electric_car_passenger_low_0.95] {\dataFeatureImportanceXGBoost};
\addplot [name path=upper_xgboost_feature_16, draw=none, forget plot] table [x=horizon, y=electric_car_passenger_high_0.95] {\dataFeatureImportanceXGBoost};
\addplot [spicy_paprika!40, fill opacity=0.75, forget plot] fill between [of=upper_xgboost_feature_16 and lower_xgboost_feature_16];
\nextgroupplot[title={\textbf{electric\_heavy\_utilitary\_vehicle}}]
\addplot [orange, mark=none, line width=1pt] table [x=horizon, y=electric_heavy_utilitary_vehicle_mean] {\dataFeatureImportanceLinear};
\addplot [name path=lower_linear_feature_17, draw=none, forget plot] table [x=horizon, y=electric_heavy_utilitary_vehicle_low_0.95] {\dataFeatureImportanceLinear};
\addplot [name path=upper_linear_feature_17, draw=none, forget plot] table [x=horizon, y=electric_heavy_utilitary_vehicle_high_0.95] {\dataFeatureImportanceLinear};
\addplot [orange!40, fill opacity=0.75, forget plot] fill between [of=upper_linear_feature_17 and lower_linear_feature_17];
\addplot [spicy_paprika, mark=none, line width=1pt] table [x=horizon, y=electric_heavy_utilitary_vehicle_mean] {\dataFeatureImportanceXGBoost};
\addplot [name path=lower_xgboost_feature_17, draw=none, forget plot] table [x=horizon, y=electric_heavy_utilitary_vehicle_low_0.95] {\dataFeatureImportanceXGBoost};
\addplot [name path=upper_xgboost_feature_17, draw=none, forget plot] table [x=horizon, y=electric_heavy_utilitary_vehicle_high_0.95] {\dataFeatureImportanceXGBoost};
\addplot [spicy_paprika!40, fill opacity=0.75, forget plot] fill between [of=upper_xgboost_feature_17 and lower_xgboost_feature_17];
\nextgroupplot[title={\textbf{electric\_public\_transports}}]
\addplot [orange, mark=none, line width=1pt] table [x=horizon, y=electric_public_transports_mean] {\dataFeatureImportanceLinear};
\addplot [name path=lower_linear_feature_18, draw=none, forget plot] table [x=horizon, y=electric_public_transports_low_0.95] {\dataFeatureImportanceLinear};
\addplot [name path=upper_linear_feature_18, draw=none, forget plot] table [x=horizon, y=electric_public_transports_high_0.95] {\dataFeatureImportanceLinear};
\addplot [orange!40, fill opacity=0.75, forget plot] fill between [of=upper_linear_feature_18 and lower_linear_feature_18];
\addplot [spicy_paprika, mark=none, line width=1pt] table [x=horizon, y=electric_public_transports_mean] {\dataFeatureImportanceXGBoost};
\addplot [name path=lower_xgboost_feature_18, draw=none, forget plot] table [x=horizon, y=electric_public_transports_low_0.95] {\dataFeatureImportanceXGBoost};
\addplot [name path=upper_xgboost_feature_18, draw=none, forget plot] table [x=horizon, y=electric_public_transports_high_0.95] {\dataFeatureImportanceXGBoost};
\addplot [spicy_paprika!40, fill opacity=0.75, forget plot] fill between [of=upper_xgboost_feature_18 and lower_xgboost_feature_18];
\addplot [twilight_indigo, mark=none, line width=1pt] table [x=horizon, y=electric_public_transports_mean] {\dataFeatureImportanceTabICL};
\addplot [name path=lower_tabicl_feature_18, draw=none, forget plot] table [x=horizon, y=electric_public_transports_low_0.95] {\dataFeatureImportanceTabICL};
\addplot [name path=upper_tabicl_feature_18, draw=none, forget plot] table [x=horizon, y=electric_public_transports_high_0.95] {\dataFeatureImportanceTabICL};
\addplot [twilight_indigo!40, fill opacity=0.75, forget plot] fill between [of=upper_tabicl_feature_18 and lower_tabicl_feature_18];
\nextgroupplot[title={\textbf{employment}}]
\addplot [spicy_paprika, mark=none, line width=1pt] table [x=horizon, y=employment_mean] {\dataFeatureImportanceXGBoost};
\addplot [name path=lower_xgboost_feature_19, draw=none, forget plot] table [x=horizon, y=employment_low_0.95] {\dataFeatureImportanceXGBoost};
\addplot [name path=upper_xgboost_feature_19, draw=none, forget plot] table [x=horizon, y=employment_high_0.95] {\dataFeatureImportanceXGBoost};
\addplot [spicy_paprika!40, fill opacity=0.75, forget plot] fill between [of=upper_xgboost_feature_19 and lower_xgboost_feature_19];
\nextgroupplot[title={\textbf{hdd}}]
\addplot [orange, mark=none, line width=1pt] table [x=horizon, y=hdd_mean] {\dataFeatureImportanceLinear};
\addplot [name path=lower_linear_feature_20, draw=none, forget plot] table [x=horizon, y=hdd_low_0.95] {\dataFeatureImportanceLinear};
\addplot [name path=upper_linear_feature_20, draw=none, forget plot] table [x=horizon, y=hdd_high_0.95] {\dataFeatureImportanceLinear};
\addplot [orange!40, fill opacity=0.75, forget plot] fill between [of=upper_linear_feature_20 and lower_linear_feature_20];
\nextgroupplot[title={\textbf{household\_income\_CVS-CJO}}]
\addplot [spicy_paprika, mark=none, line width=1pt] table [x=horizon, y=household_income_CVS-CJO_mean] {\dataFeatureImportanceXGBoost};
\addplot [name path=lower_xgboost_feature_21, draw=none, forget plot] table [x=horizon, y=household_income_CVS-CJO_low_0.95] {\dataFeatureImportanceXGBoost};
\addplot [name path=upper_xgboost_feature_21, draw=none, forget plot] table [x=horizon, y=household_income_CVS-CJO_high_0.95] {\dataFeatureImportanceXGBoost};
\addplot [spicy_paprika!40, fill opacity=0.75, forget plot] fill between [of=upper_xgboost_feature_21 and lower_xgboost_feature_21];
\nextgroupplot[title={\textbf{hybrid\_heavy\_utilitary\_vehicle}}]
\addplot [orange, mark=none, line width=1pt] table [x=horizon, y=hybrid_heavy_utilitary_vehicle_mean] {\dataFeatureImportanceLinear};
\addplot [name path=lower_linear_feature_22, draw=none, forget plot] table [x=horizon, y=hybrid_heavy_utilitary_vehicle_low_0.95] {\dataFeatureImportanceLinear};
\addplot [name path=upper_linear_feature_22, draw=none, forget plot] table [x=horizon, y=hybrid_heavy_utilitary_vehicle_high_0.95] {\dataFeatureImportanceLinear};
\addplot [orange!40, fill opacity=0.75, forget plot] fill between [of=upper_linear_feature_22 and lower_linear_feature_22];
\nextgroupplot[title={\textbf{hybrid\_public\_transports}}]
\addplot [orange, mark=none, line width=1pt] table [x=horizon, y=hybrid_public_transports_mean] {\dataFeatureImportanceLinear};
\addplot [name path=lower_linear_feature_23, draw=none, forget plot] table [x=horizon, y=hybrid_public_transports_low_0.95] {\dataFeatureImportanceLinear};
\addplot [name path=upper_linear_feature_23, draw=none, forget plot] table [x=horizon, y=hybrid_public_transports_high_0.95] {\dataFeatureImportanceLinear};
\addplot [orange!40, fill opacity=0.75, forget plot] fill between [of=upper_linear_feature_23 and lower_linear_feature_23];
\addplot [spicy_paprika, mark=none, line width=1pt] table [x=horizon, y=hybrid_public_transports_mean] {\dataFeatureImportanceXGBoost};
\addplot [name path=lower_xgboost_feature_23, draw=none, forget plot] table [x=horizon, y=hybrid_public_transports_low_0.95] {\dataFeatureImportanceXGBoost};
\addplot [name path=upper_xgboost_feature_23, draw=none, forget plot] table [x=horizon, y=hybrid_public_transports_high_0.95] {\dataFeatureImportanceXGBoost};
\addplot [spicy_paprika!40, fill opacity=0.75, forget plot] fill between [of=upper_xgboost_feature_23 and lower_xgboost_feature_23];

\end{groupplot}
\end{tikzpicture}


%% file: Plots/feature_importance_part2.tex




\pgfplotstableread[col sep=comma]{\getFeaturesImportanceDataFilePath{month}{linear}{feature_importance}{econ}}\dataFeatureImportanceLinear
\pgfplotstableread[col sep=comma]{\getFeaturesImportanceDataFilePath{month}{xgboost}{feature_importance}{econ}}\dataFeatureImportanceXGBoost
\pgfplotstableread[col sep=comma]{\getFeaturesImportanceDataFilePath{month}{tabICL}{feature_importance}{econ}}\dataFeatureImportanceTabICL

\begin{tikzpicture}
\begin{groupplot}[group style={group size=4 by 7, horizontal sep=1.5cm, vertical sep=1.75cm}, width=8cm, height=6cm, xlabel={Horizon [M]}, ylabel={Feature Importance}, legend columns=1,
    legend style={at={(2.75,-0.75), font=\Large}, anchor=north west}, /pgfplots/scale ticks below exponent={-3},
/pgfplots/scale ticks above exponent={2}]
\nextgroupplot[title={\textbf{is\_public\_holiday}}]
\addplot [orange, mark=none, line width=1pt] table [x=horizon, y=is_public_holiday_mean] {\dataFeatureImportanceLinear};
\addplot [name path=lower_linear_feature_24, draw=none, forget plot] table [x=horizon, y=is_public_holiday_low_0.95] {\dataFeatureImportanceLinear};
\addplot [name path=upper_linear_feature_24, draw=none, forget plot] table [x=horizon, y=is_public_holiday_high_0.95] {\dataFeatureImportanceLinear};
\addplot [orange!40, fill opacity=0.75, forget plot] fill between [of=upper_linear_feature_24 and lower_linear_feature_24];
\addplot [spicy_paprika, mark=none, line width=1pt] table [x=horizon, y=is_public_holiday_mean] {\dataFeatureImportanceXGBoost};
\addplot [name path=lower_xgboost_feature_24, draw=none, forget plot] table [x=horizon, y=is_public_holiday_low_0.95] {\dataFeatureImportanceXGBoost};
\addplot [name path=upper_xgboost_feature_24, draw=none, forget plot] table [x=horizon, y=is_public_holiday_high_0.95] {\dataFeatureImportanceXGBoost};
\addplot [spicy_paprika!40, fill opacity=0.75, forget plot] fill between [of=upper_xgboost_feature_24 and lower_xgboost_feature_24];
\addplot [twilight_indigo, mark=none, line width=1pt] table [x=horizon, y=is_public_holiday_mean] {\dataFeatureImportanceTabICL};
\addplot [name path=lower_tabicl_feature_24, draw=none, forget plot] table [x=horizon, y=is_public_holiday_low_0.95] {\dataFeatureImportanceTabICL};
\addplot [name path=upper_tabicl_feature_24, draw=none, forget plot] table [x=horizon, y=is_public_holiday_high_0.95] {\dataFeatureImportanceTabICL};
\addplot [twilight_indigo!40, fill opacity=0.75, forget plot] fill between [of=upper_tabicl_feature_24 and lower_tabicl_feature_24];
\nextgroupplot[title={\textbf{is\_school\_holiday}}]
\addplot [orange, mark=none, line width=1pt] table [x=horizon, y=is_school_holiday_mean] {\dataFeatureImportanceLinear};
\addplot [name path=lower_linear_feature_25, draw=none, forget plot] table [x=horizon, y=is_school_holiday_low_0.95] {\dataFeatureImportanceLinear};
\addplot [name path=upper_linear_feature_25, draw=none, forget plot] table [x=horizon, y=is_school_holiday_high_0.95] {\dataFeatureImportanceLinear};
\addplot [orange!40, fill opacity=0.75, forget plot] fill between [of=upper_linear_feature_25 and lower_linear_feature_25];
\nextgroupplot[title={\textbf{is\_weekend}}]
\addplot [orange, mark=none, line width=1pt] table [x=horizon, y=is_weekend_mean] {\dataFeatureImportanceLinear};
\addplot [name path=lower_linear_feature_26, draw=none, forget plot] table [x=horizon, y=is_weekend_low_0.95] {\dataFeatureImportanceLinear};
\addplot [name path=upper_linear_feature_26, draw=none, forget plot] table [x=horizon, y=is_weekend_high_0.95] {\dataFeatureImportanceLinear};
\addplot [orange!40, fill opacity=0.75, forget plot] fill between [of=upper_linear_feature_26 and lower_linear_feature_26];
\addplot [spicy_paprika, mark=none, line width=1pt] table [x=horizon, y=is_weekend_mean] {\dataFeatureImportanceXGBoost};
\addplot [name path=lower_xgboost_feature_26, draw=none, forget plot] table [x=horizon, y=is_weekend_low_0.95] {\dataFeatureImportanceXGBoost};
\addplot [name path=upper_xgboost_feature_26, draw=none, forget plot] table [x=horizon, y=is_weekend_high_0.95] {\dataFeatureImportanceXGBoost};
\addplot [spicy_paprika!40, fill opacity=0.75, forget plot] fill between [of=upper_xgboost_feature_26 and lower_xgboost_feature_26];
\addplot [twilight_indigo, mark=none, line width=1pt] table [x=horizon, y=is_weekend_mean] {\dataFeatureImportanceTabICL};
\addplot [name path=lower_tabicl_feature_26, draw=none, forget plot] table [x=horizon, y=is_weekend_low_0.95] {\dataFeatureImportanceTabICL};
\addplot [name path=upper_tabicl_feature_26, draw=none, forget plot] table [x=horizon, y=is_weekend_high_0.95] {\dataFeatureImportanceTabICL};
\addplot [twilight_indigo!40, fill opacity=0.75, forget plot] fill between [of=upper_tabicl_feature_26 and lower_tabicl_feature_26];
\nextgroupplot[title={\textbf{moy\_cos}}]
\addplot [orange, mark=none, line width=1pt] table [x=horizon, y=moy_cos_mean] {\dataFeatureImportanceLinear};
\addplot [name path=lower_linear_feature_27, draw=none, forget plot] table [x=horizon, y=moy_cos_low_0.95] {\dataFeatureImportanceLinear};
\addplot [name path=upper_linear_feature_27, draw=none, forget plot] table [x=horizon, y=moy_cos_high_0.95] {\dataFeatureImportanceLinear};
\addplot [orange!40, fill opacity=0.75, forget plot] fill between [of=upper_linear_feature_27 and lower_linear_feature_27];
\addplot [spicy_paprika, mark=none, line width=1pt] table [x=horizon, y=moy_cos_mean] {\dataFeatureImportanceXGBoost};
\addplot [name path=lower_xgboost_feature_27, draw=none, forget plot] table [x=horizon, y=moy_cos_low_0.95] {\dataFeatureImportanceXGBoost};
\addplot [name path=upper_xgboost_feature_27, draw=none, forget plot] table [x=horizon, y=moy_cos_high_0.95] {\dataFeatureImportanceXGBoost};
\addplot [spicy_paprika!40, fill opacity=0.75, forget plot] fill between [of=upper_xgboost_feature_27 and lower_xgboost_feature_27];
\addplot [twilight_indigo, mark=none, line width=1pt] table [x=horizon, y=moy_cos_mean] {\dataFeatureImportanceTabICL};
\addplot [name path=lower_tabicl_feature_27, draw=none, forget plot] table [x=horizon, y=moy_cos_low_0.95] {\dataFeatureImportanceTabICL};
\addplot [name path=upper_tabicl_feature_27, draw=none, forget plot] table [x=horizon, y=moy_cos_high_0.95] {\dataFeatureImportanceTabICL};
\addplot [twilight_indigo!40, fill opacity=0.75, forget plot] fill between [of=upper_tabicl_feature_27 and lower_tabicl_feature_27];
\nextgroupplot[title={\textbf{moy\_sin}}]
\addplot [orange, mark=none, line width=1pt] table [x=horizon, y=moy_sin_mean] {\dataFeatureImportanceLinear};
\addplot [name path=lower_linear_feature_28, draw=none, forget plot] table [x=horizon, y=moy_sin_low_0.95] {\dataFeatureImportanceLinear};
\addplot [name path=upper_linear_feature_28, draw=none, forget plot] table [x=horizon, y=moy_sin_high_0.95] {\dataFeatureImportanceLinear};
\addplot [orange!40, fill opacity=0.75, forget plot] fill between [of=upper_linear_feature_28 and lower_linear_feature_28];
\addplot [spicy_paprika, mark=none, line width=1pt] table [x=horizon, y=moy_sin_mean] {\dataFeatureImportanceXGBoost};
\addplot [name path=lower_xgboost_feature_28, draw=none, forget plot] table [x=horizon, y=moy_sin_low_0.95] {\dataFeatureImportanceXGBoost};
\addplot [name path=upper_xgboost_feature_28, draw=none, forget plot] table [x=horizon, y=moy_sin_high_0.95] {\dataFeatureImportanceXGBoost};
\addplot [spicy_paprika!40, fill opacity=0.75, forget plot] fill between [of=upper_xgboost_feature_28 and lower_xgboost_feature_28];
\addplot [twilight_indigo, mark=none, line width=1pt] table [x=horizon, y=moy_sin_mean] {\dataFeatureImportanceTabICL};
\addplot [name path=lower_tabicl_feature_28, draw=none, forget plot] table [x=horizon, y=moy_sin_low_0.95] {\dataFeatureImportanceTabICL};
\addplot [name path=upper_tabicl_feature_28, draw=none, forget plot] table [x=horizon, y=moy_sin_high_0.95] {\dataFeatureImportanceTabICL};
\addplot [twilight_indigo!40, fill opacity=0.75, forget plot] fill between [of=upper_tabicl_feature_28 and lower_tabicl_feature_28];
\nextgroupplot[title={\textbf{population}}]
\addplot [spicy_paprika, mark=none, line width=1pt] table [x=horizon, y=population_mean] {\dataFeatureImportanceXGBoost};
\addplot [name path=lower_xgboost_feature_29, draw=none, forget plot] table [x=horizon, y=population_low_0.95] {\dataFeatureImportanceXGBoost};
\addplot [name path=upper_xgboost_feature_29, draw=none, forget plot] table [x=horizon, y=population_high_0.95] {\dataFeatureImportanceXGBoost};
\addplot [spicy_paprika!40, fill opacity=0.75, forget plot] fill between [of=upper_xgboost_feature_29 and lower_xgboost_feature_29];
\addplot [twilight_indigo, mark=none, line width=1pt] table [x=horizon, y=population_mean] {\dataFeatureImportanceTabICL};
\addplot [name path=lower_tabicl_feature_29, draw=none, forget plot] table [x=horizon, y=population_low_0.95] {\dataFeatureImportanceTabICL};
\addplot [name path=upper_tabicl_feature_29, draw=none, forget plot] table [x=horizon, y=population_high_0.95] {\dataFeatureImportanceTabICL};
\addplot [twilight_indigo!40, fill opacity=0.75, forget plot] fill between [of=upper_tabicl_feature_29 and lower_tabicl_feature_29];
\nextgroupplot[title={\textbf{qoy\_cos}}]
\addplot [orange, mark=none, line width=1pt] table [x=horizon, y=qoy_cos_mean] {\dataFeatureImportanceLinear};
\addplot [name path=lower_linear_feature_30, draw=none, forget plot] table [x=horizon, y=qoy_cos_low_0.95] {\dataFeatureImportanceLinear};
\addplot [name path=upper_linear_feature_30, draw=none, forget plot] table [x=horizon, y=qoy_cos_high_0.95] {\dataFeatureImportanceLinear};
\addplot [orange!40, fill opacity=0.75, forget plot] fill between [of=upper_linear_feature_30 and lower_linear_feature_30];
\addplot [spicy_paprika, mark=none, line width=1pt] table [x=horizon, y=qoy_cos_mean] {\dataFeatureImportanceXGBoost};
\addplot [name path=lower_xgboost_feature_30, draw=none, forget plot] table [x=horizon, y=qoy_cos_low_0.95] {\dataFeatureImportanceXGBoost};
\addplot [name path=upper_xgboost_feature_30, draw=none, forget plot] table [x=horizon, y=qoy_cos_high_0.95] {\dataFeatureImportanceXGBoost};
\addplot [spicy_paprika!40, fill opacity=0.75, forget plot] fill between [of=upper_xgboost_feature_30 and lower_xgboost_feature_30];
\addplot [twilight_indigo, mark=none, line width=1pt] table [x=horizon, y=qoy_cos_mean] {\dataFeatureImportanceTabICL};
\addplot [name path=lower_tabicl_feature_30, draw=none, forget plot] table [x=horizon, y=qoy_cos_low_0.95] {\dataFeatureImportanceTabICL};
\addplot [name path=upper_tabicl_feature_30, draw=none, forget plot] table [x=horizon, y=qoy_cos_high_0.95] {\dataFeatureImportanceTabICL};
\addplot [twilight_indigo!40, fill opacity=0.75, forget plot] fill between [of=upper_tabicl_feature_30 and lower_tabicl_feature_30];
\nextgroupplot[title={\textbf{qoy\_sin}}]
\addplot [orange, mark=none, line width=1pt] table [x=horizon, y=qoy_sin_mean] {\dataFeatureImportanceLinear};
\addplot [name path=lower_linear_feature_31, draw=none, forget plot] table [x=horizon, y=qoy_sin_low_0.95] {\dataFeatureImportanceLinear};
\addplot [name path=upper_linear_feature_31, draw=none, forget plot] table [x=horizon, y=qoy_sin_high_0.95] {\dataFeatureImportanceLinear};
\addplot [orange!40, fill opacity=0.75, forget plot] fill between [of=upper_linear_feature_31 and lower_linear_feature_31];
\addplot [spicy_paprika, mark=none, line width=1pt] table [x=horizon, y=qoy_sin_mean] {\dataFeatureImportanceXGBoost};
\addplot [name path=lower_xgboost_feature_31, draw=none, forget plot] table [x=horizon, y=qoy_sin_low_0.95] {\dataFeatureImportanceXGBoost};
\addplot [name path=upper_xgboost_feature_31, draw=none, forget plot] table [x=horizon, y=qoy_sin_high_0.95] {\dataFeatureImportanceXGBoost};
\addplot [spicy_paprika!40, fill opacity=0.75, forget plot] fill between [of=upper_xgboost_feature_31 and lower_xgboost_feature_31];
\addplot [twilight_indigo, mark=none, line width=1pt] table [x=horizon, y=qoy_sin_mean] {\dataFeatureImportanceTabICL};
\addplot [name path=lower_tabicl_feature_31, draw=none, forget plot] table [x=horizon, y=qoy_sin_low_0.95] {\dataFeatureImportanceTabICL};
\addplot [name path=upper_tabicl_feature_31, draw=none, forget plot] table [x=horizon, y=qoy_sin_high_0.95] {\dataFeatureImportanceTabICL};
\addplot [twilight_indigo!40, fill opacity=0.75, forget plot] fill between [of=upper_tabicl_feature_31 and lower_tabicl_feature_31];
\nextgroupplot[title={\textbf{rh}}]
\addplot [orange, mark=none, line width=1pt] table [x=horizon, y=rh_mean] {\dataFeatureImportanceLinear};
\addplot [name path=lower_linear_feature_32, draw=none, forget plot] table [x=horizon, y=rh_low_0.95] {\dataFeatureImportanceLinear};
\addplot [name path=upper_linear_feature_32, draw=none, forget plot] table [x=horizon, y=rh_high_0.95] {\dataFeatureImportanceLinear};
\addplot [orange!40, fill opacity=0.75, forget plot] fill between [of=upper_linear_feature_32 and lower_linear_feature_32];
\addplot [spicy_paprika, mark=none, line width=1pt] table [x=horizon, y=rh_mean] {\dataFeatureImportanceXGBoost};
\addplot [name path=lower_xgboost_feature_32, draw=none, forget plot] table [x=horizon, y=rh_low_0.95] {\dataFeatureImportanceXGBoost};
\addplot [name path=upper_xgboost_feature_32, draw=none, forget plot] table [x=horizon, y=rh_high_0.95] {\dataFeatureImportanceXGBoost};
\addplot [spicy_paprika!40, fill opacity=0.75, forget plot] fill between [of=upper_xgboost_feature_32 and lower_xgboost_feature_32];
\addplot [twilight_indigo, mark=none, line width=1pt] table [x=horizon, y=rh_mean] {\dataFeatureImportanceTabICL};
\addplot [name path=lower_tabicl_feature_32, draw=none, forget plot] table [x=horizon, y=rh_low_0.95] {\dataFeatureImportanceTabICL};
\addplot [name path=upper_tabicl_feature_32, draw=none, forget plot] table [x=horizon, y=rh_high_0.95] {\dataFeatureImportanceTabICL};
\addplot [twilight_indigo!40, fill opacity=0.75, forget plot] fill between [of=upper_tabicl_feature_32 and lower_tabicl_feature_32];
\nextgroupplot[title={\textbf{scpi}}]
\addplot [orange, mark=none, line width=1pt] table [x=horizon, y=scpi_mean] {\dataFeatureImportanceLinear};
\addplot [name path=lower_linear_feature_33, draw=none, forget plot] table [x=horizon, y=scpi_low_0.95] {\dataFeatureImportanceLinear};
\addplot [name path=upper_linear_feature_33, draw=none, forget plot] table [x=horizon, y=scpi_high_0.95] {\dataFeatureImportanceLinear};
\addplot [orange!40, fill opacity=0.75, forget plot] fill between [of=upper_linear_feature_33 and lower_linear_feature_33];
\nextgroupplot[title={\textbf{seas\_cos}}]
\addplot [orange, mark=none, line width=1pt] table [x=horizon, y=seas_cos_mean] {\dataFeatureImportanceLinear};
\addplot [name path=lower_linear_feature_34, draw=none, forget plot] table [x=horizon, y=seas_cos_low_0.95] {\dataFeatureImportanceLinear};
\addplot [name path=upper_linear_feature_34, draw=none, forget plot] table [x=horizon, y=seas_cos_high_0.95] {\dataFeatureImportanceLinear};
\addplot [orange!40, fill opacity=0.75, forget plot] fill between [of=upper_linear_feature_34 and lower_linear_feature_34];
\addplot [spicy_paprika, mark=none, line width=1pt] table [x=horizon, y=seas_cos_mean] {\dataFeatureImportanceXGBoost};
\addplot [name path=lower_xgboost_feature_34, draw=none, forget plot] table [x=horizon, y=seas_cos_low_0.95] {\dataFeatureImportanceXGBoost};
\addplot [name path=upper_xgboost_feature_34, draw=none, forget plot] table [x=horizon, y=seas_cos_high_0.95] {\dataFeatureImportanceXGBoost};
\addplot [spicy_paprika!40, fill opacity=0.75, forget plot] fill between [of=upper_xgboost_feature_34 and lower_xgboost_feature_34];
\nextgroupplot[title={\textbf{seas\_sin}}]
\addplot [orange, mark=none, line width=1pt] table [x=horizon, y=seas_sin_mean] {\dataFeatureImportanceLinear};
\addplot [name path=lower_linear_feature_35, draw=none, forget plot] table [x=horizon, y=seas_sin_low_0.95] {\dataFeatureImportanceLinear};
\addplot [name path=upper_linear_feature_35, draw=none, forget plot] table [x=horizon, y=seas_sin_high_0.95] {\dataFeatureImportanceLinear};
\addplot [orange!40, fill opacity=0.75, forget plot] fill between [of=upper_linear_feature_35 and lower_linear_feature_35];
\addplot [spicy_paprika, mark=none, line width=1pt] table [x=horizon, y=seas_sin_mean] {\dataFeatureImportanceXGBoost};
\addplot [name path=lower_xgboost_feature_35, draw=none, forget plot] table [x=horizon, y=seas_sin_low_0.95] {\dataFeatureImportanceXGBoost};
\addplot [name path=upper_xgboost_feature_35, draw=none, forget plot] table [x=horizon, y=seas_sin_high_0.95] {\dataFeatureImportanceXGBoost};
\addplot [spicy_paprika!40, fill opacity=0.75, forget plot] fill between [of=upper_xgboost_feature_35 and lower_xgboost_feature_35];
\nextgroupplot[title={\textbf{ssrd}}]
\addplot [orange, mark=none, line width=1pt] table [x=horizon, y=ssrd_mean] {\dataFeatureImportanceLinear};
\addplot [name path=lower_linear_feature_36, draw=none, forget plot] table [x=horizon, y=ssrd_low_0.95] {\dataFeatureImportanceLinear};
\addplot [name path=upper_linear_feature_36, draw=none, forget plot] table [x=horizon, y=ssrd_high_0.95] {\dataFeatureImportanceLinear};
\addplot [orange!40, fill opacity=0.75, forget plot] fill between [of=upper_linear_feature_36 and lower_linear_feature_36];
\addplot [spicy_paprika, mark=none, line width=1pt] table [x=horizon, y=ssrd_mean] {\dataFeatureImportanceXGBoost};
\addplot [name path=lower_xgboost_feature_36, draw=none, forget plot] table [x=horizon, y=ssrd_low_0.95] {\dataFeatureImportanceXGBoost};
\addplot [name path=upper_xgboost_feature_36, draw=none, forget plot] table [x=horizon, y=ssrd_high_0.95] {\dataFeatureImportanceXGBoost};
\addplot [spicy_paprika!40, fill opacity=0.75, forget plot] fill between [of=upper_xgboost_feature_36 and lower_xgboost_feature_36];
\nextgroupplot[title={\textbf{t2m}}]
\addplot [orange, mark=none, line width=1pt] table [x=horizon, y=t2m_mean] {\dataFeatureImportanceLinear};
\addlegendentry{Linear};
\addplot [name path=lower_linear_feature_37, draw=none, forget plot] table [x=horizon, y=t2m_low_0.95] {\dataFeatureImportanceLinear};
\addplot [name path=upper_linear_feature_37, draw=none, forget plot] table [x=horizon, y=t2m_high_0.95] {\dataFeatureImportanceLinear};
\addplot [orange!40, fill opacity=0.75, forget plot] fill between [of=upper_linear_feature_37 and lower_linear_feature_37];
\addplot [spicy_paprika, mark=none, line width=1pt] table [x=horizon, y=t2m_mean] {\dataFeatureImportanceXGBoost};
\addlegendentry{XGBoost};
\addplot [name path=lower_xgboost_feature_37, draw=none, forget plot] table [x=horizon, y=t2m_low_0.95] {\dataFeatureImportanceXGBoost};
\addplot [name path=upper_xgboost_feature_37, draw=none, forget plot] table [x=horizon, y=t2m_high_0.95] {\dataFeatureImportanceXGBoost};
\addplot [spicy_paprika!40, fill opacity=0.75, forget plot] fill between [of=upper_xgboost_feature_37 and lower_xgboost_feature_37];
\addplot [twilight_indigo, mark=none, line width=1pt] table [x=horizon, y=t2m_mean] {\dataFeatureImportanceTabICL};
\addlegendentry{TabICL};
\addplot [name path=lower_tabicl_feature_37, draw=none, forget plot] table [x=horizon, y=t2m_low_0.95] {\dataFeatureImportanceTabICL};
\addplot [name path=upper_tabicl_feature_37, draw=none, forget plot] table [x=horizon, y=t2m_high_0.95] {\dataFeatureImportanceTabICL};
\addplot [twilight_indigo!40, fill opacity=0.75, forget plot] fill between [of=upper_tabicl_feature_37 and lower_tabicl_feature_37];
\nextgroupplot[title={\textbf{tcc}}]
\addplot [orange, mark=none, line width=1pt] table [x=horizon, y=tcc_mean] {\dataFeatureImportanceLinear};
\addplot [name path=lower_linear_feature_38, draw=none, forget plot] table [x=horizon, y=tcc_low_0.95] {\dataFeatureImportanceLinear};
\addplot [name path=upper_linear_feature_38, draw=none, forget plot] table [x=horizon, y=tcc_high_0.95] {\dataFeatureImportanceLinear};
\addplot [orange!40, fill opacity=0.75, forget plot] fill between [of=upper_linear_feature_38 and lower_linear_feature_38];
\addplot [spicy_paprika, mark=none, line width=1pt] table [x=horizon, y=tcc_mean] {\dataFeatureImportanceXGBoost};
\addplot [name path=lower_xgboost_feature_38, draw=none, forget plot] table [x=horizon, y=tcc_low_0.95] {\dataFeatureImportanceXGBoost};
\addplot [name path=upper_xgboost_feature_38, draw=none, forget plot] table [x=horizon, y=tcc_high_0.95] {\dataFeatureImportanceXGBoost};
\addplot [spicy_paprika!40, fill opacity=0.75, forget plot] fill between [of=upper_xgboost_feature_38 and lower_xgboost_feature_38];
\addplot [twilight_indigo, mark=none, line width=1pt] table [x=horizon, y=tcc_mean] {\dataFeatureImportanceTabICL};
\addplot [name path=lower_tabicl_feature_38, draw=none, forget plot] table [x=horizon, y=tcc_low_0.95] {\dataFeatureImportanceTabICL};
\addplot [name path=upper_tabicl_feature_38, draw=none, forget plot] table [x=horizon, y=tcc_high_0.95] {\dataFeatureImportanceTabICL};
\addplot [twilight_indigo!40, fill opacity=0.75, forget plot] fill between [of=upper_tabicl_feature_38 and lower_tabicl_feature_38];
\nextgroupplot[title={\textbf{tourism\_nights}}]
\addplot [orange, mark=none, line width=1pt] table [x=horizon, y=tourism_nights_mean] {\dataFeatureImportanceLinear};
\addplot [name path=lower_linear_feature_39, draw=none, forget plot] table [x=horizon, y=tourism_nights_low_0.95] {\dataFeatureImportanceLinear};
\addplot [name path=upper_linear_feature_39, draw=none, forget plot] table [x=horizon, y=tourism_nights_high_0.95] {\dataFeatureImportanceLinear};
\addplot [orange!40, fill opacity=0.75, forget plot] fill between [of=upper_linear_feature_39 and lower_linear_feature_39];
\addplot [spicy_paprika, mark=none, line width=1pt] table [x=horizon, y=tourism_nights_mean] {\dataFeatureImportanceXGBoost};
\addplot [name path=lower_xgboost_feature_39, draw=none, forget plot] table [x=horizon, y=tourism_nights_low_0.95] {\dataFeatureImportanceXGBoost};
\addplot [name path=upper_xgboost_feature_39, draw=none, forget plot] table [x=horizon, y=tourism_nights_high_0.95] {\dataFeatureImportanceXGBoost};
\addplot [spicy_paprika!40, fill opacity=0.75, forget plot] fill between [of=upper_xgboost_feature_39 and lower_xgboost_feature_39];
\addplot [twilight_indigo, mark=none, line width=1pt] table [x=horizon, y=tourism_nights_mean] {\dataFeatureImportanceTabICL};
\addplot [name path=lower_tabicl_feature_39, draw=none, forget plot] table [x=horizon, y=tourism_nights_low_0.95] {\dataFeatureImportanceTabICL};
\addplot [name path=upper_tabicl_feature_39, draw=none, forget plot] table [x=horizon, y=tourism_nights_high_0.95] {\dataFeatureImportanceTabICL};
\addplot [twilight_indigo!40, fill opacity=0.75, forget plot] fill between [of=upper_tabicl_feature_39 and lower_tabicl_feature_39];
\nextgroupplot[title={\textbf{tp}}]
\addplot [orange, mark=none, line width=1pt] table [x=horizon, y=tp_mean] {\dataFeatureImportanceLinear};
\addplot [name path=lower_linear_feature_40, draw=none, forget plot] table [x=horizon, y=tp_low_0.95] {\dataFeatureImportanceLinear};
\addplot [name path=upper_linear_feature_40, draw=none, forget plot] table [x=horizon, y=tp_high_0.95] {\dataFeatureImportanceLinear};
\addplot [orange!40, fill opacity=0.75, forget plot] fill between [of=upper_linear_feature_40 and lower_linear_feature_40];
\addplot [spicy_paprika, mark=none, line width=1pt] table [x=horizon, y=tp_mean] {\dataFeatureImportanceXGBoost};
\addplot [name path=lower_xgboost_feature_40, draw=none, forget plot] table [x=horizon, y=tp_low_0.95] {\dataFeatureImportanceXGBoost};
\addplot [name path=upper_xgboost_feature_40, draw=none, forget plot] table [x=horizon, y=tp_high_0.95] {\dataFeatureImportanceXGBoost};
\addplot [spicy_paprika!40, fill opacity=0.75, forget plot] fill between [of=upper_xgboost_feature_40 and lower_xgboost_feature_40];
\addplot [twilight_indigo, mark=none, line width=1pt] table [x=horizon, y=tp_mean] {\dataFeatureImportanceTabICL};
\addplot [name path=lower_tabicl_feature_40, draw=none, forget plot] table [x=horizon, y=tp_low_0.95] {\dataFeatureImportanceTabICL};
\addplot [name path=upper_tabicl_feature_40, draw=none, forget plot] table [x=horizon, y=tp_high_0.95] {\dataFeatureImportanceTabICL};
\addplot [twilight_indigo!40, fill opacity=0.75, forget plot] fill between [of=upper_tabicl_feature_40 and lower_tabicl_feature_40];
\nextgroupplot[title={\textbf{u10}}]
\addplot [orange, mark=none, line width=1pt] table [x=horizon, y=u10_mean] {\dataFeatureImportanceLinear};
\addplot [name path=lower_linear_feature_41, draw=none, forget plot] table [x=horizon, y=u10_low_0.95] {\dataFeatureImportanceLinear};
\addplot [name path=upper_linear_feature_41, draw=none, forget plot] table [x=horizon, y=u10_high_0.95] {\dataFeatureImportanceLinear};
\addplot [orange!40, fill opacity=0.75, forget plot] fill between [of=upper_linear_feature_41 and lower_linear_feature_41];
\addplot [twilight_indigo, mark=none, line width=1pt] table [x=horizon, y=u10_mean] {\dataFeatureImportanceTabICL};
\addplot [name path=lower_tabicl_feature_41, draw=none, forget plot] table [x=horizon, y=u10_low_0.95] {\dataFeatureImportanceTabICL};
\addplot [name path=upper_tabicl_feature_41, draw=none, forget plot] table [x=horizon, y=u10_high_0.95] {\dataFeatureImportanceTabICL};
\addplot [twilight_indigo!40, fill opacity=0.75, forget plot] fill between [of=upper_tabicl_feature_41 and lower_tabicl_feature_41];
\nextgroupplot[title={\textbf{v10}}]
\addplot [spicy_paprika, mark=none, line width=1pt] table [x=horizon, y=v10_mean] {\dataFeatureImportanceXGBoost};
\addplot [name path=lower_xgboost_feature_42, draw=none, forget plot] table [x=horizon, y=v10_low_0.95] {\dataFeatureImportanceXGBoost};
\addplot [name path=upper_xgboost_feature_42, draw=none, forget plot] table [x=horizon, y=v10_high_0.95] {\dataFeatureImportanceXGBoost};
\addplot [spicy_paprika!40, fill opacity=0.75, forget plot] fill between [of=upper_xgboost_feature_42 and lower_xgboost_feature_42];

\end{groupplot}
\end{tikzpicture}


%% file: Plots/context_importance.tex



\pgfplotstableread[col sep=comma]{\getImportanceDataFilePath{month}{linear}{context_importance}{econ}}\dataContextImportanceLinear
\pgfplotstableread[col sep=comma]{\getImportanceDataFilePath{month}{xgboost}{context_importance}{econ}}\dataContextImportanceXGBoost
\pgfplotstableread[col sep=comma]{\getImportanceDataFilePath{month}{tabICL}{context_importance}{econ}}\dataContextImportanceTabICL



\begin{tikzpicture}
  \begin{axis}[
      ridge base,
      at={(0,0)}, anchor=south west,
      axis x line*=bottom,
      xtick={0,5,10,15},
      xlabel={Block Number},
      xlabel style={font=\large},
    ]
    \addplot[draw = orange!50!black,  fill=orange, fill opacity=0.3] table[x=block_number, y=h48] {\dataContextImportanceLinear};
    \label{plot:Linear}
  \end{axis}
  \node[anchor=west, font=\huge, rotate = 90] at (-1.5cm, 8) {Context Importance };
 
  \begin{axis}[
      ridge base,
      at={(0,\vsep)}, anchor=south west,
      axis x line=none,
    ]
    \addplot[draw = orange!50!black,  fill=orange, fill opacity=0.4] table[x=block_number, y=h42] {\dataContextImportanceLinear};
  \end{axis}

  \begin{axis}[
      ridge base,
      at={(0,2*\vsep)}, anchor=south west,
      axis x line=none,
    ]
    \addplot[draw = orange!50!black,  fill=orange, fill opacity=0.5] table[x=block_number, y=h36] {\dataContextImportanceLinear};
  \end{axis}

  \begin{axis}[
      ridge base,
      at={(0,3*\vsep)}, anchor=south west,
      axis x line=none,
    ]
    \addplot[draw = orange!50!black,  fill=orange, fill opacity=0.6] table[x=block_number, y=h30] {\dataContextImportanceLinear};
  \end{axis}

  \begin{axis}[
      ridge base,
      at={(0,4*\vsep)}, anchor=south west,
      axis x line=none,
    ]
    \addplot[draw = orange!50!black,  fill=orange, fill opacity=0.7] table[x=block_number, y=h24] {\dataContextImportanceLinear};
  \end{axis}

  \begin{axis}[
      ridge base,
      at={(0,5*\vsep)}, anchor=south west,
      axis x line=none,
    ]
    \addplot[draw = orange!50!black,  fill=orange, fill opacity=0.8] table[x=block_number, y=h18] {\dataContextImportanceLinear};
  \end{axis}

  \begin{axis}[
      ridge base,
      at={(0,6*\vsep)}, anchor=south west,
      axis x line=none,
    ]
    \addplot[draw = orange!50!black,  fill=orange, fill opacity=0.9] table[x=block_number, y=h12] {\dataContextImportanceLinear};
  \end{axis}

  \begin{axis}[
      ridge base,
      at={(0,7*\vsep)}, anchor=south west,
      axis x line=none,
    ]
    \addplot[draw = orange!50!black,  fill=orange, fill opacity=1] table[x=block_number, y=h6] {\dataContextImportanceLinear};
  \end{axis}

  \begin{axis}[
      ridge base,
      at={(\ridgewidth + \hsep,0)}, anchor=south west,
      axis x line*=bottom,
      xtick={0,5,10,15},
      xlabel={Block Number},
      xlabel style={font=\large},
    ]
    \addplot[draw=spicy_paprika!50!black,  fill=spicy_paprika, fill opacity=0.3] table[x=block_number, y=h48] {\dataContextImportanceXGBoost};
    \label{plot:XGBoost}
  \end{axis}

  \begin{axis}[
      ridge base,
      at={(\ridgewidth + \hsep,\vsep)}, anchor=south west,
      axis x line=none,
    ]
    \addplot[draw=spicy_paprika!50!black,  fill=spicy_paprika, fill opacity=0.4] table[x=block_number, y=h42] {\dataContextImportanceXGBoost};
  \end{axis}
 
  \begin{axis}[
      ridge base,
      at={(\ridgewidth + \hsep,2*\vsep)}, anchor=south west,
      axis x line=none,
    ]
    \addplot[draw=spicy_paprika!50!black,  fill=spicy_paprika, fill opacity=0.5] table[x=block_number, y=h36] {\dataContextImportanceXGBoost};
  \end{axis}

  \begin{axis}[
      ridge base,
      at={(\ridgewidth + \hsep,3*\vsep)}, anchor=south west,
      axis x line=none,
    ]
    \addplot[draw=spicy_paprika!50!black,  fill=spicy_paprika, fill opacity=0.6] table[x=block_number, y=h30] {\dataContextImportanceXGBoost};
  \end{axis}

  \begin{axis}[
      ridge base,
      at={(\ridgewidth + \hsep,4*\vsep)}, anchor=south west,
      axis x line=none,
    ]
    \addplot[draw=spicy_paprika!50!black,  fill=spicy_paprika, fill opacity=0.7] table[x=block_number, y=h24] {\dataContextImportanceXGBoost};
  \end{axis}

  \begin{axis}[
      ridge base,
      at={(\ridgewidth + \hsep,5*\vsep)}, anchor=south west,
      axis x line=none,
    ]
    \addplot[draw=spicy_paprika!50!black,  fill=spicy_paprika, fill opacity=0.8] table[x=block_number, y=h18] {\dataContextImportanceXGBoost};
  \end{axis}

  \begin{axis}[
      ridge base,
      at={(\ridgewidth + \hsep,6*\vsep)}, anchor=south west,
      axis x line=none,
    ]
    \addplot[draw=spicy_paprika!50!black,  fill=spicy_paprika, fill opacity=0.9] table[x=block_number, y=h12] {\dataContextImportanceXGBoost};
  \end{axis}

  \begin{axis}[
      ridge base,
      at={(\ridgewidth + \hsep,7*\vsep)}, anchor=south west,
      axis x line=none,
    ]
    \addplot[draw=spicy_paprika!50!black,  fill=spicy_paprika, fill opacity=1] table[x=block_number, y=h6] {\dataContextImportanceXGBoost};
  \end{axis}

  \begin{axis}[
      ridge base,
      at={(2*\ridgewidth + 2*\hsep,0)}, anchor=south west,
      axis x line*=bottom,
      xtick={0,5,10,15},
      xlabel={Block Number},
      xlabel style={font=\large},
    ]
    \addplot[draw=twilight_indigo!50!black,  fill=twilight_indigo, fill opacity=0.3] table[x=block_number, y=h48] {\dataContextImportanceTabICL};
    \label{plot:TabICL}
  \end{axis}
  \node[anchor=east, font=\large] at (30.75cm, 0.55*\ridgeheight) {\textbf{48 Months}};

  \begin{axis}[
      ridge base,
      at={(2*\ridgewidth + 2*\hsep,\vsep)}, anchor=south west,
      axis x line=none,
    ]
    \addplot[draw=twilight_indigo!50!black,  fill=twilight_indigo, fill opacity=0.4] table[x=block_number, y=h42] {\dataContextImportanceTabICL};
  \end{axis}
  \node[anchor=east, font=\large] at (30.75cm, \vsep + 0.55*\ridgeheight) {\textbf{42 Months}};
 
  \begin{axis}[
      ridge base,
      at={(2*\ridgewidth + 2*\hsep,2*\vsep)}, anchor=south west,
      axis x line=none,
    ]
    \addplot[draw=twilight_indigo!50!black,  fill=twilight_indigo, fill opacity=0.5] table[x=block_number, y=h36] {\dataContextImportanceTabICL};
  \end{axis}
  \node[anchor=east, font=\large] at (30.75cm, 2*\vsep + 0.55*\ridgeheight) {\textbf{36 Months}};

  \begin{axis}[
      ridge base,
      at={(2*\ridgewidth + 2*\hsep,3*\vsep)}, anchor=south west,
      axis x line=none,
    ]
    \addplot[draw=twilight_indigo!50!black,  fill=twilight_indigo, fill opacity=0.6] table[x=block_number, y=h30] {\dataContextImportanceTabICL};
  \end{axis}
  \node[anchor=east, font=\large] at (30.75cm, 3*\vsep + 0.55*\ridgeheight) {\textbf{30 Months}};

  \begin{axis}[
      ridge base,
      at={(2*\ridgewidth + 2*\hsep,4*\vsep)}, anchor=south west,
      axis x line=none,
    ]
    \addplot[draw=twilight_indigo!50!black,  fill=twilight_indigo, fill opacity=0.7] table[x=block_number, y=h24] {\dataContextImportanceTabICL};
  \end{axis}
  \node[anchor=east, font=\large] at (30.75cm, 4*\vsep + 0.55*\ridgeheight) {\textbf{24 Months}};

  \begin{axis}[
      ridge base,
      at={(2*\ridgewidth + 2*\hsep,5*\vsep)}, anchor=south west,
      axis x line=none,
    ]
    \addplot[draw=twilight_indigo!50!black,  fill=twilight_indigo, fill opacity=0.8] table[x=block_number, y=h18] {\dataContextImportanceTabICL};
  \end{axis}
  \node[anchor=east, font=\large] at (30.75cm, 5*\vsep + 0.55*\ridgeheight) {\textbf{18 Months}};

  \begin{axis}[
      ridge base,
      at={(2*\ridgewidth + 2*\hsep,6*\vsep)}, anchor=south west,
      axis x line=none,
    ]
    \addplot[draw=twilight_indigo!50!black,  fill=twilight_indigo, fill opacity=0.9] table[x=block_number, y=h12] {\dataContextImportanceTabICL};
  \end{axis}
  \node[anchor=east, font=\large] at (30.75cm, 6*\vsep + 0.55*\ridgeheight) {\textbf{12 Months}};

  \begin{axis}[
      ridge base,
      at={(2*\ridgewidth + 2*\hsep,7*\vsep)}, anchor=south west,
      axis x line=none,
    ]
    \addplot[draw=twilight_indigo!50!black,  fill=twilight_indigo, fill opacity=1] table[x=block_number, y=h6] {\dataContextImportanceTabICL};
  \end{axis}
  \node[anchor=east, font=\large] at (30.75cm, 7*\vsep+0.55*\ridgeheight) {\textbf{6 Months}};

    \node[
      draw,
      fill=white,
      anchor=north,
      inner xsep=6pt,
      inner ysep=4pt
    ] at (14,-1.5) {
      \ref{plot:Linear} \Large Linear \qquad \ref{plot:XGBoost} \Large XGBoost \qquad \ref{plot:TabICL} \Large TabICL
    };
\end{tikzpicture}

%% file: Tables/table_features.tex


\begin{landscape}
\begin{center}
\renewcommand{\arraystretch}{2} 
\begin{longtable}{p{4cm}|p{4cm}|p{0.75cm}|p{4.75cm}|p{1.5cm}|p{1.5cm}|p{1.5cm}|p{1.5cm}|p{1.65cm}|p{1.65cm}}
\caption{Description of the predictors available in the dataset. Each category is color-coded for easier readability.}
\label{tab:features}\\
\toprule
\textbf{Covariate \newline Abbreviation} & \textbf{Covariate \newline Full Name}                        & \textbf{Unit}  & \textbf{Description}                                                                                            & \textbf{Source}                 & \textbf{Covariate \newline Category} & \textbf{Temporal \newline Frequency} & \textbf{Spatial \newline Frequency}   & \textbf{Temporal \newline Aggregation \newline Rule} & \textbf{Spatial \newline Aggregation \newline Rule} \\
\midrule
\midrule
\endfirsthead

\toprule
\textbf{Covariate \newline Abbreviation} & \textbf{Covariate \newline Full Name}                        & \textbf{Unit}  & \textbf{Description}                                                                                            & \textbf{Source}                 & \textbf{Covariate \newline Category} & \textbf{Temporal \newline Frequency} & \textbf{Spatial \newline Frequency}   & \textbf{Temporal \newline Aggregation \newline Rule} & \textbf{Spatial \newline Aggregation \newline Rule} \\
\midrule
\midrule
\endhead

\rowcolor{apricot_cream!40} t2m                    & Temperature of air at 2m above the surface & \SI{}{\kelvin}     & Temperature of air at 2m above the surface                                                             & \acrshort{era5}                            & Weather            & Hourly             & Gridded 0.25x0.25& Mean                      & Mean                     \\
\rowcolor{apricot_cream!40} hdd                    & Heating Degree Days                        & \SI{}{\kelvin}     & Measurement of the energy quantity needed to heat a building                                           & Computed from \newline \acrshort{era5}/RTE & Weather            & Daily              & Gridded 0.25°x0.25° & Mean                      & Mean                     \\
\rowcolor{apricot_cream!40} cdd                    & Cooling Degree Days                        & \SI{}{\kelvin}     & Measurement of the energy quantity needed to cool a building                                           & Computed from \newline \acrshort{era5}/RTE & Weather            & Daily              & Gridded 0.25°x0.25° & Mean                      & Mean                     \\
\rowcolor{apricot_cream!40} rh                     & Relative Humidity                          & \%    & Concentration of water vapor present in the air                                                                     & Computed from \newline \acrshort{era5}     & Weather            & Hourly             & Gridded 0.25°x0.25° & Mean                      & Mean                     \\
\rowcolor{apricot_cream!40} ssrd                   & Surface Solar Radiation Downwards          & \unit{\joule\per\square\meter} & Amount of solar radiation (direct and diffuse) reaching a horizontal plane at the surface of the Earth & \acrshort{era5}                   & Weather            & Hourly             & Gridded 0.25°x0.25° & Mean                      & Mean                     \\
\rowcolor{apricot_cream!40} tcc                    & Total Cloud Cover                          & 0-1   & Proportion of a grid cell covered by clouds                                                            & \acrshort{era5}                   & Weather            & Hourly             & Gridded 0.25°x0.25° & Mean                      & Mean                     \\
\rowcolor{apricot_cream!40} tp                     & Total Precipitation                        & \unit{\meter}     & Accumulated liquid and frozen water that falls to the Earth's surface                                  & \acrshort{era5}                   & Weather            & Hourly             & Gridded 0.25°x0.25° & Mean                      & Mean                     \\
\rowcolor{apricot_cream!40} u10                    & 10-meter U Wind Speed                      & \unit{\meter\per\second} & Northward component of the wind speed at 10 meters                                                     & \acrshort{era5}                   & Weather            & Hourly             & Gridded 0.25°x0.25° & Mean                      & Mean                     \\
\rowcolor{apricot_cream!40} v10                    & 10-meter V Wind Speed                      & \unit{\meter\per\second} & Eastward component of the wind speed at 10 meters                                                      & \acrshort{era5}                   & Weather            & Hourly             & Gridded 0.25°x0.25° & Mean                      & Mean  \\                 
\midrule
\rowcolor{burnt_peach!30} dow                 & Day of the Week                              & {[}-{]} & Day of the Week as an integer from 0 to 6                                                    & {[}-{]}           & Calendar & Daily     &            &  &  \\
\rowcolor{burnt_peach!30} dow\_cos            & Cosine transformation of Day of the Week     & {[}-{]} & Transform Day of the Week as cos/sin to ensure adjacent days are recognized as such          & {[}-{]}           & Calendar & Daily     &            &  &  \\
\rowcolor{burnt_peach!30} dow\_sin            & Sine transformation of Day of the Week       & {[}-{]} & Transform Day of the Week as cos/sin to ensure adjacent days are recognized as such          & {[}-{]}           & Calendar & Daily     &            &  &  \\
\rowcolor{burnt_peach!30} doy                 & Day of the Year                              & {[}-{]} & Day of the Year as an integer from 0 to 365                                                  & {[}-{]}           & Calendar & Daily     &            &  &  \\
\rowcolor{burnt_peach!30} doy\_cos            & Cosine transformation of Day of the Year     & {[}-{]} & Transform Day of the Year as cos/sin to ensure adjacent days are recognized as such          & {[}-{]}           & Calendar & Monthly   &            &  &  \\
\rowcolor{burnt_peach!30} doy\_sin            & Sine transformation of Day of the Year       & {[}-{]} & Transform Day of the Year as cos/sin to ensure adjacent days are recognized as such          & {[}-{]}           & Calendar & Monthly   &            &  &  \\
\rowcolor{burnt_peach!30} moy                 & Month of the Year                            & {[}-{]} & Month as an integer from 1 to 12                                                             &                   & Calendar & Monthly   &            &  &  \\
\rowcolor{burnt_peach!30} moy\_cos            & Cosine transformation of Month of the Year   & {[}-{]} & Transform Month of the Year as cos/sin to ensure adjacent months are recognized as such      & {[}-{]}           & Calendar & Monthly   &            &  &  \\
\rowcolor{burnt_peach!30} moy\_sin            & Sine transformation of Month of the Year     & {[}-{]} & Transform Month of the Year as cos/sin to ensure adjacent months are recognized as such      & {[}-{]}           & Calendar & Monthly   &            &  &  \\
\rowcolor{burnt_peach!30} qoy                 & Quarter of the Year                          & {[}-{]} & Quarter as an integer from 1 to 4                                                            & {[}-{]}           & Calendar & Quarterly &            &  &  \\
\rowcolor{burnt_peach!30} qoy\_cos            & Cosine transformation of Quarter of the Year & {[}-{]} & Transform Quarter of the Year as cos/sin to ensure adjacent quarters are recognized as such  & {[}-{]}           & Calendar & Quarterly &            &  &  \\
\rowcolor{burnt_peach!30} qoy\_sin            & Sine transformation of Quarter of the Year   & {[}-{]} & Transform Quarter of the Year as cos/sin to ensure adjacent quarters are recognized as such  & {[}-{]}           & Calendar & Quarterly &            &  &  \\
\rowcolor{burnt_peach!30} seas                & Season                                       & {[}-{]} & Season (Winter, Spring, Summer, Autumn) as an integer from 1 to 4                            & {[}-{]}           & Calendar & Semester  &            &  &  \\
\rowcolor{burnt_peach!30} seas\_cos           & Cosine transformation of Season              & {[}-{]} & Transform Season as cos/sin to ensure adjacent seasons are recognized as such                & {[}-{]}           & Calendar & Semester  &            &  &  \\
\rowcolor{burnt_peach!30} seas\_sin           & Sine transformation of Season                & {[}-{]} & Transform Season as cos/sin to ensure adjacent seasons are recognized as such                & {[}-{]}           & Calendar & Semester  &            &  &  \\
\rowcolor{burnt_peach!30} is\_weekend         & Weekend                                      & {[}-{]} & 1 if the day is a Saturday or a Sunday,  0 otherwise                                         & {[}-{]}           & Calendar & Daily     &            & Sum  &  \\
\rowcolor{burnt_peach!30} is\_public\_holiday & Public Holiday (Bank Holiday)                & {[}-{]} & 1 if the day is a bank or public holiday (Christmas, New Year, Whit Monday ...), 0 otherwise & French Government & Calendar & Daily     &            & Sum  &  \\
\rowcolor{burnt_peach!30} is\_school\_holiday & School Holiday                               & {[}-{]} & 1 if the day is a school holiday for an academy in France, 0 otherwise                       & French Government & Calendar & Daily     & French Academy & Sum  & Sum  \\
\midrule
\rowcolor{muted_teal!30} B\_CVS-CJO                          & NACE B Industrial Production Index corrected from seasonal and calendar effects  & {[}-{]} & Industrial Production Index Base 2021 for NACE Class B: Mining \& Quarrying                                                    & \acrshort{insee}              & Economic & Monthly   &   &  &  \\
\rowcolor{muted_teal!30} C\_CVS-CJO                          & NACE C Industrial Production Index corrected from seasonal and calendar effects  & {[}-{]} & Industrial Production Index Base 2021 for NACE Class C: Manufacturing                                                          & \acrshort{insee}              & Economic & Monthly   &   &  &  \\
\rowcolor{muted_teal!30} D\_CVS-CJO                          & NACE D Industrial Production Index corrected from seasonal and calendar effects  & {[}-{]} & Industrial Production Index Base 2021 for NACE Class D: Electricity, Gas, Steam \& Air Conditioning                            & \acrshort{insee}              & Economic & Monthly   &   &  &  \\
\rowcolor{muted_teal!30} E\_CVS-CJO                          & NACE E Industrial Production Index corrected from seasonal and calendar effects  & {[}-{]} & Industrial Production Index Base 2021 for NACE Class E: Water Supply, Sewerage, Waste Management \& Remediation Activities     & \acrshort{insee}              & Economic & Monthly   &   &  &  \\
\rowcolor{muted_teal!30} F\_CVS-CJO                          & NACE F Industrial Production Index corrected from seasonal and calendar effects  & {[}-{]} & Industrial Production Index Base 2021 for NACE Class F: Construction                                                           & \acrshort{insee}              & Economic & Monthly   &   &  &  \\
\rowcolor{muted_teal!30} H\_CVS-CJO                          & NACE H Service Production Index corrected from seasonal and calendar effects     & {[}-{]} & Service Production Index Base 2021 for NACE Class H: Transportation \& Storage                                                 & \acrshort{insee}              & Economic & Monthly   &   &  &  \\
\rowcolor{muted_teal!30} I\_CVS-CJO                          & NACE I Service Production Index corrected from seasonal and calendar effects     & {[}-{]} & Service Production Index Base 2021 for NACE Class I: Accomodation \& Food Services                                             & \acrshort{insee}              & Economic & Monthly   &   &  &  \\
\rowcolor{muted_teal!30} J\_CVS-CJO                          & NACE J Service Production Index corrected from seasonal and calendar effects     & {[}-{]} & Service Production Index Base 2021 for NACE Class J: Publishing, Broadcasting \& Content Production \& Distribution Activities & \acrshort{insee}              & Economic & Monthly   &   &  &  \\
\rowcolor{muted_teal!30} L\_CVS-CJO                          & NACE L Service Production Index corrected from seasonal and calendar effects     & {[}-{]} & Service Production Index Base 2021 for NACE Class L: Financial \& Insurance Activities                                         & \acrshort{insee}              & Economic & Monthly   &   &  &  \\
\rowcolor{muted_teal!30} M\_CVS-CJO                          & NACE M Service Production Index corrected from seasonal and calendar effects     & {[}-{]} & Service Production Index Base 2021 for NACE Class M: Real Estate Activities                                                    & \acrshort{insee}              & Economic & Monthly   &   &  &  \\
\rowcolor{muted_teal!30} N\_CVS-CJO                          & NACE N Service Production Index corrected from seasonal and calendar effects     & {[}-{]} & Service Production Index Base 2021 for NACE Class N: Professional, Scientific \& Technical Activities                          & \acrshort{insee}              & Economic & Monthly   &   &  &  \\
\rowcolor{muted_teal!30} R\_CVS-CJO                          & NACE R Service Production Index corrected from seasonal and calendar effects     & {[}-{]} & Service Production Index Base 2021 for NACE Class R: Human Health \& Social Work Activities                                    & \acrshort{insee}              & Economic & Monthly   &   &  &  \\
\rowcolor{muted_teal!30} spi\_CVS-CJO                        & Service Production Index corrected from seasonal and calendar effects            & {[}-{]} & Service Production Index Base 2021 for all NACE Service Classes                                                                & \acrshort{insee}              & Economic & Monthly   &   &  &  \\
\rowcolor{muted_teal!30} gdp\_CVS-CJO                        & Gross Domestic Product corrected from seasonal and calendar effects              & \EURdig      & Gross Domestic Product                                                                                                         & \acrshort{insee}              & Economic & Quarterly &   &  &  \\
\rowcolor{muted_teal!30} hcpi                                & Harmonized Consumer Price Index                                                  & {[}-{]} & Harmonized Consumer Price Index Base 2021                                                                                      & Eurostat           & Economic & Monthly   &   &  &  \\
\rowcolor{muted_teal!30} cpi\_w/o\_energy                    & Consumer Price Index without the Energy category                                 & {[}-{]} & Consumer Price Index Base 2025 for all households and all categories except Energy                                             & \acrshort{insee}              & Economic & Monthly   &   &  &  \\
\rowcolor{muted_teal!30} ecpi                                & Consumer Price Index for the Energy category                                     & {[}-{]} & Consumer Price Index Base 2025 for all households for the Energy category                                                      & \acrshort{insee}              & Economic & Monthly   &   &  &  \\
\rowcolor{muted_teal!30} scpi                                & Consumer Price Index for the Service Category                                    & {[}-{]} & Consumer Price Index Base 2025 for all households for the Service Category                                                     & Eurostat           & Economic & Monthly   &   &  &  \\
\rowcolor{muted_teal!30} employment                          & Number of resident persons in employment                                         & {[}-{]} & Number of resident persons in employment (National Accounts)                                                                   & Eurostat           & Economic & Quarterly &   &  &  \\
\rowcolor{muted_teal!30} population                          & Number of resident persons                                                       & {[}-{]} & Number of resident persons (National Accounts)                                                                                 & Eurostat           & Economic & Quarterly &   &  &  \\
\rowcolor{muted_teal!30} household\_income\_CVS-CJO          & Average Gross Disposable Income for Households, seasonally and calendar adjusted & \EURdig       & Total money available for spending and saving after taxes and social contributions                                             & Eurostat           & Economic & Quarterly &   &  &  \\
\rowcolor{muted_teal!30} tourism\_nights                     & Number of nights spent in tourism accomodations                                  & {[}-{]} & Number of nights spent by tourists in a tourism accomodation (hotels, camping site, lodges...).                                & Eurostat           & Economic & Monthly   & NUTS2 &  & Sum  \\
\rowcolor{muted_teal!30} PAC\_air/air                        & Number of installed aerothermal Heatpumps                                        & {[}-{]} & Number of installed aerothermal heatpumps                                                                                      & \acrshort{sdes}               & Economic & Yearly    &   &  &  \\
\rowcolor{muted_teal!30} PAC\_air/eau                        & Number of installed hydrothermal Heatpumps                                       & {[}-{]} & Number of installed hydrothermal heatpumps                                                                                     & \acrshort{sdes}               & Economic & Yearly    &   &  &  \\
\rowcolor{muted_teal!30} PAC\_geo                            & Number of installed geothermal Heatpumps                                         & {[}-{]} & Number of installed geothermal heatpumps                                                                                       & \acrshort{sdes}               & Economic & Yearly    &   &  &  \\
\rowcolor{muted_teal!30} PAC                                 & Number of installed Heatpumps                                                    & {[}-{]} & Total Number of install heatpumps (all types)                                                                                  & Computed \newline from \acrshort{sdes} & Economic & Yearly    &   &  &  \\
\rowcolor{muted_teal!30} electric\_car\_passenger            & Number of fully electric car passenger                                           & {[}-{]} & Number of fully electric car passenger registered on the 1st of January                                                        & \acrshort{sdes}               & Economic & Yearly    & City  &  &  Sum\\
\rowcolor{muted_teal!30} electric\_light\_utility & Number of fully electric light utilitary vehicle (van, pickup...)                & {[}-{]} & Number of fully electric light utility vehicles (van, small truck...) registered on the 1st of January                         & \acrshort{sdes}               & Economic & Yearly    & City  &  &  Sum\\
\rowcolor{muted_teal!30} electric\_heavy\_utility & Number of fully electric heavy utilitary vehicle (trucks, tractors...)           & {[}-{]} & Number of fully electric heavy utility vehicles (truck, lorry...) registered on the 1st of January                             & \acrshort{sdes}               & Economic & Yearly    & City  &  &  Sum\\
\rowcolor{muted_teal!30} electric\_public\_transports        & Number of fully electric public transports (buses...)                            & {[}-{]} & Number of fully electric public transports vehicles (buses, coach ...) registered on the 1st of January                        & \acrshort{sdes}               & Economic & Yearly    & City  &  &  Sum\\
\rowcolor{muted_teal!30} electric\_vehicles                  & Number of fully electric vehicles                                                & {[}-{]} & Total number of registered fully electric vehicles (all types)                                                                 & Computed from \newline \acrshort{sdes} & Economic & Yearly    & City  &  &  Sum\\
\rowcolor{muted_teal!30} hybrid\_car\_passenger              & Number of pluggable car passenger                                                & {[}-{]} & Number of pluggable hybrid car passenger registered on the 1st of January                                                      & \acrshort{sdes}               & Economic & Yearly    & City  &  &  Sum\\
\rowcolor{muted_teal!30} hybrid\_light\_utility   & Number of pluggable light utility vehicle                                        & {[}-{]} & Number of pluggable hybrid light utility vehicles (van, small truck...) registered on the 1st of January                       & \acrshort{sdes}               & Economic & Yearly    & City  &  &  Sum\\
\rowcolor{muted_teal!30} hybrid\_heavy\_utility   & Number of pluggable heavy utility vehicles                                       & {[}-{]} & Number of pluggable heavy utility vehicles (truck, lorry...) registered on the 1st of January                                  & \acrshort{sdes}               & Economic & Yearly    & City  &  &  Sum\\
\rowcolor{muted_teal!30} hybrid\_public\_transports          & Number of pluggable public transports                                            & {[}-{]} & Number of pluggable hybrid public transports vehicles (buses, coach ...) registered on the 1st of January                      & \acrshort{sdes}               & Economic & Yearly    & City  &  &  Sum\\
\rowcolor{muted_teal!30} hybrid\_vehicles                    & Number of pluggable vehicles                                                     & {[}-{]} & Total number of registered pluggable hybrid vehicles (all types)                                                               & Computed from \newline \acrshort{sdes} & Economic & Yearly    & City  &  &  Sum\\
\bottomrule
\end{longtable}
\end{center}
\end{landscape}

%% file: Plots/MAPE_horizon.tex




\pgfplotstableread[col sep=comma]{\getErrorDataFilePath{\resolutionMonth}{\linear}{\errorFolder}{\Econ}{\mape}}\dataMAPELinearMonthEcon
\pgfplotstableread[col sep=comma]{\getErrorDataFilePath{\resolutionMonth}{\linear}{\errorFolder}{\noEcon}{\mape}}\dataMAPELinearMonthNoEcon

\pgfplotstableread[col sep=comma]{\getErrorDataFilePath{\resolutionDay}{\linear}{\errorFolder}{\Econ}{\mape}}\dataMAPELinearDayEcon
\pgfplotstableread[col sep=comma]{\getErrorDataFilePath{\resolutionDay}{\linear}{\errorFolder}{\noEcon}{\mape}}\dataMAPELinearDayNoEcon

\pgfplotstableread[col sep=comma]{\getErrorDataFilePath{\resolutionMonth}{\xgboost}{\errorFolder}{\Econ}{\mape}}\dataMAPEXGBoostMonthEcon
\pgfplotstableread[col sep=comma]{\getErrorDataFilePath{\resolutionMonth}{\xgboost}{\errorFolder}{\noEcon}{\mape}}\dataMAPEXGBoostMonthNoEcon

\pgfplotstableread[col sep=comma]{\getErrorDataFilePath{\resolutionDay}{\xgboost}{\errorFolder}{\Econ}{\mape}}\dataMAPEXGBoostDayEcon
\pgfplotstableread[col sep=comma]{\getErrorDataFilePath{\resolutionDay}{\xgboost}{\errorFolder}{\noEcon}{\mape}}\dataMAPEXGBoostDayNoEcon

\pgfplotstableread[col sep=comma]{\getErrorDataFilePath{\resolutionMonth}{\tabicl}{\errorFolder}{\Econ}{\mape}}\dataMAPETabICLMonthEcon
\pgfplotstableread[col sep=comma]{\getErrorDataFilePath{\resolutionMonth}{\tabicl}{\errorFolder}{\noEcon}{\mape}}\dataMAPETabICLMonthNoEcon

\pgfplotstableread[col sep=comma]{\getErrorDataFilePath{\resolutionDay}{\tabicl}{\errorFolder}{\Econ}{\mape}}\dataMAPETabICLDayEcon
\pgfplotstableread[col sep=comma]{\getErrorDataFilePath{\resolutionDay}{\tabicl}{\errorFolder}{\noEcon}{\mape}}\dataMAPETabICLDayNoEcon

\begin{tikzpicture}
    \begin{groupplot}[
    group style={
        group size=3 by 2,        
        horizontal sep=1.8cm,
        vertical sep=1.8cm,
    },
    width=6cm,
    height=5cm,
    grid=major,
    scaled ticks=false,
    xtick={1, 12, 24, 36, 48},
    xticklabels={1, 12, 24, 36, 48},
    legend columns = 2,
    legend style={at={(1.15,-2.1)}, anchor=south west},
]
 
\nextgroupplot[ymin=1.5, ymax=5.75, title={\textbf{Linear | Month}}, xlabel={Horizon [M]}, ylabel={$MAPE$}]
\addplot[mark=none, line width=1pt, color=light_indigo_purple]
        table [x=horizon, y=mean, col sep=comma] {\dataMAPELinearMonthEcon};
\addlegendentry{With Economic};
\addplot[mark=none, line width=1pt, color=light_light_apricot]
        table [x=horizon, y=mean, col sep=comma] {\dataMAPELinearMonthNoEcon};
\addlegendentry{Without Economic};
\addplot[name path=lower_econ_linear_month, fill=none, draw=none, mark=none, forget plot]
    table[x=horizon, y=low_0.95, col sep=comma, header=true]{\dataMAPELinearMonthEcon};
\addplot[name path=upper_econ_linear_month, fill=none, draw=none, mark=none, forget plot]
    table[x=horizon, y=high_0.95, col sep=comma, header=true]{\dataMAPELinearMonthEcon};
\addplot[light_indigo_purple!40, fill opacity=0.75, forget plot] fill between[of=lower_econ_linear_month and upper_econ_linear_month];

\addplot[name path=lower_no_econ_linear_month, fill=none, draw=none, mark=none, forget plot]
    table[x=horizon, y=low_0.95, col sep=comma, header=true]{\dataMAPELinearMonthNoEcon};
\addplot[name path=upper_no_econ_linear_month, fill=none, draw=none, mark=none, forget plot]
    table[x=horizon, y=high_0.95, col sep=comma, header=true]{\dataMAPELinearMonthNoEcon};
\addplot[light_light_apricot!40, fill opacity=0.75, forget plot] fill between[of=lower_no_econ_linear_month and upper_no_econ_linear_month];

\nextgroupplot[title={\textbf{XGBoost | Month}}, xlabel={Horizon [M]}, ylabel={$MAPE$}]
\addplot[mark=none, line width=1pt, color=light_indigo_purple]
        table [x=horizon, y=mean, col sep=comma] {\dataMAPEXGBoostMonthEcon};
\addplot[mark=none, line width=1pt, color=light_light_apricot]
        table [x=horizon, y=mean, col sep=comma] {\dataMAPEXGBoostMonthNoEcon};
\addplot[name path=lower_econ_xgboost_month, fill=none, draw=none, mark=none, forget plot]
    table[x=horizon, y=low_0.95, col sep=comma, header=true]{\dataMAPEXGBoostMonthEcon};
\addplot[name path=upper_econ_xgboost_month, fill=none, draw=none, mark=none, forget plot]
    table[x=horizon, y=high_0.95, col sep=comma, header=true]{\dataMAPEXGBoostMonthEcon};
\addplot[light_indigo_purple!40, fill opacity=0.75, forget plot] fill between[of=lower_econ_xgboost_month and upper_econ_xgboost_month];

\addplot[name path=lower_no_econ_xgboost_month, fill=none, draw=none, mark=none, forget plot]
    table[x=horizon, y=low_0.95, col sep=comma, header=true]{\dataMAPEXGBoostMonthNoEcon};
\addplot[name path=upper_no_econ_xgboost_month, fill=none, draw=none, mark=none, forget plot]
    table[x=horizon, y=high_0.95, col sep=comma, header=true]{\dataMAPEXGBoostMonthNoEcon};
\addplot[light_light_apricot!40, fill opacity=0.75, forget plot] fill between[of=lower_no_econ_xgboost_month and upper_no_econ_xgboost_month];

\nextgroupplot[title={\textbf{TabICL | Month}}, xlabel={Horizon [M]}, ylabel={$MAPE$}]
\addplot[mark=none, line width=1pt, color=light_indigo_purple]
        table [x=horizon, y=mean, col sep=comma] {\dataMAPETabICLMonthEcon};
\addplot[mark=none, line width=1pt, color=light_light_apricot]
        table [x=horizon, y=mean, col sep=comma] {\dataMAPETabICLMonthNoEcon};
\addplot[name path=lower_econ_tabICL_month, fill=none, draw=none, mark=none, forget plot]
    table[x=horizon, y=low_0.95, col sep=comma, header=true]{\dataMAPETabICLMonthEcon};
\addplot[name path=upper_econ_tabICL_month, fill=none, draw=none, mark=none, forget plot]
    table[x=horizon, y=high_0.95, col sep=comma, header=true]{\dataMAPETabICLMonthEcon};
\addplot[light_indigo_purple!40, fill opacity=0.75, forget plot] fill between[of=lower_econ_tabICL_month and upper_econ_tabICL_month];

\addplot[name path=lower_no_econ_tabICL_month, fill=none, draw=none, mark=none, forget plot]
    table[x=horizon, y=low_0.95, col sep=comma, header=true]{\dataMAPETabICLMonthNoEcon};
\addplot[name path=upper_no_econ_tabICL_month, fill=none, draw=none, mark=none, forget plot]
    table[x=horizon, y=high_0.95, col sep=comma, header=true]{\dataMAPETabICLMonthNoEcon};
\addplot[light_light_apricot!40, fill opacity=0.75, forget plot] fill between[of=lower_no_econ_tabICL_month and upper_no_econ_tabICL_month];

\nextgroupplot[ymin = 3.25, ymax=7.75, title={\textbf{Linear | Day}}, xlabel={Horizon [M]}, ylabel={$MAPE$}]
\addplot[mark=none, line width=1pt, color=light_indigo_purple]
        table [x=horizon, y=mean, col sep=comma] {\dataMAPELinearDayEcon};
\addplot[mark=none, line width=1pt, color=light_light_apricot]
        table [x=horizon, y=mean, col sep=comma] {\dataMAPELinearDayNoEcon};
\addplot[name path=lower_econ_linear_day, fill=none, draw=none, mark=none, forget plot]
    table[x=horizon, y=low_0.95, col sep=comma, header=true]{\dataMAPELinearDayEcon};
\addplot[name path=upper_econ_linear_day, fill=none, draw=none, mark=none, forget plot]
    table[x=horizon, y=high_0.95, col sep=comma, header=true]{\dataMAPELinearDayEcon};
\addplot[light_indigo_purple!40, fill opacity=0.75, forget plot] fill between[of=lower_econ_linear_day and upper_econ_linear_day];

\addplot[name path=lower_no_econ_linear_day, fill=none, draw=none, mark=none, forget plot]
    table[x=horizon, y=low_0.95, col sep=comma, header=true]{\dataMAPELinearDayNoEcon};
\addplot[name path=upper_no_econ_linear_day, fill=none, draw=none, mark=none, forget plot]
    table[x=horizon, y=high_0.95, col sep=comma, header=true]{\dataMAPELinearDayNoEcon};
\addplot[light_light_apricot!40, fill opacity=0.75, forget plot] fill between[of=lower_no_econ_linear_day and upper_no_econ_linear_day];

\nextgroupplot[title={\textbf{XGBoost | Day}}, xlabel={Horizon [M]}, ylabel={$MAPE$}]
\addplot[mark=none, line width=1pt, color=light_indigo_purple]
        table [x=horizon, y=mean, col sep=comma] {\dataMAPEXGBoostDayEcon};
\addplot[mark=none, line width=1pt, color=light_light_apricot]
        table [x=horizon, y=mean, col sep=comma] {\dataMAPEXGBoostDayNoEcon};
\addplot[name path=lower_econ_xgboost_day, fill=none, draw=none, mark=none, forget plot]
    table[x=horizon, y=low_0.95, col sep=comma, header=true]{\dataMAPEXGBoostDayEcon};
\addplot[name path=upper_econ_xgboost_day, fill=none, draw=none, mark=none, forget plot]
    table[x=horizon, y=high_0.95, col sep=comma, header=true]{\dataMAPEXGBoostDayEcon};
\addplot[light_indigo_purple!40, fill opacity=0.75, forget plot] fill between[of=lower_econ_xgboost_day and upper_econ_xgboost_day];

\addplot[name path=lower_no_econ_xgboost_day, fill=none, draw=none, mark=none, forget plot]
    table[x=horizon, y=low_0.95, col sep=comma, header=true]{\dataMAPEXGBoostDayNoEcon};
\addplot[name path=upper_no_econ_xgboost_day, fill=none, draw=none, mark=none, forget plot]
    table[x=horizon, y=high_0.95, col sep=comma, header=true]{\dataMAPEXGBoostDayNoEcon};
\addplot[light_light_apricot!40, fill opacity=0.75, forget plot] fill between[of=lower_no_econ_xgboost_day and upper_no_econ_xgboost_day];

\nextgroupplot[title={\textbf{TabICL | Day}}, xlabel={Horizon [M]}, ylabel={$MAPE$}]
\addplot[mark=none, line width=1pt, color=light_indigo_purple]
        table [x=horizon, y=mean, col sep=comma] {\dataMAPETabICLDayEcon};
\addplot[mark=none, line width=1pt, color=light_light_apricot]
        table [x=horizon, y=mean, col sep=comma] {\dataMAPETabICLDayNoEcon};
\addplot[name path=lower_econ_tabICL_day, fill=none, draw=none, mark=none, forget plot]
    table[x=horizon, y=low_0.95, col sep=comma, header=true]{\dataMAPETabICLDayEcon};
\addplot[name path=upper_econ_tabICL_day, fill=none, draw=none, mark=none, forget plot]
    table[x=horizon, y=high_0.95, col sep=comma, header=true]{\dataMAPETabICLDayEcon};
\addplot[light_indigo_purple!40, fill opacity=0.75, forget plot] fill between[of=lower_econ_tabICL_day and upper_econ_tabICL_day];

\addplot[name path=lower_no_econ_tabICL_day, fill=none, draw=none, mark=none, forget plot]
    table[x=horizon, y=low_0.95, col sep=comma, header=true]{\dataMAPETabICLDayNoEcon};
\addplot[name path=upper_no_econ_tabICL_day, fill=none, draw=none, mark=none, forget plot]
    table[x=horizon, y=high_0.95, col sep=comma, header=true]{\dataMAPETabICLDayNoEcon};
\addplot[light_light_apricot!40, fill opacity=0.75, forget plot] fill between[of=lower_no_econ_tabICL_day and upper_no_econ_tabICL_day];
 
\end{groupplot}

\end{tikzpicture}

%% file: Plots/BIAS_horizon.tex




\pgfplotstableread[col sep=comma]{\getErrorDataFilePath{\resolutionMonth}{\linear}{\errorFolder}{\Econ}{\bias}}\dataBIASLinearMonthEcon
\pgfplotstableread[col sep=comma]{\getErrorDataFilePath{\resolutionMonth}{\linear}{\errorFolder}{\noEcon}{\bias}}\dataBIASLinearMonthNoEcon

\pgfplotstableread[col sep=comma]{\getErrorDataFilePath{\resolutionDay}{\linear}{\errorFolder}{\Econ}{\bias}}\dataBIASLinearDayEcon
\pgfplotstableread[col sep=comma]{\getErrorDataFilePath{\resolutionDay}{\linear}{\errorFolder}{\noEcon}{\bias}}\dataBIASLinearDayNoEcon

\pgfplotstableread[col sep=comma]{\getErrorDataFilePath{\resolutionMonth}{\xgboost}{\errorFolder}{\Econ}{\bias}}\dataBIASXGBoostMonthEcon
\pgfplotstableread[col sep=comma]{\getErrorDataFilePath{\resolutionMonth}{\xgboost}{\errorFolder}{\noEcon}{\bias}}\dataBIASXGBoostMonthNoEcon

\pgfplotstableread[col sep=comma]{\getErrorDataFilePath{\resolutionDay}{\xgboost}{\errorFolder}{\Econ}{\bias}}\dataBIASXGBoostDayEcon
\pgfplotstableread[col sep=comma]{\getErrorDataFilePath{\resolutionDay}{\xgboost}{\errorFolder}{\noEcon}{\bias}}\dataBIASXGBoostDayNoEcon

\pgfplotstableread[col sep=comma]{\getErrorDataFilePath{\resolutionMonth}{\tabicl}{\errorFolder}{\Econ}{\bias}}\dataBIASTabICLMonthEcon
\pgfplotstableread[col sep=comma]{\getErrorDataFilePath{\resolutionMonth}{\tabicl}{\errorFolder}{\noEcon}{\bias}}\dataBIASTabICLMonthNoEcon

\pgfplotstableread[col sep=comma]{\getErrorDataFilePath{\resolutionDay}{\tabicl}{\errorFolder}{\Econ}{\bias}}\dataBIASTabICLDayEcon
\pgfplotstableread[col sep=comma]{\getErrorDataFilePath{\resolutionDay}{\tabicl}{\errorFolder}{\noEcon}{\bias}}\dataBIASTabICLDayNoEcon

\begin{tikzpicture}
    \begin{groupplot}[
    group style={
        group size=3 by 2,        
        horizontal sep=1.8cm,
        vertical sep=1.8cm,
    },
    width=6cm,
    height=5cm,
    grid=major,
    scaled y ticks=true,
    xtick={1, 12, 24, 36, 48},
    xticklabels={1, 12, 24, 36, 48},
    legend columns = 2,
    legend style={at={(1.15,-2.1)}, anchor=south west},
    /pgfplots/scale ticks below exponent={-3},
    /pgfplots/scale ticks above exponent={2}    
]
 
\nextgroupplot[ymin=-2300, ymax=2200, title={\textbf{Linear | Month}}, xlabel={Horizon [M]}, ylabel={$Bias$}]
\addplot[mark=none, line width=1pt, color=light_indigo_purple]
        table [x=horizon, y=mean, col sep=comma] {\dataBIASLinearMonthEcon};
\addlegendentry{With Economic};
\addplot[mark=none, line width=1pt, color=light_light_apricot]
        table [x=horizon, y=mean, col sep=comma] {\dataBIASLinearMonthNoEcon};
\addlegendentry{Without Economic};
\addplot[name path=lower_econ_linear_month, fill=none, draw=none, mark=none, forget plot]
    table[x=horizon, y=low_0.95, col sep=comma, header=true]{\dataBIASLinearMonthEcon};
\addplot[name path=upper_econ_linear_month, fill=none, draw=none, mark=none, forget plot]
    table[x=horizon, y=high_0.95, col sep=comma, header=true]{\dataBIASLinearMonthEcon};
\addplot[light_indigo_purple!40, fill opacity=0.75, forget plot] fill between[of=lower_econ_linear_month and upper_econ_linear_month];

\addplot[name path=lower_no_econ_linear_month, fill=none, draw=none, mark=none, forget plot]
    table[x=horizon, y=low_0.95, col sep=comma, header=true]{\dataBIASLinearMonthNoEcon};
\addplot[name path=upper_no_econ_linear_month, fill=none, draw=none, mark=none, forget plot]
    table[x=horizon, y=high_0.95, col sep=comma, header=true]{\dataBIASLinearMonthNoEcon};
\addplot[light_light_apricot!40, fill opacity=0.75, forget plot] fill between[of=lower_no_econ_linear_month and upper_no_econ_linear_month];

\nextgroupplot[title={\textbf{XGBoost | Month}}, xlabel={Horizon [M]}, ylabel={$Bias$}]
\addplot[mark=none, line width=1pt, color=light_indigo_purple]
        table [x=horizon, y=mean, col sep=comma] {\dataBIASXGBoostMonthEcon};
\addplot[mark=none, line width=1pt, color=light_light_apricot]
        table [x=horizon, y=mean, col sep=comma] {\dataBIASXGBoostMonthNoEcon};
\addplot[name path=lower_econ_xgboost_month, fill=none, draw=none, mark=none, forget plot]
    table[x=horizon, y=low_0.95, col sep=comma, header=true]{\dataBIASXGBoostMonthEcon};
\addplot[name path=upper_econ_xgboost_month, fill=none, draw=none, mark=none, forget plot]
    table[x=horizon, y=high_0.95, col sep=comma, header=true]{\dataBIASXGBoostMonthEcon};
\addplot[light_indigo_purple!40, fill opacity=0.75, forget plot] fill between[of=lower_econ_xgboost_month and upper_econ_xgboost_month];

\addplot[name path=lower_no_econ_xgboost_month, fill=none, draw=none, mark=none, forget plot]
    table[x=horizon, y=low_0.95, col sep=comma, header=true]{\dataBIASXGBoostMonthNoEcon};
\addplot[name path=upper_no_econ_xgboost_month, fill=none, draw=none, mark=none, forget plot]
    table[x=horizon, y=high_0.95, col sep=comma, header=true]{\dataBIASXGBoostMonthNoEcon};
\addplot[light_light_apricot!40, fill opacity=0.75, forget plot] fill between[of=lower_no_econ_xgboost_month and upper_no_econ_xgboost_month];

\nextgroupplot[title={\textbf{TabICL | Month}}, xlabel={Horizon [M]}, ylabel={$Bias$}]
\addplot[mark=none, line width=1pt, color=light_indigo_purple]
        table [x=horizon, y=mean, col sep=comma] {\dataBIASTabICLMonthEcon};
\addplot[mark=none, line width=1pt, color=light_light_apricot]
        table [x=horizon, y=mean, col sep=comma] {\dataBIASTabICLMonthNoEcon};
\addplot[name path=lower_econ_tabICL_month, fill=none, draw=none, mark=none, forget plot]
    table[x=horizon, y=low_0.95, col sep=comma, header=true]{\dataBIASTabICLMonthEcon};
\addplot[name path=upper_econ_tabICL_month, fill=none, draw=none, mark=none, forget plot]
    table[x=horizon, y=high_0.95, col sep=comma, header=true]{\dataBIASTabICLMonthEcon};
\addplot[light_indigo_purple!40, fill opacity=0.75, forget plot] fill between[of=lower_econ_tabICL_month and upper_econ_tabICL_month];

\addplot[name path=lower_no_econ_tabICL_month, fill=none, draw=none, mark=none, forget plot]
    table[x=horizon, y=low_0.95, col sep=comma, header=true]{\dataBIASTabICLMonthNoEcon};
\addplot[name path=upper_no_econ_tabICL_month, fill=none, draw=none, mark=none, forget plot]
    table[x=horizon, y=high_0.95, col sep=comma, header=true]{\dataBIASTabICLMonthNoEcon};
\addplot[light_light_apricot!40, fill opacity=0.75, forget plot] fill between[of=lower_no_econ_tabICL_month and upper_no_econ_tabICL_month];

\nextgroupplot[ymin = -3200, ymax=200, title={\textbf{Linear | Day}}, xlabel={Horizon [M]}, ylabel={$Bias$}]
\addplot[mark=none, line width=1pt, color=light_indigo_purple]
        table [x=horizon, y=mean, col sep=comma] {\dataBIASLinearDayEcon};
\addplot[mark=none, line width=1pt, color=light_light_apricot]
        table [x=horizon, y=mean, col sep=comma] {\dataBIASLinearDayNoEcon};
\addplot[name path=lower_econ_linear_day, fill=none, draw=none, mark=none, forget plot]
    table[x=horizon, y=low_0.95, col sep=comma, header=true]{\dataBIASLinearDayEcon};
\addplot[name path=upper_econ_linear_day, fill=none, draw=none, mark=none, forget plot]
    table[x=horizon, y=high_0.95, col sep=comma, header=true]{\dataBIASLinearDayEcon};
\addplot[light_indigo_purple!40, fill opacity=0.75, forget plot] fill between[of=lower_econ_linear_day and upper_econ_linear_day];

\addplot[name path=lower_no_econ_linear_day, fill=none, draw=none, mark=none, forget plot]
    table[x=horizon, y=low_0.95, col sep=comma, header=true]{\dataBIASLinearDayNoEcon};
\addplot[name path=upper_no_econ_linear_day, fill=none, draw=none, mark=none, forget plot]
    table[x=horizon, y=high_0.95, col sep=comma, header=true]{\dataBIASLinearDayNoEcon};
\addplot[light_light_apricot!40, fill opacity=0.75, forget plot] fill between[of=lower_no_econ_linear_day and upper_no_econ_linear_day];

\nextgroupplot[title={\textbf{XGBoost | Day}}, xlabel={Horizon [M]}, ylabel={$Bias$}]
\addplot[mark=none, line width=1pt, color=light_indigo_purple]
        table [x=horizon, y=mean, col sep=comma] {\dataBIASXGBoostDayEcon};
\addplot[mark=none, line width=1pt, color=light_light_apricot]
        table [x=horizon, y=mean, col sep=comma] {\dataBIASXGBoostDayNoEcon};
\addplot[name path=lower_econ_xgboost_day, fill=none, draw=none, mark=none, forget plot]
    table[x=horizon, y=low_0.95, col sep=comma, header=true]{\dataBIASXGBoostDayEcon};
\addplot[name path=upper_econ_xgboost_day, fill=none, draw=none, mark=none, forget plot]
    table[x=horizon, y=high_0.95, col sep=comma, header=true]{\dataBIASXGBoostDayEcon};
\addplot[light_indigo_purple!40, fill opacity=0.75, forget plot] fill between[of=lower_econ_xgboost_day and upper_econ_xgboost_day];

\addplot[name path=lower_no_econ_xgboost_day, fill=none, draw=none, mark=none, forget plot]
    table[x=horizon, y=low_0.95, col sep=comma, header=true]{\dataBIASXGBoostDayNoEcon};
\addplot[name path=upper_no_econ_xgboost_day, fill=none, draw=none, mark=none, forget plot]
    table[x=horizon, y=high_0.95, col sep=comma, header=true]{\dataBIASXGBoostDayNoEcon};
\addplot[light_light_apricot!40, fill opacity=0.75, forget plot] fill between[of=lower_no_econ_xgboost_day and upper_no_econ_xgboost_day];

\nextgroupplot[title={\textbf{TabICL | Day}}, xlabel={Horizon [M]}, ylabel={$Bias$}]
\addplot[mark=none, line width=1pt, color=light_indigo_purple]
        table [x=horizon, y=mean, col sep=comma] {\dataBIASTabICLDayEcon};
\addplot[mark=none, line width=1pt, color=light_light_apricot]
        table [x=horizon, y=mean, col sep=comma] {\dataBIASTabICLDayNoEcon};
\addplot[name path=lower_econ_tabICL_day, fill=none, draw=none, mark=none, forget plot]
    table[x=horizon, y=low_0.95, col sep=comma, header=true]{\dataBIASTabICLDayEcon};
\addplot[name path=upper_econ_tabICL_day, fill=none, draw=none, mark=none, forget plot]
    table[x=horizon, y=high_0.95, col sep=comma, header=true]{\dataBIASTabICLDayEcon};
\addplot[light_indigo_purple!40, fill opacity=0.75, forget plot] fill between[of=lower_econ_tabICL_day and upper_econ_tabICL_day];

\addplot[name path=lower_no_econ_tabICL_day, fill=none, draw=none, mark=none, forget plot]
    table[x=horizon, y=low_0.95, col sep=comma, header=true]{\dataBIASTabICLDayNoEcon};
\addplot[name path=upper_no_econ_tabICL_day, fill=none, draw=none, mark=none, forget plot]
    table[x=horizon, y=high_0.95, col sep=comma, header=true]{\dataBIASTabICLDayNoEcon};
\addplot[light_light_apricot!40, fill opacity=0.75, forget plot] fill between[of=lower_no_econ_tabICL_day and upper_no_econ_tabICL_day];
 
\end{groupplot}

\end{tikzpicture}

%% file: Plots/VARIANCE_horizon.tex




\pgfplotstableread[col sep=comma]{\getErrorDataFilePath{\resolutionMonth}{\linear}{\errorFolder}{\Econ}{\variance}}\dataVARIANCELinearMonthEcon
\pgfplotstableread[col sep=comma]{\getErrorDataFilePath{\resolutionMonth}{\linear}{\errorFolder}{\noEcon}{\variance}}\dataVARIANCELinearMonthNoEcon

\pgfplotstableread[col sep=comma]{\getErrorDataFilePath{\resolutionDay}{\linear}{\errorFolder}{\Econ}{\variance}}\dataVARIANCELinearDayEcon
\pgfplotstableread[col sep=comma]{\getErrorDataFilePath{\resolutionDay}{\linear}{\errorFolder}{\noEcon}{\variance}}\dataVARIANCELinearDayNoEcon

\pgfplotstableread[col sep=comma]{\getErrorDataFilePath{\resolutionMonth}{\xgboost}{\errorFolder}{\Econ}{\variance}}\dataVARIANCEXGBoostMonthEcon
\pgfplotstableread[col sep=comma]{\getErrorDataFilePath{\resolutionMonth}{\xgboost}{\errorFolder}{\noEcon}{\variance}}\dataVARIANCEXGBoostMonthNoEcon

\pgfplotstableread[col sep=comma]{\getErrorDataFilePath{\resolutionDay}{\xgboost}{\errorFolder}{\Econ}{\variance}}\dataVARIANCEXGBoostDayEcon
\pgfplotstableread[col sep=comma]{\getErrorDataFilePath{\resolutionDay}{\xgboost}{\errorFolder}{\noEcon}{\variance}}\dataVARIANCEXGBoostDayNoEcon

\pgfplotstableread[col sep=comma]{\getErrorDataFilePath{\resolutionMonth}{\tabicl}{\errorFolder}{\Econ}{\variance}}\dataVARIANCETabICLMonthEcon
\pgfplotstableread[col sep=comma]{\getErrorDataFilePath{\resolutionMonth}{\tabicl}{\errorFolder}{\noEcon}{\variance}}\dataVARIANCETabICLMonthNoEcon

\pgfplotstableread[col sep=comma]{\getErrorDataFilePath{\resolutionDay}{\tabicl}{\errorFolder}{\Econ}{\variance}}\dataVARIANCETabICLDayEcon
\pgfplotstableread[col sep=comma]{\getErrorDataFilePath{\resolutionDay}{\tabicl}{\errorFolder}{\noEcon}{\variance}}\dataVARIANCETabICLDayNoEcon

\begin{tikzpicture}
    \begin{groupplot}[
    group style={
        group size=3 by 2,        
        horizontal sep=1.8cm,
        vertical sep=1.8cm,
    },
    width=6cm,
    height=5cm,
    grid=major,
    scaled y ticks=true,
    xtick={1, 12, 24, 36, 48},
    xticklabels={1, 12, 24, 36, 48},
    legend columns = 2,
    legend style={at={(1.15,-2.1)}, anchor=south west},
    /pgfplots/scale ticks below exponent={-3},
    /pgfplots/scale ticks above exponent={2},
]
 
\nextgroupplot[ymax = 2000000, title={\textbf{Linear | Month}}, xlabel={Horizon [M]}, ylabel={$Variance$}]
\addplot[mark=none, line width=1pt, color=light_indigo_purple]
        table [x=horizon, y=mean, col sep=comma] {\dataVARIANCELinearMonthEcon};
\addlegendentry{With Economic};
\addplot[mark=none, line width=1pt, color=light_light_apricot]
        table [x=horizon, y=mean, col sep=comma] {\dataVARIANCELinearMonthNoEcon};
\addlegendentry{Without Economic};
\addplot[name path=lower_econ_linear_month, fill=none, draw=none, mark=none, forget plot]
    table[x=horizon, y=low_0.95, col sep=comma, header=true]{\dataVARIANCELinearMonthEcon};
\addplot[name path=upper_econ_linear_month, fill=none, draw=none, mark=none, forget plot]
    table[x=horizon, y=high_0.95, col sep=comma, header=true]{\dataVARIANCELinearMonthEcon};
\addplot[light_indigo_purple!40, fill opacity=0.75, forget plot] fill between[of=lower_econ_linear_month and upper_econ_linear_month];

\addplot[name path=lower_no_econ_linear_month, fill=none, draw=none, mark=none, forget plot]
    table[x=horizon, y=low_0.95, col sep=comma, header=true]{\dataVARIANCELinearMonthNoEcon};
\addplot[name path=upper_no_econ_linear_month, fill=none, draw=none, mark=none, forget plot]
    table[x=horizon, y=high_0.95, col sep=comma, header=true]{\dataVARIANCELinearMonthNoEcon};
\addplot[light_light_apricot!40, fill opacity=0.75, forget plot] fill between[of=lower_no_econ_linear_month and upper_no_econ_linear_month];

\nextgroupplot[title={\textbf{XGBoost | Month}}, xlabel={Horizon [M]}, ylabel={$Variance$}]
\addplot[mark=none, line width=1pt, color=light_indigo_purple]
        table [x=horizon, y=mean, col sep=comma] {\dataVARIANCEXGBoostMonthEcon};
\addplot[mark=none, line width=1pt, color=light_light_apricot]
        table [x=horizon, y=mean, col sep=comma] {\dataVARIANCEXGBoostMonthNoEcon};
\addplot[name path=lower_econ_xgboost_month, fill=none, draw=none, mark=none, forget plot]
    table[x=horizon, y=low_0.95, col sep=comma, header=true]{\dataVARIANCEXGBoostMonthEcon};
\addplot[name path=upper_econ_xgboost_month, fill=none, draw=none, mark=none, forget plot]
    table[x=horizon, y=high_0.95, col sep=comma, header=true]{\dataVARIANCEXGBoostMonthEcon};
\addplot[light_indigo_purple!40, fill opacity=0.75, forget plot] fill between[of=lower_econ_xgboost_month and upper_econ_xgboost_month];

\addplot[name path=lower_no_econ_xgboost_month, fill=none, draw=none, mark=none, forget plot]
    table[x=horizon, y=low_0.95, col sep=comma, header=true]{\dataVARIANCEXGBoostMonthNoEcon};
\addplot[name path=upper_no_econ_xgboost_month, fill=none, draw=none, mark=none, forget plot]
    table[x=horizon, y=high_0.95, col sep=comma, header=true]{\dataVARIANCEXGBoostMonthNoEcon};
\addplot[light_light_apricot!40, fill opacity=0.75, forget plot] fill between[of=lower_no_econ_xgboost_month and upper_no_econ_xgboost_month];

\nextgroupplot[title={\textbf{TabICL | Month}}, xlabel={Horizon [M]}, ylabel={$Variance$}]
\addplot[mark=none, line width=1pt, color=light_indigo_purple]
        table [x=horizon, y=mean, col sep=comma] {\dataVARIANCETabICLMonthEcon};
\addplot[mark=none, line width=1pt, color=light_light_apricot]
        table [x=horizon, y=mean, col sep=comma] {\dataVARIANCETabICLMonthNoEcon};
\addplot[name path=lower_econ_tabICL_month, fill=none, draw=none, mark=none, forget plot]
    table[x=horizon, y=low_0.95, col sep=comma, header=true]{\dataVARIANCETabICLMonthEcon};
\addplot[name path=upper_econ_tabICL_month, fill=none, draw=none, mark=none, forget plot]
    table[x=horizon, y=high_0.95, col sep=comma, header=true]{\dataVARIANCETabICLMonthEcon};
\addplot[light_indigo_purple!40, fill opacity=0.75, forget plot] fill between[of=lower_econ_tabICL_month and upper_econ_tabICL_month];

\addplot[name path=lower_no_econ_tabICL_month, fill=none, draw=none, mark=none, forget plot]
    table[x=horizon, y=low_0.95, col sep=comma, header=true]{\dataVARIANCETabICLMonthNoEcon};
\addplot[name path=upper_no_econ_tabICL_month, fill=none, draw=none, mark=none, forget plot]
    table[x=horizon, y=high_0.95, col sep=comma, header=true]{\dataVARIANCETabICLMonthNoEcon};
\addplot[light_light_apricot!40, fill opacity=0.75, forget plot] fill between[of=lower_no_econ_tabICL_month and upper_no_econ_tabICL_month];

\nextgroupplot[ymax=10000000, title={\textbf{Linear | Day}}, xlabel={Horizon [M]}, ylabel={$Variance$}]
\addplot[mark=none, line width=1pt, color=light_indigo_purple]
        table [x=horizon, y=mean, col sep=comma] {\dataVARIANCELinearDayEcon};
\addplot[mark=none, line width=1pt, color=light_light_apricot]
        table [x=horizon, y=mean, col sep=comma] {\dataVARIANCELinearDayNoEcon};
\addplot[name path=lower_econ_linear_day, fill=none, draw=none, mark=none, forget plot]
    table[x=horizon, y=low_0.95, col sep=comma, header=true]{\dataVARIANCELinearDayEcon};
\addplot[name path=upper_econ_linear_day, fill=none, draw=none, mark=none, forget plot]
    table[x=horizon, y=high_0.95, col sep=comma, header=true]{\dataVARIANCELinearDayEcon};
\addplot[light_indigo_purple!40, fill opacity=0.75, forget plot] fill between[of=lower_econ_linear_day and upper_econ_linear_day];

\addplot[name path=lower_no_econ_linear_day, fill=none, draw=none, mark=none, forget plot]
    table[x=horizon, y=low_0.95, col sep=comma, header=true]{\dataVARIANCELinearDayNoEcon};
\addplot[name path=upper_no_econ_linear_day, fill=none, draw=none, mark=none, forget plot]
    table[x=horizon, y=high_0.95, col sep=comma, header=true]{\dataVARIANCELinearDayNoEcon};
\addplot[light_light_apricot!40, fill opacity=0.75, forget plot] fill between[of=lower_no_econ_linear_day and upper_no_econ_linear_day];

\nextgroupplot[title={\textbf{XGBoost | Day}}, xlabel={Horizon [M]}, ylabel={$Variance$}]
\addplot[mark=none, line width=1pt, color=light_indigo_purple]
        table [x=horizon, y=mean, col sep=comma] {\dataVARIANCEXGBoostDayEcon};
\addplot[mark=none, line width=1pt, color=light_light_apricot]
        table [x=horizon, y=mean, col sep=comma] {\dataVARIANCEXGBoostDayNoEcon};
\addplot[name path=lower_econ_xgboost_day, fill=none, draw=none, mark=none, forget plot]
    table[x=horizon, y=low_0.95, col sep=comma, header=true]{\dataVARIANCEXGBoostDayEcon};
\addplot[name path=upper_econ_xgboost_day, fill=none, draw=none, mark=none, forget plot]
    table[x=horizon, y=high_0.95, col sep=comma, header=true]{\dataVARIANCEXGBoostDayEcon};
\addplot[light_indigo_purple!40, fill opacity=0.75, forget plot] fill between[of=lower_econ_xgboost_day and upper_econ_xgboost_day];

\addplot[name path=lower_no_econ_xgboost_day, fill=none, draw=none, mark=none, forget plot]
    table[x=horizon, y=low_0.95, col sep=comma, header=true]{\dataVARIANCEXGBoostDayNoEcon};
\addplot[name path=upper_no_econ_xgboost_day, fill=none, draw=none, mark=none, forget plot]
    table[x=horizon, y=high_0.95, col sep=comma, header=true]{\dataVARIANCEXGBoostDayNoEcon};
\addplot[light_light_apricot!40, fill opacity=0.75, forget plot] fill between[of=lower_no_econ_xgboost_day and upper_no_econ_xgboost_day];

\nextgroupplot[title={\textbf{TabICL | Day}}, xlabel={Horizon [M]}, ylabel={$Variance$}]
\addplot[mark=none, line width=1pt, color=light_indigo_purple]
        table [x=horizon, y=mean, col sep=comma] {\dataVARIANCETabICLDayEcon};
\addplot[mark=none, line width=1pt, color=light_light_apricot]
        table [x=horizon, y=mean, col sep=comma] {\dataVARIANCETabICLDayNoEcon};
\addplot[name path=lower_econ_tabICL_day, fill=none, draw=none, mark=none, forget plot]
    table[x=horizon, y=low_0.95, col sep=comma, header=true]{\dataVARIANCETabICLDayEcon};
\addplot[name path=upper_econ_tabICL_day, fill=none, draw=none, mark=none, forget plot]
    table[x=horizon, y=high_0.95, col sep=comma, header=true]{\dataVARIANCETabICLDayEcon};
\addplot[light_indigo_purple!40, fill opacity=0.75, forget plot] fill between[of=lower_econ_tabICL_day and upper_econ_tabICL_day];

\addplot[name path=lower_no_econ_tabICL_day, fill=none, draw=none, mark=none, forget plot]
    table[x=horizon, y=low_0.95, col sep=comma, header=true]{\dataVARIANCETabICLDayNoEcon};
\addplot[name path=upper_no_econ_tabICL_day, fill=none, draw=none, mark=none, forget plot]
    table[x=horizon, y=high_0.95, col sep=comma, header=true]{\dataVARIANCETabICLDayNoEcon};
\addplot[light_light_apricot!40, fill opacity=0.75, forget plot] fill between[of=lower_no_econ_tabICL_day and upper_no_econ_tabICL_day];
 
\end{groupplot}

\end{tikzpicture}

%% file: Plots/RMSE_skill_score.tex



\pgfplotstableread[col sep=comma]{\getErrorDataFilePath{\resolutionMonth}{\linear}{\errorFolder}{\Econ}{\rmsess}}\dataRMSESSLinearMonth
\pgfplotstableread[col sep=comma]{\getErrorDataFilePath{\resolutionDay}{\linear}{\errorFolder}{\Econ}{\rmsess}}\dataRMSESSLinearDay

\pgfplotstableread[col sep=comma]{\getErrorDataFilePath{\resolutionMonth}{\xgboost}{\errorFolder}{\Econ}{\rmsess}}\dataRMSESSXGBoostMonth
\pgfplotstableread[col sep=comma]{\getErrorDataFilePath{\resolutionDay}{\xgboost}{\errorFolder}{\Econ}{\rmsess}}\dataRMSESSXGBoostDay

\pgfplotstableread[col sep=comma]{\getErrorDataFilePath{\resolutionMonth}{\tabicl}{\errorFolder}{\Econ}{\rmsess}}\dataRMSESSTabICLMonth
\pgfplotstableread[col sep=comma]{\getErrorDataFilePath{\resolutionDay}{\tabicl}{\errorFolder}{\Econ}{\rmsess}}\dataRMSESSTabICLDay

\begin{tikzpicture}
    \begin{groupplot}[
    group style={
        group size=2 by 1,        
        horizontal sep=1.8cm,
        group name= myplots,
    },
    width=7cm,
    height=5cm,
    scale only axis,
    grid=major,
    scaled ticks=false,
    xtick={1, 12, 24, 36, 48},
    xticklabels={1, 12, 24, 36, 48},
    legend columns=3,
    legend style={at={(1.1,-0.25)}, anchor=north}, 
]
 
\nextgroupplot[ymin = -0.25, ymax = 0.35, restrict y to domain*=-0.25:0.35, title={\textbf{Monthly Resolution}}, xlabel={Horizon [M]}, ylabel={Skill Score $\left[RMSESS\right]$}]
\addplot[mark=none, line width=1.5pt, color=orange]
        table [x=horizon, y =mean, col sep=comma] {\dataRMSESSLinearMonth};
\addlegendentry{Linear}

\addplot[mark=none, line width=1.5pt, color=spicy_paprika]
        table [x=horizon, y=mean, col sep=comma] {\dataRMSESSXGBoostMonth};
\addlegendentry{XGBoost} 

\addplot[mark=none, line width=1.5pt, color=twilight_indigo]
        table [x=horizon, y=mean, col sep=comma] {\dataRMSESSTabICLMonth};
\addlegendentry{TabICL}

\addplot[mark=none, line width=1.5pt, color=black, style = dashed, domain = 1:48, samples=100]{0};      
\addplot[name path=lower_linear_month, fill=none, draw=none, mark=none, forget plot]
    table[x=horizon, y=low_0.95, col sep=comma, header=true]{\dataRMSESSLinearMonth};
\addplot[name path=upper_linear_month, fill=none, draw=none, mark=none, forget plot]
    table[x=horizon, y=high_0.95, col sep=comma, header=true]{\dataRMSESSLinearMonth};
\addplot[orange!40, fill opacity=0.75, forget plot] fill between[of=lower_linear_month and upper_linear_month];

\addplot[name path=lower_xgboost_month, fill=none, draw=none, mark=none, forget plot]
    table[x=horizon, y=low_0.95, col sep=comma, header=true]{\dataRMSESSXGBoostMonth};
\addplot[name path=upper_xgboost_month, fill=none, draw=none, mark=none, forget plot]
    table[x=horizon, y=high_0.95, col sep=comma, header=true]{\dataRMSESSXGBoostMonth};
\addplot[spicy_paprika!40, fill opacity=0.75, forget plot] fill between[of=lower_xgboost_month and upper_xgboost_month];

\addplot[name path=lower_tabicl_month, fill=none, draw=none, mark=none, forget plot]
    table[x=horizon, y=low_0.95, col sep=comma, header=true]{\dataRMSESSTabICLMonth};
\addplot[name path=upper_tabicl_month, fill=none, draw=none, mark=none, forget plot]
    table[x=horizon, y=high_0.95, col sep=comma, header=true]{\dataRMSESSTabICLMonth};
\addplot[twilight_indigo!40, fill opacity=0.75, forget plot] fill between[of=lower_tabicl_month and upper_tabicl_month];

\nextgroupplot[ ymin = -0.35, ymax=0.45,  title={\textbf{Daily Resolution}}, xlabel={Horizon [M]}, ylabel={Skill Score $\left[RMSESS\right]$}]
\addplot[mark=none, line width=1.5pt, color=orange]
        table [x=horizon, y=mean, col sep=comma] {\dataRMSESSLinearDay};

\addplot[mark=none, line width=1.5pt, color=spicy_paprika]
        table [x=horizon, y=mean, col sep=comma] {\dataRMSESSXGBoostDay};

\addplot[mark=none, line width=1.5pt, color=twilight_indigo]
        table [x=horizon, y=mean, col sep=comma] {\dataRMSESSTabICLDay};

\addplot[mark=none, line width=1.5pt, color=black, style = dashed, domain = 1:48, samples=100]{0}; 
\addplot[name path=lower_linear_day, fill=none, draw=none, mark=none, forget plot]
    table[x=horizon, y=low_0.95, col sep=comma, header=true]{\dataRMSESSLinearDay};
\addplot[name path=upper_linear_day, fill=none, draw=none, mark=none, forget plot]
    table[x=horizon, y=high_0.95, col sep=comma, header=true]{\dataRMSESSLinearDay};
\addplot[orange!40, fill opacity=0.75, forget plot] fill between[of=lower_linear_day and upper_linear_day];

\addplot[name path=lower_xgboost_day, fill=none, draw=none, mark=none, forget plot]
    table[x=horizon, y=low_0.95, col sep=comma, header=true]{\dataRMSESSXGBoostDay};
\addplot[name path=upper_xgboost_day, fill=none, draw=none, mark=none, forget plot]
    table[x=horizon, y=high_0.95, col sep=comma, header=true]{\dataRMSESSXGBoostDay};
\addplot[spicy_paprika!40, fill opacity=0.75, forget plot] fill between[of=lower_xgboost_day and upper_xgboost_day];

\addplot[name path=lower_tabicl_day, fill=none, draw=none, mark=none, forget plot]
    table[x=horizon, y=low_0.95, col sep=comma, header=true]{\dataRMSESSTabICLDay};
\addplot[name path=upper_tabicl_day, fill=none, draw=none, mark=none, forget plot]
    table[x=horizon, y=high_0.95, col sep=comma, header=true]{\dataRMSESSTabICLDay};
\addplot[twilight_indigo!40, fill opacity=0.75, forget plot] fill between[of=lower_tabicl_day and upper_tabicl_day];

\end{groupplot}
\end{tikzpicture}

%% file: Plots/MAPE_skill_score.tex



\pgfplotstableread[col sep=comma]{\getErrorDataFilePath{\resolutionMonth}{\linear}{\errorFolder}{\Econ}{\mapess}}\dataMAPESSLinearMonth
\pgfplotstableread[col sep=comma]{\getErrorDataFilePath{\resolutionDay}{\linear}{\errorFolder}{\Econ}{\mapess}}\dataMAPESSLinearDay

\pgfplotstableread[col sep=comma]{\getErrorDataFilePath{\resolutionMonth}{\xgboost}{\errorFolder}{\Econ}{\mapess}}\dataMAPESSXGBoostMonth
\pgfplotstableread[col sep=comma]{\getErrorDataFilePath{\resolutionDay}{\xgboost}{\errorFolder}{\Econ}{\mapess}}\dataMAPESSXGBoostDay

\pgfplotstableread[col sep=comma]{\getErrorDataFilePath{\resolutionMonth}{\tabicl}{\errorFolder}{\Econ}{\mapess}}\dataMAPESSTabICLMonth
\pgfplotstableread[col sep=comma]{\getErrorDataFilePath{\resolutionDay}{\tabicl}{\errorFolder}{\Econ}{\mapess}}\dataMAPESSTabICLDay

\begin{tikzpicture}
    \begin{groupplot}[
    group style={
        group size=2 by 1,        
        horizontal sep=1.8cm,
        group name= myplots,
    },
    width=7cm,
    height=5cm,
    scale only axis,
    grid=major,
    scaled ticks=false,
    xtick={1, 12, 24, 36, 48},
    xticklabels={1, 12, 24, 36, 48},
    legend columns=3,
    legend style={at={(1.1,-0.25)}, anchor=north}, 
]
 
\nextgroupplot[ymin=-0.25, ymax=0.35, title={\textbf{Monthly Resolution}}, xlabel={Horizon [M]}, ylabel={Skill Score $\left[MAPESS\right]$}]
\addplot[mark=none, line width=1.5pt, color=orange]
        table [x=horizon, y =mean, col sep=comma] {\dataMAPESSLinearMonth};
\addlegendentry{Linear}

\addplot[mark=none, line width=1.5pt, color=spicy_paprika]
        table [x=horizon, y=mean, col sep=comma] {\dataMAPESSXGBoostMonth};
\addlegendentry{XGBoost} 

\addplot[mark=none, line width=1.5pt, color=twilight_indigo]
        table [x=horizon, y=mean, col sep=comma] {\dataMAPESSTabICLMonth};
\addlegendentry{TabICL}

\addplot[mark=none, line width=1.5pt, color=black, style = dashed, domain = 1:48, samples=100]{0};      
\addplot[name path=lower_linear_month, fill=none, draw=none, mark=none, forget plot]
    table[x=horizon, y=low_0.95, col sep=comma, header=true]{\dataMAPESSLinearMonth};
\addplot[name path=upper_linear_month, fill=none, draw=none, mark=none, forget plot]
    table[x=horizon, y=high_0.95, col sep=comma, header=true]{\dataMAPESSLinearMonth};
\addplot[orange!40, fill opacity=0.75, forget plot] fill between[of=lower_linear_month and upper_linear_month];

\addplot[name path=lower_xgboost_month, fill=none, draw=none, mark=none, forget plot]
    table[x=horizon, y=low_0.95, col sep=comma, header=true]{\dataMAPESSXGBoostMonth};
\addplot[name path=upper_xgboost_month, fill=none, draw=none, mark=none, forget plot]
    table[x=horizon, y=high_0.95, col sep=comma, header=true]{\dataMAPESSXGBoostMonth};
\addplot[spicy_paprika!40, fill opacity=0.75, forget plot] fill between[of=lower_xgboost_month and upper_xgboost_month];

\addplot[name path=lower_tabicl_month, fill=none, draw=none, mark=none, forget plot]
    table[x=horizon, y=low_0.95, col sep=comma, header=true]{\dataMAPESSTabICLMonth};
\addplot[name path=upper_tabicl_month, fill=none, draw=none, mark=none, forget plot]
    table[x=horizon, y=high_0.95, col sep=comma, header=true]{\dataMAPESSTabICLMonth};
\addplot[twilight_indigo!40, fill opacity=0.75, forget plot] fill between[of=lower_tabicl_month and upper_tabicl_month];

\nextgroupplot[ymin=-0.35, ymax=0.45,title={\textbf{Daily Resolution}}, xlabel={Horizon [M]}, ylabel={Skill Score $\left[MAPESS\right]$}]
\addplot[mark=none, line width=1.5pt, color=orange]
        table [x=horizon, y=mean, col sep=comma] {\dataMAPESSLinearDay};

\addplot[mark=none, line width=1.5pt, color=spicy_paprika]
        table [x=horizon, y=mean, col sep=comma] {\dataMAPESSXGBoostDay};

\addplot[mark=none, line width=1.5pt, color=twilight_indigo]
        table [x=horizon, y=mean, col sep=comma] {\dataMAPESSTabICLDay};

\addplot[mark=none, line width=1.5pt, color=black, style = dashed, domain = 1:48, samples=100]{0}; 
\addplot[name path=lower_linear_day, fill=none, draw=none, mark=none, forget plot]
    table[x=horizon, y=low_0.95, col sep=comma, header=true]{\dataMAPESSLinearDay};
\addplot[name path=upper_linear_day, fill=none, draw=none, mark=none, forget plot]
    table[x=horizon, y=high_0.95, col sep=comma, header=true]{\dataMAPESSLinearDay};
\addplot[orange!40, fill opacity=0.75, forget plot] fill between[of=lower_linear_day and upper_linear_day];

\addplot[name path=lower_xgboost_day, fill=none, draw=none, mark=none, forget plot]
    table[x=horizon, y=low_0.95, col sep=comma, header=true]{\dataMAPESSXGBoostDay};
\addplot[name path=upper_xgboost_day, fill=none, draw=none, mark=none, forget plot]
    table[x=horizon, y=high_0.95, col sep=comma, header=true]{\dataMAPESSXGBoostDay};
\addplot[spicy_paprika!40, fill opacity=0.75, forget plot] fill between[of=lower_xgboost_day and upper_xgboost_day];

\addplot[name path=lower_tabicl_day, fill=none, draw=none, mark=none, forget plot]
    table[x=horizon, y=low_0.95, col sep=comma, header=true]{\dataMAPESSTabICLDay};
\addplot[name path=upper_tabicl_day, fill=none, draw=none, mark=none, forget plot]
    table[x=horizon, y=high_0.95, col sep=comma, header=true]{\dataMAPESSTabICLDay};
\addplot[twilight_indigo!40, fill opacity=0.75, forget plot] fill between[of=lower_tabicl_day and upper_tabicl_day];

\end{groupplot}
\end{tikzpicture}

%% file: Plots/RMSE_horizon_comparison.tex




\begin{tikzpicture}
    \begin{groupplot}[
    group style={
        group size=2 by 2,        
        horizontal sep=1.8cm,
        vertical sep=1.8cm,
    },
    width=8cm,
    height=6cm,
    grid=major,
    scaled ticks=false,
    xtick={1, 12, 24, 36, 48},
    xticklabels={1, 12, 24, 36, 48},
    legend columns = 3,
    legend style={at={(-0.6, -1.8)}, anchor=south west},
]
 
\nextgroupplot[title={\textbf{Without Economic | Month}}, xlabel={Horizon [M]}, ylabel={$RMSE$}]
\addplot[mark=none, line width=1pt, color=orange, style = dashed]
        table [x=horizon, y=mean, col sep=comma] {\getErrorDataFilePath{\resolutionMonth}{\linear}{\errorFolder}{\noEcon}{\rmse}};
\addplot[mark=none, line width=1pt, color=spicy_paprika, style = dashed]
        table [x=horizon, y=mean, col sep=comma] {\getErrorDataFilePath{\resolutionMonth}{\xgboost}{\errorFolder}{\noEcon}{\rmse}};
\addplot[mark=none, line width=1pt, color=twilight_indigo, style = dashed]
        table [x=horizon, y=mean, col sep=comma] {\getErrorDataFilePath{\resolutionMonth}{\tabicl}{\errorFolder}{\noEcon}{\rmse}};

\addplot[name path=lower_no_econ_linear_month, fill=none, draw=none, mark=none, forget plot]
    table[x=horizon, y=low_0.95, col sep=comma, header=true]{\getErrorDataFilePath{\resolutionMonth}{\linear}{\errorFolder}{\noEcon}{\rmse}};
\addplot[name path=upper_no_econ_linear_month, fill=none, draw=none, mark=none, forget plot]
    table[x=horizon, y=high_0.95, col sep=comma, header=true]{\getErrorDataFilePath{\resolutionMonth}{\linear}{\errorFolder}{\noEcon}{\rmse}};
\addplot[orange!40, fill opacity=0.75, forget plot] fill between[of=lower_no_econ_linear_month and upper_no_econ_linear_month];

\addplot[name path=lower_no_econ_xgboost_month, fill=none, draw=none, mark=none, forget plot]
    table[x=horizon, y=low_0.95, col sep=comma, header=true]{\getErrorDataFilePath{\resolutionMonth}{\xgboost}{\errorFolder}{\noEcon}{\rmse}};
\addplot[name path=upper_no_econ_xgboost_month, fill=none, draw=none, mark=none, forget plot]
    table[x=horizon, y=high_0.95, col sep=comma, header=true]{\getErrorDataFilePath{\resolutionMonth}{\xgboost}{\errorFolder}{\noEcon}{\rmse}};
\addplot[spicy_paprika!40, fill opacity=0.75, forget plot] fill between[of=lower_no_econ_xgboost_month and upper_no_econ_xgboost_month];

\addplot[name path=lower_no_econ_tabICL_month, fill=none, draw=none, mark=none, forget plot]
    table[x=horizon, y=low_0.95, col sep=comma, header=true]{\getErrorDataFilePath{\resolutionMonth}{\tabicl}{\errorFolder}{\noEcon}{\rmse}};
\addplot[name path=upper_no_econ_tabICL_month, fill=none, draw=none, mark=none, forget plot]
    table[x=horizon, y=high_0.95, col sep=comma, header=true]{\getErrorDataFilePath{\resolutionMonth}{\tabicl}{\errorFolder}{\noEcon}{\rmse}};
\addplot[twilight_indigo!40, fill opacity=0.75, forget plot] fill between[of=lower_no_econ_tabICL_month and upper_no_econ_tabICL_month];

\nextgroupplot[ymin=400, ymax=2550,, title={\textbf{With Economic | Month}}, xlabel={Horizon [M]}, ylabel={$RMSE$}]
\addplot[mark=none, line width=1pt, color=orange]
        table [x=horizon, y=mean, col sep=comma] {\getErrorDataFilePath{\resolutionMonth}{\linear}{\errorFolder}{\Econ}{\rmse}};
\addlegendentry{Linear};
\addplot[mark=none, line width=1pt, color=spicy_paprika]
        table [x=horizon, y=mean, col sep=comma] {\getErrorDataFilePath{\resolutionMonth}{\xgboost}{\errorFolder}{\Econ}{\rmse}};
\addlegendentry{XGBoost};
\addplot[mark=none, line width=1pt, color=twilight_indigo]
        table [x=horizon, y=mean, col sep=comma] {\getErrorDataFilePath{\resolutionMonth}{\tabicl}{\errorFolder}{\Econ}{\rmse}};
\addlegendentry{TabICL};

\addplot[name path=lower_econ_linear_month, fill=none, draw=none, mark=none, forget plot]
    table[x=horizon, y=low_0.95, col sep=comma, header=true]{\getErrorDataFilePath{\resolutionMonth}{\linear}{\errorFolder}{\Econ}{\rmse}};
\addplot[name path=upper_econ_linear_month, fill=none, draw=none, mark=none, forget plot]
    table[x=horizon, y=high_0.95, col sep=comma, header=true]{\getErrorDataFilePath{\resolutionMonth}{\linear}{\errorFolder}{\Econ}{\rmse}};
\addplot[orange!40, fill opacity=0.75, forget plot] fill between[of=lower_econ_linear_month and upper_econ_linear_month];

\addplot[name path=lower_econ_xgboost_month, fill=none, draw=none, mark=none, forget plot]
    table[x=horizon, y=low_0.95, col sep=comma, header=true]{\getErrorDataFilePath{\resolutionMonth}{\xgboost}{\errorFolder}{\Econ}{\rmse}};
\addplot[name path=upper_econ_xgboost_month, fill=none, draw=none, mark=none, forget plot]
    table[x=horizon, y=high_0.95, col sep=comma, header=true]{\getErrorDataFilePath{\resolutionMonth}{\xgboost}{\errorFolder}{\Econ}{\rmse}};
\addplot[spicy_paprika!40, fill opacity=0.75, forget plot] fill between[of=lower_econ_xgboost_month and upper_econ_xgboost_month];

\addplot[name path=lower_econ_tabICL_month, fill=none, draw=none, mark=none, forget plot]
    table[x=horizon, y=low_0.95, col sep=comma, header=true]{\getErrorDataFilePath{\resolutionMonth}{\tabicl}{\errorFolder}{\Econ}{\rmse}};
\addplot[name path=upper_econ_tabICL_month, fill=none, draw=none, mark=none, forget plot]
    table[x=horizon, y=high_0.95, col sep=comma, header=true]{\getErrorDataFilePath{\resolutionMonth}{\tabicl}{\errorFolder}{\Econ}{\rmse}};
\addplot[twilight_indigo!40, fill opacity=0.75, forget plot] fill between[of=lower_econ_tabICL_month and upper_econ_tabICL_month];

\nextgroupplot[title={\textbf{Without Economic | Day}}, xlabel={Horizon [M]}, ylabel={$RMSE$}]
\addplot[mark=none, line width=1pt, color=orange, style = dashed]
        table [x=horizon, y=mean, col sep=comma] {\getErrorDataFilePath{\resolutionDay}{\linear}{\errorFolder}{\noEcon}{\rmse}};
\addplot[mark=none, line width=1pt, color=spicy_paprika, style = dashed]
        table [x=horizon, y=mean, col sep=comma] {\getErrorDataFilePath{\resolutionDay}{\xgboost}{\errorFolder}{\noEcon}{\rmse}};
\addplot[mark=none, line width=1pt, color=twilight_indigo, style = dashed]
        table [x=horizon, y=mean, col sep=comma] {\getErrorDataFilePath{\resolutionDay}{\tabicl}{\errorFolder}{\noEcon}{\rmse}};
           

\addplot[name path=lower_no_econ_linear_day, fill=none, draw=none, mark=none, forget plot]
    table[x=horizon, y=low_0.95, col sep=comma, header=true]{\getErrorDataFilePath{\resolutionDay}{\linear}{\errorFolder}{\noEcon}{\rmse}};
\addplot[name path=upper_no_econ_linear_day, fill=none, draw=none, mark=none, forget plot]
    table[x=horizon, y=high_0.95, col sep=comma, header=true]{\getErrorDataFilePath{\resolutionDay}{\linear}{\errorFolder}{\noEcon}{\rmse}};
\addplot[orange!40, fill opacity=0.75, forget plot] fill between[of=lower_no_econ_linear_day and upper_no_econ_linear_day];

\addplot[name path=lower_no_econ_xgboost_day, fill=none, draw=none, mark=none, forget plot]
    table[x=horizon, y=low_0.95, col sep=comma, header=true]{\getErrorDataFilePath{\resolutionDay}{\xgboost}{\errorFolder}{\noEcon}{\rmse}};
\addplot[name path=upper_no_econ_xgboost_day, fill=none, draw=none, mark=none, forget plot]
    table[x=horizon, y=high_0.95, col sep=comma, header=true]{\getErrorDataFilePath{\resolutionDay}{\xgboost}{\errorFolder}{\noEcon}{\rmse}};
\addplot[spicy_paprika!40, fill opacity=0.75, forget plot] fill between[of=lower_no_econ_xgboost_day and upper_no_econ_xgboost_day];

\addplot[name path=lower_no_econ_tabICL_day, fill=none, draw=none, mark=none, forget plot]
    table[x=horizon, y=low_0.95, col sep=comma, header=true]{\getErrorDataFilePath{\resolutionDay}{\tabicl}{\errorFolder}{\noEcon}{\rmse}};
\addplot[name path=upper_no_econ_tabICL_day, fill=none, draw=none, mark=none, forget plot]
    table[x=horizon, y=high_0.95, col sep=comma, header=true]{\getErrorDataFilePath{\resolutionDay}{\tabicl}{\errorFolder}{\noEcon}{\rmse}};
\addplot[twilight_indigo!40, fill opacity=0.75, forget plot] fill between[of=lower_no_econ_tabICL_day and upper_no_econ_tabICL_day];

\nextgroupplot[ymin = 1400, ymax=4200, title={\textbf{With Economic | Day}}, xlabel={Horizon [M]}, ylabel={$RMSE$}]
\addplot[mark=none, line width=1pt, color=orange]
        table [x=horizon, y=mean, col sep=comma] {\getErrorDataFilePath{\resolutionDay}{\linear}{\errorFolder}{\Econ}{\rmse}};
\addplot[mark=none, line width=1pt, color=spicy_paprika]
        table [x=horizon, y=mean, col sep=comma] {\getErrorDataFilePath{\resolutionDay}{\xgboost}{\errorFolder}{\Econ}{\rmse}};
\addplot[mark=none, line width=1pt, color=twilight_indigo]
        table [x=horizon, y=mean, col sep=comma] {\getErrorDataFilePath{\resolutionDay}{\tabicl}{\errorFolder}{\Econ}{\rmse}};
\addplot[name path=lower_econ_linear_day, fill=none, draw=none, mark=none, forget plot]
    table[x=horizon, y=low_0.95, col sep=comma, header=true]{\getErrorDataFilePath{\resolutionDay}{\linear}{\errorFolder}{\Econ}{\rmse}};
\addplot[name path=upper_econ_linear_day, fill=none, draw=none, mark=none, forget plot]
    table[x=horizon, y=high_0.95, col sep=comma, header=true]{\getErrorDataFilePath{\resolutionDay}{\linear}{\errorFolder}{\Econ}{\rmse}};
\addplot[orange!40, fill opacity=0.75, forget plot] fill between[of=lower_econ_linear_day and upper_econ_linear_day];

\addplot[name path=lower_econ_xgboost_day, fill=none, draw=none, mark=none, forget plot]
    table[x=horizon, y=low_0.95, col sep=comma, header=true]{\getErrorDataFilePath{\resolutionDay}{\xgboost}{\errorFolder}{\Econ}{\rmse}};
\addplot[name path=upper_econ_xgboost_day, fill=none, draw=none, mark=none, forget plot]
    table[x=horizon, y=high_0.95, col sep=comma, header=true]{\getErrorDataFilePath{\resolutionDay}{\xgboost}{\errorFolder}{\Econ}{\rmse}};
\addplot[spicy_paprika!40, fill opacity=0.75, forget plot] fill between[of=lower_econ_xgboost_day and upper_econ_xgboost_day];

\addplot[name path=lower_econ_tabICL_day, fill=none, draw=none, mark=none, forget plot]
    table[x=horizon, y=low_0.95, col sep=comma, header=true]{\getErrorDataFilePath{\resolutionDay}{\tabicl}{\errorFolder}{\Econ}{\rmse}};
\addplot[name path=upper_econ_tabICL_day, fill=none, draw=none, mark=none, forget plot]
    table[x=horizon, y=high_0.95, col sep=comma, header=true]{\getErrorDataFilePath{\resolutionDay}{\tabicl}{\errorFolder}{\Econ}{\rmse}};
\addplot[twilight_indigo!40, fill opacity=0.75, forget plot] fill between[of=lower_econ_tabICL_day and upper_econ_tabICL_day];
 
\end{groupplot}

\end{tikzpicture}

%% file: Plots/MAPE_horizon_comparison.tex




\begin{tikzpicture}
    \begin{groupplot}[
    group style={
        group size=2 by 2,        
        horizontal sep=1.8cm,
        vertical sep=1.8cm,
    },
    width=8cm,
    height=6cm,
    grid=major,
    scaled ticks=false,
    xtick={1, 12, 24, 36, 48},
    xticklabels={1, 12, 24, 36, 48},
    legend columns = 3,
    legend style={at={(-0.6,-1.8)}, anchor=south west},
]
 
\nextgroupplot[title={\textbf{Without Economic | Month}}, xlabel={Horizon [M]}, ylabel={$MAPE$}]
\addplot[mark=none, line width=1pt, color=orange, style = dashed]
        table [x=horizon, y=mean, col sep=comma] {\getErrorDataFilePath{\resolutionMonth}{\linear}{\errorFolder}{\noEcon}{\mape}};
\addplot[mark=none, line width=1pt, color=spicy_paprika, style = dashed]
        table [x=horizon, y=mean, col sep=comma] {\getErrorDataFilePath{\resolutionMonth}{\xgboost}{\errorFolder}{\noEcon}{\mape}};
\addplot[mark=none, line width=1pt, color=twilight_indigo, style = dashed]
        table [x=horizon, y=mean, col sep=comma] {\getErrorDataFilePath{\resolutionMonth}{\tabicl}{\errorFolder}{\noEcon}{\mape}};

\addplot[name path=lower_no_econ_linear_month, fill=none, draw=none, mark=none, forget plot]
    table[x=horizon, y=low_0.95, col sep=comma, header=true]{\getErrorDataFilePath{\resolutionMonth}{\linear}{\errorFolder}{\noEcon}{\mape}};
\addplot[name path=upper_no_econ_linear_month, fill=none, draw=none, mark=none, forget plot]
    table[x=horizon, y=high_0.95, col sep=comma, header=true]{\getErrorDataFilePath{\resolutionMonth}{\linear}{\errorFolder}{\noEcon}{\mape}};
\addplot[orange!40, fill opacity=0.75, forget plot] fill between[of=lower_no_econ_linear_month and upper_no_econ_linear_month];

\addplot[name path=lower_no_econ_xgboost_month, fill=none, draw=none, mark=none, forget plot]
    table[x=horizon, y=low_0.95, col sep=comma, header=true]{\getErrorDataFilePath{\resolutionMonth}{\xgboost}{\errorFolder}{\noEcon}{\mape}};
\addplot[name path=upper_no_econ_xgboost_month, fill=none, draw=none, mark=none, forget plot]
    table[x=horizon, y=high_0.95, col sep=comma, header=true]{\getErrorDataFilePath{\resolutionMonth}{\xgboost}{\errorFolder}{\noEcon}{\mape}};
\addplot[spicy_paprika!40, fill opacity=0.75, forget plot] fill between[of=lower_no_econ_xgboost_month and upper_no_econ_xgboost_month];

\addplot[name path=lower_no_econ_tabICL_month, fill=none, draw=none, mark=none, forget plot]
    table[x=horizon, y=low_0.95, col sep=comma, header=true]{\getErrorDataFilePath{\resolutionMonth}{\tabicl}{\errorFolder}{\noEcon}{\mape}};
\addplot[name path=upper_no_econ_tabICL_month, fill=none, draw=none, mark=none, forget plot]
    table[x=horizon, y=high_0.95, col sep=comma, header=true]{\getErrorDataFilePath{\resolutionMonth}{\tabicl}{\errorFolder}{\noEcon}{\mape}};
\addplot[twilight_indigo!40, fill opacity=0.75, forget plot] fill between[of=lower_no_econ_tabICL_month and upper_no_econ_tabICL_month];

\nextgroupplot[ymin=1, ymax=5.5, title={\textbf{With Economic | Month}}, xlabel={Horizon [M]}, ylabel={$MAPE$}]
\addplot[mark=none, line width=1pt, color=orange]
        table [x=horizon, y=mean, col sep=comma] {\getErrorDataFilePath{\resolutionMonth}{\linear}{\errorFolder}{\Econ}{\mape}};
\addlegendentry{Linear};
\addplot[mark=none, line width=1pt, color=spicy_paprika]
        table [x=horizon, y=mean, col sep=comma] {\getErrorDataFilePath{\resolutionMonth}{\xgboost}{\errorFolder}{\Econ}{\mape}};
\addlegendentry{XGBoost};
\addplot[mark=none, line width=1pt, color=twilight_indigo]
        table [x=horizon, y=mean, col sep=comma] {\getErrorDataFilePath{\resolutionMonth}{\tabicl}{\errorFolder}{\Econ}{\mape}};
\addlegendentry{TabICL};

\addplot[name path=lower_econ_linear_month, fill=none, draw=none, mark=none, forget plot]
    table[x=horizon, y=low_0.95, col sep=comma, header=true]{\getErrorDataFilePath{\resolutionMonth}{\linear}{\errorFolder}{\Econ}{\mape}};
\addplot[name path=upper_econ_linear_month, fill=none, draw=none, mark=none, forget plot]
    table[x=horizon, y=high_0.95, col sep=comma, header=true]{\getErrorDataFilePath{\resolutionMonth}{\linear}{\errorFolder}{\Econ}{\mape}};
\addplot[orange!40, fill opacity=0.75, forget plot] fill between[of=lower_econ_linear_month and upper_econ_linear_month];

\addplot[name path=lower_econ_xgboost_month, fill=none, draw=none, mark=none, forget plot]
    table[x=horizon, y=low_0.95, col sep=comma, header=true]{\getErrorDataFilePath{\resolutionMonth}{\xgboost}{\errorFolder}{\Econ}{\mape}};
\addplot[name path=upper_econ_xgboost_month, fill=none, draw=none, mark=none, forget plot]
    table[x=horizon, y=high_0.95, col sep=comma, header=true]{\getErrorDataFilePath{\resolutionMonth}{\xgboost}{\errorFolder}{\Econ}{\mape}};
\addplot[spicy_paprika!40, fill opacity=0.75, forget plot] fill between[of=lower_econ_xgboost_month and upper_econ_xgboost_month];

\addplot[name path=lower_econ_tabICL_month, fill=none, draw=none, mark=none, forget plot]
    table[x=horizon, y=low_0.95, col sep=comma, header=true]{\getErrorDataFilePath{\resolutionMonth}{\tabicl}{\errorFolder}{\Econ}{\mape}};
\addplot[name path=upper_econ_tabICL_month, fill=none, draw=none, mark=none, forget plot]
    table[x=horizon, y=high_0.95, col sep=comma, header=true]{\getErrorDataFilePath{\resolutionMonth}{\tabicl}{\errorFolder}{\Econ}{\mape}};
\addplot[twilight_indigo!40, fill opacity=0.75, forget plot] fill between[of=lower_econ_tabICL_month and upper_econ_tabICL_month];

\nextgroupplot[title={\textbf{Without Economic | Day}}, xlabel={Horizon [M]}, ylabel={$MAPE$}]
\addplot[mark=none, line width=1pt, color=orange, style = dashed]
        table [x=horizon, y=mean, col sep=comma] {\getErrorDataFilePath{\resolutionDay}{\linear}{\errorFolder}{\noEcon}{\mape}};
\addplot[mark=none, line width=1pt, color=spicy_paprika, style = dashed]
        table [x=horizon, y=mean, col sep=comma] {\getErrorDataFilePath{\resolutionDay}{\xgboost}{\errorFolder}{\noEcon}{\mape}};
\addplot[mark=none, line width=1pt, color=twilight_indigo, style = dashed]
        table [x=horizon, y=mean, col sep=comma] {\getErrorDataFilePath{\resolutionDay}{\tabicl}{\errorFolder}{\noEcon}{\mape}};
           

\addplot[name path=lower_no_econ_linear_day, fill=none, draw=none, mark=none, forget plot]
    table[x=horizon, y=low_0.95, col sep=comma, header=true]{\getErrorDataFilePath{\resolutionDay}{\linear}{\errorFolder}{\noEcon}{\mape}};
\addplot[name path=upper_no_econ_linear_day, fill=none, draw=none, mark=none, forget plot]
    table[x=horizon, y=high_0.95, col sep=comma, header=true]{\getErrorDataFilePath{\resolutionDay}{\linear}{\errorFolder}{\noEcon}{\mape}};
\addplot[orange!40, fill opacity=0.75, forget plot] fill between[of=lower_no_econ_linear_day and upper_no_econ_linear_day];

\addplot[name path=lower_no_econ_xgboost_day, fill=none, draw=none, mark=none, forget plot]
    table[x=horizon, y=low_0.95, col sep=comma, header=true]{\getErrorDataFilePath{\resolutionDay}{\xgboost}{\errorFolder}{\noEcon}{\mape}};
\addplot[name path=upper_no_econ_xgboost_day, fill=none, draw=none, mark=none, forget plot]
    table[x=horizon, y=high_0.95, col sep=comma, header=true]{\getErrorDataFilePath{\resolutionDay}{\xgboost}{\errorFolder}{\noEcon}{\mape}};
\addplot[spicy_paprika!40, fill opacity=0.75, forget plot] fill between[of=lower_no_econ_xgboost_day and upper_no_econ_xgboost_day];

\addplot[name path=lower_no_econ_tabICL_day, fill=none, draw=none, mark=none, forget plot]
    table[x=horizon, y=low_0.95, col sep=comma, header=true]{\getErrorDataFilePath{\resolutionDay}{\tabicl}{\errorFolder}{\noEcon}{\mape}};
\addplot[name path=upper_no_econ_tabICL_day, fill=none, draw=none, mark=none, forget plot]
    table[x=horizon, y=high_0.95, col sep=comma, header=true]{\getErrorDataFilePath{\resolutionDay}{\tabicl}{\errorFolder}{\noEcon}{\mape}};
\addplot[twilight_indigo!40, fill opacity=0.75, forget plot] fill between[of=lower_no_econ_tabICL_day and upper_no_econ_tabICL_day];

\nextgroupplot[ymin = 1.5, ymax=6.5, title={\textbf{With Economic | Day}}, xlabel={Horizon [M]}, ylabel={$MAPE$}]
\addplot[mark=none, line width=1pt, color=orange]
        table [x=horizon, y=mean, col sep=comma] {\getErrorDataFilePath{\resolutionDay}{\linear}{\errorFolder}{\Econ}{\mape}};
\addplot[mark=none, line width=1pt, color=spicy_paprika]
        table [x=horizon, y=mean, col sep=comma] {\getErrorDataFilePath{\resolutionDay}{\xgboost}{\errorFolder}{\Econ}{\mape}};
\addplot[mark=none, line width=1pt, color=twilight_indigo]
        table [x=horizon, y=mean, col sep=comma] {\getErrorDataFilePath{\resolutionDay}{\tabicl}{\errorFolder}{\Econ}{\mape}};
\addplot[name path=lower_econ_linear_day, fill=none, draw=none, mark=none, forget plot]
    table[x=horizon, y=low_0.95, col sep=comma, header=true]{\getErrorDataFilePath{\resolutionDay}{\linear}{\errorFolder}{\Econ}{\mape}};
\addplot[name path=upper_econ_linear_day, fill=none, draw=none, mark=none, forget plot]
    table[x=horizon, y=high_0.95, col sep=comma, header=true]{\getErrorDataFilePath{\resolutionDay}{\linear}{\errorFolder}{\Econ}{\mape}};
\addplot[orange!40, fill opacity=0.75, forget plot] fill between[of=lower_econ_linear_day and upper_econ_linear_day];

\addplot[name path=lower_econ_xgboost_day, fill=none, draw=none, mark=none, forget plot]
    table[x=horizon, y=low_0.95, col sep=comma, header=true]{\getErrorDataFilePath{\resolutionDay}{\xgboost}{\errorFolder}{\Econ}{\mape}};
\addplot[name path=upper_econ_xgboost_day, fill=none, draw=none, mark=none, forget plot]
    table[x=horizon, y=high_0.95, col sep=comma, header=true]{\getErrorDataFilePath{\resolutionDay}{\xgboost}{\errorFolder}{\Econ}{\mape}};
\addplot[spicy_paprika!40, fill opacity=0.75, forget plot] fill between[of=lower_econ_xgboost_day and upper_econ_xgboost_day];

\addplot[name path=lower_econ_tabICL_day, fill=none, draw=none, mark=none, forget plot]
    table[x=horizon, y=low_0.95, col sep=comma, header=true]{\getErrorDataFilePath{\resolutionDay}{\tabicl}{\errorFolder}{\Econ}{\mape}};
\addplot[name path=upper_econ_tabICL_day, fill=none, draw=none, mark=none, forget plot]
    table[x=horizon, y=high_0.95, col sep=comma, header=true]{\getErrorDataFilePath{\resolutionDay}{\tabicl}{\errorFolder}{\Econ}{\mape}};
\addplot[twilight_indigo!40, fill opacity=0.75, forget plot] fill between[of=lower_econ_tabICL_day and upper_econ_tabICL_day];
 
\end{groupplot}

\end{tikzpicture}

%% file: Plots/RMSE_horizon_covid.tex

\pgfplotstableread[col sep=comma]{\getCrisisErrorDataFilePath{\resolutionMonth}{\linear}{\errorFolder}{\Econ}{\rmse}{\Covid}}\dataRMSELinearMonthEconCovid
\pgfplotstableread[col sep=comma]{\getCrisisErrorDataFilePath{\resolutionMonth}{\linear}{\errorFolder}{\noEcon}{\rmse}{\Covid}}\dataRMSELinearMonthNoEconCovid

\pgfplotstableread[col sep=comma]{\getCrisisErrorDataFilePath{\resolutionDay}{\linear}{\errorFolder}{\Econ}{\rmse}{\Covid}}\dataRMSELinearDayEconCovid
\pgfplotstableread[col sep=comma]{\getCrisisErrorDataFilePath{\resolutionDay}{\linear}{\errorFolder}{\noEcon}{\rmse}{\Covid}}\dataRMSELinearDayNoEconCovid

\pgfplotstableread[col sep=comma]{\getCrisisErrorDataFilePath{\resolutionMonth}{\xgboost}{\errorFolder}{\Econ}{\rmse}{\Covid}}\dataRMSEXGBoostMonthEconCovid
\pgfplotstableread[col sep=comma]{\getCrisisErrorDataFilePath{\resolutionMonth}{\xgboost}{\errorFolder}{\noEcon}{\rmse}{\Covid}}\dataRMSEXGBoostMonthNoEconCovid

\pgfplotstableread[col sep=comma]{\getCrisisErrorDataFilePath{\resolutionDay}{\xgboost}{\errorFolder}{\Econ}{\rmse}{\Covid}}\dataRMSEXGBoostDayEconCovid
\pgfplotstableread[col sep=comma]{\getCrisisErrorDataFilePath{\resolutionDay}{\xgboost}{\errorFolder}{\noEcon}{\rmse}{\Covid}}\dataRMSEXGBoostDayNoEconCovid

\pgfplotstableread[col sep=comma]{\getCrisisErrorDataFilePath{\resolutionMonth}{\tabicl}{\errorFolder}{\Econ}{\rmse}{\Covid}}\dataRMSETabICLMonthEconCovid
\pgfplotstableread[col sep=comma]{\getCrisisErrorDataFilePath{\resolutionMonth}{\tabicl}{\errorFolder}{\noEcon}{\rmse}{\Covid}}\dataRMSETabICLMonthNoEconCovid

\pgfplotstableread[col sep=comma]{\getCrisisErrorDataFilePath{\resolutionDay}{\tabicl}{\errorFolder}{\Econ}{\rmse}{\Covid}}\dataRMSETabICLDayEconCovid
\pgfplotstableread[col sep=comma]{\getCrisisErrorDataFilePath{\resolutionDay}{\tabicl}{\errorFolder}{\noEcon}{\rmse}{\Covid}}\dataRMSETabICLDayNoEconCovid

\begin{tikzpicture}
    \begin{groupplot}[
    group style={
        group size=3 by 2,        
        horizontal sep=1.8cm,
        vertical sep=1.8cm,
    },
    width=6cm,
    height=5cm,
    grid=major,
    scaled ticks=false,
    xtick={1, 12, 24, 36, 48},
    xticklabels={1, 12, 24, 36, 48},
    legend columns = 2,
    legend style={at={(1.15,-2.1)}, anchor=south west},
]
 
\nextgroupplot[title={\textbf{Linear | Month}}, xlabel={Horizon [M]}, ylabel={$RMSE$}]
\addplot[mark=none, line width=1pt, color=light_indigo_purple]
        table [x=horizon, y=mean, col sep=comma] {\dataRMSELinearMonthEconCovid};
\addlegendentry{With Economic};
\addplot[mark=none, line width=1pt, color=light_light_apricot]
        table [x=horizon, y=mean, col sep=comma] {\dataRMSELinearMonthNoEconCovid};
\addlegendentry{Without Economic};
\addplot[name path=lower_econ_linear_month_covid, fill=none, draw=none, mark=none, forget plot]
    table[x=horizon, y=low_0.95, col sep=comma, header=true]{\dataRMSELinearMonthEconCovid};
\addplot[name path=upper_econ_linear_month_covid, fill=none, draw=none, mark=none, forget plot]
    table[x=horizon, y=high_0.95, col sep=comma, header=true]{\dataRMSELinearMonthEconCovid};
\addplot[light_indigo_purple!40, fill opacity=0.75, forget plot] fill between[of=lower_econ_linear_month_covid and upper_econ_linear_month_covid];

\addplot[name path=lower_no_econ_linear_month_covid, fill=none, draw=none, mark=none, forget plot]
    table[x=horizon, y=low_0.95, col sep=comma, header=true]{\dataRMSELinearMonthNoEconCovid};
\addplot[name path=upper_no_econ_linear_month_covid, fill=none, draw=none, mark=none, forget plot]
    table[x=horizon, y=high_0.95, col sep=comma, header=true]{\dataRMSELinearMonthNoEconCovid};
\addplot[light_light_apricot!40, fill opacity=0.75, forget plot] fill between[of=lower_no_econ_linear_month_covid and upper_no_econ_linear_month_covid];

\nextgroupplot[title={\textbf{XGBoost | Month}}, xlabel={Horizon [M]}, ylabel={$RMSE$}]
\addplot[mark=none, line width=1pt, color=light_indigo_purple]
        table [x=horizon, y=mean, col sep=comma] {\dataRMSEXGBoostMonthEconCovid};
\addplot[mark=none, line width=1pt, color=light_light_apricot]
        table [x=horizon, y=mean, col sep=comma] {\dataRMSEXGBoostMonthNoEconCovid};
\addplot[name path=lower_econ_xgboost_month_covid, fill=none, draw=none, mark=none, forget plot]
    table[x=horizon, y=low_0.95, col sep=comma, header=true]{\dataRMSEXGBoostMonthEconCovid};
\addplot[name path=upper_econ_xgboost_month_covid, fill=none, draw=none, mark=none, forget plot]
    table[x=horizon, y=high_0.95, col sep=comma, header=true]{\dataRMSEXGBoostMonthEconCovid};
\addplot[light_indigo_purple!40, fill opacity=0.75, forget plot] fill between[of=lower_econ_xgboost_month_covid and upper_econ_xgboost_month_covid];

\addplot[name path=lower_no_econ_xgboost_month_covid, fill=none, draw=none, mark=none, forget plot]
    table[x=horizon, y=low_0.95, col sep=comma, header=true]{\dataRMSEXGBoostMonthNoEconCovid};
\addplot[name path=upper_no_econ_xgboost_month_covid, fill=none, draw=none, mark=none, forget plot]
    table[x=horizon, y=high_0.95, col sep=comma, header=true]{\dataRMSEXGBoostMonthNoEconCovid};
\addplot[light_light_apricot!40, fill opacity=0.75, forget plot] fill between[of=lower_no_econ_xgboost_month_covid and upper_no_econ_xgboost_month_covid];

\nextgroupplot[title={\textbf{TabICL | Month}}, xlabel={Horizon [M]}, ylabel={$RMSE$}]
\addplot[mark=none, line width=1pt, color=light_indigo_purple]
        table [x=horizon, y=mean, col sep=comma] {\dataRMSETabICLMonthEconCovid};
\addplot[mark=none, line width=1pt, color=light_light_apricot]
        table [x=horizon, y=mean, col sep=comma] {\dataRMSETabICLMonthNoEconCovid};
\addplot[name path=lower_econ_tabICL_month_covid, fill=none, draw=none, mark=none, forget plot]
    table[x=horizon, y=low_0.95, col sep=comma, header=true]{\dataRMSETabICLMonthEconCovid};
\addplot[name path=upper_econ_tabICL_month_covid, fill=none, draw=none, mark=none, forget plot]
    table[x=horizon, y=high_0.95, col sep=comma, header=true]{\dataRMSETabICLMonthEconCovid};
\addplot[light_indigo_purple!40, fill opacity=0.75, forget plot] fill between[of=lower_econ_tabICL_month_covid and upper_econ_tabICL_month_covid];

\addplot[name path=lower_no_econ_tabICL_month_covid, fill=none, draw=none, mark=none, forget plot]
    table[x=horizon, y=low_0.95, col sep=comma, header=true]{\dataRMSETabICLMonthNoEconCovid};
\addplot[name path=upper_no_econ_tabICL_month_covid, fill=none, draw=none, mark=none, forget plot]
    table[x=horizon, y=high_0.95, col sep=comma, header=true]{\dataRMSETabICLMonthNoEconCovid};
\addplot[light_light_apricot!40, fill opacity=0.75, forget plot] fill between[of=lower_no_econ_tabICL_month_covid and upper_no_econ_tabICL_month_covid];

\nextgroupplot[title={\textbf{Linear | Day}}, xlabel={Horizon [M]}, ylabel={$RMSE$}]
\addplot[mark=none, line width=1pt, color=light_indigo_purple]
        table [x=horizon, y=mean, col sep=comma] {\dataRMSELinearDayEconCovid};
\addplot[mark=none, line width=1pt, color=light_light_apricot]
        table [x=horizon, y=mean, col sep=comma] {\dataRMSELinearDayNoEconCovid};
\addplot[name path=lower_econ_linear_day_covid, fill=none, draw=none, mark=none, forget plot]
    table[x=horizon, y=low_0.95, col sep=comma, header=true]{\dataRMSELinearDayEconCovid};
\addplot[name path=upper_econ_linear_day_covid, fill=none, draw=none, mark=none, forget plot]
    table[x=horizon, y=high_0.95, col sep=comma, header=true]{\dataRMSELinearDayEconCovid};
\addplot[light_indigo_purple!40, fill opacity=0.75, forget plot] fill between[of=lower_econ_linear_day_covid and upper_econ_linear_day_covid];

\addplot[name path=lower_no_econ_linear_day_covid, fill=none, draw=none, mark=none, forget plot]
    table[x=horizon, y=low_0.95, col sep=comma, header=true]{\dataRMSELinearDayNoEconCovid};
\addplot[name path=upper_no_econ_linear_day_covid, fill=none, draw=none, mark=none, forget plot]
    table[x=horizon, y=high_0.95, col sep=comma, header=true]{\dataRMSELinearDayNoEconCovid};
\addplot[light_light_apricot!40, fill opacity=0.75, forget plot] fill between[of=lower_no_econ_linear_day_covid and upper_no_econ_linear_day_covid];

\nextgroupplot[title={\textbf{XGBoost | Day}}, xlabel={Horizon [M]}, ylabel={$RMSE$}]
\addplot[mark=none, line width=1pt, color=light_indigo_purple]
        table [x=horizon, y=mean, col sep=comma] {\dataRMSEXGBoostDayEconCovid};
\addplot[mark=none, line width=1pt, color=light_light_apricot]
        table [x=horizon, y=mean, col sep=comma] {\dataRMSEXGBoostDayNoEconCovid};
\addplot[name path=lower_econ_xgboost_day_covid, fill=none, draw=none, mark=none, forget plot]
    table[x=horizon, y=low_0.95, col sep=comma, header=true]{\dataRMSEXGBoostDayEconCovid};
\addplot[name path=upper_econ_xgboost_day_covid, fill=none, draw=none, mark=none, forget plot]
    table[x=horizon, y=high_0.95, col sep=comma, header=true]{\dataRMSEXGBoostDayEconCovid};
\addplot[light_indigo_purple!40, fill opacity=0.75, forget plot] fill between[of=lower_econ_xgboost_day_covid and upper_econ_xgboost_day_covid];

\addplot[name path=lower_no_econ_xgboost_day_covid, fill=none, draw=none, mark=none, forget plot]
    table[x=horizon, y=low_0.95, col sep=comma, header=true]{\dataRMSEXGBoostDayNoEconCovid};
\addplot[name path=upper_no_econ_xgboost_day_covid, fill=none, draw=none, mark=none, forget plot]
    table[x=horizon, y=high_0.95, col sep=comma, header=true]{\dataRMSEXGBoostDayNoEconCovid};
\addplot[light_light_apricot!40, fill opacity=0.75, forget plot] fill between[of=lower_no_econ_xgboost_day_covid and upper_no_econ_xgboost_day_covid];

\nextgroupplot[title={\textbf{TabICL | Day}}, xlabel={Horizon [M]}, ylabel={$RMSE$}]
\addplot[mark=none, line width=1pt, color=light_indigo_purple]
        table [x=horizon, y=mean, col sep=comma] {\dataRMSETabICLDayEconCovid};
\addplot[mark=none, line width=1pt, color=light_light_apricot]
        table [x=horizon, y=mean, col sep=comma] {\dataRMSETabICLDayNoEconCovid};
\addplot[name path=lower_econ_tabICL_day_covid, fill=none, draw=none, mark=none, forget plot]
    table[x=horizon, y=low_0.95, col sep=comma, header=true]{\dataRMSETabICLDayEconCovid};
\addplot[name path=upper_econ_tabICL_day_covid, fill=none, draw=none, mark=none, forget plot]
    table[x=horizon, y=high_0.95, col sep=comma, header=true]{\dataRMSETabICLDayEconCovid};
\addplot[light_indigo_purple!40, fill opacity=0.75, forget plot] fill between[of=lower_econ_tabICL_day_covid and upper_econ_tabICL_day_covid];

\addplot[name path=lower_no_econ_tabICL_day_covid, fill=none, draw=none, mark=none, forget plot]
    table[x=horizon, y=low_0.95, col sep=comma, header=true]{\dataRMSETabICLDayNoEconCovid};
\addplot[name path=upper_no_econ_tabICL_day_covid, fill=none, draw=none, mark=none, forget plot]
    table[x=horizon, y=high_0.95, col sep=comma, header=true]{\dataRMSETabICLDayNoEconCovid};
\addplot[light_light_apricot!40, fill opacity=0.75, forget plot] fill between[of=lower_no_econ_tabICL_day_covid and upper_no_econ_tabICL_day_covid];
 
\end{groupplot}

\end{tikzpicture}

%% file: Plots/RMSE_horizon_sobriety.tex

\pgfplotstableread[col sep=comma]{\getCrisisErrorDataFilePath{\resolutionMonth}{\linear}{\errorFolder}{\Econ}{\rmse}{\Sobriety}}\dataRMSELinearMonthEconSobriety
\pgfplotstableread[col sep=comma]{\getCrisisErrorDataFilePath{\resolutionMonth}{\linear}{\errorFolder}{\noEcon}{\rmse}{\Sobriety}}\dataRMSELinearMonthNoEconSobriety

\pgfplotstableread[col sep=comma]{\getCrisisErrorDataFilePath{\resolutionDay}{\linear}{\errorFolder}{\Econ}{\rmse}{\Sobriety}}\dataRMSELinearDayEconSobriety
\pgfplotstableread[col sep=comma]{\getCrisisErrorDataFilePath{\resolutionDay}{\linear}{\errorFolder}{\noEcon}{\rmse}{\Sobriety}}\dataRMSELinearDayNoEconSobriety

\pgfplotstableread[col sep=comma]{\getCrisisErrorDataFilePath{\resolutionMonth}{\xgboost}{\errorFolder}{\Econ}{\rmse}{\Sobriety}}\dataRMSEXGBoostMonthEconSobriety
\pgfplotstableread[col sep=comma]{\getCrisisErrorDataFilePath{\resolutionMonth}{\xgboost}{\errorFolder}{\noEcon}{\rmse}{\Sobriety}}\dataRMSEXGBoostMonthNoEconSobriety

\pgfplotstableread[col sep=comma]{\getCrisisErrorDataFilePath{\resolutionDay}{\xgboost}{\errorFolder}{\Econ}{\rmse}{\Sobriety}}\dataRMSEXGBoostDayEconSobriety
\pgfplotstableread[col sep=comma]{\getCrisisErrorDataFilePath{\resolutionDay}{\xgboost}{\errorFolder}{\noEcon}{\rmse}{\Sobriety}}\dataRMSEXGBoostDayNoEconSobriety

\pgfplotstableread[col sep=comma]{\getCrisisErrorDataFilePath{\resolutionMonth}{\tabicl}{\errorFolder}{\Econ}{\rmse}{\Sobriety}}\dataRMSETabICLMonthEconSobriety
\pgfplotstableread[col sep=comma]{\getCrisisErrorDataFilePath{\resolutionMonth}{\tabicl}{\errorFolder}{\noEcon}{\rmse}{\Sobriety}}\dataRMSETabICLMonthNoEconSobriety

\pgfplotstableread[col sep=comma]{\getCrisisErrorDataFilePath{\resolutionDay}{\tabicl}{\errorFolder}{\Econ}{\rmse}{\Sobriety}}\dataRMSETabICLDayEconSobriety
\pgfplotstableread[col sep=comma]{\getCrisisErrorDataFilePath{\resolutionDay}{\tabicl}{\errorFolder}{\noEcon}{\rmse}{\Sobriety}}\dataRMSETabICLDayNoEconSobriety

\begin{tikzpicture}
    \begin{groupplot}[
    group style={
        group size=3 by 2,        
        horizontal sep=1.8cm,
        vertical sep=1.8cm,
    },
    width=6cm,
    height=5cm,
    grid=major,
    scaled ticks=false,
    xtick={1, 12, 24, 36, 48},
    xticklabels={1, 12, 24, 36, 48},
    legend columns = 2,
    legend style={at={(1.15,-2.1)}, anchor=south west},
]
 
\nextgroupplot[title={\textbf{Linear | Month}}, xlabel={Horizon [M]}, ylabel={$RMSE$}]
\addplot[mark=none, line width=1pt, color=light_indigo_purple]
        table [x=horizon, y=mean, col sep=comma] {\dataRMSELinearMonthEconSobriety};
\addlegendentry{With Economic};
\addplot[mark=none, line width=1pt, color=light_light_apricot]
        table [x=horizon, y=mean, col sep=comma] {\dataRMSELinearMonthNoEconSobriety};
\addlegendentry{Without Economic};
\addplot[name path=lower_econ_linear_month_sobriety, fill=none, draw=none, mark=none, forget plot]
    table[x=horizon, y=low_0.95, col sep=comma, header=true]{\dataRMSELinearMonthEconSobriety};
\addplot[name path=upper_econ_linear_month_sobriety, fill=none, draw=none, mark=none, forget plot]
    table[x=horizon, y=high_0.95, col sep=comma, header=true]{\dataRMSELinearMonthEconSobriety};
\addplot[light_indigo_purple!40, fill opacity=0.75, forget plot] fill between[of=lower_econ_linear_month_sobriety and upper_econ_linear_month_sobriety];

\addplot[name path=lower_no_econ_linear_month_sobriety, fill=none, draw=none, mark=none, forget plot]
    table[x=horizon, y=low_0.95, col sep=comma, header=true]{\dataRMSELinearMonthNoEconSobriety};
\addplot[name path=upper_no_econ_linear_month_sobriety, fill=none, draw=none, mark=none, forget plot]
    table[x=horizon, y=high_0.95, col sep=comma, header=true]{\dataRMSELinearMonthNoEconSobriety};
\addplot[light_light_apricot!40, fill opacity=0.75, forget plot] fill between[of=lower_no_econ_linear_month_sobriety and upper_no_econ_linear_month_sobriety];

\nextgroupplot[title={\textbf{XGBoost | Month}}, xlabel={Horizon [M]}, ylabel={$RMSE$}]
\addplot[mark=none, line width=1pt, color=light_indigo_purple]
        table [x=horizon, y=mean, col sep=comma] {\dataRMSEXGBoostMonthEconSobriety};
\addplot[mark=none, line width=1pt, color=light_light_apricot]
        table [x=horizon, y=mean, col sep=comma] {\dataRMSEXGBoostMonthNoEconSobriety};
\addplot[name path=lower_econ_xgboost_month_sobriety, fill=none, draw=none, mark=none, forget plot]
    table[x=horizon, y=low_0.95, col sep=comma, header=true]{\dataRMSEXGBoostMonthEconSobriety};
\addplot[name path=upper_econ_xgboost_month_sobriety, fill=none, draw=none, mark=none, forget plot]
    table[x=horizon, y=high_0.95, col sep=comma, header=true]{\dataRMSEXGBoostMonthEconSobriety};
\addplot[light_indigo_purple!40, fill opacity=0.75, forget plot] fill between[of=lower_econ_xgboost_month_sobriety and upper_econ_xgboost_month_sobriety];

\addplot[name path=lower_no_econ_xgboost_month_sobriety, fill=none, draw=none, mark=none, forget plot]
    table[x=horizon, y=low_0.95, col sep=comma, header=true]{\dataRMSEXGBoostMonthNoEconSobriety};
\addplot[name path=upper_no_econ_xgboost_month_sobriety, fill=none, draw=none, mark=none, forget plot]
    table[x=horizon, y=high_0.95, col sep=comma, header=true]{\dataRMSEXGBoostMonthNoEconSobriety};
\addplot[light_light_apricot!40, fill opacity=0.75, forget plot] fill between[of=lower_no_econ_xgboost_month_sobriety and upper_no_econ_xgboost_month_sobriety];

\nextgroupplot[title={\textbf{TabICL | Month}}, xlabel={Horizon [M]}, ylabel={$RMSE$}]
\addplot[mark=none, line width=1pt, color=light_indigo_purple]
        table [x=horizon, y=mean, col sep=comma] {\dataRMSETabICLMonthEconSobriety};
\addplot[mark=none, line width=1pt, color=light_light_apricot]
        table [x=horizon, y=mean, col sep=comma] {\dataRMSETabICLMonthNoEconSobriety};
\addplot[name path=lower_econ_tabICL_month_sobriety, fill=none, draw=none, mark=none, forget plot]
    table[x=horizon, y=low_0.95, col sep=comma, header=true]{\dataRMSETabICLMonthEconSobriety};
\addplot[name path=upper_econ_tabICL_month_sobriety, fill=none, draw=none, mark=none, forget plot]
    table[x=horizon, y=high_0.95, col sep=comma, header=true]{\dataRMSETabICLMonthEconSobriety};
\addplot[light_indigo_purple!40, fill opacity=0.75, forget plot] fill between[of=lower_econ_tabICL_month_sobriety and upper_econ_tabICL_month_sobriety];

\addplot[name path=lower_no_econ_tabICL_month_sobriety, fill=none, draw=none, mark=none, forget plot]
    table[x=horizon, y=low_0.95, col sep=comma, header=true]{\dataRMSETabICLMonthNoEconSobriety};
\addplot[name path=upper_no_econ_tabICL_month_sobriety, fill=none, draw=none, mark=none, forget plot]
    table[x=horizon, y=high_0.95, col sep=comma, header=true]{\dataRMSETabICLMonthNoEconSobriety};
\addplot[light_light_apricot!40, fill opacity=0.75, forget plot] fill between[of=lower_no_econ_tabICL_month_sobriety and upper_no_econ_tabICL_month_sobriety];

\nextgroupplot[title={\textbf{Linear | Day}}, xlabel={Horizon [M]}, ylabel={$RMSE$}]
\addplot[mark=none, line width=1pt, color=light_indigo_purple]
        table [x=horizon, y=mean, col sep=comma] {\dataRMSELinearDayEconSobriety};
\addplot[mark=none, line width=1pt, color=light_light_apricot]
        table [x=horizon, y=mean, col sep=comma] {\dataRMSELinearDayNoEconSobriety};
\addplot[name path=lower_econ_linear_day_sobriety, fill=none, draw=none, mark=none, forget plot]
    table[x=horizon, y=low_0.95, col sep=comma, header=true]{\dataRMSELinearDayEconSobriety};
\addplot[name path=upper_econ_linear_day_sobriety, fill=none, draw=none, mark=none, forget plot]
    table[x=horizon, y=high_0.95, col sep=comma, header=true]{\dataRMSELinearDayEconSobriety};
\addplot[light_indigo_purple!40, fill opacity=0.75, forget plot] fill between[of=lower_econ_linear_day_sobriety and upper_econ_linear_day_sobriety];

\addplot[name path=lower_no_econ_linear_day_sobriety, fill=none, draw=none, mark=none, forget plot]
    table[x=horizon, y=low_0.95, col sep=comma, header=true]{\dataRMSELinearDayNoEconSobriety};
\addplot[name path=upper_no_econ_linear_day_sobriety, fill=none, draw=none, mark=none, forget plot]
    table[x=horizon, y=high_0.95, col sep=comma, header=true]{\dataRMSELinearDayNoEconSobriety};
\addplot[light_light_apricot!40, fill opacity=0.75, forget plot] fill between[of=lower_no_econ_linear_day_sobriety and upper_no_econ_linear_day_sobriety];

\nextgroupplot[title={\textbf{XGBoost | Day}}, xlabel={Horizon [M]}, ylabel={$RMSE$}]
\addplot[mark=none, line width=1pt, color=light_indigo_purple]
        table [x=horizon, y=mean, col sep=comma] {\dataRMSEXGBoostDayEconSobriety};
\addplot[mark=none, line width=1pt, color=light_light_apricot]
        table [x=horizon, y=mean, col sep=comma] {\dataRMSEXGBoostDayNoEconSobriety};
\addplot[name path=lower_econ_xgboost_day_sobriety, fill=none, draw=none, mark=none, forget plot]
    table[x=horizon, y=low_0.95, col sep=comma, header=true]{\dataRMSEXGBoostDayEconSobriety};
\addplot[name path=upper_econ_xgboost_day_sobriety, fill=none, draw=none, mark=none, forget plot]
    table[x=horizon, y=high_0.95, col sep=comma, header=true]{\dataRMSEXGBoostDayEconSobriety};
\addplot[light_indigo_purple!40, fill opacity=0.75, forget plot] fill between[of=lower_econ_xgboost_day_sobriety and upper_econ_xgboost_day_sobriety];

\addplot[name path=lower_no_econ_xgboost_day_sobriety, fill=none, draw=none, mark=none, forget plot]
    table[x=horizon, y=low_0.95, col sep=comma, header=true]{\dataRMSEXGBoostDayNoEconSobriety};
\addplot[name path=upper_no_econ_xgboost_day_sobriety, fill=none, draw=none, mark=none, forget plot]
    table[x=horizon, y=high_0.95, col sep=comma, header=true]{\dataRMSEXGBoostDayNoEconSobriety};
\addplot[light_light_apricot!40, fill opacity=0.75, forget plot] fill between[of=lower_no_econ_xgboost_day_sobriety and upper_no_econ_xgboost_day_sobriety];

\nextgroupplot[title={\textbf{TabICL | Day}}, xlabel={Horizon [M]}, ylabel={$RMSE$}]
\addplot[mark=none, line width=1pt, color=light_indigo_purple]
        table [x=horizon, y=mean, col sep=comma] {\dataRMSETabICLDayEconSobriety};
\addplot[mark=none, line width=1pt, color=light_light_apricot]
        table [x=horizon, y=mean, col sep=comma] {\dataRMSETabICLDayNoEconSobriety};
\addplot[name path=lower_econ_tabICL_day_sobriety, fill=none, draw=none, mark=none, forget plot]
    table[x=horizon, y=low_0.95, col sep=comma, header=true]{\dataRMSETabICLDayEconSobriety};
\addplot[name path=upper_econ_tabICL_day_sobriety, fill=none, draw=none, mark=none, forget plot]
    table[x=horizon, y=high_0.95, col sep=comma, header=true]{\dataRMSETabICLDayEconSobriety};
\addplot[light_indigo_purple!40, fill opacity=0.75, forget plot] fill between[of=lower_econ_tabICL_day_sobriety and upper_econ_tabICL_day_sobriety];

\addplot[name path=lower_no_econ_tabICL_day_sobriety, fill=none, draw=none, mark=none, forget plot]
    table[x=horizon, y=low_0.95, col sep=comma, header=true]{\dataRMSETabICLDayNoEconSobriety};
\addplot[name path=upper_no_econ_tabICL_day_sobriety, fill=none, draw=none, mark=none, forget plot]
    table[x=horizon, y=high_0.95, col sep=comma, header=true]{\dataRMSETabICLDayNoEconSobriety};
\addplot[light_light_apricot!40, fill opacity=0.75, forget plot] fill between[of=lower_no_econ_tabICL_day_sobriety and upper_no_econ_tabICL_day_sobriety];
 
\end{groupplot}

\end{tikzpicture}

%% file: Plots/selected_features_day.tex





\pgfplotstableread[col sep=comma]{\getSelectedFeaturesDataFilePath{day}{linear}{selected_features}{econ}{calendar}}\dataCalendarLinear
\pgfplotstableread[col sep=comma]{\getSelectedFeaturesDataFilePath{day}{xgboost}{selected_features}{econ}{calendar}}\dataCalendarXGBoost
\pgfplotstableread[col sep=comma]{\getSelectedFeaturesDataFilePath{day}{tabICL}{selected_features}{econ}{calendar}}\dataCalendarTabICL
\pgfplotstableread[col sep=comma]{\getSelectedFeaturesDataFilePath{day}{linear}{selected_features}{econ}{weather}}\dataWeatherLinear
\pgfplotstableread[col sep=comma]{\getSelectedFeaturesDataFilePath{day}{xgboost}{selected_features}{econ}{weather}}\dataWeatherXGBoost
\pgfplotstableread[col sep=comma]{\getSelectedFeaturesDataFilePath{day}{tabICL}{selected_features}{econ}{weather}}\dataWeatherTabICL
\pgfplotstableread[col sep=comma]{\getSelectedFeaturesDataFilePath{day}{linear}{selected_features}{econ}{economic}}\dataEconomicLinear
\pgfplotstableread[col sep=comma]{\getSelectedFeaturesDataFilePath{day}{xgboost}{selected_features}{econ}{economic}}\dataEconomicXGBoost
\pgfplotstableread[col sep=comma]{\getSelectedFeaturesDataFilePath{day}{tabICL}{selected_features}{econ}{economic}}\dataEconomicTabICL

\begin{tikzpicture}

\begin{axis}[
    ybar,
    bar width=5pt,
    width=8.5cm,
    height=6.5cm,
    x=1cm,
    ymin=0,
    ymax=1.15,
    ymajorgrids,
    axis lines*=left,
    ylabel={Selection Ratio},
    enlarge x limits=0.075,
    xtick=data,
    x tick label style={rotate=45, anchor=east},
    name=plotCalendar,
    title={\large\textbf{Calendar Features}},
    at={(0,0)},
    anchor=north west,
    symbolic x coords={doy\_cos, doy\_sin, dow\_cos, dow\_sin, moy\_cos, moy\_sin,  is\_weekend, is\_public\_holiday, is\_school\_holiday},
    legend columns = 3,
    legend style={at={(0.85,-2.25)}, anchor=north west,
    font=\Large, /tikz/column sep=5pt}
  ]
  \addplot[color = orange, fill=orange, fill opacity=0.75] table[x=features, y=occurence] {\dataCalendarLinear};
  \addplot[color = spicy_paprika, fill=spicy_paprika, fill opacity=0.75]   table[x=features, y=occurence] {\dataCalendarXGBoost};
  \addplot[color = twilight_indigo, , fill=twilight_indigo, fill opacity=0.75]      table[x=features, y=occurence] {\dataCalendarTabICL};
  \legend{Linear, XGBoost, TabICL}
\end{axis}

\begin{axis}[
    ybar,
    bar width=5pt,
    width=8.5cm,
    height=6.5cm,
    x=1cm,
    ymin=0,
    ymax=1.15,
    ymajorgrids,
    axis lines*=left,
    ylabel={Selection Ratio},
    enlarge x limits=0.075,
    xtick=data,
    x tick label style={rotate=45, anchor=east},
    name=plotWeather,
    title={\large\textbf{Weather Features}},
    at={(plotCalendar.north east)},
    anchor=north west,
    xshift=2cm,
    xtick={t2m, hdd, cdd, rh,  ssrd, tcc, tp, u10, v10},
    xticklabels={t2m, hdd, cdd, rh,  ssrd, tcc, tp, u10, v10},
    symbolic x coords={t2m, hdd, cdd, rh,  ssrd, tcc, tp, u10, v10},
  ]
  \addplot[color = orange, fill=orange, fill opacity=0.75]table[x=features, y=occurence] {\dataWeatherLinear};
  \addplot[color = spicy_paprika, fill=spicy_paprika, fill opacity=0.75] table[x=features, y=occurence] {\dataWeatherXGBoost};
  \addplot[color = twilight_indigo, , fill=twilight_indigo, fill opacity=0.75] table[x=features, y=occurence] {\dataWeatherTabICL};
\end{axis}

\begin{axis}[
    ybar,
    bar width=5pt,
    width=8.5cm,
    height=6.5cm,
    x=1cm,
    ymin=0,
    ymax=1.15,
    ymajorgrids,
    axis lines*=left,
    ylabel={Selection Ratio},
    enlarge x limits=0.075,
    xtick=data,
    x tick label style={rotate=45, anchor=east},
    name=plotEconomic,
    title={\large\textbf{Economic Features}},
    at={($(plotCalendar.south west)!0.5!(plotWeather.south east)$)},
    anchor=north,
    yshift=-2.25cm,
    enlarge x limits=0.015,
    xtick={B\_CVS-CJO, C\_CVS-CJO, D\_CVS-CJO, E\_CVS-CJO, H\_CVS-CJO, I\_CVS-CJO, J\_CVS-CJO, L\_CVS-CJO, M\_CVS-CJO, N\_CVS-CJO, S\_CVS-CJO, cpi\_w/o\_energy, ecpi, scpi, employment, population, household\_income\_CVS-CJO, tourism\_nights, PAC\_air/air, PAC\_air/eau, PAC\_geo, electric\_car\_passenger, electric\_light\_utilitary\_vehicle, electric\_heavy\_utilitary\_vehicle, electric\_public\_transports, electric\_vehicles, hybrid\_car\_passenger, hybrid\_light\_utilitary\_vehicle, hybrid\_heavy\_utilitary\_vehicle, hybrid\_public\_transports},
    xticklabels={B\_CVS-CJO, C\_CVS-CJO, D\_CVS-CJO, E\_CVS-CJO, H\_CVS-CJO, I\_CVS-CJO, J\_CVS-CJO, L\_CVS-CJO, M\_CVS-CJO, N\_CVS-CJO, S\_CVS-CJO, cpi\_w/o\_energy, ecpi, scpi, employment, population, household\_income\_CVS-CJO, tourism\_nights, PAC\_air/air, PAC\_air/eau, PAC\_geo, electric\_car\_passenger, electric\_light\_utilitary\_vehicle, electric\_heavy\_utilitary\_vehicle, electric\_public\_transports, electric\_vehicles, hybrid\_car\_passenger, hybrid\_light\_utilitary\_vehicle, hybrid\_heavy\_utilitary\_vehicle, hybrid\_public\_transports},
    symbolic x coords={B\_CVS-CJO, C\_CVS-CJO, D\_CVS-CJO, E\_CVS-CJO, H\_CVS-CJO, I\_CVS-CJO, J\_CVS-CJO, L\_CVS-CJO, M\_CVS-CJO, N\_CVS-CJO, S\_CVS-CJO, cpi\_w/o\_energy, ecpi, scpi, employment, population, household\_income\_CVS-CJO, tourism\_nights, PAC\_air/air, PAC\_air/eau, PAC\_geo, electric\_car\_passenger, electric\_light\_utilitary\_vehicle, electric\_heavy\_utilitary\_vehicle, electric\_public\_transports, electric\_vehicles, hybrid\_car\_passenger, hybrid\_light\_utilitary\_vehicle, hybrid\_heavy\_utilitary\_vehicle, hybrid\_public\_transports},
  ]
  \addplot[color = orange, fill=orange, fill opacity=0.75] table[x=features, y=occurence] {\dataEconomicLinear};
  \addplot[color = spicy_paprika, fill=spicy_paprika, fill opacity=0.75] table[x=features, y=occurence] {\dataEconomicXGBoost};
  \addplot[color = twilight_indigo, , fill=twilight_indigo, fill opacity=0.75] table[x=features, y=occurence] {\dataEconomicTabICL};
\end{axis}

\end{tikzpicture}


%% file: Plots/RMSE_feature_selection.tex




\pgfplotstableread[col sep=comma]{\getErrorDataFilePath{\resolutionMonth}{\linear}{\errorFolder}{\Econ}{\rmse}}\dataRMSELinearMonthEcon
\pgfplotstableread[col sep=comma]{\getErrorDataFilePath{\resolutionMonth}{\linear}{\errorFolder}{\noEcon}{\rmse}}\dataRMSELinearMonthNoEcon
\pgfplotstableread[col sep=comma]{\getFullErrorDataFilePath{\resolutionMonth}{\linear}{\errorFolder}{\Econ}{\rmse}}\dataRMSELinearMonthEconNoFS
\pgfplotstableread[col sep=comma]{\getFullErrorDataFilePath{\resolutionMonth}{\linear}{\errorFolder}{\noEcon}{\rmse}}\dataRMSELinearMonthNoEconNoFS

\pgfplotstableread[col sep=comma]{\getErrorDataFilePath{\resolutionMonth}{\xgboost}{\errorFolder}{\Econ}{\rmse}}\dataRMSEXGBoostMonthEcon
\pgfplotstableread[col sep=comma]{\getErrorDataFilePath{\resolutionMonth}{\xgboost}{\errorFolder}{\noEcon}{\rmse}}\dataRMSEXGBoostMonthNoEcon
\pgfplotstableread[col sep=comma]{\getFullErrorDataFilePath{\resolutionMonth}{\xgboost}{\errorFolder}{\Econ}{\rmse}}\dataRMSEXGBoostMonthEconNoFS
\pgfplotstableread[col sep=comma]{\getFullErrorDataFilePath{\resolutionMonth}{\xgboost}{\errorFolder}{\noEcon}{\rmse}}\dataRMSEXGBoostMonthNoEconNoFS

\pgfplotstableread[col sep=comma]{\getErrorDataFilePath{\resolutionMonth}{\tabicl}{\errorFolder}{\Econ}{\rmse}}\dataRMSETabICLMonthEcon
\pgfplotstableread[col sep=comma]{\getErrorDataFilePath{\resolutionMonth}{\tabicl}{\errorFolder}{\noEcon}{\rmse}}\dataRMSETabICLMonthNoEcon
\pgfplotstableread[col sep=comma]{\getFullErrorDataFilePath{\resolutionMonth}{\tabicl}{\errorFolder}{\Econ}{\rmse}}\dataRMSETabICLMonthEconNoFS
\pgfplotstableread[col sep=comma]{\getFullErrorDataFilePath{\resolutionMonth}{\tabicl}{\errorFolder}{\noEcon}{\rmse}}\dataRMSETabICLMonthNoEconNoFS

\pgfplotstableread[col sep=comma]{\getErrorDataFilePath{\resolutionDay}{\linear}{\errorFolder}{\Econ}{\rmse}}\dataRMSELinearDayEcon
\pgfplotstableread[col sep=comma]{\getErrorDataFilePath{\resolutionDay}{\linear}{\errorFolder}{\noEcon}{\rmse}}\dataRMSELinearDayNoEcon
\pgfplotstableread[col sep=comma]{\getFullErrorDataFilePath{\resolutionDay}{\linear}{\errorFolder}{\Econ}{\rmse}}\dataRMSELinearDayEconNoFS
\pgfplotstableread[col sep=comma]{\getFullErrorDataFilePath{\resolutionDay}{\linear}{\errorFolder}{\noEcon}{\rmse}}\dataRMSELinearDayNoEconNoFS

\pgfplotstableread[col sep=comma]{\getErrorDataFilePath{\resolutionDay}{\xgboost}{\errorFolder}{\Econ}{\rmse}}\dataRMSEXGBoostDayEcon
\pgfplotstableread[col sep=comma]{\getErrorDataFilePath{\resolutionDay}{\xgboost}{\errorFolder}{\noEcon}{\rmse}}\dataRMSEXGBoostDayNoEcon
\pgfplotstableread[col sep=comma]{\getFullErrorDataFilePath{\resolutionDay}{\xgboost}{\errorFolder}{\Econ}{\rmse}}\dataRMSEXGBoostDayEconNoFS
\pgfplotstableread[col sep=comma]{\getFullErrorDataFilePath{\resolutionDay}{\xgboost}{\errorFolder}{\noEcon}{\rmse}}\dataRMSEXGBoostDayNoEconNoFS

\pgfplotstableread[col sep=comma]{\getErrorDataFilePath{\resolutionDay}{\tabicl}{\errorFolder}{\Econ}{\rmse}}\dataRMSETabICLDayEcon
\pgfplotstableread[col sep=comma]{\getErrorDataFilePath{\resolutionDay}{\tabicl}{\errorFolder}{\noEcon}{\rmse}}\dataRMSETabICLDayNoEcon
\pgfplotstableread[col sep=comma]{\getFullErrorDataFilePath{\resolutionDay}{\tabicl}{\errorFolder}{\Econ}{\rmse}}\dataRMSETabICLDayEconNoFS
\pgfplotstableread[col sep=comma]{\getFullErrorDataFilePath{\resolutionDay}{\tabicl}{\errorFolder}{\noEcon}{\rmse}}\dataRMSETabICLDayNoEconNoFS

\begin{tikzpicture}
    \begin{groupplot}[
    group style={
        group size=3 by 2,        
        horizontal sep=1.8cm,
        vertical sep=1.8cm,
    },
    width=6cm,
    height=5cm,
    grid=major,
    scaled ticks=false,
    xtick={1, 12, 24, 36, 48},
    xticklabels={1, 12, 24, 36, 48},
    legend columns = 2,
    legend style={at={(-0.2,-2.2)}, anchor=south west},
]
 
\nextgroupplot[ymax = 300000, title={\textbf{Linear | Month}}, xlabel={Horizon [M]}, ylabel={$RMSE$}]

\addplot[mark=none, line width=1pt, color=teal]
        table [x=horizon, y=mean, col sep=comma] {\dataRMSELinearMonthEcon};
\addlegendentry{Feature Selection on all Weather+Calendar+Economic covariates};
\addplot[mark=none, line width=1pt, color=dark_teal]
        table [x=horizon, y=mean, col sep=comma] {\dataRMSELinearMonthNoEcon};
\addlegendentry{Feature Selection on all Weather+Calendar covariates};
\addplot[mark=none, line width=1pt, color=scarlet_rush]
        table [x=horizon, y=mean, col sep=comma] {\dataRMSELinearMonthEconNoFS};
\addlegendentry{All Weather+Calendar+Economic covariates};
\addplot[mark=none, line width=1pt, color=sunflower_gold]
        table [x=horizon, y=mean, col sep=comma] {\dataRMSELinearMonthNoEconNoFS};
\addlegendentry{All Weather+Calendar covariates};

\addplot[name path=lower_econ_linear_month_fs, fill=none, draw=none, mark=none, forget plot]
    table[x=horizon, y=low_0.95, col sep=comma, header=true]{\dataRMSELinearMonthEcon};
\addplot[name path=upper_econ_linear_month_fs, fill=none, draw=none, mark=none, forget plot]
    table[x=horizon, y=high_0.95, col sep=comma, header=true]{\dataRMSELinearMonthEcon};
\addplot[teal!40, fill opacity=0.75, forget plot] fill between[of=lower_econ_linear_month_fs and upper_econ_linear_month_fs];

\addplot[name path=lower_no_econ_linear_month_fs, fill=none, draw=none, mark=none, forget plot]
    table[x=horizon, y=low_0.95, col sep=comma, header=true]{\dataRMSELinearMonthNoEcon};
\addplot[name path=upper_no_econ_linear_month_fs, fill=none, draw=none, mark=none, forget plot]
    table[x=horizon, y=high_0.95, col sep=comma, header=true]{\dataRMSELinearMonthNoEcon};
\addplot[dark_teal!40, fill opacity=0.75, forget plot] fill between[of=lower_no_econ_linear_month_fs and upper_no_econ_linear_month_fs];

\addplot[name path=lower_econ_linear_month_no_fs, fill=none, draw=none, mark=none, forget plot]
    table[x=horizon, y=low_0.95, col sep=comma, header=true]{\dataRMSELinearMonthEconNoFS};
\addplot[name path=upper_econ_linear_month_no_fs, fill=none, draw=none, mark=none, forget plot]
    table[x=horizon, y=high_0.95, col sep=comma, header=true]{\dataRMSELinearMonthEconNoFS};
\addplot[scarlet_rush!40, fill opacity=0.75, forget plot] fill between[of=lower_econ_linear_month_no_fs and upper_econ_linear_month_no_fs];

\addplot[name path=lower_no_econ_linear_month_no_fs, fill=none, draw=none, mark=none, forget plot]
    table[x=horizon, y=low_0.95, col sep=comma, header=true]{\dataRMSELinearMonthNoEconNoFS};
\addplot[name path=upper_no_econ_linear_month_no_fs, fill=none, draw=none, mark=none, forget plot]
    table[x=horizon, y=high_0.95, col sep=comma, header=true]{\dataRMSELinearMonthNoEconNoFS};
\addplot[sunflower_gold!40, fill opacity=0.75, forget plot] fill between[of=lower_no_econ_linear_month_no_fs and upper_no_econ_linear_month_no_fs];

\nextgroupplot[title={\textbf{XGBoost | Month}}, xlabel={Horizon [M]}, ylabel={$RMSE$}]
\addplot[mark=none, line width=1pt, color=teal]
        table [x=horizon, y=mean, col sep=comma] {\dataRMSEXGBoostMonthEcon};
\addplot[mark=none, line width=1pt, color=dark_teal]
        table [x=horizon, y=mean, col sep=comma] {\dataRMSEXGBoostMonthNoEcon};
\addplot[mark=none, line width=1pt, color=scarlet_rush]
        table [x=horizon, y=mean, col sep=comma] {\dataRMSEXGBoostMonthEconNoFS};
\addplot[mark=none, line width=1pt, color=sunflower_gold]
        table [x=horizon, y=mean, col sep=comma] {\dataRMSEXGBoostMonthNoEconNoFS};

\addplot[name path=lower_econ_xgboost_month_fs, fill=none, draw=none, mark=none, forget plot]
    table[x=horizon, y=low_0.95, col sep=comma, header=true]{\dataRMSEXGBoostMonthEcon};
\addplot[name path=upper_econ_xgboost_month_fs, fill=none, draw=none, mark=none, forget plot]
    table[x=horizon, y=high_0.95, col sep=comma, header=true]{\dataRMSEXGBoostMonthEcon};
\addplot[teal!40, fill opacity=0.75, forget plot] fill between[of=lower_econ_xgboost_month_fs and upper_econ_xgboost_month_fs];

\addplot[name path=lower_no_econ_xgboost_month_fs, fill=none, draw=none, mark=none, forget plot]
    table[x=horizon, y=low_0.95, col sep=comma, header=true]{\dataRMSEXGBoostMonthNoEcon};
\addplot[name path=upper_no_econ_xgboost_month_fs, fill=none, draw=none, mark=none, forget plot]
    table[x=horizon, y=high_0.95, col sep=comma, header=true]{\dataRMSEXGBoostMonthNoEcon};
\addplot[dark_teal!40, fill opacity=0.75, forget plot] fill between[of=lower_no_econ_xgboost_month_fs and upper_no_econ_xgboost_month_fs];

\addplot[name path=lower_econ_xgboost_month_no_fs, fill=none, draw=none, mark=none, forget plot]
    table[x=horizon, y=low_0.95, col sep=comma, header=true]{\dataRMSEXGBoostMonthEconNoFS};
\addplot[name path=upper_econ_xgboost_month_no_fs, fill=none, draw=none, mark=none, forget plot]
    table[x=horizon, y=high_0.95, col sep=comma, header=true]{\dataRMSEXGBoostMonthEconNoFS};
\addplot[scarlet_rush!40, fill opacity=0.75, forget plot] fill between[of=lower_econ_xgboost_month_no_fs and upper_econ_xgboost_month_no_fs];

\addplot[name path=lower_no_econ_xgboost_month_no_fs, fill=none, draw=none, mark=none, forget plot]
    table[x=horizon, y=low_0.95, col sep=comma, header=true]{\dataRMSEXGBoostMonthNoEconNoFS};
\addplot[name path=upper_no_econ_xgboost_month_no_fs, fill=none, draw=none, mark=none, forget plot]
    table[x=horizon, y=high_0.95, col sep=comma, header=true]{\dataRMSEXGBoostMonthNoEconNoFS};
\addplot[sunflower_gold!40, fill opacity=0.75, forget plot] fill between[of=lower_no_econ_xgboost_month_no_fs and upper_no_econ_xgboost_month_no_fs];

\nextgroupplot[title={\textbf{TabICL | Month}}, xlabel={Horizon [M]}, ylabel={$RMSE$}]
\addplot[mark=none, line width=1pt, color=teal]
        table [x=horizon, y=mean, col sep=comma] {\dataRMSETabICLMonthEcon};
\addplot[mark=none, line width=1pt, color=dark_teal]
        table [x=horizon, y=mean, col sep=comma] {\dataRMSETabICLMonthNoEcon};
\addplot[mark=none, line width=1pt, color=scarlet_rush]
        table [x=horizon, y=mean, col sep=comma] {\dataRMSETabICLMonthEconNoFS};
\addplot[mark=none, line width=1pt, color=sunflower_gold]
        table [x=horizon, y=mean, col sep=comma] {\dataRMSETabICLMonthNoEconNoFS};

\addplot[name path=lower_econ_tabicl_month_fs, fill=none, draw=none, mark=none, forget plot]
    table[x=horizon, y=low_0.95, col sep=comma, header=true]{\dataRMSETabICLMonthEcon};
\addplot[name path=upper_econ_tabicl_month_fs, fill=none, draw=none, mark=none, forget plot]
    table[x=horizon, y=high_0.95, col sep=comma, header=true]{\dataRMSETabICLMonthEcon};
\addplot[teal!40, fill opacity=0.75, forget plot] fill between[of=lower_econ_tabicl_month_fs and upper_econ_tabicl_month_fs];

\addplot[name path=lower_no_econ_tabicl_month_fs, fill=none, draw=none, mark=none, forget plot]
    table[x=horizon, y=low_0.95, col sep=comma, header=true]{\dataRMSETabICLMonthNoEcon};
\addplot[name path=upper_no_econ_tabicl_month_fs, fill=none, draw=none, mark=none, forget plot]
    table[x=horizon, y=high_0.95, col sep=comma, header=true]{\dataRMSETabICLMonthNoEcon};
\addplot[dark_teal!40, fill opacity=0.75, forget plot] fill between[of=lower_no_econ_tabicl_month_fs and upper_no_econ_tabicl_month_fs];

\addplot[name path=lower_econ_tabicl_month_no_fs, fill=none, draw=none, mark=none, forget plot]
    table[x=horizon, y=low_0.95, col sep=comma, header=true]{\dataRMSETabICLMonthEconNoFS};
\addplot[name path=upper_econ_tabicl_month_no_fs, fill=none, draw=none, mark=none, forget plot]
    table[x=horizon, y=high_0.95, col sep=comma, header=true]{\dataRMSETabICLMonthEconNoFS};
\addplot[scarlet_rush!40, fill opacity=0.75, forget plot] fill between[of=lower_econ_tabicl_month_no_fs and upper_econ_tabicl_month_no_fs];

\addplot[name path=lower_no_econ_tabicl_month_no_fs, fill=none, draw=none, mark=none, forget plot]
    table[x=horizon, y=low_0.95, col sep=comma, header=true]{\dataRMSETabICLMonthNoEconNoFS};
\addplot[name path=upper_no_econ_tabicl_month_no_fs, fill=none, draw=none, mark=none, forget plot]
    table[x=horizon, y=high_0.95, col sep=comma, header=true]{\dataRMSETabICLMonthNoEconNoFS};
\addplot[sunflower_gold!40, fill opacity=0.75, forget plot] fill between[of=lower_no_econ_tabicl_month_no_fs and upper_no_econ_tabicl_month_no_fs];

\nextgroupplot[ymax = 300000, title={\textbf{Linear | Day}}, xlabel={Horizon [M]}, ylabel={$RMSE$}]
\addplot[mark=none, line width=1pt, color=teal]
        table [x=horizon, y=mean, col sep=comma] {\dataRMSELinearDayEcon};
\addplot[mark=none, line width=1pt, color=dark_teal]
        table [x=horizon, y=mean, col sep=comma] {\dataRMSELinearDayNoEcon};
\addplot[mark=none, line width=1pt, color=scarlet_rush]
        table [x=horizon, y=mean, col sep=comma] {\dataRMSELinearDayEconNoFS};
\addplot[mark=none, line width=1pt, color=sunflower_gold]
        table [x=horizon, y=mean, col sep=comma] {\dataRMSELinearDayNoEconNoFS};

\addplot[name path=lower_econ_linear_day_fs, fill=none, draw=none, mark=none, forget plot]
    table[x=horizon, y=low_0.95, col sep=comma, header=true]{\dataRMSELinearDayEcon};
\addplot[name path=upper_econ_linear_day_fs, fill=none, draw=none, mark=none, forget plot]
    table[x=horizon, y=high_0.95, col sep=comma, header=true]{\dataRMSELinearDayEcon};
\addplot[teal!40, fill opacity=0.75, forget plot] fill between[of=lower_econ_linear_day_fs and upper_econ_linear_day_fs];

\addplot[name path=lower_no_econ_linear_day_fs, fill=none, draw=none, mark=none, forget plot]
    table[x=horizon, y=low_0.95, col sep=comma, header=true]{\dataRMSELinearDayNoEcon};
\addplot[name path=upper_no_econ_linear_day_fs, fill=none, draw=none, mark=none, forget plot]
    table[x=horizon, y=high_0.95, col sep=comma, header=true]{\dataRMSELinearDayNoEcon};
\addplot[dark_teal!40, fill opacity=0.75, forget plot] fill between[of=lower_no_econ_linear_day_fs and upper_no_econ_linear_day_fs];

\addplot[name path=lower_econ_linear_day_no_fs, fill=none, draw=none, mark=none, forget plot]
    table[x=horizon, y=low_0.95, col sep=comma, header=true]{\dataRMSELinearDayEconNoFS};
\addplot[name path=upper_econ_linear_day_no_fs, fill=none, draw=none, mark=none, forget plot]
    table[x=horizon, y=high_0.95, col sep=comma, header=true]{\dataRMSELinearDayEconNoFS};
\addplot[scarlet_rush!40, fill opacity=0.75, forget plot] fill between[of=lower_econ_linear_day_no_fs and upper_econ_linear_day_no_fs];

\addplot[name path=lower_no_econ_linear_day_no_fs, fill=none, draw=none, mark=none, forget plot]
    table[x=horizon, y=low_0.95, col sep=comma, header=true]{\dataRMSELinearDayNoEconNoFS};
\addplot[name path=upper_no_econ_linear_day_no_fs, fill=none, draw=none, mark=none, forget plot]
    table[x=horizon, y=high_0.95, col sep=comma, header=true]{\dataRMSELinearDayNoEconNoFS};
\addplot[sunflower_gold!40, fill opacity=0.75, forget plot] fill between[of=lower_no_econ_linear_day_no_fs and upper_no_econ_linear_day_no_fs];

\nextgroupplot[title={\textbf{XGBoost | Day}}, xlabel={Horizon [M]}, ylabel={$RMSE$}]
\addplot[mark=none, line width=1pt, color=teal]
        table [x=horizon, y=mean, col sep=comma] {\dataRMSEXGBoostDayEcon};
\addplot[mark=none, line width=1pt, color=dark_teal]
        table [x=horizon, y=mean, col sep=comma] {\dataRMSEXGBoostDayNoEcon};
\addplot[mark=none, line width=1pt, color=scarlet_rush]
        table [x=horizon, y=mean, col sep=comma] {\dataRMSEXGBoostDayEconNoFS};
\addplot[mark=none, line width=1pt, color=sunflower_gold]
        table [x=horizon, y=mean, col sep=comma] {\dataRMSEXGBoostDayNoEconNoFS};

\addplot[name path=lower_econ_xgboost_day_fs, fill=none, draw=none, mark=none, forget plot]
    table[x=horizon, y=low_0.95, col sep=comma, header=true]{\dataRMSEXGBoostDayEcon};
\addplot[name path=upper_econ_xgboost_day_fs, fill=none, draw=none, mark=none, forget plot]
    table[x=horizon, y=high_0.95, col sep=comma, header=true]{\dataRMSEXGBoostDayEcon};
\addplot[teal!40, fill opacity=0.75, forget plot] fill between[of=lower_econ_xgboost_day_fs and upper_econ_xgboost_day_fs];

\addplot[name path=lower_no_econ_xgboost_day_fs, fill=none, draw=none, mark=none, forget plot]
    table[x=horizon, y=low_0.95, col sep=comma, header=true]{\dataRMSEXGBoostDayNoEcon};
\addplot[name path=upper_no_econ_xgboost_day_fs, fill=none, draw=none, mark=none, forget plot]
    table[x=horizon, y=high_0.95, col sep=comma, header=true]{\dataRMSEXGBoostDayNoEcon};
\addplot[dark_teal!40, fill opacity=0.75, forget plot] fill between[of=lower_no_econ_xgboost_day_fs and upper_no_econ_xgboost_day_fs];

\addplot[name path=lower_econ_xgboost_day_no_fs, fill=none, draw=none, mark=none, forget plot]
    table[x=horizon, y=low_0.95, col sep=comma, header=true]{\dataRMSEXGBoostDayEconNoFS};
\addplot[name path=upper_econ_xgboost_day_no_fs, fill=none, draw=none, mark=none, forget plot]
    table[x=horizon, y=high_0.95, col sep=comma, header=true]{\dataRMSEXGBoostDayEconNoFS};
\addplot[scarlet_rush!40, fill opacity=0.75, forget plot] fill between[of=lower_econ_xgboost_day_no_fs and upper_econ_xgboost_day_no_fs];

\addplot[name path=lower_no_econ_xgboost_day_no_fs, fill=none, draw=none, mark=none, forget plot]
    table[x=horizon, y=low_0.95, col sep=comma, header=true]{\dataRMSEXGBoostDayNoEconNoFS};
\addplot[name path=upper_no_econ_xgboost_day_no_fs, fill=none, draw=none, mark=none, forget plot]
    table[x=horizon, y=high_0.95, col sep=comma, header=true]{\dataRMSEXGBoostDayNoEconNoFS};
\addplot[sunflower_gold!40, fill opacity=0.75, forget plot] fill between[of=lower_no_econ_xgboost_day_no_fs and upper_no_econ_xgboost_day_no_fs];

\nextgroupplot[title={\textbf{TabICL | Day}}, xlabel={Horizon [M]}, ylabel={$RMSE$}]
\addplot[mark=none, line width=1pt, color=teal]
        table [x=horizon, y=mean, col sep=comma] {\dataRMSETabICLDayEcon};
\addplot[mark=none, line width=1pt, color=dark_teal]
        table [x=horizon, y=mean, col sep=comma] {\dataRMSETabICLDayNoEcon};
\addplot[mark=none, line width=1pt, color=scarlet_rush]
        table [x=horizon, y=mean, col sep=comma] {\dataRMSETabICLDayEconNoFS};
\addplot[mark=none, line width=1pt, color=sunflower_gold]
        table [x=horizon, y=mean, col sep=comma] {\dataRMSETabICLDayNoEconNoFS};

\addplot[name path=lower_econ_tabicl_day_fs, fill=none, draw=none, mark=none, forget plot]
    table[x=horizon, y=low_0.95, col sep=comma, header=true]{\dataRMSETabICLDayEcon};
\addplot[name path=upper_econ_tabicl_day_fs, fill=none, draw=none, mark=none, forget plot]
    table[x=horizon, y=high_0.95, col sep=comma, header=true]{\dataRMSETabICLDayEcon};
\addplot[teal!40, fill opacity=0.75, forget plot] fill between[of=lower_econ_tabicl_day_fs and upper_econ_tabicl_day_fs];

\addplot[name path=lower_no_econ_tabicl_day_fs, fill=none, draw=none, mark=none, forget plot]
    table[x=horizon, y=low_0.95, col sep=comma, header=true]{\dataRMSETabICLDayNoEcon};
\addplot[name path=upper_no_econ_tabicl_day_fs, fill=none, draw=none, mark=none, forget plot]
    table[x=horizon, y=high_0.95, col sep=comma, header=true]{\dataRMSETabICLDayNoEcon};
\addplot[dark_teal!40, fill opacity=0.75, forget plot] fill between[of=lower_no_econ_tabicl_day_fs and upper_no_econ_tabicl_day_fs];

\addplot[name path=lower_econ_tabicl_day_no_fs, fill=none, draw=none, mark=none, forget plot]
    table[x=horizon, y=low_0.95, col sep=comma, header=true]{\dataRMSETabICLDayEconNoFS};
\addplot[name path=upper_econ_tabicl_day_no_fs, fill=none, draw=none, mark=none, forget plot]
    table[x=horizon, y=high_0.95, col sep=comma, header=true]{\dataRMSETabICLDayEconNoFS};
\addplot[scarlet_rush!40, fill opacity=0.75, forget plot] fill between[of=lower_econ_tabicl_day_no_fs and upper_econ_tabicl_day_no_fs];

\addplot[name path=lower_no_econ_tabicl_day_no_fs, fill=none, draw=none, mark=none, forget plot]
    table[x=horizon, y=low_0.95, col sep=comma, header=true]{\dataRMSETabICLDayNoEconNoFS};
\addplot[name path=upper_no_econ_tabicl_day_no_fs, fill=none, draw=none, mark=none, forget plot]
    table[x=horizon, y=high_0.95, col sep=comma, header=true]{\dataRMSETabICLDayNoEconNoFS};
\addplot[sunflower_gold!40, fill opacity=0.75, forget plot] fill between[of=lower_no_econ_tabicl_day_no_fs and upper_no_econ_tabicl_day_no_fs];
\end{groupplot}

\end{tikzpicture}

%% file: Plots/RMSE_context_ablation.tex
%

\pgfplotstableread[col sep=comma]{\getContextAblationDataFilePath{\resolutionMonth}{\tabicl}{\contextimpFolder}{\Econ}{\rmse}}\dataContextAblationTabICLRMSE

\begin{tikzpicture}
\begin{axis}[
        title = {\textbf{TabICL | Month}},
        width=15cm,
        height=10cm,
        grid=major,
        thick,
        mark size=2pt,
        xlabel={Horizon [M]},
        ylabel={$RMSE$},
        xmin=0,
        xmax=49,
        xtick={1,6,12,18,24,30,36,42,48},
        xticklabels={1,6,12,18,24,30,36,42,48},
        legend columns = 3,
        legend style={at={(1.0, 0)}, anchor=south east},
        ]
\addplot+[mark=none, line width=1.5pt, color=black, style=dashed]
        table [x=horizon, y=1.0_mean, col sep=comma] {\dataContextAblationTabICLRMSE};
\addlegendentry{$\SI{100}{\percent}$};
\addplot+[mark=none, line width=1pt, color=dusk_blue, style=solid]
        table [x=horizon, y=0.9_mean, col sep=comma] {\dataContextAblationTabICLRMSE};
\addlegendentry{$\SI{90}{\percent}$};
\addplot+[mark=none, line width=1pt, color=dusty_lavender, style=solid]
        table [x=horizon, y=0.8_mean, col sep=comma] {\dataContextAblationTabICLRMSE};
\addlegendentry{$\SI{80}{\percent}$};
\addplot+[mark=none, line width=1pt, color=rosewood, style=solid]
        table [x=horizon, y=0.7_mean, col sep=comma] {\dataContextAblationTabICLRMSE};
\addlegendentry{$\SI{70}{\percent}$};
\addplot+[mark=none, line width=1pt, color=light_coral, style=solid]
        table [x=horizon, y=0.6_mean, col sep=comma] {\dataContextAblationTabICLRMSE};
\addlegendentry{$\SI{60}{\percent}$};
\addplot+[mark=none, line width=1pt, color=light_bronze, style=solid]
        table [x=horizon, y=0.5_mean, col sep=comma] {\dataContextAblationTabICLRMSE};
\addlegendentry{$\SI{50}{\percent}$};
\end{axis}
\end{tikzpicture}

%% file: Plots/complexity.tex
%

\pgfplotstableread[col sep=comma]{\detokenize{Data/month/tabICL/complexity_month.txt}}\dataComplexityTabICLMonth
\pgfplotstableread[col sep=comma]{\detokenize{Data/day/tabICL/complexity_day.txt}}\dataComplexityTabICLDay

\begin{tikzpicture}

\begin{groupplot}[
    group style={
        group size=2 by 1,        
        horizontal sep=1.8cm,
    },
    width=7cm,
    height=5cm,
    scale only axis,
    grid=major,
    scaled ticks=false,
    legend columns=4,
    legend style={at={(0.25,-0.5)}, anchor=south west}, 
]
\nextgroupplot[ymin = 0, ymax = 1, title={\textbf{TabICL | Month}}, xlabel={Number of rows $n$ in context + inference data}, ylabel={Complexity $\mathcal{O}\left(n^{2} + nm^{2}\right)$ (Normalized)}]
\addplot+[mark=none, line width=1.5pt, color=dusk_blue, style=solid]
        table [x=n_samples, y=complexity_55_features_norm, col sep=comma] {\dataComplexityTabICLMonth};
\addlegendentry{$\SI{100}{\percent}$ | $m=55$};

\addplot+[mark=none, line width=1.5pt, color=dusty_lavender, style=solid]
        table [x=n_samples, y=complexity_15_features_norm, col sep=comma] {\dataComplexityTabICLMonth};
\addlegendentry{$\SI{100}{\percent}$ | $m=15$};

\addplot+[mark=none, line width=1.5pt, color=rosewood, style=solid]
        table [x=n_samples, y=complexity_25_features_norm, col sep=comma] {\dataComplexityTabICLMonth};
\addlegendentry{$\SI{100}{\percent}$ | $m=25$};

\addplot+[mark=none, line width=1.5pt, color=light_coral, style=solid]
        table [x=n_samples, y=complexity_35_features_norm, col sep=comma] {\dataComplexityTabICLMonth};
\addlegendentry{$\SI{100}{\percent}$ | $m=35$};

\addplot+[mark=none, line width=1.5pt, color=dusk_blue, style=dashed]
        table [x=n_samples, y=complexity_55_features_0.8_norm, col sep=comma] {\dataComplexityTabICLMonth};
\addlegendentry{$\SI{80}{\percent}$ | $m=55$};

\addplot+[mark=none, line width=1.5pt, color=dusty_lavender, style=dashed]
        table [x=n_samples, y=complexity_15_features_0.8_norm, col sep=comma] {\dataComplexityTabICLMonth};
\addlegendentry{$\SI{80}{\percent}$ | $m=15$};

\addplot+[mark=none, line width=1.5pt, color=rosewood, style=dashed]
        table [x=n_samples, y=complexity_25_features_0.8_norm, col sep=comma] {\dataComplexityTabICLMonth};
\addlegendentry{$\SI{80}{\percent}$ | $m=25$};

\addplot+[mark=none, line width=1.5pt, color=light_coral, style=dashed]
        table [x=n_samples, y=complexity_35_features_0.8_norm, col sep=comma] {\dataComplexityTabICLMonth};
\addlegendentry{$\SI{80}{\percent}$ | $m=35$};

\nextgroupplot[ymin = 0, ymax = 1, title={\textbf{TabICL | Day}}, xlabel={Number of rows $n$ in context + inference data}, ylabel={Complexity $\mathcal{O}\left(n^{2} + nm^{2}\right)$ (Normalized)}]
\addplot+[mark=none, line width=1.5pt, color=dusk_blue, style=solid]
        table [x=n_samples, y=complexity_55_features_norm, col sep=comma] {\dataComplexityTabICLDay};
\addplot+[mark=none, line width=1.5pt, color=dusty_lavender, style=solid]
        table [x=n_samples, y=complexity_15_features_norm, col sep=comma] {\dataComplexityTabICLDay};
\addplot+[mark=none, line width=1.5pt, color=rosewood, style=solid]
        table [x=n_samples, y=complexity_25_features_norm, col sep=comma] {\dataComplexityTabICLDay};
\addplot+[mark=none, line width=1.5pt, color=light_coral, style=solid]
        table [x=n_samples, y=complexity_35_features_norm, col sep=comma] {\dataComplexityTabICLDay};
\addplot+[mark=none, line width=1.5pt, color=dusk_blue, style=dashed]
        table [x=n_samples, y=complexity_55_features_0.8_norm, col sep=comma] {\dataComplexityTabICLDay};
\addplot+[mark=none, line width=1.5pt, color=dusty_lavender, style=dashed]
        table [x=n_samples, y=complexity_15_features_0.8_norm, col sep=comma] {\dataComplexityTabICLDay};
\addplot+[mark=none, line width=1.5pt, color=rosewood, style=dashed]
        table [x=n_samples, y=complexity_25_features_0.8_norm, col sep=comma] {\dataComplexityTabICLDay};
\addplot+[mark=none, line width=1.5pt, color=light_coral, style=dashed]
        table [x=n_samples, y=complexity_35_features_0.8_norm, col sep=comma] {\dataComplexityTabICLDay};
\end{groupplot}
\end{tikzpicture}